\documentclass[a4paper,11pt]{book}
\usepackage[a4paper,
            bindingoffset=0.2in,
            left=1.25in,
            right=1.25in,
            top=1.1in,
            bottom=1.1in,
            footskip=.25in]{geometry}

\usepackage[authoryear,colon,square]{natbib}
\usepackage{style}
\usepackage{macros}
\usepackage{algo}

\begin{document}

\pagecolor{anthracite}\afterpage{\nopagecolor}

\definecolor{poste_color}{RGB}{222, 184, 65}

\begin{center}
\color{white}
\begin{minipage}{0.99\linewidth}
    \thispagestyle{empty}
    \centering
    \tikz[remember picture,overlay] \node[opacity=0.3,inner sep=0pt] at (current page.center){\includegraphics[width=\paperwidth,height=\paperheight]{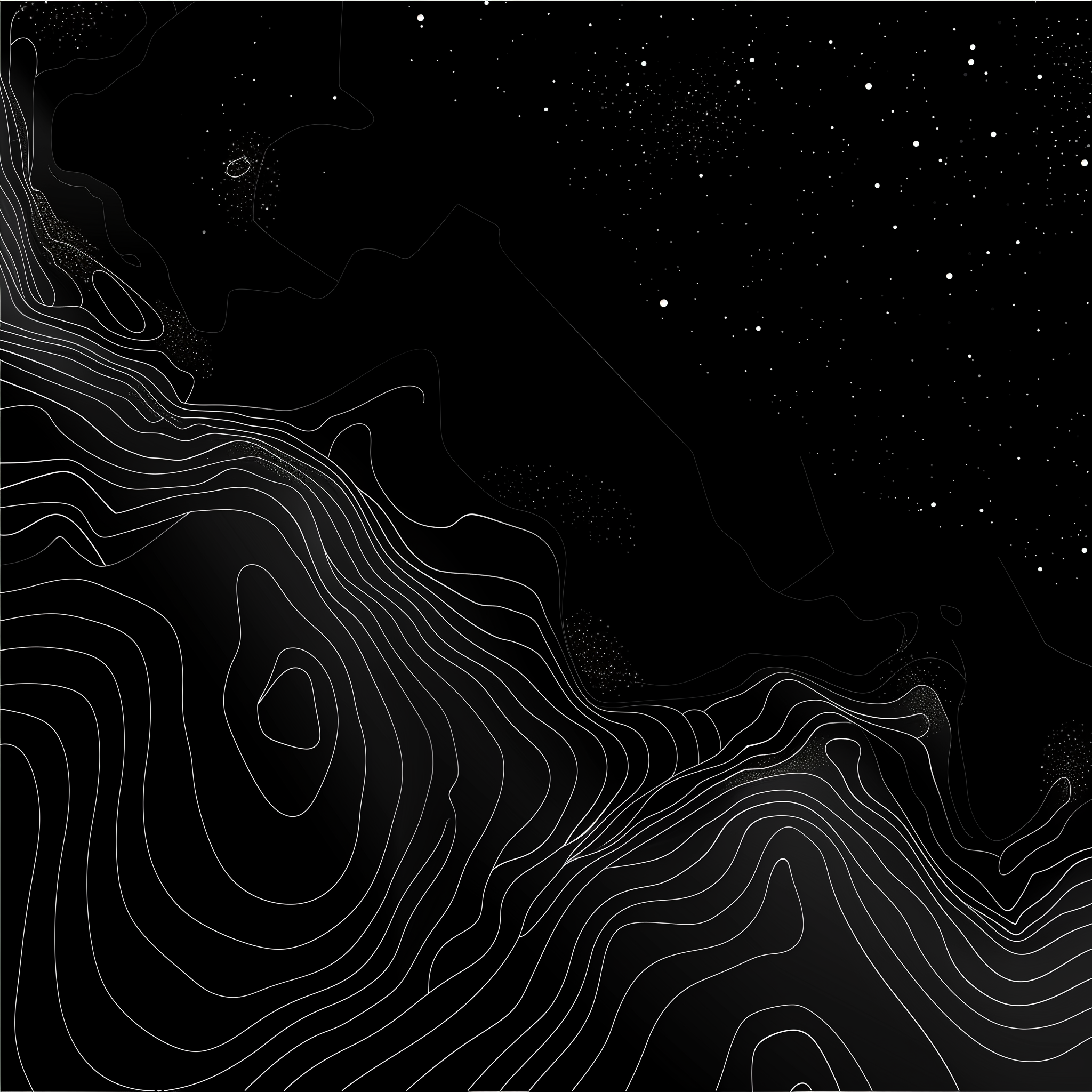}};

    \vspace{-1cm}

    \vspace{1cm}
    
    {\Huge {\textbf{Sparks of Explainability}}\\ Recent Advancements in Explaining Large Vision Models\\[0.5em]}

    \vspace{2cm}
    
    {
    Presented by\\
    {\textbf{\Large Thomas Fel}}\\[1em]

    Supervised by\\
    \textbf{Prof. Thomas Serre}\\[4em]
    }

    \vspace{1cm}
    
    {
    \raggedright
    Presented and publicly defended on July 25, 2024\\
    }
    \vspace{0em}
    \begin{tabbing}
        \hspace{8cm} \= \kill
        Prof. \textbf{George A. Alvarez} \> Reviewer\\
        \textit{\textcolor{poste_color}{Professor, Harvard University}} \\[0.5em]
        
        Prof. \textbf{Céline Hudelot} \> Reviewer\\
        \textit{\textcolor{poste_color}{Professor, Centrale Paris}} \\[0.5em]

        Dr. \textbf{Robert Geirhos} \> Examiner\\
        \textit{\textcolor{poste_color}{Research Scientist, Google}} \\[0.5em]
        
        Prof. \textbf{Ruth Fong} \> Examiner\\
        \textit{\textcolor{poste_color}{Professor, Princeton University}} \\[0.5em]
        
        Prof. \textbf{Rufin Van Rullen} \> Examiner\\
        \textit{\textcolor{poste_color}{Professor, Cerco, CNRS}} \\[0.5em]

        \\
        
        \textbf{Prof. Thomas Serre} \> Thesis Director\\
        \textit{\textcolor{poste_color}{Professor, Brown University \& ANITI}} \\[0.5em]
    \end{tabbing}

    \vspace{1.5cm}
    
    {\Large\textbf{DOCTORAL THESIS}\\[0.5em]}
    {\large Doctoral School of Mathematics, Computer Science, and Telecommunications of Toulouse}\\[2em]
    
\end{minipage}
\end{center}

\clearpage

\thispagestyle{empty}

\vspace*{5cm}

\setlength{\epigraphwidth}{0.5\textwidth}

{
\epigraph{\large``Understanding the world for a man is reducing it to the human, stamping it with his seal.''.}{\textit{\large Albert Camus}}
}

\clearpage

\begin{center}
{\Large Résumé \\ \adforn{21}}
\end{center}

Cette thèse doctorale vise à pousser les frontières de l'état de l'art ainsi que le développement d'outils dans le domaine de l'explicabilité en vision par ordinateur. Elle se focalise spécifiquement sur la construction d'un ensemble d'outils destinés à améliorer notre compréhension des caractéristiques (ou \textit{features}) exploitées par les réseaux de neurones profonds actuellement employés dans des tâches de vision. L'explicabilité représente un domaine clé pour améliorer les interactions entre les humains et les systèmes d'intelligence artificielle, pour certifier ces même système dans des applications critique mais aussi d'un point de vue scientifique pour décrypter un nouveau type d'intelligence: l'intelligence artificielle. 
De manière plus concrète, la complexité et le manque de transparence de ces modèles constituent un obstacle majeur à leur adoption dans des applications qui demande un haut niveau de sécurité et de confiance. L'explicabilité est au coeur de ces problématiques, et les réponses que ce champ de recherche doit apporter sont attendus et permettraient des avancées technologiques significatives et une adoption plus rapide de l'intelligence artificielle. À travers ce manuscrit, nous explorons et proposons de nouvelles méthodes d'explicabilité, apportant chacune une contribution à la compréhension des modèles de vision.

Nous entamons cette thèse par une analyse détaillée des méthodes d'attribution, également connues sous le nom de cartes de saillance. Ces techniques révèlent où le modèle porte son attention pour prendre une décision, grâce à l'utilisation de cartes thermiques. La première section propose une métrique, inspirée de la stabilité algorithmique, qui se fonde sur ces attributions pour évaluer la qualité des explications fournies par les modèles, permettant ainsi d'identifier ceux offrant les meilleures explications. Nous introduisons ensuite une nouvelle méthode d'attribution, inspirée du champ de l'Analyse de Sensibilité Globale, basée sur les indices de Sobol. Cette approche de type boîte noire, soutenue par un fondement théorique solide, permet de réduire de moitié le temps de calcul par rapport à l'état de l'art grâce à l'utilisation de séquences quasi-Monte Carlo. Nous poursuivons en présentant la première méthode d'attribution dotée de garanties formelles, \eva, qui repose sur l'analyse de perturbation vérifiée.

De manière surprenante, nous avons constaté que ces méthodes, lorsqu'elles sont testées dans des cas d'usage réels avec de véritables utilisateurs, s'avèrent peu utiles pour comprendre les modèles. Plus spécifiquement, dans des scénarios complexes, ces techniques se révèlent inefficaces, alors qu'elles suffisent pour identifier des biais dans des contextes plus simples. Deux hypothèses sont alors formulées pour surmonter ces défis : la première suggère la nécessité d'aligner les modèles de vision par ordinateur avec le raisonnement humain, remettant ainsi en question le modèle lui-même ; la seconde avance que les méthodes d'attribution actuelles ne sont pas suffisantes et ne révèlent pas assez d'informations. Ces hypothèses sont ensuite examinées dans des chapitres dédiés.

Pour aborder la première hypothèse, nous proposons une nouvelle routine d'entraînement qui vise non seulement à minimiser la fonction de coût habituelle mais aussi à imiter les explications humaines, autrement dit, à avoir raison pour les bonnes raisons. Étonnamment, non seulement les modèles parviennent à généraliser, adoptant des stratégies humaines, mais leur précision augmente également. Nous explorons ensuite une seconde approche visant à aligner les modèles non pas par régularisation mais par contrainte, optimisant dans un espace fonctionnel restreint : celui des fonctions 1-Lipschitz. L'analyse établit un lien entre la robustesse des modèles, notamment ceux caractérisés par une propriété 1-Lipschitz, et leur capacité à fournir des explications alignées avec le raisonnement humain.

Ensuite, nous examinons la seconde hypothèse, selon laquelle les méthodes d'attribution actuelles sont insuffisantes car elles révèlent uniquement \textit{où} le modèle porte son attention, sans expliciter \textit{ce qu'il perçoit}. Nous adoptons une approche d'explicabilité basée sur les concepts, évoluant de la focalisation sur le « où » vers une compréhension du « quoi » perçu par le modèle. Cette transition est concrétisée par la méthode \craft, qui automatise l'extraction des concepts utilisés par un modèle et évalue ensuite l'importance de chaque concept extrait. Nous analysons en profondeur les composantes des méthodes actuelles d'extraction de concept et démontrons qu'elles comprennent deux phases : une phase d'extraction et une phase d'estimation de l'importance. Nous unifions ensuite les différentes approches de la littérature en montrant que la phase d'extraction peut être conceptualisée comme un problème d'apprentissage de dictionnaire, et que la phase d'estimation d'importance utilise implicitement des méthodes d'attribution. Après avoir établi ce cadre unificateur, nous introduisons \maco, une méthode de visualisation des caractéristiques que nous appliquons aux concepts, permettant de visualiser les concepts extraits. Nous concluons en intégrant ces différentes méthodes dans une démonstration interactive, qui offre une exploration et une compréhension des concepts les plus importants pour les 1000 classes d'ImageNet d'un modèle ResNet.

La thèse se termine par une réflexion approfondie sur les méthodes développées, les progrès réalisés et les défis rencontrés, ouvrant des perspectives sur les futures directions de recherche en explicabilité en Intelligence Artificielle (IA). Nous soulignons l'importance de poursuivre la recherche de synergies entre les différentes méthodes étudiées, ainsi que les voies prometteuses pour exploiter pleinement le potentiel de l'explicabilité.

\clearpage

\begin{center}
{\Large Abstract \\ \adforn{21}}
\end{center}

This doctoral thesis aims to advance the state of the art and the development of tools in the field of explainability in computer vision. It specifically focuses on creating a set of tools designed to enhance our understanding of the features utilized by deep neural networks currently employed in vision tasks. Explainability represents a key area for improving interactions between humans and artificial intelligence systems, as well as from a scientific standpoint to decipher a new type of intelligence: artificial intelligence. More concretely, the complexity and lack of transparency of these models pose a major obstacle to their adoption in critical systems and raise crucial questions, potentially capable of leading to significant advances in our understanding of intelligence, provided their mechanisms can be deciphered. Through this manuscript, we explore several explainability methods, each contributing to the understanding and improvement of the explainability of vision models while acknowledging their respective limitations.

We begin this thesis with a detailed analysis of attribution methods, also known as saliency maps or heat maps. These techniques reveal where the model focuses its attention to make a decision, through the use of heatmaps. The first paper proposes a metric inspired by algorithmic stability that is based on these attributions to assess the quality of explanations provided by the models, thus identifying those offering the best explanations. We then introduce a new attribution method inspired by the field of Global Sensitivity Analysis based on Sobol indices. This black-box approach, supported by a solid theoretical foundation, allows for halving the computation time compared to the state of the art through the use of quasi-Monte Carlo sequences. We continue by presenting the first attribution method with formal guarantees, \eva, which relies on verified perturbation analysis.

Surprisingly, we found that these methods, when tested in real-use cases with actual users, prove to be of little use in understanding the models. More specifically, in complex scenarios, these techniques prove ineffective, while they are sufficient to identify biases in simpler contexts. Two hypotheses are then formulated to overcome these challenges: the first suggests the need to align computer vision models with human reasoning, thereby questioning the model itself; the second advances that current attribution methods are not sufficient and do not reveal enough information. These hypotheses are then examined in dedicated chapters.

To address the first hypothesis, we propose a new training routine aimed not only at minimizing the usual cost function but also at mimicking human explanations, in other words, being right for the right reasons. Surprisingly, not only do the models manage to generalize, adopting human strategies, but their accuracy also increases. We then explore a second approach aimed at aligning models not through regularization but through constraint, optimizing in a restricted functional space: that of 1-Lipschitz functions. The analysis establishes a link between the robustness of the models, especially those characterized by a 1-Lipschitz property, and their ability to provide explanations aligned with human reasoning.

Next, we examine the second hypothesis, according to which current attribution methods are insufficient because they reveal only \textit{where} the model focuses its attention, without specifying \textit{what it perceives}. We adopt an explainability approach based on concepts, moving from focusing on the "where" to understanding the "what" perceived by the model. This transition is materialized by the \craft~method, which automates the extraction of concepts used by a model and then assesses the importance of each extracted concept. We thoroughly analyze the components of current concept extraction methods and demonstrate that they include two phases: an extraction phase and an importance estimation phase. We then unify the different approaches in the literature by showing that the extraction phase can be conceptualized as a dictionary learning problem, and that the importance estimation phase implicitly uses attribution methods. After establishing this unifying framework, we introduce \maco, a feature visualization method that we apply to concepts, allowing the visualization of extracted concepts. We conclude by integrating these different methods into an interactive demonstration, offering exploration and understanding of the most important concepts for the 1000 ImageNet classes of a ResNet model.

The thesis concludes with a thorough reflection on the developed methods, the progress made, and the challenges encountered, opening perspectives on future research directions in AI explainability. We emphasize the importance of continuing the search for synergies between the different methods studied, as well as promising avenues for fully leveraging the potential of explainable AI.

\clearpage

\section*{Remerciements}

Au seuil de cette aventure qu'a été ma thèse, je me trouve face à la tâche délicate de traduire en mots simples, mais chargés de sens, toute l'étendue de ma gratitude envers tous ceux qui m'ont accompagné durant ces trois ans.

En premier lieu, mes pensées se dirigent vers le Professeur Thomas Serre. Sans son soutien constant, les pages de cette thèse seraient restées désespérément blanches. Les nombreuses réflexions que nous avons partagées durant ces trois années ont été essentielles à cette aventure et m'ont permis de vivre une expérience incroyablement enrichissante, plongé avec passion dans le monde fascinant de l'explicabilité. Pour tout cela, Thomas, je te remercie du fond du cœur.

Je tiens à remercier les Professeurs George A. Alvarez et Céline Hudelot d'avoir accepté d'évaluer ma thèse de doctorat. Je remercie également Ruth Fong, Robert Geirhos et Ruffin Van Rullen  pour avoir accepté de faire partie de mon comité de thèse.

Je tiens ensuite à rendre un hommage chaleureux à mes compagnons de route, Agustin, Louis et Thibaut. Avec vous, chaque jour de travail était une aventure ; nos escapades, du désert de Salta aux nuits polaires d'Ushuaïa, resteront à jamais gravées dans ma mémoire. Merci pour ces instants de pure fraternité, pour nos interrogations naïves, mais surtout pour votre générosité qui a été un soutien inestimable.

Je ne saurai passer sous silence la gratitude que je porte à Rémi. Ta bonté et ton calme sont un phare pour ceux qui ont le privilège de te connaître. Cette thèse a été l'occasion de croiser ton chemin. Tes conseils, ton temps généreusement offert, et nos discussions, tantôt profondes, tantôt très absurdes, m'ont accompagné tout au long de ce voyage.

Un merci tout particulier à Gregory, Laurent et Claire qui ont été les artisans discrets de cette quête, me soutenant à chaque pas et veillant sur ma liberté intellectuelle. Leur encouragement a été un don précieux, permettant à cette recherche de s'épanouir. Laurent, nos conversations nourriront ma réflexion pour longtemps.
David Vigouroux mérite une mention spéciale. Tu as été l'étincelle initiale de mon épanouissement à l'IRT, un mentor dont l'intelligence et le soutien, souvent en coulisses, ont été déterminants. Franck, ta gentillesse, ta sagesse et ton calme ont été pour moi une source d'inspiration constante.
À toute l'équipe DEEL, Ana, David B., Adil et Paul, nos échanges, nos rires et nos soirées resteront parmi mes meilleurs souvenirs. À Justin et Mikaël, pour avoir gardé nos serveurs à flot durant ces trois années, ce qui n'a pas été facile, mais sans qui rien n'aurait été possible.
Mon parcours m'a ensuite mené à Brown, où j'ai eu l'immense chance de rencontrer un autre mentor exceptionnel, Drew Linsley. Ta guidance, empreinte de bienveillance, a été un cadeau. Ivan, nos sessions de codage, d'apprentissage et nos discussions transatlantiques resteront gravées dans ma mémoire.
Je suis profondément reconnaissant envers Katherine pour son influence bienveillante et intelligente, qui m’a permis d’envisager la suite de cette thèse avec confiance. Merci pour ta guidance précieuse durant ces moments de passage.
Victor, ta contribution à la dernière étape de ma thèse a été source d'inspiration sous bien des aspects. Tu m'as donné une vision claire du chercheur que j'aspire à devenir, merci. Mélanie, ta profonde expertise n'a d'égal que ton humilité. Tes formations et tes conseils ont été très précieux durant cette aventure.
Enfin, un merci du fond du cœur à tous ceux qui ont jalonné ce voyage, enrichissant chaque étape de leur présence. Merci à Mathieu pour son intelligence et son humour, à Léo pour m'avoir fait découvrir la prédiction conforme, à Lucas et Antonin pour les bons moments passé à développer Xplique, à Sabine pour les moments à Brown, et à Julien pour toutes les discussions enrichissantes et ces soirées à discuter d'explicabilité. Enfin, je voudrais remercier chaleureusement toute l'équipe de l'IA à la SNCF, qui m'a si bien accueilli.
Mes amis, piliers du quotidien, ont été d'un soutien constant. Un immense merci à JL pour avoir toujours su me remonter le moral. Un merci tout particulier également à Théo, Anthony, Rayane, Hamza, Bruno, Roxane, Bastien, Martin, Lucas, Damien, Théo et tous les autres, qui ont partagé avec moi les hauts et les bas de cette quête. Alessandra, ton accompagnement durant cette année charnière a été précieux, un véritable trésor.

Pour conclure cette section de remerciements, je tiens à dédier mes ultimes mots à mon frère Arthur ainsi qu'à mes parents. Ils ont été le socle solide sur lequel j'ai bâti mes rêves et ma persévérance. Sans votre amour et votre foi constante, bien des chemins auraient été plus ardus. Par-delà l'apport académique de cette thèse se cache une ambition, plus vulgaire, mais que je sais partager avec beaucoup : celle de voir, même brièvement, vos yeux s'illuminer d'un éclair de fierté.

\clearpage

\section*{Notations}
\label{notations}

\renewcommand{\arraystretch}{1.2}
\begin{tabular}{p{2.0cm}|p{15cm}}
$\sx$ & Set. \\
$\mathbb{N}$ & Set of integers. \\
$\Real$ & Set of real numbers. \\
$\subset$ & Strict inclusion. \\
$\subseteq$ & Inclusion with possible equality. \\

\\
$\abs{\cdot}$ & Absolute value of a real number. \\
$\norm{\cdot}$ & Euclidean norm of a vector. \\
$x$ & A real scalar belonging to $\Real$. \\
$\vx$ & A vector in $\Real^n$. \\
$\mx$ & A matrix in $\Real^{n \times d}$. \\
$\Id_n$ & Identity matrix of size $n$. \\
$\mx^\tr$ & Transpose of $\mx$. \\
$\mx^{-1}, \mx^\pinv$ & Inverse and Moore–Penrose pseudo-inverse of $\mx$. \\
$\odot$ & Hadamard product (element-wise product). \\

\\
$\rx$ & A random scalar variable. \\
$\rvx$ & A random vector in $\Real^n$. \\
$\rmx$ & A random matrix in $\Real^{n \times d}$. \\
$\P(\rx), \P(\rvx)$ & Probability distribution of $\rx$ (resp. joint probability distribution of $\rvx$). \\
$\rx \sim \P_{\rx}$ & Random variable $\rx$ follows a probability distribution $\P_{\rx}$. \\
$\E(\rx)$ & Expectation of $\rx$ with respect to $\P(\rx)$. \\
$\V(\rx)$ & Variance of $\rx$ under $\P(\rx)$. \\

\\
$x_i$ & $i$-th element of the vector $\vx$. \\
$X_{i,j}$ & Element at row $i$, column $j$ of the matrix $\mx$. \\
$\mx_{i, :}$ & $i$-th row of the matrix $\mx$. \\
$\mx_{:, i}$ & $i$-th column of the matrix $\mx$. \\
$\vx_{\v{u}}$ & Subvector of $\vx$ with indices $\v{u}$ such that $\vx_{\v{u}} = (x_i)_{i \in \v{u}}$. \\
$\vx_{\complementary\v{u}}$ & Complementary subvector of $\vx_{\v{u}}$ with indices $\complementary\v{u}$ such that $\vx_{\complementary\v{u}} = (x_i)_{i \notin \v{u}}$. \\

\\
$\f(\vx; \v{\theta})$ & A function of $\vx$ parametrized by $\v{\theta}$, we sometimes write $\f(\vx)$, omitting $\v{\theta}$ for brevity. \\
$\f \compose \g$ & Composition of functions $\f$ and $\g$. \\
$dy / dx$ & Derivative of $y$ with respect to $x$. \\
$\pd y / \pd x$ & Partial derivative of $y$ with respect to $x$. \\
$\grad_{\vx} y$ & Gradient of $y$ with respect to $\vx$. \\
$\pd \f / \pd \vx$ & Jacobian matrix $\m{J} \in \Real^{n \times p}$ of $\f : \Real^n \to \Real^p$.
\end{tabular}
\renewcommand{\arraystretch}{1}

\vspace{0.2cm}

Sometimes we use a function $\f$ whose argument is a scalar but apply it to a vector $\f(\vx)$ or a matrix $\f(\mx)$. This denotes the application of $\f$ to the array element-wise. 

\clearpage

\tableofcontents

\chapter{General Introduction}
\label{chap:intro}

\epigraph{``One sits down on a desert sand dune, sees nothing, hears nothing. Yet through the silence something throbs, and gleams.''}{\textit{Antoine de Saint-Exupéry}}

Some mysteries are meant to remain unsolved; Deep Learning is not one of them. At the entrance of this document, before exploring the subject of Deep Learning—where the \textit{silence} of our understanding contrasts sharply with the powerful \textit{noise} of its achievements—let us take a moment to reflect on the origins of AI.

The concept of creating thinking machines emerged during the Dartmouth Workshop in 1956~\cite{mccarthy2006proposal}\footnote{Interestingly, Lloyd Shapley, who we will encounter later in this manuscript, was invited and is already mentioned in the workshop proposal.}, a seminal event that marked the inception of Artificial Intelligence (AI) as a formal research discipline. This gathering laid the foundational stones for exploring the potential of developing machines endowed with intelligent capabilities. Since then, AI has traversed through various evolutionary stages, characterized by alternating waves of enthusiasm spurred by significant breakthroughs. Alan Turing's prediction regarding the ascent of artificial intelligence appears to have been prophetic, as evidenced by the successive emergence of Machine Learning and subsequently Deep Learning, two branches of statistical learning.

Approximately a decade ago, AI witnessed a transformation with the advent of Deep Learning (DL)~\cite{lecun2015deep,Serre2019DeepLT}. Deep Learning methodologies, rooted in deep neural networks, have catalyzed revolutionary advancements across diverse domains by showcasing exceptional aptitude in discerning complex patterns and behaviors from large datasets. The surge in Deep Learning's prominence can be attributed to several factors, including the exponential increase of data, advancements in hardware and software for machine learning, and pivotal breakthroughs in research methodologies.

A pivotal moment in the adoption of Deep Learning occurred in 2012, when the Computer Vision (CV) community witnessed a groundbreaking development. The winning solution of the ImageNet Large Scale Visual Recognition Challenge (ILSVRC), spearheaded by ~\cite{krizhevsky2012imagenet}, showcased the prowess of deep neural networks in image classification tasks. For the first time, a deep learning model outperformed traditional handcrafted methods by automatically learning rich and discriminative features directly from raw pixel data. This achievement marked a paradigm shift in computer vision, setting the stage for subsequent advancements in deep learning-based image analysis.
The success of deep learning methods extended beyond image classification, encompassing a broad spectrum of visual tasks, including object detection and segmentation. These techniques, empowered by the sheer complexity and expressiveness of deep neural networks, surpassed conventional approaches, exceeding expectations and inspiring further innovation.

Despite their remarkable achievements, deep learning models often operate as black boxes, with their decision-making processes obscured by their immense complexity. Moreover, they are susceptible to errors and can exhibit undesirable behaviors, such as learning shortcuts~\cite{geirhos2020shortcut} to achieve high accuracy on specific tasks. Recognizing these challenges, the need for eXplainable AI (XAI) methodologies has emerged~\cite{doshivelez2017rigorous}, aiming to elucidate the inner workings of deep learning models and enhance their transparency and trustworthiness.

This chapter aims to provide a succinct introduction and essential background in both deep learning and explainability, which will be useful for understanding the remainder of the manuscript. It is not intended as a comprehensive review of the state of the art but will offer key insights for grasping the thesis structure. The chapter is organized as follows: \autoref{sec:intro:deep_learning} revisits statistical learning fundamentals, deep learning, and its application in computer vision, while \autoref{sec:intro:xai_landscape} presents an overview of the explainability approaches in AI. We will conclude with \autoref{sec:intro:contrib}, where we describe the structure of the manuscript and the contributions it builds upon.

\newcommand{\risk}{\s{R}}
\newcommand{\emprisk}{{\risk_{\text{emp}}}}

\section{Deep Learning Background}
\label{sec:intro:deep_learning}

In this section, we will briefly revisit the framework of this study, namely deep neural networks. To do so, we will briefly review the statistical learning framework in which we operate, and then we will explore the different components of neural network architectures for vision tasks. For a more comprehensive background, we encourage the reader to refer to \cite{goodfellow2016deep}.

\subsection{Statistical Learning}

Deep learning methodologies are firmly rooted in the principles of statistical learning theory~\cite{vapnik1999overview}. Consider measurable spaces $\sx \subset \Real^d$ and $\sy \subset \Real$, representing the input and output spaces respectively\footnote{All topological spaces are equipped with their Borel $\sigma$-algebra}. In supervised learning, we are presented with a dataset of labeled instances:
\[
\s{D} = \{ (\vx_1, y_1), \ldots, (\vx_n, y_n) \} \in (\sx \times \sy)^n.
\]
Our objective in this context is to ascertain the best approximation $\f$, the stochastic relationship between input $\vx \in \sx$ and its corresponding label $y \in \sy$, which is formalized as the conditional probability measure $\E(y | \vx)$ under the probability measure $\P_{\rvx,\ry}$ on $(\sx \times \sy)$.

Achieving the best approximation entails specifying a hypothesis space $\fspace$ comprising potential predictors and defining an appropriate loss function $\ell : \sy \times \sy \to \Real$ that quantifies the discrepancy between a predictor and the true label. The concept of a loss function originated in statistical decision theory, pioneered by~\cite{wald1949statistical}, with roots tracing back to Laplace’s theory of errors.

For instance, in a dataset containing images of dogs and cats, $\sx$ denotes the space of images and $\sy$ represents the labels $\{+1, -1\}$, where $+1$ denotes the presence of a dog and $-1$ denotes the presence of a cat. A conceivable loss function could measure the Euclidean distance between the predicted label and the ground truth.

In essence, the learning problem can be formulated as:
\[
\risk(\f) \defas \underset{(\rvx, \ry) \sim \P_{\rvx,\ry}}{\E}\ell(\f(\rvx), \ry)
~~~~ \text{and} ~~~~
\f^\star = \argmin_{\f \in \fspace} \risk(\f).
\]

With a $\fspace$ the set of all possible functions from $\sx$ to $\sy$. In practical scenarios, access to the true distribution $\P_{\rvx,\ry}$ is typically unavailable\footnote{As described in the Notations section, we use $(\rvx, \ry)$ to denote random variables representing theoretical constructs of inputs and outputs from the probability distribution, and $(\vx, y)$ to denote specific samples or instances from our dataset. This notation clarifies the distinction between theoretical models and empirical data in our analysis.}. Hence, we resort to approximating the learning problem using the available training set $\s{D}$ to minimize the so-called \textit{empirical risk} $\emprisk(\f)$:

\begin{definition}[Empirical risk.]
For a training dataset $\s{D} = \{(\vx_1, y_1), \ldots,  (\vx_n, y_n) \}$ and a function $\f : \sx \to \sy$, the empirical risk with respect to the loss $\ell$ is defined as:
\[
\emprisk(\f) \defas \frac{1}{n}\sum_{i=1}^n \ell(\f(\vx_i), y_i).
\]
\end{definition}

The objective is to minimize the average loss over the training data. This fundamental learning approach is known as \textit{empirical risk minimization}(ERM):

\begin{definition}[ERM learning algorithm]
Given a hypothesis set $\fspace$, the ERM selects $\f^\star$, which minimizes the empirical risk within $\fspace$:
\[
\f^\star = \argmin_{\f \in \fspace} \emprisk(\f)
\]
\end{definition}

However, we face two challenges here. Firstly, the problem is not always convex (unless employing a simple model like linear regression), rendering empirical risk minimization computationally intractable in practice. Secondly, this approach does not guarantee minimization of errors on unseen data points, leading to overfitting issues that hinder generalization and necessitate regularization.

\paragraph{Regularization}

Minimizing $\emprisk(\cdot)$ alone does not suffice for achieving robust generalization. A common strategy involves augmenting the objective with a regularization term $\Omega(\f)$:
\[
\f^\star = \argmin_{\f \in \fspace} \emprisk(\f) + \Omega(\f).
\]

Here, $\Omega(\f)$ regulates the complexity of the function. Optimization of the adjusted loss function mitigates overfitting and facilitates better generalization.

\paragraph{Stochastic Gradient Descent (SGD)}

In the pursuit of optimizing the empirical risk, our functions within the hypothesis space $\fspace$ are usually parameterized by a set of parameters $\parameters \in \Theta$. The goal of learning in this context becomes the optimization of these parameters to minimize the loss function, effectively finding the best approximation $\f^\star$ that represents our model. A cornerstone for this optimization of empirical risk, especially within the realm of deep learning, is \textit{Stochastic Gradient Descent} (SGD). This method stands in contrast to the classical Gradient Descent approach, which necessitates computing the gradient of the loss function $\ell$ across the entire dataset to execute a single parameter update. Instead, SGD opts for a more dynamic and computationally efficient route by iteratively adjusting the model parameters utilizing a randomly selected subset of the data at each iteration. This strategy markedly diminishes computational demands, thereby enabling the training of sophisticated models on voluminous datasets.

\begin{definition}[SGD]
Given a loss function $\ell$, a learning rate $\eta$, a training dataset $\s{D}$ and a mini-batch $\s{B} = \{ \vx_i, y_i \}_{i=0}^{|\s{B}|} \subset \s{D}$, SGD iteratively updates the model's parameters $\parameters$ by calculating the gradient of $\ell$ with respect to $\parameters$ on $\s{B}$:
\[
\parameters_{t+1} = \parameters_t - \eta \sum_i^{|\s{B}|} \nabla_{\parameters}  \ell(\f(\vx_i; \parameters_t), y_i),
\]
where $\parameters_{t}$ represents the successive parameters, $\f(\cdot; \parameters)$ ny prediction function parametrized by $\parameters$ and $t$ the current iteration step.
\end{definition}

This process of iterative parameter adjustment via SGD is a direct application of the empirical risk minimization principle, adapted for the practical challenges of training deep neural networks. It allows for efficient computation and robust search through the parameter space, even in the face of complex models and large datasets.

The element of randomness in SGD, by way of selecting data points, injects a beneficial noise into the optimization trajectory. This aspect can aid in circumventing local minima, thus potentially steering the optimization towards more optimal solutions in the complex, non-convex problem spaces typical of deep neural networks. Moreover, the ability of SGD to operate efficiently with mini-batches underscores its indispensability for deep learning models. This is particularly relevant in scenarios where the sheer scale of the dataset and the model's complexity render full-batch processing impractical.

For more detail on statistical learning theory, we refer the reader to \cite{hastie2009elements}. Having discuss the learning framework in which we operate, we proceed to introduce the focal point of this work: neural networks.

\subsection{Neural Networks}

In this section, we revisit the core components of deep learning: neural networks, emphasizing their parameterization and the pivotal role of convolution operations, particularly for image data. Neural networks, parameterized by weights and biases collectively denoted as $\parameters$, are foundational to deep learning's success in various domains.

\begin{definition}[Neuron]
A neuron is a function $\neuron: \Real^d \to \Real$, parameterized by $\parameters = \{\v{w}, b\}$, and defined as:
\[
\neuron(\vx; \parameters) \defas \sigma(\v{w}^\tr \vx + b),
\]
where $\v{w} \in \Real^d$ is the weight vector, $b \in \Real$ is the bias, and $\sigma: \Real \to \Real$ is a non-linear activation function.
\end{definition}

Neurons aggregate input signals linearly weighted by $\v{w}$, add a bias $b$, and apply a non-linear function $\sigma$ to produce an output. This process enables the model to learn complex relationships between inputs and outputs.

A neural network combines multiple neurons in layers, and multiple layers can be stacked to form a deep neural network. Let's denote $\parameters$ as the collection of all parameters across the network. Then, a fully connected feedforward neural network (FCNN) can be defined as follows:

\begin{definition}[Fully Connected Feedforward Neural Network (FCNN)]
A FCNN $\f(\vx; \parameters)$ of $L$ layers is defined as a composition of layers of neurons:
\[
\f(\vx; \parameters) \defas (\layer^{(L)} \circ \ldots \circ \layer^{(1)})(\vx),
\]
where $\layer^{(i)}(\v{a}; \parameters^{(i)}) = \sigma(\m{W}_{(i)}^\tr \v{a} + \v{b}_{(i)})$ denotes the $i$-th layer function, with $\parameters^{(i)} = \{\m{W}_{(i)}, \v{b}_{(i)}\}$ being the parameters of the $i$-th layer, and $\v{a}$ the activations from the previous layer.
\end{definition}

At the core, the ensemble of neural networks we will study in this work is characterized by this structured aggregation of distinct layers, or "blocks," each serving a unique computational purpose. For the rest of this work, we will refer to this architecture interchangeably as FCNN or MLP.
Among these, certain blocks hold particular relevance to our investigation. Consequently, we will dedicate the concluding segment of this section to a description of these components. Specifically, our focus will encompass convolutional blocks, residual connections, batch normalization and finally, we will delve into the attention block.

\paragraph{Convolution layer.} Those layers are particularly adept at handling grid-like data, such as images, through the use of convolution operations. Convolution leverages the spatial structure of data, allowing the network to learn filters that capture local patterns, and stacking convolution able the model to build more global features such as shape.

\begin{definition}[Convolution Operation]
The convolution of an input $\vx$ with a filter $\v{w}$, parameterized by $\parameters = \{\v{w}, b\}$, for a single channel, is defined as:
\[
(\vx \bigotimes \v{w})_{i,j} \defas \sum_{m}\sum_{n} \vx_{m,n} \cdot \v{w}_{i-m, j-n} + b,
\]
where $\bigotimes$ denotes the convolution operation. For multichannel inputs, this operation is performed independently for each channel and summed to produce a single output.
\end{definition}

Alternatively, the convolution operation can be understood in the frequency domain through the Fourier Transform~\cite{chi2020fast}, which translates the convolution into a point-wise product in the frequency space. Following a convolution operation in a CNN, the output is typically passed through a non-linear activation function, (e.g., ReLU), to introduce non-linearity into the model. The result of applying a convolution followed by an activation function is known as an activation or \textit{feature map}. Each feature map has a dimensionality of $W \times H \times C$, where $W$ and $H$ are the width and height of the map, respectively, and $C$ refers to the number of channels. These dimensions correspond to the spatial dimensions of the image being processed and the depth of the feature map, which represents the number of filters applied during the convolution. 

Modern neural networks often cascade multiple convolution layers, alternating them with pooling layers and activation functions. Pooling layers reduce the spatial dimensions ($W$ and $H$) of the feature maps, helping to decrease the computational load and increase the \textit{receptive field} of the features. The combination of convolution, activation, and pooling layers allows the network to learn hierarchical representations of the input data, where higher-level features are composed of lower-level ones.

\paragraph{Residual Connections.}
The introduction of residual connections marked a significant advancement in deep learning architectures. Residual connections was introduced in 
\cite{he2016deep} to address the \textit{vanishing gradient} problem~\cite{hochreiter1998vanishing} that arises in very deep networks by allowing gradients to flow through a shortcut path. It consists in re-applying activations of previous layer into the next layer: 

\begin{definition}[Residual Connection]
A residual connection in a neural network allows the input of a layer to be added to its output, facilitating the learning of an identity function. This is defined as:
\[
\f(\vx; \parameters) \defas \vx + \layer(\vx; \parameters),
\]
where $\f$ represents the function implemented by the layer with residual connection, $\layer$ is the layer's original transformation function, and $\vx$ is the input to the layer. The parameters $\parameters$ denote the weights and biases of $\layer$.
\end{definition}

It turns out that allowing information to bypass one or more layers facilitate the backpropagation, thus ensuring that deeper networks can still learn effectively. This innovation has been fundamental in the development of state-of-the-art architectures.

\paragraph{Batch Normalization.} Still in the purpose of enhancing the training stability of deep neural networks, Batch Normalization~\cite{ioffe2015batch} emerges as a crucial innovation. This technique propose to adjust the internal covariate shift -- the distribution of each layer's inputs changes during training, as the parameters of the previous layers change. To do so, Batch Normalization standardizes the inputs to a layer for each mini-batch, thus stabilizing the learning process and allowing for higher learning rates and quicker convergence.

\begin{definition}[Batch Normalization]
Given a mini-batch of inputs $\s{B} = \{\vx_1, \ldots, \vx_n \}$, Batch Normalization normalizes the input of each feature to have zero mean and unit variance. Additionally, it introduces two trainable parameters, $\v{\gamma}$ and $\v{\beta}$, to scale and shift the normalized value. Mathematically, for an input feature $\vx$, the Batch Normalization transform is defined as:
\[
\text{BN}_{\v{\gamma}, \v{\beta}}(\vx) \defas \v{\gamma} \left(\frac{\vx - \v{\mu}_{\s{B}}}{\sqrt{\v{\sigma}_{\s{B}}^2 + \varepsilon}}\right) + \v{\beta},
\]
where $\v{\mu}_{\s{B}} = \frac{1}{n} \sum_i^n \v{x}_i$ is the empirical mean over the mini-batch $\s{B}$ ($\v{\sigma}_{\s{B}}^2$ the empirical variance), and $\varepsilon$ is a small constant\footnote{This constant may have a real impact on the training (see ~\cite{nado2020evaluating}) and are not usually well defined: $1e^{-3}$ for Tensorflow~\cite{tensorflow2015} and $1e^{-5}$ for Pytorch~\cite{paszke2019pytorch}.} added for numerical stability. The parameters $\v{\gamma}$ and $\v{\beta}$ are learned along with the original model parameters, allowing the network to undo the normalization if it is found to be counter-productive for the learning of certain layers.
\end{definition}

The $(\v{\gamma}, \v{\beta})$ parameters are usually of size $p$ with $p$ the number of features, which means that $\text{BN}_{\v{\gamma}, \v{\beta}}(\cdot)$ control the mean and variance on each channels for convolution neural net, or neurons for a MLP.

As previously stated, batch Normalization not only accelerates the training process by reducing the number of epochs required to train deep networks but also mitigates the problem of gradient vanishing/exploding, making it easier to train deep networks with saturating non-linearities. It has since become a standard component in the architecture of modern neural networks. However, lately new kind of normalization have emerged such as \textit{LayerNorm}~\cite{ba2016layer}.

\paragraph{Attention Mechanisms in Vision.} Attention mechanisms were introduced in~\cite{vaswani2017attention} and have had a profound impact across the deep learning community. Originating in NLP with the advent of Large Language Models (LLMs), these mechanisms have also significantly influenced the field of computer vision with the ViT architecture~\cite{dosovitskiy2020image,zhai2022scaling,steiner2021train}. The Attention, as initially described by~\cite{vaswani2017attention}, involves dynamically computing weights --- attention weights -- among multiple \textit{tokens} (e.g., words in a sentence, patches in an image) to facilitate their "mixing" to create a feature. This interaction among all input variables is often not possible with a single convolution (e.g., when we use filters smaller than the image size, the top-left pixel does not interact with the bottom-right pixel). From this perspective, convolution imposes an inductive bias of local interactions, whereas attention mechanisms enable all sorts of interactions, even between distant image patches. Formally, an image $\vx$ is divided into patches called tokens $\{\v{t}_1, \ldots, \v{t}_n\}, \v{t}_i \in \Real^{p}$ of dimension $p$. Each of these tokens is then processed through multiple MLPs to reduce their dimensions to $p' << p$, producing three matrices $\m{Q}, \m{K}, \m{V}$, termed key, query, and value, upon which the attention operation is then applied:

\begin{definition}
    
Given an input image $\vx$, segmented into a sequence of tokens $\m{T} = \{ \v{t}_1, \ldots, \v{t}_n \}$, where each $\v{t}_i \in \Real^p$ represents a patch of the image. These tokens are then processed through three separate MLPs, each one differently parametrized. The attention mechanism is then applied. Formally:

\begin{align*}
&\m{Q} \defas \text{MLP}_Q(\m{T}; \parameters_Q) ~~~
\m{K} \defas \text{MLP}_K(\m{T}; \parameters_K) ~~~
\m{V} \defas \text{MLP}_V(\m{T}; \parameters_V) \\
&\text{Attention}(\m{Q}, \m{K}, \m{V}) \defas \text{softmax}\left(\frac{\m{Q}\m{K}^\top}{\sqrt{p'}}\right)\m{V},
\end{align*}

where $\m{Q}$, $\m{K}$, and $\m{V}$ all lie in $\Real^{n \times p'}$. The softmax operation is applied to the rows of the resulting matrix, allowing the model to dynamically allocate attention across different regions of the input based on the relevance of each token to another token.

\end{definition}

This mechanism is especially advantageous in vision for its capacity to adaptively enhance the receptive field, enabling extensive interactions (as necessary for recognizing shapes, for instance). Leveraging attention allows models to process large volumes of visual data efficiently, focusing computational resources on the most informative parts of an image. This approach has led to the development of Transformer models, such as the Vision Transformer (ViT)~\cite{dosovitskiy2020image}, which is now considered as state-of-the-art across a wide array of computer vision tasks.

However, the attention mechanism's computational efficiency is hampered by the quadratic growth of the matrix-matrix multiplication $\m{Q}\m{K}^\top$ cost in relation to the number of tokens. This issue limits its scalability, particularly for high-resolution images or large datasets. In response, subsequent research has focused on devising strategies to mitigate this computational burden. Alternative approaches, such as sparse attention patterns, low-rank approximations, and locality-sensitive hashing, have been proposed to reduce the complexity from quadratic to sub-quadratic or even linear, with respect to the number of tokens. For a more in-depth discussion on these solutions, readers are encouraged to refer to \cite{zhang2023cab}.

\paragraph{Closing Note.} Recognizing the critical role of parameters ($\parameters$) in the various blocks we've discussed is essential. Deep neural networks are incredibly effective across numerous domains, largely due to their extensive parameterization -— for instance, ResNet50 with 25 million parameters and ViT-H boasting 632 million. This complex parametrization does not only boost their performance but also obscures their decision-making processes, making them \textit{black boxes}. This opacity underscores the necessity for Explainable AI (XAI). In the following section, we'll delve into the motivations behind XAI and explore how it can reveal the inner workings of these complex models, making their operations more transparent and understandable.

\section{Explainability Landscape}
\label{sec:intro:xai_landscape}

The objective of this section is twofold: firstly, to underscore the imperative of explainability within machine learning, and secondly, to delineate a concise overview of the diverse methodologies underpinning explainability -- to say it otherwise, to ``flag'' the existing sub-fields. We aim to acquaint the reader with pivotal terms and explainability methods discussed throughout this manuscript. To do so, we propose a taxonomy categorizing explainability methods into three dimensions: methods that explain individual predictions, those studying the model internal mechanics, and those interpreting the data's influence. This classification, albeit simplistic, facilitates a structured introduction to the landscape of explainability.

\begin{figure}[ht]
    \centering
    \includegraphics[width=0.8\textwidth]{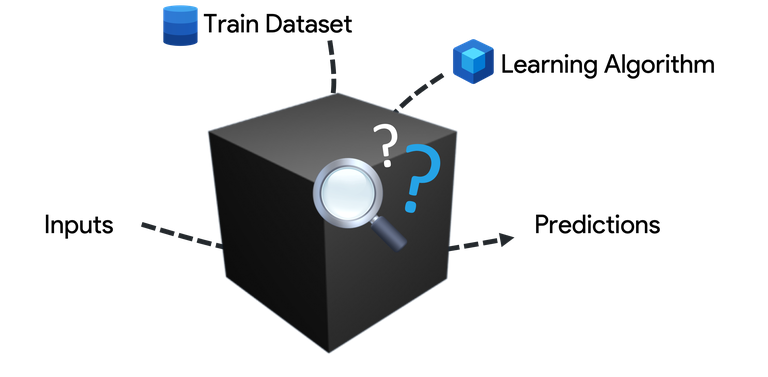}
    \caption{\textbf{Illustration of the Black-Box Problem.} Neural networks undergo training on a \textit{Training Dataset} through a specific \textit{Learning Algorithm}. After training, the model performs inferences using the learned parameters to make \textit{Predictions}. However, the multitude of operations from \textit{Input} to prediction is excessively complex for human comprehension, thus the name Black-box.}
    \label{fig:intro:blackbox}
\end{figure}

\subsection{Motivation}
\label{sec:intro:motivation}

Tracing the origins of explainability in AI is akin to exploring the very essence of science, as the pursuit of explanations, particularly within the realm of scientific thought, has historically been a foundational pillar, as highlighted by~\cite{hospers1946explanation}, suggesting that the impetus for explanation is deeply rooted in the fabric of scientific discourse itself. 
A closer intellectual lineage to modern explainability could be found back over half a century, finding ground in the domain of mathematical logic~\cite{hempel1948studies}.
However, it was not until the advent of deep learning, that the modern conceptualization of explainability -- as it is addressed within this manuscript -- emerged. Unsurprisingly, it is deep learning that has been the catalyst for the establishment of this burgeoning research field, and it is important to delineate the goal of XAI as well as its expected outcomes prior to examining the existing body of work.

It would typically be prudent to start with a definition; however, the quest for a formal definition of explainability is unlikely to be straightforward. \cite{lombrozo2006structure} notes that explanations serve as the currency of our belief systems, a medium through which we exchange and interrogate our understanding of the world. This discourse raises fundamental questions about the nature of explanations and the criteria that distinguish more effective explanations from their less compelling counterparts. The academic community has variously characterized explanations as embodying a \textit{deductive-nomological} essence \cite{hempel1948studies}, akin to logical proofs, or as mechanisms that provide a deeper understanding of underlying processes, as proposed by \cite{bechtel2005explanation}. \cite{keil2006explanation} proposed a broader conceptualization, advocating for an understanding of explanations as embodying an implicit explanatory comprehension\footnote{Interestingly, one could interpret the essence of this article from an informational perspective on explainability as an addition of information \textit{given} a common body of knowledge.}.

Given the rich literature on this subject, attempting to distill a singular definition that encompasses the entire spectrum of use-cases and motivations within the field would be a \textit{Sisyphean} task and would take us too far. Therefore, we propose to adopts a pragmatic approach to defining explainable AI (XAI), not through abstract or absolute terms but by aligning with the specific objectives it seeks to achieve. This approach will thus have to settle for a localized and use-case specific definition of explainability, allowing us to focus on the technical aspects of the domain. Among the myriad objectives identified in the literature~\cite{jacovi2021formalizing,miller2017explanation,survey2019metrics,saeed2023explainable,weber2023beyond,antoniadi2021current,das2020opportunities}
, six primary goals could be noted, as central to the discourse on XAI:

\begin{itemize}
    \item \textbf{Building trust in model predictions.} For example, in healthcare, AI-assisted diagnostics can use explainability to highlight influential areas in medical images, helping clinicians trust and verify AI diagnoses by visually indicating regions of interest.
    
    \item \textbf{Elucidating important aspects of learned models.} The SNCF for example, could need explainability in autonomous railway systems to help engineers understand the decision-making process behind navigational actions, ensuring the AI correctly recognizes stop signs and detect obstacles.
    
    \item \textbf{Assisting in meeting regulatory requirements and facilitating the certification process.} Financial services leveraging AI for credit scoring can use explainability to detail how individual features influence credit scores, aiding in compliance with regulations like GDPR.
    
    \item \textbf{Uncovering and addressing biases or unintended effects learned by models.} Explainability can reveal if an AI recruitment tool unfairly weighs certain demographics, allowing developers to correct these biases.
    
    \item \textbf{Detecting and preempting potential failure cases.} In predictive maintenance for manufacturing, explainability reveals conditions leading to equipment failures, enabling preemptive actions to prevent or mitigate effects.
    
    \item \textbf{Debugging models to enhance training methodologies:} Still in the context of the SNCF's autonomous railway systems, explainability can help engineers and developers understand why a model might misinterpret sensor data or fail to correctly predict maintenance needs. By analyzing instances where the model's performance deviates from expectations, the teams can refine data inputs, adjust model parameters, and ultimately improve the reliability and safety of autonomous railway operations.
    
\end{itemize}

These objectives highlight the heterogeneity of aims within the field and underscore the magnitude of the challenges that confront us. Having established that XAI presents a real \textbf{conceptual challenge}, we will now see that it is also a real \textbf{technical challenge}.

\subsection{Explaining Predictions}

\begin{figure}[ht]
    \centering
    \includegraphics[width=1.\textwidth]{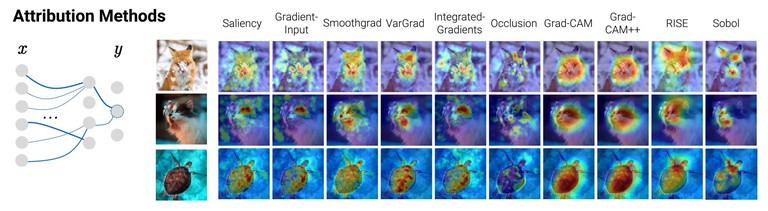}
    \caption{\textbf{Attribution Methods.} Attribution methods will be the subject of the \autoref{chap:attributions}. These methods aim to explain a specific prediction through heatmaps, where hotter areas indicate a greater significance of the pixel for the decision.}
    \label{fig:intro:attributions}
\end{figure}

The development of methods to explain model predictions has been a critical aspect of research, originating with the introduction of attribution methods~\cite{Zeiler2011}. These approaches aim to clarify the rationale behind a model's decision, whether it is the classification of an instance, the detection of an object within an image, or the prediction of a value in regression tasks. \textit{Attribution methods}, which produce a heatmap to represent the importance of each input variable (see \autoref{fig:intro:attributions}), are among the most widely used due to their straightforward implementation in automatic differentiation frameworks such as TensorFlow~\cite{tensorflow2015} and PyTorch~\cite{paszke2019pytorch}.

A broad range of attribution techniques exists, using gradients~\cite{zeiler2013visualizing,shrikumar2017learning,sundararajan2017axiomatic,smilkov2017smoothgrad}, perturbations~\cite{ancona2017better,petsiuk2018rise,Fong_2017,fel2021sobol,novello2022making}, or internal model activations~\cite{Selvaraju_2019,chattopadhay2018grad} to generate explanations. A general definition is given below:

\begin{definition}[Attribution Method.] 
\label{def:attributions}
For a model $\f : \sx \to \sy$ and an input $\vx \in \sx$, an attribution method is a functional:

\[
\explainer : \fspace \times \sx \to \Real^{|\sx|}
\]

where $\explanation = \explainer(\f, \vx)$ (with $\f \in \fspace$) represents an attribution map that explains the prediction of $\f$ for input $\vx$. The higher the scalar value in $\explanation$, the more important the variable is considered.
\end{definition}

Despite their utility, attribution methods face challenges related to reliability~\cite{adebayo2018sanity,sixt2020explanations,ghorbani2017interpretation,slack2021counterfactual,sturmfels2020visualizing,hsieh2020evaluations,hase2021out}, computational efficiency~\cite{novello2022making}, and the implicit assumptions about importance~\cite{eva2}. A dedicated chapter (\autoref{chap:attributions}) further explores these methods, addressing their complexities and constraints.

\subsection{Explaining the Model}

Explaining a model involves uncovering the internal mechanics that drive its predictions. This can be approached through various methodologies, each aiming to make the model's operations or internal states more transparent.

\begin{figure}[ht]
    \centering
    \includegraphics[width=1.\textwidth]{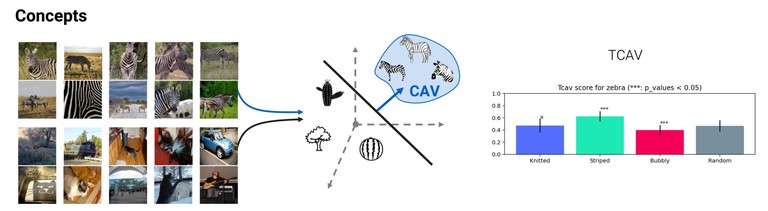}
    \caption{\textbf{Concept Activation Vector (CAV).} An example of extracting the "striped" concept using images featuring this concept and random images. A classifier in the intermediate space is utilized to identify the CAV as the vector orthogonal to the decision boundary. Methods for analyzing concepts will be discussed in \autoref{chap:concepts}.}
    \label{fig:intro:concepts}
\end{figure}

\paragraph{Concept-based Explainability.} Recent developments in explainability have underscored the need to go beyond attribution methods~\cite{doshivelez2017rigorous}. A flagship of these methods is \textit{concept-based} explainability~\cite{kim2018interpretability}, which involves identifying human-understandable concepts within a model. Briefly, this method compares two datasets, one containing the concept of interest and a 'random' dataset used for one-class detection with a linear model. The orthogonal to the decision boundary is called a concept vector, see \autoref{fig:intro:concepts}. Further methods have been proposed to not just retrieve human-defined concepts, but to study concepts utilized by the model itself~\cite{ghorbani2019towards,fel2023craft,lrpconcepts,graziani2023concept,zhang2021invertible,fel2023holistic}. Unlike attribution methods that provide a heatmap of input importance, concept-based explainability seeks to discover "what" triggers a feature. A general approach to defining a concept within a model's operational framework is as follows:

\begin{definition}[Concept Vector.]
\label{def:intro:cav}
Given a Fully Connected Feedforward Neural Network (FCNN) $\f : \sx \to \actspace \subseteq \Real^d$, a concept vector $\v{v} \in \Real^d$ is identified as a vector representing a concept in the activation space $\actspace$ of the FCNN. Depending on the context, the alignment or dot product between an activation $\v{a} \in \actspace$ and the concept vector $\v{v}$ indicates the extent to which $\v{a}$ it embodies the concept.
\end{definition}

An entire chapter is dedicated to these methods, offering a more nuanced understanding of what the model has learned.

\begin{figure}[ht]
    \centering
    \includegraphics[width=1.\textwidth]{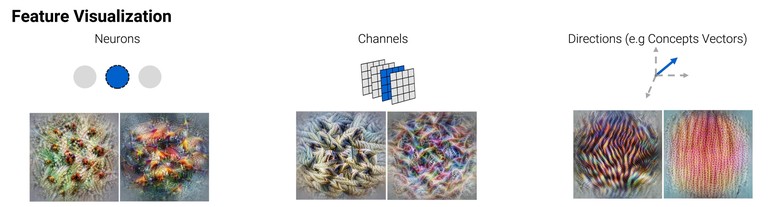}
    \caption{\textbf{Illustration of Feature Visualization (FV).} An example of visualization for neurons (ladybug and goldfish), channels of a convolutional network as well as for CAV using FV. Feature Visualizations will be discussed in \autoref{chap:concepts}.}
    \label{fig:intro:fviz}
\end{figure}

\paragraph{Feature Visualization.}
Feature visualization~\cite{olah2017feature} aims to generate images that maximally activate specific parts of the network, providing insights into the kinds of features to which the network is sensitive and identifying what each component of the network is looking for in its inputs.

\begin{definition}[Feature Visualization]
\label{def:intro:feature_viz}
Given a FCNN $\f : \sx \to \Real^d$ and a target structure (neurons, channels, concept) $\v{v} \in \Real^d$, the feature visualization $\vx^\star$ associated with $\v{v}$ is defined as:
\[
\vx^\star = \argmax_{\vx \in \sx} \langle \f(\vx), \v{v} \rangle - \Omega(\vx),
\]
where $\Omega(\vx)$ typically applies a penalty to ensure the resulting image remains within a natural image manifold, and $\v{v}$ can be a one-hot vector targeting a specific neuron or a more complex structure representing a concept.
\end{definition}

Feature visualization techniques have been instrumental in uncovering fascinating phenomena and features in convolutional models~\cite{nguyen2016multifaceted,nguyen2019understanding} and especially~\cite{cammarata2020thread}. \autoref{chap:concepts} has a section dedicated to this subject and proposes improvements to existing feature visualization methods.

\paragraph{Interpretability by Design.} Creating models with interpretability as a foundational goal entails architecting models to output not only predictions but also an explanation understandable to humans. 

Recently, a promising methodology~\cite{bohle2022b,bohle2023holistically} has been proposed, it involves training models that dynamically adjust their internal parameters in response to input data, reminiscent of synaptic plasticity~\cite{abraham1996metaplasticity}. Specifically, for a given input point $\vx$, these models generate a unique set of linear weights that directly map $\vx$ to its prediction $y = \f(\vx) \vx^\tr$ with $\f : \sx \to \Real^{|\sx|}$. This innovative approach promises not only enhanced interpretability, but also a direct mechanism for solving (at least locally) the problem of prediction specific explanations.

Historically, other methodologies have aimed at achieving interpretability by adhering to specific desiderata during model construction. For instance, the seminal work by~\cite{alvarezmelis2018robust} focused on developing models that are both robust and interpretable by design, ensuring that the model's behavior remains consistent and faithful to the data it was trained on, thereby enhancing trustworthiness and reliability.

Furthermore, the concept of prototypical networks, as discussed by~\cite{rudin2019stop}, introduces a framework where predictions are based on the similarity of input features to prototype examples. This methodology not only simplifies the interpretability of predictions by anchoring them to recognizable instances but also facilitates a more intuitive understanding by comparing new inputs to known, labeled examples.

In summary, focusing on making models interpretable has the literature to propose promising methods. This approach is a serious candidate to make deep neural networks work in a manner that humans can understand. We will  dive deeper into these methods in the chapter dedicated to Alignment (\autoref{chap:alignment}).

\subsection{Explaining through Data}

Understanding model behavior extends to examining the influence of training data on the model's learning and predictions. Influence functions~\cite{cook1980characterizations} are a key tool in this domain, enabling the estimation of how the model's parameters or predictions would change if a particular data point were \textit{removed}\footnote{The actual formulation is expressing the difference in the parameter space for an infinitesimal perturbation.} from the training set.

\begin{definition}[Influence Function]
Given a learning algorithm $\mathcal{A} : \sx^n \times \Real^n \to \bm{\Theta}$, where $\parameters \in \bm{\Theta}$ represents the parameter space of a predictor $\f(\cdot; \parameters)$. A dataset $\s{D} = \{(\vx_1, y_1), \ldots, (\vx_n, y_n)\} \subseteq \mathcal{X}^n$, and a vector of weights $\v{w} = (w_1, \ldots, w_n)$ for the data points in $\mathcal{D}$, the influence function approximates the effect on the parameter vector $\parameters$ when the weight of the data point indexed by $i$ is infinitesimally perturbed by an amount $\xi$. This is mathematically expressed as:
\[
\mathcal{I}(i ; \mathcal{A}, \s{D}, \v{w}) \defas \lim_{\xi \to 0} \frac{1}{\xi} \left( \mathcal{A}(\s{D}, \v{w} + \xi \mathbf{e}_i) - \mathcal{A}(\s{D}, \v{w}) \right),
\]
where $\mathbf{e}_i$ is the canonical vector with respect to the $i$-th data point's weight.
\end{definition}

Influence functions trace their roots to robust statistics, offering a lens to examine the sensitivity of parameter estimates to changes in the underlying data distribution. This concept has been instrumental in identifying leverage points and outliers in data analysis, where the influence of such points on statistical estimations can lead to biased or misleading conclusions.

Recent works~\cite{koh2017understanding} have significantly extended and refined the application of influence functions, offering more possibility to shape better insights into the data's role in shaping model behavior.

\section{Application: FRSign Dataset}
\label{sec:intro:frsign}

During this thesis, we propose the exploration and application of our work on the FRSign dataset~\cite{2020frsign}, a recent railway dataset, as a case study to monitor the progression of work and the development of new tools. The FRSign dataset is introduced and presented, as well as models trained on it, and will be referenced in subsequent chapters, specifically in Chapter \autoref{chap:attributions} and Chapter \autoref{chap:concepts}.

\subsection{Introduction to the FRSign Dataset}

The FRSign dataset is a recent open-source dataset, released in~\cite{2020frsign}, with the aim of pushing advancements in autonomous transportation, particularly in the less-explored area of rail systems. Despite the prevalent focus on datasets tailored for autonomous driving applications in recent years, alternative modes of transportation, such as railways, have not received comparable attention. FRSign tries to address this gap by providing a meticulously collected dataset from various locations across France, focusing exclusively on railway infrastructure. This vision-based dataset is primarily geared towards enhancing the detection and recognition capabilities concerning railway traffic lights, and thus adapted as an application for our work on Explainable AI for vision.

The dataset labelisation benefits from detailed, hand-labeled annotations,
encompassing over 100,000 images. Each image is labeled over six distinct types of French railway traffic lights, complete with metadata such as acquisition date, time, sensor parameters, and bounding boxes. This dataset was developed collaboratively by IRT SystemX as a part of the TAS (Safe Autonomous Land Transport) project, in conjunction with industry leaders in railway traffic SNCF.

\subsection{Detailed Dataset Statistics}

\begin{figure}[ht]
    \centering
    \includegraphics[width=0.9\textwidth]{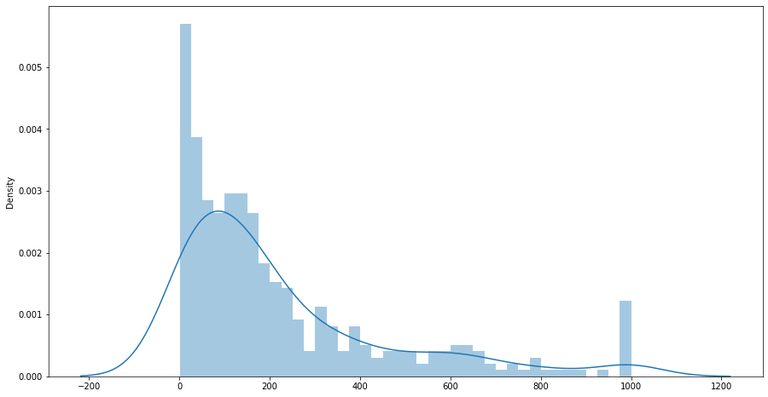}
    \caption{\textbf{FRSign Dataset Statistics.} Distribution of Images per Video Sequence. The FRSign dataset exhibits a general trend of having a modest number of images per sequence, despite the presence of an outlier sequence containing more than 5000 images. Interestingly, there is a notable peak in the distribution, with a significant number of sequences having around 1000 images each.}
    \label{fig:frsign:stats}
\end{figure}

The FRSign dataset is voluminous, with a total size of 310GB, comprising 393 sequences that depict the journey of trains from one station to another. Each sequence, captured in video format, can be decomposed into individual images, leading to an average of 469 images per sequence when sampled at one image per second. The distribution of images per sequence exhibits a high degree of variability, with a standard deviation of 469, following an exponential distribution pattern, see~\autoref{fig:frsign:stats}. This distribution features a predominance of sequences with a relatively small image count (ranging from 1 to 10) to sequences with the highest image count, reaching up to 5200 images. It is noteworthy that a significant proportion of sequences contain approximately 1000 images. 

The sequence composition of the dataset underscores the importance of considering the non-independent and identically distributed (i.i.d.) nature of the images when training machine learning models. To address this, we propose a tailored train/test split strategy. But first, we describe briefly the type of data.

\paragraph{Data Types and Structure}

\begin{figure}[ht]
    \centering
    \includegraphics[width=0.9\textwidth]{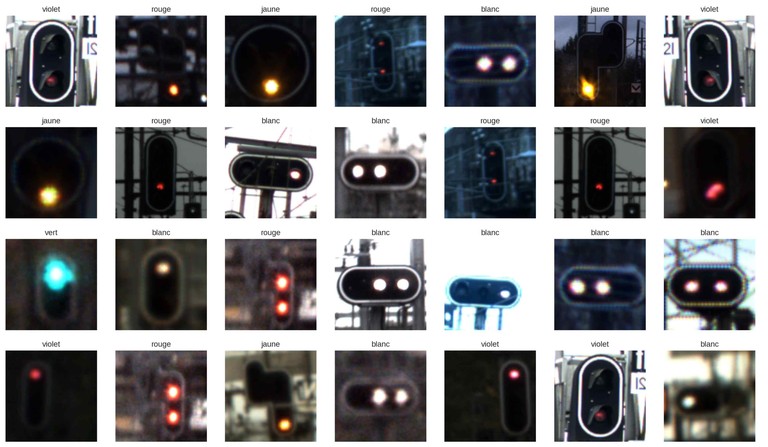}
    \caption{\textbf{FRSign Dataset Samples.} Examples of cropped images derived from the dataset. These images exhibit varying resolutions and collectively represent the six possible classes of railway signals within the dataset.}
    \label{fig:frsign:samples}
\end{figure}

At its core, the FRSign dataset comprises tuples of images and labels, with the images representing cropped segments of railway signaling lights at varying resolutions. These cropped images, derived from original footage with a uniform resolution of $1980\times1080$, vary in size due to the differing dimensions of the region of interest across sequences. This variability introduces a unique challenge in maintaining consistency across the dataset. Each cropped image is associated with one of 6 distinct classes, encompassing various states of railway signals (e.g., red light, yellow-red light). An illustrative sample of these cropped images is provided in \ref{fig:frsign:samples}.

\paragraph{Strategic Train/Test Split}

\begin{figure}[ht]
    \centering
    \includegraphics[width=0.4\textwidth]{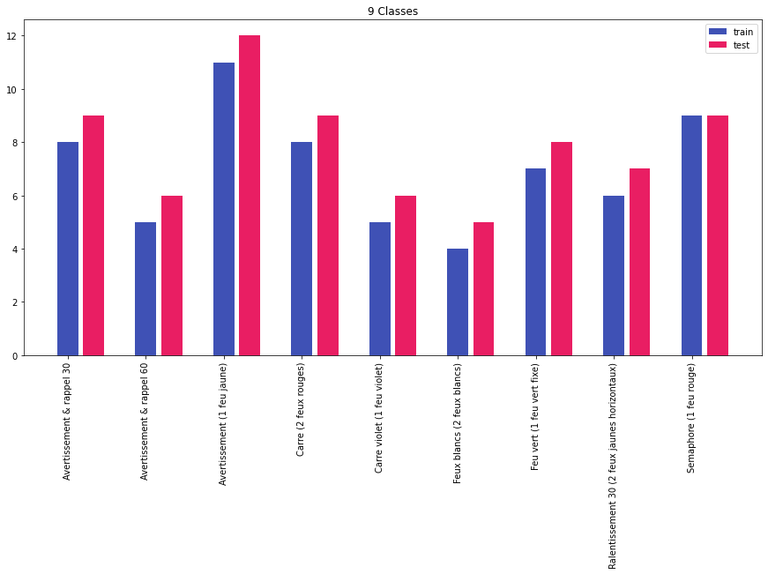}
    \includegraphics[width=0.4\textwidth]{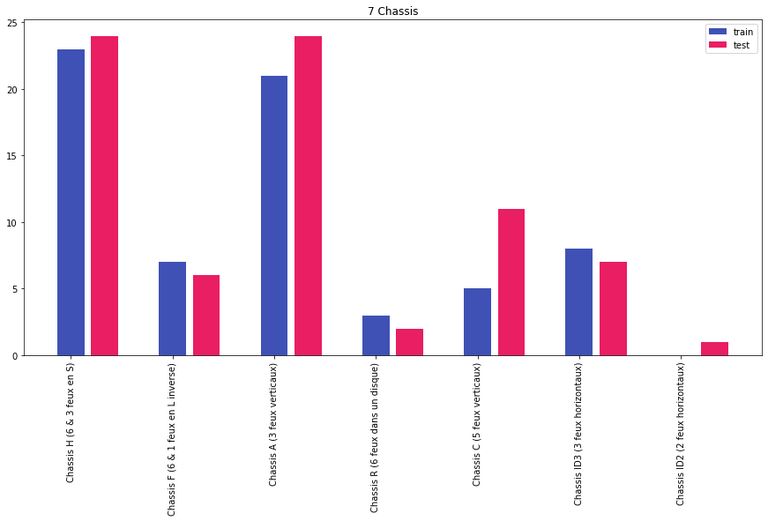}
    \caption{\textbf{FRSign Dataset Splits:} Statistics of the training and testing splits utilized for the dataset. A deliberate effort was made to consider both the classes and the distribution of different types of railway signals (``chassis'') to achieve a balanced representation across splits. Despite these efforts, there remains a degree of class imbalance.}
    \label{fig:frsign:splits}
\end{figure}

Given the sequential nature of the images within the FRSign dataset, employing a traditional random train/test split could potentially introduce significant biases and negatively impact the performance of models trained on this data. Such biases arise because images within a sequence are not independent; rather, they are closely related in both time and appearance, which could lead to overfitting if images from the same sequence are distributed across both training and testing sets.

Consequently, we advocate for a sequence-based split strategy, ensuring that all images from a given sequence are exclusively assigned to either the training or testing set, and not mixed between them. This approach not only preserves the integrity of the dataset's sequential nature but also facilitates a more balanced and effective model evaluation process by minimizing data leakage and ensuring that the model is tested on truly unseen data.

Formally, for a given sequence \(i\) containing \(n\) data points \(\mathcal{S}^{(i)} = \{ (\vx_1^{(i)}, y_1^{(i)}), \ldots, (\vx_n^{(i)}, y_n^{(i)}) \}\), each point in this sequence will be placed in the same split. For the 400 sequences \(\mathcal{S}^{(1)}, \ldots, \mathcal{S}^{(400)}\), we propose a split strategy that not only balances the classes present in each split but also takes into consideration the distribution of different lights models (``chassis''), as seen in \autoref{fig:frsign:splits}. This nuanced approach ensures that the split is not only balanced in terms of the number of images or sequences but also in terms of labels and representativeness of the various spurious cues. 

\subsection{Model Training and Evaluation}

A key focus of this thesis is the comprehensive evaluation of various models trained on the FRSign dataset, with an emphasis on examining different explanatory methods and their impact on the field of autonomous transportation. The models were subjected to a series of data augmentation techniques, including mild geometric transformations and noise addition, to enhance the robustness and generalizability of the trained models. Specific details on the augmentation strategies and their implementation will be provided in the following sections.

We examined three distinct vision models for their performance on the FRSign dataset:

\subsection{Models}

We have trained several models that we will use to showcase our explainability methods and track the progression in this manuscript. The primary objective with these models is not necessarily to achieve the highest performance, but rather to observe and compare the benefits of XAI tools. We will now describe the set of models trained on this dataset. For each of these models, a mild form of data augmentation was applied, consisting of left-right flipping, stochastic noise addition with a probability of $0.5$ from a Gaussian distribution:

\[
\v{\delta}(\r{u}) = 
\begin{cases} 
\mathcal{N}(0, \Id\sigma^2) & \text{if } \r{u} > 0.5, \\
0 & \text{otherwise}.
\end{cases}
\]

for $\r{u} \sim \mathcal{U}([0,1])$ and $\sigma = 0.1$. Additionally, we implemented slight cropping augmentation that crops from $0.8$ to $1.0$ of the original image size and contrast augmentation $\r{c} \sim \s{U}([0.8, 1.2])$. Overall, we focused on three vision models for our experiments, described as follows:

\begin{itemize}
    
    \item \textbf{VGG-16.} The first model trained is a modified version of the classic VGG-16, which accepts input images of size $128\times128$. This variant of VGG-16 includes Batch Normalization added after each convolutional layer and before ReLU activation functions. The architecture head is a global average pooling layer, omitting the original dual dense layers due to their significant memory consumption without a corresponding increase in performance. The model is trained with a dropout rate of $0.7$ and a weight decay of $1e-5$, using AdamW optimizer and cosine annealing scheduling with a warmup over 500 iterations (half an epoch) across a total of 60 epochs. The learning rate varies from a maximum of $1e^{-3}$ to a minimum of $1e^{-5}$. This model achieved an accuracy of 85\% in testing.
    
    \item \textbf{ResNet-50.} The second model is a ResNet50, configured to accept input images of size $224\times224$. The architecture adheres to the original ResNet design, the head being a global average pooling layer followed by a linear layer at the top. This model was trained without dropout but with a weight decay of $1e-5$, using the AdamW optimizer and a cosine annealing schedule with a warmup phase spanning 500 iterations (half an epoch) over a total of 60 epochs. The learning rate ranges from $1e^{-2}$ at its highest to $1e^{-5}$ at its lowest. This configuration led to a testing accuracy of 92\%.

    \item \textbf{ViT-S-32.} The third model trained is a Vision Transformer (ViT-S), designed for input images of size $224\times224$. The architecture consists of 12 blocks, with a width of 384 and 6 attention heads for each attention block, totaling 22M parameters. Additionally, the patch size used is 32. This model incorporates a stochastic depth probability of $0.1$ and a weight decay of $5e-5$, employing the AdamW optimizer and cosine annealing scheduling with a warmup over 500 iterations (half an epoch) across a total of 60 epochs. The learning rate varies from $3e^{-3}$ at its highest to $1e^{-5}$ at its lowest. The ViT model reached a testing accuracy of 90\%.
\end{itemize}

\section{Outline \& Contributions}
\label{sec:intro:contrib}

This doctoral dissertation is organized to further the field of explainability for computer vision. It delves into a variety of specific methodologies across several chapters, outlined as follows:

\paragraph{\autoref{chap:intro}: Introduction.} This initial section provides a concise introduction to deep learning and explainability, establishing the foundational knowledge necessary for the remainder of this document.

\paragraph{\autoref{chap:attributions}: Attribution Methods.} The second chapter is dedicated to attribution methods. It begins by illustrating the feasibility of identifying models that offer superior explanations through the lens of algorithmic stability of its attributions maps. Subsequently, a state-of-the-art black-box attribution method based on \textit{Sobol} indices and Quasi-Monte Carlo sampling is introduced, which reduces computational costs by a factor of two compared to its predecessors. The discussion progresses to the development of an attribution method, \eva, that offers strong formal guarantees using perturbation verification analysis. The practical applicability of these methods, particularly in real-world scenarios and from a human perspective, is subsequently evaluated. It is found that while attribution methods prove to be sufficient and highly useful for straightforward scenarios, their utility vanish when faced with more intricate situations. The chapter concludes by proposing two hypotheses to address these limitations: (1) the need for models that better align with human reasoning, and (2) the necessity to go beyond current attribution methods. These hypotheses are then investigated in \autoref{chap:alignment} and \autoref{chap:concepts}, respectively.

\paragraph{\autoref{chap:alignment}: Model Alignment.} This chapter explores how explainable AI (XAI) can be employed to align current vision models with human cognition through novel training routines. It highlights a trend towards decreasing alignment between models and human understanding and demonstrates how the proposed routine can counter this trend, even improving accuracy. The chapter concludes by noting an intriguing link between model robustness and explainability, exemplified by 1-Lipschitz models.

\paragraph{\autoref{chap:concepts}: Concept based Explainability.} Building on the insights from \autoref{chap:attributions}, this section argues for a shift from explaining solely on \where~a model looks to understanding \what~it sees (\what features the model recognizes at its focal points). A novel method, \craft, is introduced for automatically extracting and evaluating the importance of concepts within trained networks, offering both global and local (heatmap-based) explanations. This approach significantly addresses the issues raised in \autoref{chap:attributions} and opens up new avenues for explainability. The discussion extends to competing methods, proposing a unified framework for automatic concept extraction under the paradigm of \textit{dictionary learning}. It is shown that existing concept importance methods are essentially specific cases of attribution methods applied to concepts. The chapter concludes with \maco~a novel method of feature visualization that scale to deep network for better representation of extracted concepts.

\paragraph{\autoref{chap:conclusion}: Concluding Remarks and Future Directions.} The dissertation concludes with a comprehensive review of the developed methods and tools, reflections on the journey, and thoughts on the future of explainability in AI.

\section{Related publications}

This dissertation integrates and builds upon a series of peer-reviewed publications, open-source projects, and contributions to the wider research community. Below, we categorize these works based on their relevance to the core chapters of this thesis and additional contributions that extend beyond the thesis scope\footnote{The symbol \equal denote equal contributions.}.

\subsection{Foundational Contributions}

This section outlines the peer-reviewed publications that form the backbone of the thesis, organized by the relevant chapters.

\paragraph{Attributions (\autoref{chap:attributions})}

\begin{itemize}
    \item \textbf{Thomas Fel}, David Vigouroux, Remi Cadene, Thomas Serre (2022). \textit{``How Good is your Explanation? Algorithmic Stability Measures to Assess the Quality of Explanations for Deep Neural Networks''.} In: \textit{Proceedings of the Winter Conference on Computer Vision} (\textcolor{confcolor}{WACV})

    \item \textbf{Thomas Fel}\equal, Remi Cadene\equal, Mathieu Chalvidal, Matthieu Cord, David Vigouroux, Thomas Serre, (2021). \textit{``Look at the Variance! Efficient Black-box Explanations with Sobol-based Sensitivity Analysis''.} In: \textit{Advances in Neural Information Processing Systems}  (\textcolor{confcolor}{NeurIPS})

    \item \textbf{Thomas Fel}\equal, Melanie Ducoffe\equal, David Vigouroux\equal, Remi Cadene, Mikael Capelle, Claire Nicodeme, Thomas Serre, (2023). \textit{``Don't Lie to Me! Robust and Efficient Explainability with Verified Perturbation Analysis''.} In: \textit{IEEE/CVF Conference on Computer Vision and Pattern Recognition}  (\textcolor{confcolor}{CVPR})

    \item Julien Colin\equal, \textbf{Thomas Fel}\equal, Remi Cadène, Thomas Serre, (2021). \textit{``What I Cannot Predict, I Do Not Understand: A Human-Centered Evaluation Framework for Explainability Methods''.} In: \textit{Advances in Neural Information Processing Systems} (\textcolor{confcolor}{NeurIPS})

\end{itemize}

\paragraph{Model Alignment (\autoref{chap:alignment})}

\begin{itemize}

    \item \textbf{Thomas Fel}\equal, Ivan F Rodriguez\equal, Drew Linsley\equal, Thomas Serre, (2022). \textit{``Harmonizing the object recognition strategies of deep neural networks with humans''.} In: \textit{Advances in Neural Information Processing Systems}  (\textcolor{confcolor}{NeurIPS})

    \item Mathieu Serrurier, Franck Mamalet, \textbf{Thomas Fel}, Louis Béthune, Thibaut Boissin, (2023). \textit{``On the explainable properties of 1-Lipschitz Neural Networks: An Optimal Transport Perspective''.} In: \textit{Advances in Neural Information Processing Systems} (\textcolor{confcolor}{NeurIPS})

\end{itemize}

\paragraph{Concept-based Explainability (\autoref{chap:concepts})}

\begin{itemize}
    \item \textbf{Thomas Fel}\equal, Agustin Picard\equal, Louis Bethune\equal, Thibaut Boissin\equal, David Vigouroux, Julien Colin, Rémi Cadène, Thomas Serre, (2023). \textit{``CRAFT: Concept Recursive Activation FacTorization for Explainability''.} In: \textit{IEEE/CVF Conference on Computer Vision and Pattern Recognition} (\textcolor{confcolor}{CVPR})

    \item \textbf{Thomas Fel}\equal, Victor Boutin\equal, Mazda Moayeri, Rémi Cadène, Louis Bethune, Mathieu Chalvidal, Thomas Serre, et al., (2023). \textit{``A Holistic Approach to Unifying Automatic Concept Extraction and Concept Importance Estimation''.} In: \textit{Advances in Neural Information Processing Systems}  (\textcolor{confcolor}{NeurIPS})

    \item \textbf{Thomas Fel}\equal, Thibaut Boissin\equal, Victor Boutin\equal, Agustin Picard\equal, Paul Novello\equal, Julien Colin, Drew Linsley, Tom Rousseau, Rémi Cadène, Lore Goetschalckx, et al., (2024). \textit{``Unlocking feature visualization for deep network with MAgnitude constrained optimization''.} In: \textit{Advances in Neural Information Processing Systems}  (\textcolor{confcolor}{NeurIPS})
\end{itemize}

\subsection{Open Source Contributions}
\label{sec:open-source-contributions}

Throughout my PhD, I have actively contributed to the open-source community, leading to the development and maintenance of several projects, notably \xplique~which now count 500+ stars on GitHub and implement more than 50 articles in Explainability and lead to the following publication:

\begin{itemize}
    \item \textbf{Thomas Fel}\equal, Lucas Hervier\equal, David Vigouroux, Antonin Poche, Justin Plakoo, Remi Cadene, Mathieu Chalvidal, Julien Colin, Thibaut Boissin, Louis Bethune, Agustin Picard, Claire Nicodeme, Laurent Gardes, Gregory Flandin, Thomas Serre, (2022). \textit{``Xplique: A Deep Learning Explainability Toolbox''.} In: \textit{Workshop on Explainable Artificial Intelligence for Computer Vision} (\textcolor{confcolor}{CVPR} W.)
\end{itemize}

in total, I open sourced and mainteaned 5 projects, all of them available on GitHub:
\begin{itemize}

    \item \textbf{Xplique:}, an open source Explainability toolbox implementing more than 50 papers of explainability, with a proper documentation, tutorials and notebooks. Available at \url{https://github.com/deel-ai/xplique} or available through \lstinline[language=bash]{pip install xplique}.

    \item \textbf{CRAFT:} an open source repo to reproduce our work on Concept based explainability (see \autoref{chap:concepts}), in Tensorflow and Pytorch, with tutoriels. Available at \url{https://github.com/deel-ai/Craft} or available through \lstinline[language=bash]{pip install craft-xai}.

    \item \textbf{Harmonization:} an open source zoo of harmonized models (see \autoref{chap:alignment}) trained as well as notebook and tutorial to evaluate other models, in Tensorflow and Pytorch. Available at \url{https://github.com/serre-lab/Harmonization} or available through \lstinline[language=bash]{pip install harmonization}.

    \item \textbf{Sobol: } an open source version of Sobol attribution method (see \autoref{chap:attributions}), in Tensorflow and Pytorch. Available at \url{ https://github.com/fel-thomas/Sobol-Attribution-Method}.

    \item \textbf{Numkdoc:} an open source parser of MkDocs for numpy style documentation, now use to build the documentation of \xplique~and other library. Available at \url{ https://github.com/fel-thomas/numkdoc}.

\end{itemize}

Lastly, I have open-sourced a public interactive demo titled \Lens, which encapsulates 3 of the research contributions presented in the final chapter. This demonstration leverages the findings discussed in the last chapter of this manuscript, showcasing the interconnections and collaborative potential among the different studies explored throughout my thesis \url{https://serre-lab.github.io/Lens/}.

\subsection{Extended Contributions}

In addition to my direct thesis work, I have also contributed to several projects that, while not the main focus of this dissertation, address related challenges in the field.

\begin{itemize}
    \item \textbf{Thomas Fel}\equal, Louis Bethune\equal, Andrew Kyle Lampinen, Thomas Serre, Katherine Hermann, (2024). \textit{``Understanding Visual Feature Reliance through the Lens of Complexity''.} In: \textit{Advances in Neural Information Processing Systems} (\textcolor{confcolor}{NeurIPS})

    \item Katherine L. Hermann, Hossein Mobahi, \textbf{Thomas Fel}, Michael C. Mozer, (2024). \textit{``On the Foundations of Shortcut Learning''.} In: \textit{Proceedings of the International Conference on Learning Representations} (\textcolor{confcolor}{ICLR})

    \item Paul Novello, \textbf{Thomas Fel}, David Vigouroux, (2022). \textit{``Making Sense of Dependence: Efficient Black-box Explanations Using Dependence Measure''.} In: \textit{Advances in Neural Information Processing Systems} (\textcolor{confcolor}{NeurIPS})

    \item Victor Boutin, \textbf{Thomas Fel}, Lakshya Singhal, Rishav Mukherji, Akash Nagaraj, Julien Colin, Thomas Serre, (2023). \textit{``Diffusion Models as Artists: Are we Closing the Gap between Humans and Machines?''.} In: \textit{Proceedings of the International Conference on Machine Learning} (\textcolor{confcolor}{ICML})

    \item Drew Linsley, Ivan F Rodriguez, \textbf{Thomas Fel}, Michael Arcaro, Saloni Sharma, Margaret Livingstone, Thomas Serre, (2023). \textit{``Performance-optimized deep neural networks are evolving into worse models of inferotemporal visual cortex''.} In: \textit{Advances in Neural Information Processing Systems} (\textcolor{confcolor}{NeurIPS}).

    \item Sabine Muzellec, \textbf{Thomas Fel}, Victor Boutin, Leo Andeol, Rufin VanRullen, Thomas Serre, (2024). \textit{``Saliency strikes back: How filtering out high frequencies improves white-box explanations''.} In: \textit{Proceedings of the International Conference on Machine Learning} (\textcolor{confcolor}{ICML})

    \item Agustin Martin Picard, Lucas Hervier, \textbf{Thomas Fel}, David Vigouroux, (2023). \textit{``Influenci{\ae}: A library for tracing the influence back to the data-points''.} In: \textit{Proceedings of World Conference on eXplainable Artificial Intelligence} (\textcolor{confcolor}{xAI}).

    \item Fanny Jourdan, Agustin Picard, \textbf{Thomas Fel}, Laurent Risser, Jean Michel Loubes, Nicholas Asher, (2023). \textit{``COCKATIEL: COntinuous Concept ranKed ATtribution with Interpretable ELements for explaining neural net classifiers on NLP tasks''.} In: \textit{Proceedings of the Association for Computational Linguistics} (\textcolor{confcolor}{ACL}).

    \item Léo Andéol, \textbf{Thomas Fel}, Florence De Grancey, Luca Mossina, (2023). \textit{``Confident Object Detection via Conformal Prediction and Conformal Risk Control: an Application to Railway Signaling''.} In: \textit{Symposium on Conformal and Probabilistic Prediction with Applications} (\textcolor{confcolor}{COPA}).

    \item Christopher J Hamblin, \textbf{Thomas Fel}, Srijani Saha, Talia Konkle, George A Alvarez, (2023). \textit{``Feature Accentuation: Explaining 'what' features respond to in natural images''}. \textit{Preprint}.

\end{itemize}

\clearpage

\chapter{Attributions Methods}
\label{chap:attributions}

\begin{chapterabstract}
\textit{
Attributions are commonly used tools to explain neural networks. They help reveal where the model is paying attention, aiding users in determining the relevance of elements deemed important by the model. In the context of images, results are often presented as heatmaps, where hotter areas indicate greater importance, while cooler areas suggest less importance.
In this chapter, we introduce four key contributions to the field of Attributions methods. First, we propose in \autoref{sec:attributions:mege}~a new explainability metric based on algorithmic stability, aimed at identifying models with more general and consistent explanations. Next, we explore in \autoref{sec:attributions:sobol}~a new black-box attribution methods by introducing a state-of-the-art method using Sobol indices and Quasi-Monte Carlo sampling. This method is notably twice as fast as existing approaches and is grounded in a strong theoretical foundation in Sensitivity Analysis. We then show that it is possible to further extend theoretical guarantees by presenting in \autoref{sec:attributions:eva}~ the first explainability method that scales to large vision models with formal guarantees. The chapter concludes with \autoref{sec:attributions:metapred}~where an evaluation of the utility of these methods is performed, revealing that they are most useful in simple scenarios. Based on these findings, we propose several hypotheses to address this limitation, which will naturally leads to the topics of the next chapters.
}
\end{chapterabstract}

The work in this chapter has led to the publication of the following conference papers:
{\small{
\begin{itemize}
    \item \textbf{Thomas Fel}, David Vigouroux, Remi Cadene, Thomas Serre (2022). \textit{``How Good is your Explanation? Algorithmic Stability Measures to Assess the Quality of Explanations for Deep Neural Networks''.} In: \textit{Proceedings of the Winter Conference on Computer Vision} (\textcolor{confcolor}{WACV})
    
    \item \textbf{Thomas Fel}\equal, Remi Cadene\equal, Mathieu Chalvidal, Matthieu Cord, David Vigouroux, Thomas Serre, (2021). \textit{``Look at the Variance! Efficient Black-box Explanations with Sobol-based Sensitivity Analysis''.} In: \textit{Advances in Neural Information Processing Systems}  (\textcolor{confcolor}{NeurIPS})
    
    \item \textbf{Thomas Fel}\equal, Melanie Ducoffe\equal, David Vigouroux\equal, Remi Cadene, Mikael Capelle, Claire Nicodeme, Thomas Serre, (2023). \textit{``Don't Lie to Me! Robust and Efficient Explainability with Verified Perturbation Analysis''.} In: \textit{IEEE/CVF Conference on Computer Vision and Pattern Recognition}  (\textcolor{confcolor}{CVPR})

    \item Julien Colin\equal, \textbf{Thomas Fel}\equal, Remi Cadène, Thomas Serre, (2021). \textit{``What I Cannot Predict, I Do Not Understand: A Human-Centered Evaluation Framework for Explainability Methods''.} In: \textit{Advances in Neural Information Processing Systems} (\textcolor{confcolor}{NeurIPS})    

\end{itemize}
}}

\minitoc
\clearpage

\section{Overview}
\label{sec:attributions:intro}
As mentioned in \autoref{chap:intro}, attribution methods (see Definition \ref{def:attributions}) aim to explain a specific prediction of a model. That is, for a topological input space $\sx \subset \Real^d$ and $\sy \subset \Real$ an output space, we study a \textit{predictor}\footnote{For brevity, we intentionally omit the parameters $\f(\vx; \parameters)$ of the predictor.} $\f : \sx \to \Real$, which is a measurable\footnote{All topological spaces are still equipped with their Borel $\sigma$-algebra.} function map any image $\vx \in \sx$ to a prediction $\f(\vx) \in \sy$.
The goal of attribution methods is to explain which variables of $\vx$ are most important in explaining the decision $\f(\vx)$. We will later see that the crux of the matter boils down to defining \textit{what we mean by importance}. 
Formally, an attribution method $\explainer : \mathfrak{F} \times \sx \to \Real^{|\sx|}$ is a functional that given a predictor and an input return a real for each variable in the input -- we note that in this definition, the score is not necessarily bounded. The higher the score, the more important the variable is considered; the lower the score, the more dispensable the variable may be. 
In this section, we aim not to exhaustively cover all attribution methods, but to highlight the most popular ones for vision models. We'll begin by examining gradient-based methods, followed by those utilizing internal states, and conclude with black box methods relying solely on forward calls to the models. Subsequently, we will recall the most common automatic metrics used for evaluating attributions explanations.

\subsection{Gradient-based methods.}

When exploring the importance of variables in a system or model, one of the primary approaches is through local sensitivity methods. These methods offer quantitative techniques for evaluating the impact of infinitesimal changes around the nominal value of an input. By studying how outputs vary with small shifts in inputs, these methods focus on partial derivatives concerning each input parameter. In essence, they allow us to understand how sensitive a system is to infinitesimal alterations in its initial conditions or parameters. 
In practice, those methods use the auto-differentiation framework and thus assume derivability, $\f$ need to be at least of class $C^1$ -- which is not strictly true for ReLU networks~\cite{bertoin2021numerical}.

\paragraph{Saliency.} It turns out that one of the first attribution methods for deep neural network, Saliency -- introduced in \cite{simonyan2013deep} -- is a local sensitivity method and is using absolute value of gradient as importance measure. Formally, Saliency (Sa) defined as:
$$
\explainer_{\text{Sa}}(\f, \vx) = \grad_{\vx} \f(\vx)
$$

In essence, as $\f(\cdot)$ often represent the logit value for a specific class, indicating which pixels in a small neighborhood need modification to most significantly impact the class score, whether positively or negatively\footnote{In the original paper, the authors propose to apply the $\ell_{\infty}$-norm over the channel in case of RGB images.}.

\paragraph{Gradient-Input.} Another close variant is the Gradient-Input (GI) method proposed by \cite{shrikumar2017learning}. This method involves element-wise multiplication of the input with the gradient of the target score. Formally:
$$
\explainer_{\text{GI}}(\f, \vx) = \vx \odot \grad_{\vx} \f(\vx).
$$
It was introduced to improve the sharpness of the attribution maps. A theoretical analysis conducted by~\cite{ancona2017better} showed that Gradient $\odot$ Input is equivalent to two other popular method $\epsilon$-LRP and DeepLIFT~\cite{shrikumar2017learning}, under certain conditions -- using a baseline of zero, and with all biases to zero. 

However, it turns out that gradient-based methods can be quite noisy when visualized. Several reasons have been identified for this phenomenon, and various methods have been proposed to address it. 

\paragraph{SmoothGrad.} One such method is SmoothGrad (SG)~\cite{smilkov2017smoothgrad}, which, as the name implies, aims to smooth out the noise in the gradients. SmoothGrad computes the average gradient over multiple points generated by small perturbations drawn independently and identically from an isotropic normal distribution with standard deviation $\sigma$ around the point of interest. The smoothing effect induced by this averaging process helps reduce visual noise, thereby improving the quality of explanations. Formally:
$$ 
\explainer_{\text{SG}}(\f, \vx) = \underset{\v{\delta} \sim \mathcal{N}(0, \Id\sigma)}{\E}(\nabla_{\vx} \f( \vx + \v{\delta}) ).
$$

\paragraph{Integrated gradients.} Another method aimed at mitigating the noise issue, based on axiomatic principles, is Integrated gradients (Ig)~\cite{sundararajan2017axiomatic}. Integrated gradients involve summing the gradient values along a path from a baseline state $\vx_0$ to the current value $\vx$. The baseline $\vx_0$ used is the zero vector, and the integral can be easily approximated by evaluating the gradient at a set of points evenly spaced between the baseline and the point of interest. Formally:
$$ 
\explainer_{\text{Ig}}(\f, \vx) = (\vx - \vx_0) \int_0^1 \grad_{\vx} \f(\vx_0 + \alpha(\vx - \vx_0)) \dif\alpha. 
$$

\paragraph{VarGrad, SquareGrad.} Other methods have been proposed as variants of SmoothGrad such as VarGrad (VG)~\cite{hooker2018benchmark} or SquareGrad~\cite{hooker2018benchmark} that resp. take the variance of the gradient or the squared gradient to diminish and reduce the noise. For an in depth study of those methods, we refer the reader to the excellent work of~\cite{seo2018noise}. Formally:

$$ 
\explainer_{\text{VG}}(\f, \vx) = \underset{\v{\delta} \sim \mathcal{N}(0, \Id\sigma)}{\V}(\nabla_{\vx} \f( \vx + \v{\delta}) ).
$$

$$ 
\explainer_{\text{S2}}(\f, \vx) = \underset{\v{\delta} \sim \mathcal{N}(0, \Id\sigma)}{\E}(\nabla_{\vx} \f( \vx + \v{\delta})^2 ).
$$

\paragraph{Meaningful Perturbation.} Ruth Fong's seminal work, presented in \cite{fong2017meaningful} and further elaborated in \cite{fong2019extremal}, introduces a novel perspective on attribution methods through the concept of \textit{meaningful perturbation}. Diverging from traditional gradient-based methods, this approach focuses on manipulating the input image to identify the smallest subset of pixels whose alteration -- be it through deletion, inpainting or  blurring -- most significantly affects the model's output. The core of this method lies in the optimization of a mask $\v{m}$, applied to the original image $\vx$, through a perturbation function $\v{\tau}$:

$$
\explainer_{\text{Mp}}(\f, \vx) = \argmin_{\v{m} \in \Real^{|\sx|}} \f(\v{\tau}(\vx, \v{m})) + \lambda \norm{\v{1} - \v{m}}_1
$$

Here, $\v{\tau}(\vx, \v{0}) = \vx$ implies the absence of perturbation: the image remains unchanged. The objective is to determine the minimal set of variables whose removal most dramatically decreases the model's confidence in its decision. The optimization of the mask $\v{m}$ is achieved through gradient descent. The article propose additional mechanisms designed to enhance and stabilize the optimization. These include total variation (TV) regularization to promote spatial coherence and smoothness in the mask, low-dimensional parameterization to reduce the high-frequency, and stochastic augmentation to ensure robustness against variations in input.  For full detail, we refer the reader to the excellent~\cite{fong2017meaningful}\footnote{We also note that the idea of Meta-predictor that we will found later in the manuscript, \autoref{sec:meta_pred}, originate from this article.}.

\vspace{0.3cm}

Gradient-based methods rely on the assumption of differentiability, but may not fully exploit the architectural components of models. In contrast, the upcoming methods we will discuss leverage these structural components to provide more faithful explanations.

\subsection{Internal methods.}

\begin{figure}[ht]
    \centering
    \includegraphics[width=1.0\textwidth]{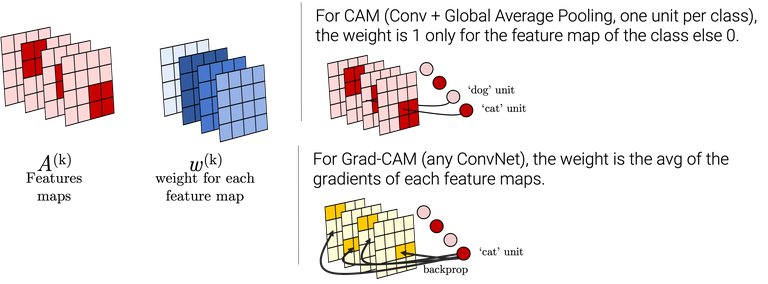}
    \caption{\textbf{Illustration of CAM and Grad-CAM.} Both methods utilize internal feature maps $\m{A}^{(k)}$ and an associated weight $w^{(k)}$ for each feature map (one per channel) to construct the explanation. In CAM, each channel is linked exclusively to a single logit, enabling the upsampling of the $k$-th feature map corresponding to the $k$-th logit. For standard ConvNets, Grad-CAM suggests using the average gradient per channel to determine the weights.}
    \label{fig:attributions:gradcam}
\end{figure}

\paragraph{Grad-CAM (GC).} The most popular method specifically designed for Convolutional Neural Networks (CNN) is Grad-CAM~\cite{Selvaraju_2019}. This method leverages the gradients and the $k$ feature maps $\m{A}^{(k)}(\vx)$ of the last convolutional layer concerning the input $\vx$. To generate the attribution map, we compute weights $w^{(k)}$, essentially scalars corresponding to each filter $\m{A}^{(k)}(\vx)$, where $k$ represents the number of filters (or channels) and $Z$ denotes a constant, which is the number of features in each feature map. The computation of $w^{(k)}(\f, \vx) = \frac{1}{Z} \sum_i \sum_j \frac{\partial \f(\vx)}{\partial \m{A}^{(k)}_{i,j}(\vx)}$ is crucial for this process. The Grad-CAM explanation is then obtained as:
$$\explainer_{\text{GC}}(\f, \vx) = \max(0, \sum_k w^{(k)}(\f,\vx) \m{A}^{(k)}(\vx)) .$$
Since the size of the resulting explanation depends on the dimensions (width, height) of the last feature map, the author performs a bilinear interpolation to match the dimensions of the input. 

\paragraph{Grad-CAM++.} Another close method is Grad-CAM++ (G++)~\cite{chattopadhay2018grad} which is an extension of Grad-CAM that combine the
positive partial derivatives of feature maps of a convolutional layer with a weighted special score. The weights $w^{(k)}$ associated to each feature map is computed as follows : 
$$w^{(k)}(\f, \vx) = 
    \sum_i \sum_j \Big(
    \frac{ \frac{\pd^2 \f(\vx) }{ (\pd \m{A}_{i,j}^{(k)})^2 } }
    { 2 \frac{\pd^2 \f(\vx) }{ (\pd \m{A}_{i,j}^{(k)})^2 } + \sum_i \sum_j \m{A}^{(k)}_{i,j}  \frac{\pd^3 \f(\vx) }{ (\pd \m{A}_{i,j}^{(k)})^3 } }
    \Big).
$$ 
These methods are generally very fast as they only require a forward pass, and the backward pass on the last convolutional layer is usually not computationally expensive. This efficiency often allows them to outperform Saliency in terms of computation time. Moreover, they address the issue of gradient noise by computing a coarse attribution map of the size of the last feature map (e.g., $7\times7$ for a ResNet50) and extrapolating it using a bilinear (or bicubic) interpolation, resulting in a smooth and non-noisy attribution.

However, they only work on a specific type of architecture, namely differentiable convolutional models. In the final part, we will explore the most agnostic methods, which require no assumptions about the model, and we will refer to them as black-box methods.

\subsection{Black-box methods.}

The final section explores black-box methods, which exclusively rely on model forward passes and manipulate input perturbations to infer variable importance. These methods, known for their causal influence on the model, often offer straightforward interpretability and mitigate confidence issues inherent in gradient-based approaches \cite{adebayo2018sanity,ghorbani2017interpretation,sixt2020explanations}. However, they typically demand extensive computational resources and exhibit poor scalability. These challenges form the core focus of the work presented in the thesis, \autoref{sec:attributions:sobol}, which proposes a novel, efficient, and theoretically sound black-box method based on Sobol indices.

One simple way to study model sensitivity through image perturbations is the One-At-a-Time (OAT) method, where each input variable is sequentially modified while keeping others at nominal values. This process observes the resulting effect on the output. OAT typically involves shifting one input variable while keeping others at a nominal value. The nominal value often represents the target image, while perturbations can span an entire space or adhere to a specific baseline state.

For instance, the Occlusion method involves setting each variable $x_i$ of the input to a baseline state $x_0$ and measuring the score difference to determine variable importance:

$$
\explainer_{\text{OC}}^{(i)}(\f, \vx) = \f(\vx) - \f(\vx_{[x_i = x_0]})
$$

Here, $\vx_{[x_i = x_0]}$ denotes the change to the baseline state for variable $x_i$. In practice, Occlusion operates not on a pixel-wise but on an entire patch-wise level to reduce computational costs and obtain coarser maps, which are later extrapolated similarly to Grad-CAM.

\paragraph{LIME (LI).} Another popular method, LIME~\cite{ribeiro2016lime}, involves dropping patches instead of pixels and fitting a linear model to deduce variable importance, formally:

$$
\explainer_{\text{Li}}(\f, \vx) = \argmin_{\v{w}} 
    \underset{\rv{u} \sim \s{U}(\{0,1\}^d)}{\E} 
        \big( \v{\pi}(\rv{u}) \norm{\f(\v{\tau}(\vx, \rv{u})) - \rv{u} \v{w}^\tr }_2 + \Omega(\v{w}) \big)
$$

Here, $\tau(\vx, \cdot)$ segments an image to return super-pixels according to the second argument $\rv{u}$, a randomly drawn binary vector in $\{0,1\}^d$ (e.g., $\v{\tau}(\vx, \v{1}) = \vx$). Furthermore, $\Omega(\cdot)$ represents a complexity penalty on the predictor, namely the weight $\v{w}$ of the linear model\footnote{It's worth noting that a more general formulation exists that does not specify the predictor's form, but a linear model is generally used to maintain interpretability.}. In the end, LIME identifies the weight of each super-pixel, effectively reconstructing each prediction score as an independent sum of super-pixels.

\paragraph{Shapley Values.} Shapley value is another popular that offer a principled approach to attributing the contribution of each variable to a model's prediction. Originating from cooperative game theory, Shapley values aim to fairly distribute the value generated by cooperation among players. In the context of vision, pixels (variables) are akin to players, and the prediction outcome represents the value generated by their cooperation. Given a prediction function $\f$ and a set of features $\vx = \{x_1, x_2, ..., x_d\}$, the Shapley value $\explainer_{Sh}^{(i)}$ for a feature $x_i$ is defined as the \textbf{weighted}\footnote{The Shapley value is not computed as the average marginal contribution, as is often perceived.} average marginal contribution of a feature $x_i$ across all possible feature combinations. Mathematically, it is expressed as:
$$
\explainer_{Sh}^{(i)}(\f, \vx) = \sum_{\v{u} \in \{0,1\}^d, u_i = 0} \frac{\norm{\v{u}}_1!(d - \norm{\v{u}}_1 - 1)!}{d!} \big(
(\f \circ \v{\tau})(\vx, \v{u} + \e_i) - (\f \circ \v{\tau})(\vx, \v{u})
\big)
$$
Here, $\v{u}$ represents a subset of features excluding $x_i$, and $\v{\tau}(\vx, .)$ is a function that selectively reveals or hides the pixel $i$ based on the presence of $u_i$. Direct computation of Shapley values involves evaluating the model for every possible subset of features, making it computationally expensive for high-dimensional data.

\paragraph{RISE (RI).} Another notable method, RISE (Randomized Input Sampling), was introduced by \cite{petsiuk2018rise} and represents a cutting-edge approach to black-box explainability in vision models. It builds upon the Occlusion method by simultaneously probing the model with multiple randomly removed patches to compute the conditional expectation of the score concerning patch presence.
In practical terms, RISE generates low-dimensional patches, typically $7 \times 7$, and extrapolates them to cover approximately half of the image. Once these images are generated, the conditional expectation of the score with respect to the presence of the patches is computed. Formally:
$$
 \explainer_{\text{RI}}^{(i)}(\f, \vx) = \E_{\rv{m} \sim \P_{\rv{m}}}(\f(\vx \odot \rv{m}) | \rv{m}_i = 1).
$$
Despite its effectiveness, RISE requires approximately 8000 forward passes for one explanation, posing challenges for real-time or efficient explanations.

\subsection{Metrics}

\subsubsection{Plausibility}
The initial metrics proposed in the field of explainable AI were based on the concept of plausibility, aligning with the terminology introduced by \cite{jacovi2021formalizing}. These metrics aim to measure the extent to which an attribution-based explanation correlates with a "ground truth" explanation, i.e., an ideal representation that precisely indicates the model's rationale. For instance, to explain a model's recognition of a cat, an ideal explanation might be a segmentation map highlighting the cat or the hottest point on a heatmap situated directly on the cat.

Among such approaches is the framework proposed by \cite{fong2017meaningful}, which, along with its associated library, simplifies the measure of plausibility. Other notable mentions include~\cite{poerner2018evaluating,lundberg2017unified}, which provides a benchmark for evaluating explanations against ground truth annotations.

However, these plausibility metrics have been critiqued for a significant limitation: a high plausibility score does not necessarily affirm the explanation method's effectiveness, but rather the quality of the explanation itself. To accurately evaluate explanation methods, the criteria should reflect how well an explanation reveals the true basis of the model's decision-making process, regardless of whether the model's decisions are correct or desirable. For example, if an explanation method uncovers that the model is using grass to identify a cat, it demonstrates the method's accuracy in capturing the model's focus but might be penalized by plausibility metrics. To address this, various fidelity metrics have been developed.

\subsubsection{Fidelity}

\begin{figure}[ht]
    \centering
    \includegraphics[width=1.0\textwidth]{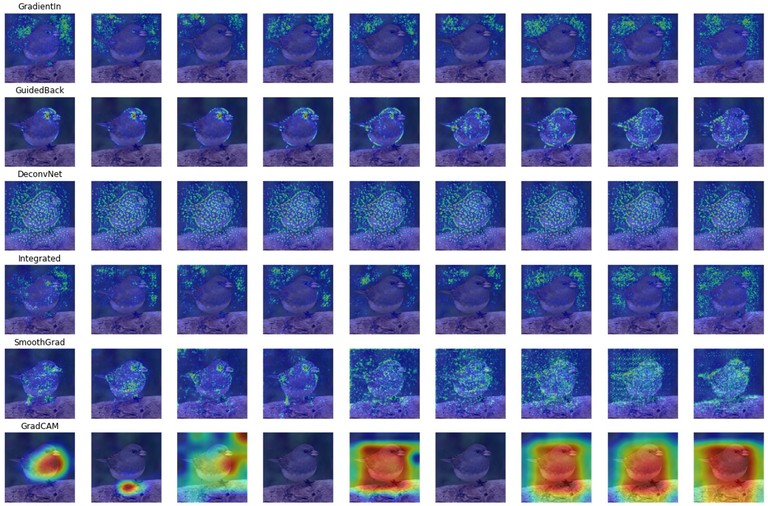}
    \caption{\textbf{Reproduction of findings from \cite{adebayo2018sanity} using Xplique \cite{fel2022xplique}.} This experiment assesses explanation methods by progressively randomizing the weights of the model's layers, culminating in a fully randomized model on the right. Interestingly, for certain explanation methods, the generated explanations remain visually coherent and plausible, even when the underlying model is random. This observation led the authors to speculate that some explanation techniques might primarily be performing contour detection rather than revealing meaningful insights into the model's decision-making process. Such a phenomenon raises concerns about the fidelity of attribution methods: the mere coherence of an explanation image does not necessarily validate its relevance in depicting the model's operational rationale.}
    \label{fig:intro:sanity}
\end{figure}

As we have described, plausibility metrics introduce a significant issue related to confirmation bias: the fact that an explanation appears coherent and plausible does not necessarily mean it accurately reflects the underlying model processes \cite{adebayo2018sanity}. A seminal study by \cite{adebayo2018sanity} demonstrated that certain explainability methods provided similar explanations for both a randomized model and a trained model. This phenomenon, illustrated in~\autoref{fig:intro:sanity}, is problematic as some methods resembles contour detection rather than a meaningful explanation, validating a critical concern: an explanation's coherence does not guarantee its relevance to the prediction's evidential basis.

This issue has led to the development of fidelity metrics that we will now describe. They should ensure that the attribution accurately transcribe what is happening within the model, regardless of whether the outcome is aesthetically pleasing or seems plausible. 

\begin{figure}[ht]
    \centering
    \includegraphics[width=1.0\textwidth]{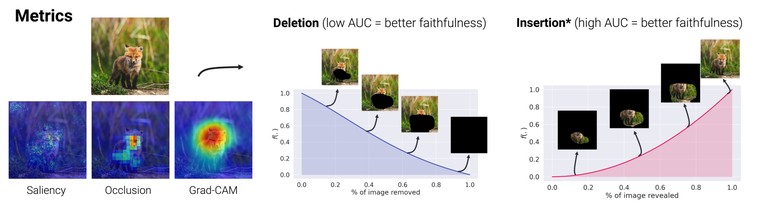}
    \caption{\textbf{Example of Two Fidelity Metrics: Deletion \& Insertion.} These two fidelity metrics operate similarly, using a heatmap to calculate a path from the original image to a baseline image. For Deletion, this is done by starting from the original image and removing the most significant areas according to the heatmap. For Insertion, the process begins from the baseline and progressively adds pixels from the most to the least important, as indicated by the heatmap. The resulting graphs are called the Deletion Curve and the Insertion Curve, respectively. Consequently, a lower Area Under Curve (AUC) value for Deletion is desirable, while a higher AUC value for Insertion is preferred.}
    \label{fig:attribution:deletion}
\end{figure}

\paragraph{Deletion.}~\cite{petsiuk2018rise} 
The first metric is Deletion, it consists in measuring the drop in the score when the important variables are set to a baseline state. Intuitively, a sharper drop indicates that the explanation method has well-identified the important variables for the decision. The operation is repeated on the whole image until all the pixels are at a baseline state. Formally, at step $k$, with $\v{u}$ the $k$-most important variables according to an attribution method, the Deletion$^{(k)}$ score is given by:

$$
\text{Deletion}^{(k)} = \f(\vx_{[x_i = x_0, i \in \v{u}]})
$$

At the initial step, we have $\text{Deletion}^{(0)} = \f(\vx)$, representing the model's output using the original input image. For the final step, $\text{Deletion}^{(d)} = \f(\vx_0)$ corresponds to the model's output when provided with the baseline image, with $\vx \in \Real^d$. The sequence of scores obtained through this process forms a curve $\mathcal{C} = \{ (0, \text{Deletion}^{(0)}), \ldots, (k, \text{Deletion}^{(k)}), \ldots, (d, \text{Deletion}^{(d)}) \}$, capturing the impact of gradually deleting information from the input on the model's performance. The final Deletion score is derived from the Area Under the Curve (AUC) of this deletion curve, denoted as $\text{Deletion} = \text{AUC}(\mathcal{C})$. The baseline is usually a scalar (e.g, $0$) but could also be a random variable drawn from a distribution (e.g. $\rvx_0 \sim \mathcal{N}(0, 1)$).

\paragraph{Insertion.}~\cite{petsiuk2018rise}
Insertion consists in performing the inverse of Deletion, starting with an image in a baseline state and then progressively adding the most important variables. Formally, at step $k$, with $\v{u}$ the $k$-most important variables according to an attribution method, the Insertion$^{(k)}$ score is given by:
$$
\text{Insertion}^{(k)} = \f(\vx_{[x_i = x_0, i \in \complementary\v{u}]})
$$

With $\complementary\v{u}$ the complementary set of $\v{u}$ on $\{1, \ldots, d\}$. The final score is also computed using the AUC of the curve, and the baselines are similar to those of Deletion.

These two metrics, therefore, generate a sequence based on the original explanation $\explainer(\f, \vx)$. In the case of the Deletion metric, the process starts with a minimal number of pixels being masked—close to zero—and progressively moves towards the image being nearly entirely replaced by the baseline state. Conversely, the third metric, \muf, adopts a different approach. Rather than gradually obscuring the image until it reaches a baseline state, aims to maintain closeness to the original image and only drops a specified percentage of pixels (or patches) at random.

\paragraph{$\mu$Fidelity}~\cite{aggregating2020} consists in measuring the correlation between the fall of the score when variables are put at a baseline state and the importance of these variables. Formally:

$$
\mu\text{Fidelity} = \underset{\v{u} \subseteq \{1, ..., d\}}{\operatorname{Corr}}\left( \sum_{i \in \v{u}} \explainer(\f, \vx)_i  , \f(\vx) - \f(\vx_{[x_i = x_0, i \in \v{u}]})\right) ~~s.t.~~ |\v{u}| = k
$$

In various studies, the parameter $k$ is often set to 20\% of the total number of variables, and the baseline used is generally consistent with those employed by the Deletion and Insertion metrics. It is noteworthy that the concepts of Deletion and Insertion metrics have been independently rediscovered and named differently across multiple studies\footnote{Surprisingly, even though literature frequently cites \cite{petsiuk2018rise} for introducing these concepts, earlier works had already proposed similar metrics, albeit under different terminologies, such as those by \cite{samek2015evaluating}, \cite{fong2017meaningful}, and \cite{kapishnikov2019xrai}.}.

Another important metric not covered in this work is Infidelity, presented in \cite{yeh2019infidelity}, which is defined as:

$$
\text{Infidelity} = \underset{\rv{m} \sim \P_{\rv{m}}}{\E} \left(( \explainer(\f, \vx) \rv{m}^\tr - (\f(\vx) - \f(\vx - \rv{m})))^2 \right).
$$

Infidelity can be seen as a variation of the \muf~metric, where the focus shifts from measuring correlation to assessing the $\ell_2$ distance. This distance is calculated between the heatmap scores of the removed variable subsets and the change in the model's output score when the features within this set are excluded. 

These metrics serve a crucial role in ensuring that an explanation accurately captures the model's decision-making process. A pertinent question arising from this is how well an explanation holds up when deviating from a specific input, $\vx$, or, more precisely, the degree of stability of this explanation across varying inputs.

\subsubsection{Stability}

Stability is often highlighted as a desirable attribute of an explanation. \cite{alvarezmelis2018robust} were among the first to formally propose a stability metric for attribution methods, advocating that ``similar inputs should lead to similar explanations''. They conceptualized the stability metric as a measure of robustness within a local neighborhood around $\vx$, denoted by $\ball = \{\v{z} : \norm{\vx - \v{z}}_p \leq \delta \}$:

$$
\text{Stability} = \underset{\v{z} \in \ball}{\max} \frac{\norm{\explainer(\f, \vx) -\explainer(\f, \v{z})}_p}{\norm{\vx - \v{z}}_p}
$$

This formulation can be interpreted as a local Lipschitz constant, not of the function $\f$ itself, but of the explanation function $\explainer$. Essentially, it seeks to quantify how significantly an explanation changes as inputs vary slightly around our original point, underlining the \textit{explanation method's sensitivity} to input perturbations. 

More recently, alternative versions of the stability metric have been proposed, which instead of focusing on maximal deviations, consider an average measure of robustness across the local neighborhood \cite{aggregating2020}:

$$
\text{Stability}_{\text{avg}} = \int_{\ball} \norm{\explainer(\f, \vx) -\explainer(\f, \v{z})}_p \diff \v{z}
$$

These two metrics thereby assess the extent to which an explanation—or a heatmap provided to a user—remains valid within a vicinity of points. This vicinity is often defined as an $\ell_2$ ball around the focal point but has been refined and rethought in more recent works, such as in~\cite{agarwal2022rethinking}. Through these measures, we gain insights into the explanation's reliability, ensuring that the importance it provides are not just accurate for a single point but hold across a set of points around $\vx$.

\paragraph{Closing note.} All these automated metrics enable a fair and objective comparison between different attribution methods. However, as we will explore in the last part of this chapter (\autoref{sec:attributions:metapred}), dedicated to human-centric metrics, the ultimate goal of explainability is to be useful to humans for a set of specific tasks. Armed with this knowledge, we are now ready to tackle the first part of this chapter that precisely seeks to expand the set of available metrics. This effort aims to establish a metric that evaluates the explanations provided by a model (and not just the attribution method), in order to identify models that offer the best explanations.

\clearpage

\section{Algorithmic Stability to find model with better explanations}
\label{sec:attributions:mege}
\newcommand{\fidelity}{\textit{Fidelity}}
\newcommand{\stability}{\textit{Stability}}
\newcommand{\comprehensibility}{\textit{Comprehensibility}}
\newcommand{\consistency}{\textit{Consistency}}
\newcommand{\representativity}{\textit{Generalizability}}
\newcommand{\mege}{MeGe}
\newcommand{\reco}{ReCo}
\newcommand{\Setp}{\boldsymbol{\mathcal{S}^{ \ne }}}
\newcommand{\Setm}{\boldsymbol{\mathcal{S}^{ = }}}
\newcommand{\Seta}{\boldsymbol{\mathcal{S}}}
\newcommand{\fold}{\mathcal{V}}
\newcommand{\deltaij}{\delta_{\vx}^{(i,j)}}

In this section, we aim to develop a new metric for explainability to identify better models. As we've seen in \autoref{sec:attributions:intro}, many attributions methods have been proposed to explain how deep neural networks make decisions, but there hasn't been much effort to ensure that the explanations they provide are objectively relevant. While several desirable \textit{properties} for trustworthy explanations have been identified, it's been challenging to come up with \textit{objective measures} for them. Here, we propose two new measures to assess explanations, borrowed from the field of algorithmic stability: mean generalizability (\mege) and relative consistency (\reco).

We'll begin by briefly reviewing related work on metrics, then we'll introduce our methods and the two metrics. Afterward, we conduct extensive experiments using various network architectures, common explainability methods, and several image datasets to showcase the advantages of these measures. We'll demonstrate that they pass sanity checks, allowing us to move on to the experimental phase where we'll show (1) that current fidelity measures are not sufficient to guarantee algorithmically stable and trustworthy explanations, (2) that our metrics can be use to select the best attribution method for a given model, and finally (3) that our metrics can be use to identify models with better explanations.

\begin{figure}[ht]
  \centering
  \includegraphics[width=0.80\textwidth]{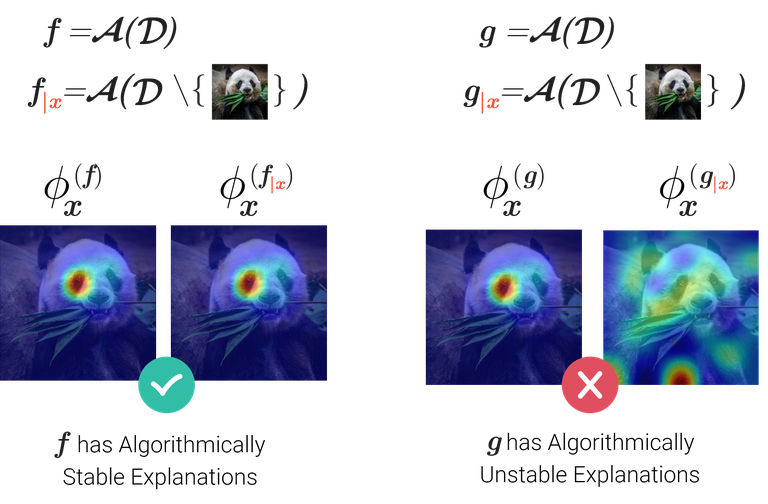}
  \caption{ 
  The trustworthiness of a predictor's explanations hinges on its algorithmic stability. This concept implies that when an image $\vx$ is excluded from the training dataset $\s{D}$, a separate predictor $\f_{\textcolor{red}{|\vx}}$ trained using the same algorithm $\s{A}$ (but without $\vx$) should yield a comparable explanation for that image. This stability indicates that the explanations are drawn from multiple points, making them more general. For instance, when classifying images as "pandas," the explanations should consistently highlight the dark areas around their eyes, with or without $\vx$ in the dataset.
  }
  \label{fig:big_picture}
\end{figure}

\subsection{Background.} 

In this section, we focus on evaluating explanations provided by explainability methods, which give insight into how a given neural network architecture reaches a particular decision~\cite{doshivelez2017rigorous}. These explainability methods produce an influence score for each variables. In the case of image classification, these methods will produce heatmaps indicating the diagnosticity of individual image regions. Most of these explainability methods rely on backpropagating the gradient with respect to a given input image ~\cite{Zeiler2011, simonyan2013deep, bach2015pixel, Fong_2017, shrikumar2017learning, sundararajan2017axiomatic, smilkov2017smoothgrad, Selvaraju_2019, hartley2021swag} or with respect to a perturbation of the input~\cite{zeiler2013visualizing, zhou2014object, ribeiro2016lime, li2016understanding, zintgraf2017visualizing, ribeiro2018anchors}. We refer the reader to \autoref{sec:attributions:intro} for more details.

Despite a wide range of explainability methods, assessing the quality and trustworthiness of these explanations is still an open problem. It is in part due to the difficulty of obtaining objective ground truths~\cite{samek2015evaluating,linsley2018learning}. Several criteria have been proposed to evaluate the quality of explanations~\cite{tintarev2007survey, miller2017explanation,robnik2018perturbation,gilpin2018explaining,alvarezmelis2018robust, survey2019metrics,ferrettini2021coalitional}. According to~\cite{survey2019metrics}, the five major properties include: \fidelity, \stability, \comprehensibility, \representativity~and \consistency. Yet, properties such as \representativity~and \consistency~do not come with a practical definition.

In order to measure these different properties, there are two main approaches currently used. The first subjective approach consists in putting the human at the heart of the process, either by explicitly asking for human feedback~\cite{Selvaraju_2019, ribeiro2016lime, lundberg2017unified}, or by indirectly measuring the performance of the human/classifier duo~\cite{lage2019evaluation, narayanan2018humans, schmidt2019quantifying}. Nevertheless, human intervention sometimes brings undesirable effects, including a possible confirmation bias~\cite{adebayo2018sanity}.

A second type of approaches has also started to emerge specifically for computer vision applications. The main idea is to build objective proxy tasks that a good explanation must be able to solve.
These measures aim to evaluate explanations based on two properties: \fidelity~and \stability. The first method to measure \fidelity~was first proposed in~\cite{samek2015evaluating} based on estimating the drop in prediction score resulting from deleting pixels deemed important by an explanation method. To ensure that the drop in score does not come from a change in distribution, ROAR~\cite{hooker2018benchmark} was proposed which re-train a classifier model between each deletion step. This boils down to measuring the correlation between the attributions for each pixel and the difference in the prediction score when they are modified and has been clearly formalized~\cite{yeh2019infidelity, aggregating2020, rieger2020irof}.
Nevertheless, it should be noted that the different fidelity metrics proposed requires defining a baseline state which might favor explainability methods that internally relies on the same baseline~\cite{sturmfels2020visualizing}.

Those \fidelity~metrics are a first step toward trustworthy explanations: by making sure that we have faithful explanations, we can then look at other criteria to quantitatively measure these explanations. 

\paragraph{Algorithmic Stability} represents a nuanced form of Sensitivity Analysis, focusing on the impact that modifications to the learning dataset have on an algorithm's output. This concept is pivotal for deriving various generalization bounds, as highlighted by Bousquet et al.~\cite{bousquet2002stability}. Essentially, an algorithm demonstrates stability if it yields consistent predictions across datasets that only differ by a single instance. More precisely, an algorithm $\mathcal{A}(\vx ; \s{D})$, which train on $\s{D}$ and output a prediction on $\vx$, is deemed $\xi$-uniformly stable if, for any two datasets $\mathcal{D}$ and $\mathcal{D}'$ differing by at most one element, the subsequent inequality is satisfied for a given loss function $\ell$:

\begin{definition}[Uniform Stability~\cite{bousquet2002stability}] An algorithm $\mathcal{A}$ exhibits uniform stability $\xi$ with respect to a loss function $\ell$ if for every input $\vx$ and output $y$, the following condition is met:

\begin{equation}
\sup_{\vx} \left| \ell(\mathcal{A}(\vx; \mathcal{D}), y) - \ell(\mathcal{A}(\vx; \mathcal{D}'), y) \right| \leq \xi
\end{equation}

\end{definition}

Here, $\xi$ denotes a small, non-negative constant. This stability criterion is integral for assessing the generalization performance of various statistical learning models, offering a theoretical basis to gauge the efficacy of a model on novel data. Algorithmic stability suggests that an algorithm's reliance on any particular training instance is minimal, thus mitigating the risk of overfitting and bolstering the model's generalization capabilities. Consequently, there is a direct correlation between algorithmic stability and generalization error, illuminating the delicate equilibrium between training data fidelity and resilience to data variability.

Below, we briefly provide some motivation that rely on this notion for the proposed  \mege~and \reco~ measures before describing a training procedure applicable to a large family of machine learning models in order to estimate these two values. 

\subsection{Algorithmic Stability measure for Explainability}

\paragraph{Notations.} We consider a standard supervised learning setting where a datapoint is denoted $\v{z} = (\vx, \vy)$ s.t. $\vx \in \sx$ is an observation (e.g., $\sx = \mathbb{R}^d$) and $\vy \in \sy$ is a class label (e.g., $\sy = \mathbb{R}^p$).
The data set is denoted as $\s{D} = \{ \v{z}_1, ..., \v{z}_m \}$,  we designate $\fold = \{ \fold_1, ..., \fold_k \}$ the set of $k$ disjoints subsets (\textit{folds}) of size $m/k$ at random where each $\fold_i \subset \s{D}$. Throughout this work, we will assume $k$ divides $m$ for convenience.
Let $\s{A}$ be a deterministic learning algorithm that maps any number of data points onto a predictor function $\s{A} : \s{D} \to (\sx \to \sy)$.
In particular, we consider the \textit{fold} $\fold_i$ and the associated predictor $\f_i = \s{A}(\fold \setminus \fold_i)$.

An explanation method is a functional, denoted $\explainer$, which, given a predictor $\f_i$ and a datapoint $\vx$, assigns an importance score for each input dimension $\explanation_{\vx}^{(i)} = \explainer(\f_i, \vx)$. Moreover, we assume a distance $d(\cdot,\cdot)$ over the explanations.
Finally, the following Boolean connectives are used: $\neg$ denotes a negation, $\land$ denotes a conjunction, and $\oplus$ denotes an exclusive or (XOR).

\paragraph{Motivation.} We first consider \representativity: we provide a definition, discuss the inherent difficulties associated with its measurement, and describe a method for estimating it.
We then motivate the need for assessing the \consistency~of an explanation and propose a measure.

\begin{definition}{\representativity}\\
A measure of how generalizable an explanation is, and the extent to which it captures the underlying patterns or features across various datasets or scenarios. 
\label{def:mege:representativity}
\end{definition}

Intuitively, a representative explanation would be an explanation that holds for a large number of samples.
To assess the number of samples that can be covered by a given explanation, it might be tempting to compute a distance between the explanations associated with those samples. However,  because of the large variations in the appearance of objects that arise because of translation, scale, and 3D rotation in natural images, two explanations can be similar (i.e., close in pixel space) without necessarily reflecting a similar visual strategy used by the predictor (for instance, decisions could be driven by the same pixel locations -- yet driven by different visual features). Conversely, two spatially distant explanations could be based on the same features that appear at different locations because of translation. 

Our proposed solution to this problem is to only use distance measured between explanations for the same sample. This constraint leads us to consider the notion of algorithmic stability as a proxy for generalization: intuitively, given a predictor and a training data set, a good explanation for a decision made for a given data point should be robust to the addition or removal of that data point from the training set. One benefit of such a characteristic is that it can be evaluated based solely on a distance between explanations from the same sample.

In what follows, we will propose a relaxed version of the algorithmic stability -- computationally more manageable -- applied to the explanations using several predictors trained on different \textit{folds}.
It is important to note that the term algorithmic stability \cite{bousquet2002stability} is not related to the \stability~of an explanation as defined in~\cite{aggregating2020}.

Following this consideration, we will be looking at how well a predictor's explanations generalize from seen to unseen data points:

\begin{equation}
    \label{eq:mege:delta}
    \deltaij = d(\explanation_{\vx}^i, \explanation_{\vx}^j) \; s.t. \; \vx \in \fold_i, \vx \notin \fold_j.
\end{equation}

\noindent By making sure that $\vx$ only belongs to the \textit{fold} $\fold_i$, we measure the distance between two explanations, one of which comes from a predictor that was not fitted to the sample $\vx$. 
By computing these distances, we hope to characterize the \representativity~of the explanations.
We now propose to study the consistency property.

\begin{definition}{\consistency} \\
The extent to which different predictors trained on the same task do not exhibit logical contradictions.
\label{def:mege:consistency}
\end{definition}

A statement, or a set of statements, is said to be logically consistent when it has no logical contradictions.
A logical contradiction occurs when both a statement and its negation are found to be true.
In logic, a fundamental law -- the law of non-contradiction -- is that a statement and its negation cannot both be true simultaneously.  
Similarly, we measure the consistency between explanations by ensuring that contradictory predictions lead to different explanations.

Following this definition, if the same explanation gets associated with two contradictory predictions the explanation is said to be inconsistent. 
This means avoiding the case where for an observation $\vx \in \fold_i$, two predictors $\f_i, \f_j$ (where $i \neq j$), trained on the same task, give the same explanation but different predictions:

\begin{align}
    \label{eq:mege:consistency}
    \f_i(\vx) \neq \f_j(\vx) \implies \explanation_{\vx}^{(i)} \neq \explanation_{\vx}^{(j)}
\end{align}

Nevertheless, we have to define what it means for two explanations to be different. For this, we use a measure of dissimilarity between explanations and a threshold to judge whether the explanations are consistent or not. This threshold will be relative to the distance between explanations when predictions are not contrary.
By measuring the rate of inconsistent explanations, we hope to capture the notion of \consistency~for explanations.

\paragraph{$k$-Fold Cross-Training}

\begin{figure*}[ht]
  \centering
  \includegraphics[width=0.99\textwidth]{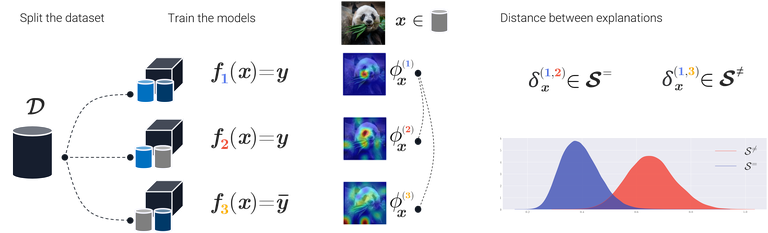}
  \caption{Application of the proposed procedure for $3$ \textit{folds}. Each predictor is trained on two of the 3 \textit{folds}, e.g, $\f_{\textcolor{blue}{1}}$ is trained on $\s{D} \setminus \fold_1$. For a given sample $\vx$ such that $\vx \in \fold_1$, the explanations for each predictors are calculated ($\explanation_{\vx}^{(\textcolor{blue}{1})}, \explanation_{\vx}^{(\textcolor{red}{2})}, \explanation_{\vx}^{(\textcolor{yellow}{3})}$). The distance between $\explanation_{\vx}^{(\textcolor{blue}{1})}$ and the  other two explanations  $\explanation_{\vx}^{(\textcolor{red}{2})}, \explanation_{\vx}^{(\textcolor{yellow}{3})}$ are computed. 
  All distances for which predictions do not contradict each other are added to $\Setm$ while the others are added to $\Setp$ (note that this is the case for $\delta^{(\textcolor{blue}{1},\textcolor{yellow}{3})}_{\vx}$ since $\f_{\textcolor{blue}{1}}(\vx) \neq \f_{\textcolor{yellow}{3}}(\vx)$).
  }
  \label{fig:mege:methodology}
\end{figure*}

We recall that our data set is divided into $k$-\textit{folds} of the same size $\s{D} = \{\fold_i\}_{i=0}^k$, and that each predictor is trained through a learning algorithm $\f_i = \s{A}(\fold \setminus \fold_i)$.
We assume that the predictors exhibit comparable performance across folds. In our experiments, we ensure a similar accuracy on the test set.

We will now measure the distances between two explanations associated with these different predictors.
To be more precise, we are really only interested in computing $\deltaij$ (see Eq.~\ref{eq:mege:delta}):, the distance between two explanations whereby one of the two predictors was not fitted on $\vx$. Otherwise, it may be trivial for two predictors that were trained on that sample to yield the same explanation -- especially if overfitting occurs (see Fig. \ref{fig:mege:methodology}).

In the case where both predictors gave a correct prediction, a small distance between the two explanations suggests that the explanations receive support from several samples. In other words, the fact that  explanations do not vary widely when adding or removing a particular sample or set of samples suggests good \representativity. Alternatively, if the two predictors give contrary predictions, the corresponding explanations should be different. Indeed, the very notion of \consistency~between explanations implies that the same explanation cannot account for two different outcomes. 

We separate distances into two sets, $\Setm$~when the predictors have made correct predictions s.t. it is desirable to have a small distance between explanations, $\Setp$~when one of the predictors have given a wrong prediction s.t. it is desirable to have higher distances between the pairs of explanations. The case where both predictors give a bad prediction is ignored (for details, see the Alg.~\ref{alg:mege:procedure} in the appendix).

\begin{align}
    \Setm &= \{ \deltaij : \f_i(\vx) = \vy \land  \f_j(\vx) = \vy \} \\
    \Setp &= \{ \deltaij : \f_i(\vx) = \vy \oplus  \f_j(\vx) = \vy \}
\end{align}
$$ \hspace{10000pt minus 1fil} \forall (i, j) \in \{1, ..., k\}^2 ~s.t.~ i \neq j, ~\forall (\vx, \vy) \in \fold_i  \hfilneg $$

\paragraph{Mean generalizability : \mege}
\label{MeGemeasure}
From Def.~\ref{def:mege:representativity}, the distance between explanations arising from predictors trained on a dataset that contained vs. did not contain a given sample should be small. 
As those distances are contained in $\Setm$, one way to measure the \representativity~of explanations is to compute the average over $\Setm$.

As a reminder, the average of $\Setm$ corresponds to the average change of explanation when the sample is removed from the training set. This change is related to the \representativity~of the explanation: the more representative an explanation is, the more it persists when we remove a point.

To ensure a high value of our metric for low distances, we define the \mege~measure as a similarity measure:
\begin{align} 
    MeGe &= \Big(1 + \frac{1}{|\Setm|}\sum_{\delta ~\in~ \Setm} \delta \Big)^{-1}
    \label{eq:mege:mege}
\end{align}

Explanations with good \representativity~ will therefore be associated with higher similarity scores between explanations (close to 1).

\paragraph{Relative consistency : \reco}
\label{ReComeasure}

From Def.~\ref{def:mege:consistency} and Eq.~\ref{eq:mege:consistency}, explanations arising from different predictors are said to be consistent if they are close when the predictions agree with one another.
As a reminder, the distance between explanations for the consistent predictions are represented by $\Setm$, and those associated with inconsistent predictions by $\Setp$. 
Visually, we seek to maximize the shift between the corresponding distributions for the sets $\Setm$ and $\Setp$.
Formally, we are looking for a distance value that separates $\Setm$ and $\Setp$, e.g., such that all the lower distances belong to $\Setm$ and the higher ones to $\Setp$. The clearer the separation, the more consistent the explanations are.
In order to find this separation, we introduce \reco, a statistical measure based on maximizing the balanced accuracy.

Where $ \Seta = \Setm \cup \Setp $ and $\lambda \in \Seta$ a fixed threshold value, we can define the true positive rate $TPR$ as the rate for which distances below a threshold come from consistent predictions among all distances below the threshold $TPR(\lambda) = \frac{|\{\delta \in \Setm : \delta \leqslant \lambda \}|} {|\{ \delta \in \Seta~:~ \delta \leqslant \lambda \}| \hfill}$. In a similar way, $TNR$ denotes the rate for which distances above a threshold come from opposite predictions among all the distances above the threshold $TNR(\lambda) = \frac{|\{ \delta \in \Setp : \delta > \lambda \}|}{|\{ \delta \in \Seta~:~ \delta > \lambda \}| \hfill}$. Basing our measure on these rates allows us to assess the quality of these explanations independently of the accuracy of the predictor, we define \reco~ as the maximal balanced accuracy:

\begin{align}
    ReCo &= \max_{\lambda \in \Seta}\ TPR(\lambda) + TNR(\lambda) - 1, 
    \label{eq:mege:reco}
\end{align}

with a score of 1 indicating perfect consistency of the predictors' explanations, and a score of 0 indicating a complete inconsistency.

\subsection{Experiments}

We carried out three sets of experiments using a variety of neural network architectures and explanation methods. 
The first one consisted in ensuring the functioning and the reliability of the measures via a simple sanity check done over a large number of predictors ($175$ in total).
The second set of experiments consisted in highlighting a limitation of the fidelity measure -- namely its \textbf{independence} with respect to the quality of the explanations. 
We developed these considerations in a dedicated section where we demonstrate an application to the selection of a method using the two new criteria \mege~and \reco.
Finally, in a third set of experiments, we showed quantitatively that some predictors are more interpretable: our analyses revealed that 1-Lipschitz neural networks yield explanations that are more representative and coherent.

\paragraph{Setup.} For all experiments, we used 5 splits ($k = 5$), i.e., $5$ predictors with comparable accuracy ($ \pm 3\%$), which allows us to study the explanations in common training conditions (80\% of the data are used for training and 20\% for testing).
For ILSVRC 2012, our predictors are based on a ResNet-50 architecture~\cite{he2016deep}, and a ResNet-18 for the other datasets.
\paragraph{Explanation methods.}
In order to produce the necessary explanations for the experiment, we used $7$ methods of explanation. 
The methods selected are those commonly found in the literature in addition to one control method (Random).
The explanations methods chosen are as follow: Saliency \textbf{(SA)}~\cite{simonyan2013deep}, Gradient $\odot$ Input \textbf{(GI)}~\cite{ancona2017better}, Integrated Gradients \textbf{(IG)}~\cite{sundararajan2017axiomatic}, SmoothGrad \textbf{(SG)}~\cite{smilkov2017smoothgrad}, Grad-CAM \textbf{(GC)}~\cite{Selvaraju_2019}, Grad-CAM++ \textbf{(G+)}~\cite{chattopadhay2018grad} and RISE \textbf{(RI)}~\cite{petsiuk2018rise}. Further information on these methods can be found in \autoref{sec:attributions:intro}.

\paragraph{Datasets.} We applied the procedure described above and evaluated the proposed measures for each of the degradations on $4$ image classification datasets: 
\textbf{ILSVRC 2012}~\cite{imagenet_cvpr09}: a subset of the ImageNet dataset from which we randomly selected $50$ classes. The size of the images considered was $224 \times 224$.
The reduced number of classes being sufficient to show that the metrics pass the test performed even in the case of high dimensional images.
\textbf{CIFAR10}~\cite{krizhevsky2009learning}: a low-resolution labeled datasets with 10 classes respectively, consisting of $60,000$ ($32 \times 32$) color images. 
\textbf{EuroSAT}~\cite{helber2019eurosat}: a labeled dataset with $10$ classes consisting of $27,000$ color images ($64 \times 64$) from the Sentinel-2 satellite.
\textbf{Fashion MNIST}~\cite{xiao2017fashion}: a dataset containing $70,000$ low-resolution ($28 \times 28$) grayscale images labeled in $10$ categories.

\begin{figure*}[ht]
    \centering
    \includegraphics[width=0.98\textwidth]{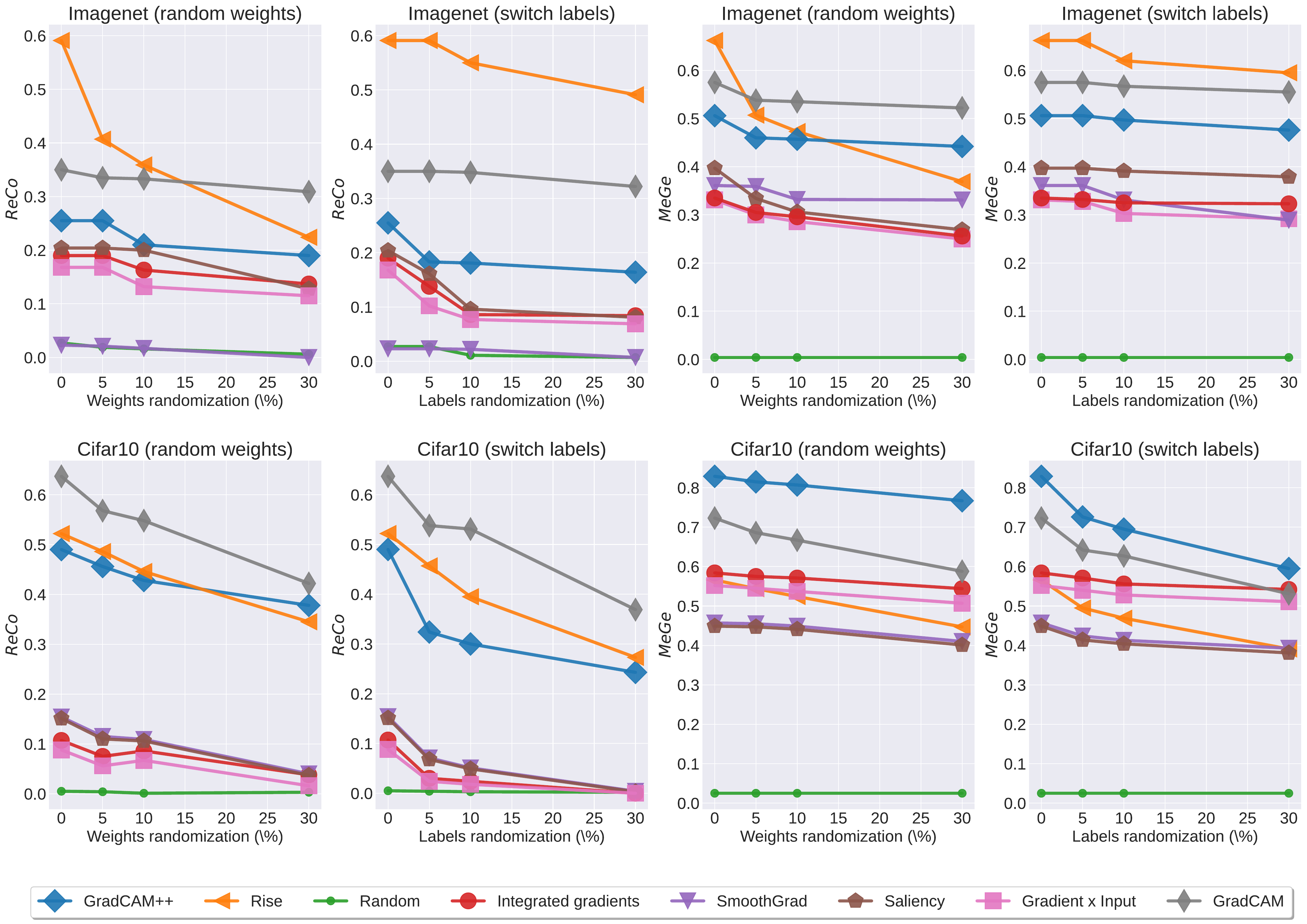}
    \caption{\textbf{\mege~ and \reco~scores} for predictors trained with no degradations (first point from the left), as well as for progressively randomized predictors and predictors trained with switched labels.
    For all the methods tested, the more the predictor is degraded, the more the \consistency ~and \representativity~scores drop, which means that the associated metrics pass the sanity check.
    \textbf{Top} ImageNet. \textbf{Bottom} Cifar-10.
    }
    \label{fig:mege:sanity_check}
\end{figure*}

\paragraph{Distance over explanations.} The procedure introduced in the previous section requires to define a distance between two explanations derived for the same sample. 
Since a feature attribution consists of ranking the features most sensitive to the predictor's decision, it seems natural to consider the  Spearman rank correlation~\cite{spearman1904measure} to compare the similarity between explanations. Several authors have provided theoretical and experimental arguments in line with this choice~\cite{ghorbani2017interpretation, adebayo2018sanity, tomsett2019sanity}. However, it is important to note that the problem of measuring similarity between explanations is still an open problem. We conduct two sanity checks: spatial correlation, and noise test on several candidates distances to ensure they could respond to the problem.  
The distances tested were built from: 1-Wasserstein distance (the Earth mover distance from~\cite{flamary2017pot}), Sørensen–Dice~\cite{dice1945} coefficient, Spearman rank correlation, SSIM~\cite{ssim2004}, and $\ell_1$ and $\ell_2$ norms.
The results of those sanity checks can be found in Appendix \autoref{ap:mege:distances}.
In line with prior work, we chose to use one minus the absolute value of the Spearman rank correlation (see~Appendix \autoref{ap:mege:distances} for more details).

\paragraph{Sanity check for explanation measures.} Our first set of experiments aims to ensure that the propose metrics approximate the desired quantities by performing a sanity check: on average, as the learning is degraded, we expect to see an overall increase in the number of inconsistent explanations.
To ensure that the metric captures these notions, we applied two different types of degradation on the predictors for each data set: weight randomizations and label shuffling.
\begin{itemize}
\setlength\itemsep{-0.2em}
\item Randomizing the weights, inspired by~\cite{adebayo2018sanity}. We gradually randomize $5$\%, $10$\% and $30$\% of the predictor layers by adding Gaussian noise. By degrading the weights learned by the network, we expect to find degraded explanations. 
\item Shuffling of labels, inspired by~\cite{neyshabur2017exploring, adebayo2018sanity} the predictors are trained on a data set with $5$\%, $10$\% and $30$\% of bad labels. By artificially breaking the relationship between the labels, we expect the explanations to lose their consistency.
\end{itemize}

The \mege~measure encodes the \representativity~of the explanations, which is related to the ability of the predictor to derive general strategies. 
Thus, the degradation of the parameters of a predictor directly affects these strategies.
Fig.~\ref{fig:mege:sanity_check} shows the correlation of the measures with the intensity of the degradation applied: \mege~and \reco~capture the degradation of the explanation and pass the sanity check.

We note that all the tested methods perform better than the random baseline (random). However, the drop in score, is not the same and some methods are more sensitive to predictor changes, such as Grad-CAM or RISE, in accordance with previous work~\cite{adebayo2018sanity, sixt2020explanations}. 
It was subsequently observed that this sensitivity seems to translate into a better \fidelity~score for the methods.
Nevertheless, it should be noted that this sanity test is a \textit{necessary but not sufficient} condition for a \representativity~and \consistency~metric.

\paragraph{The implications of the fidelity metric.} To mark the difference between the proposed measures and the \fidelity, we applied the \muf~measure from~\cite{aggregating2020} (see \autoref{sec:attributions:intro}) to the normally trained predictors and those progressively degraded. We observe that this metric does not pass the sanity check: the fidelity measure is invariant to the performancee of the predictor as well as to the quality of its explanations. For \muf, the score obtained is averaged over $10,000$ test samples, and the size of the subset is $15$\% of the image.

\begin{figure}[ht]
    \centering
    \includegraphics[width=0.70\textwidth]{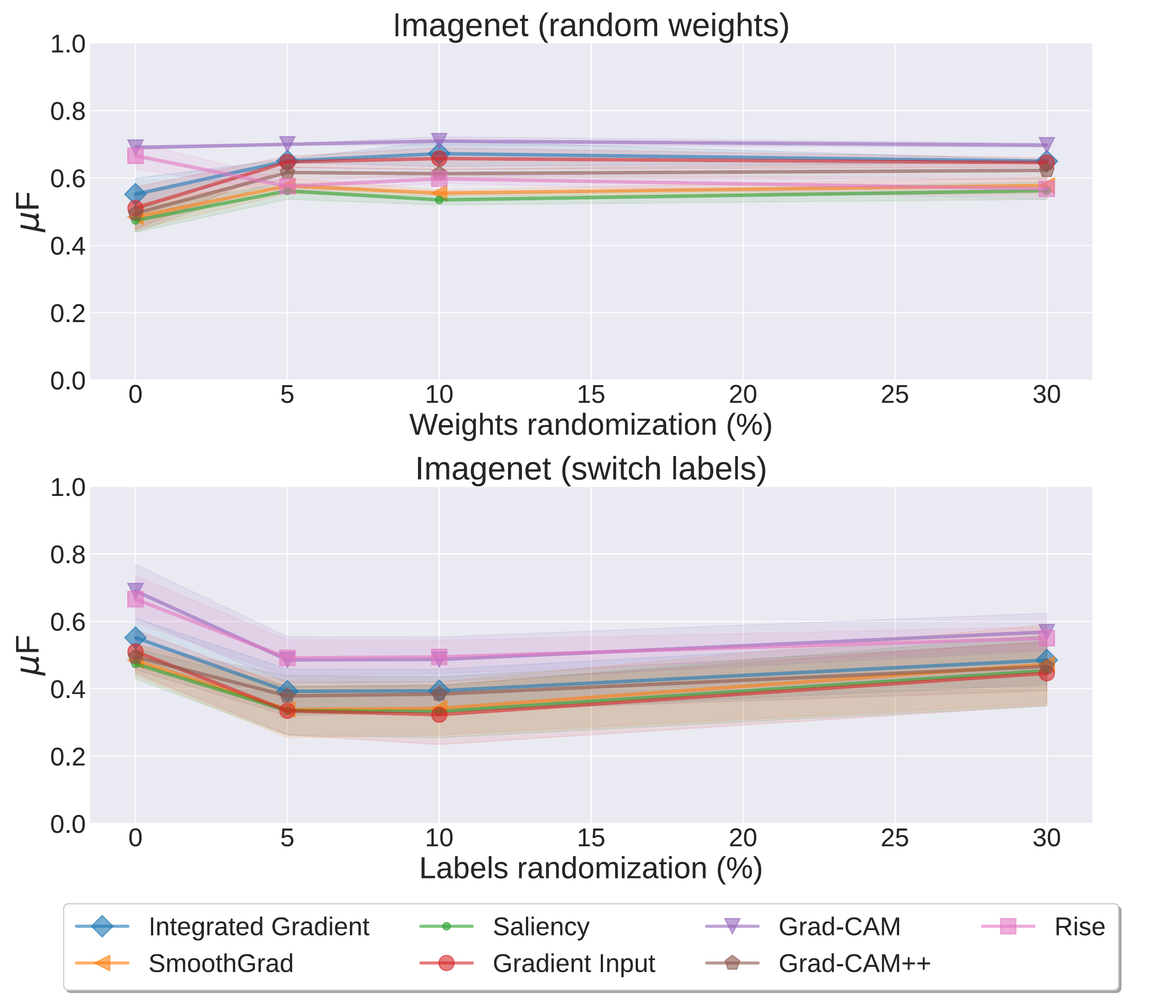}
    \caption{
    \fidelity~scores on ImageNet for normally trained ResNet-50 predictors (first point on the left) as well as for progressively randomized predictors and predictors trained with switched labels.
    Even a strong degradation of the predictor does not impact the \fidelity~of the tested methods. Hence, the \fidelity~is intended to ensure that the explanations correctly reflect the underlying strategies of the model, regardless of whether these strategies are general or consistent.
    }
    \label{fig:fidelity}
\end{figure}

As shown in Fig.~\ref{fig:fidelity}, predictor degradation does not impact the \fidelity~ metric on the methods tested. 
The \fidelity~property is essential in a good explanation since it allows us to make sure that we are studying the strategies of the predictor.
However, it is not sufficient: if the explanation reflects well the strategies of the predictor, the latter may use specific and inconsistent strategies. In that, the \fidelity~measure is only a first step towards a good explanation.

\paragraph{Method selection criterion.} The \mege~and \reco~measures can be used as additional criteria for choosing an explainability method.  
As a reminder, a good method should provide explanations that are as faithful as possible and, if possible, consistent and representative.
Thus, the tested methods can be compared using the scores obtained for these measures. We note that these measures are complementary in that the fidelity score can be interpreted as a confidence bound on the other measures performed on the explanations.

\begin{table}[h]
    \centering
    \scalebox{0.85}{
        \begin{tabular}{l lllllll}
        \toprule
        \textbf{ImageNet} & SA & GI & IG & SG & GC & G+ & RI \\
        \midrule
        $\mu F$ & 0.47 & 0.51 & 0.55 & 0.48 & \textbf{0.69} & 0.49 & \underline{0.67} \\
        \mege     & 0.40 & 0.50 & \underline{0.58} & 0.36 & 0.34 & 0.33 & \textbf{0.66} \\
        \reco     & 0.20 & 0.17 & 0.16 & 0.02 & \underline{0.35} & 0.26 & \textbf{0.59} \\
        \bottomrule \\
        \end{tabular}
    }
    \caption{\consistency, \representativity~and \fidelity~score for ResNet-50 models on ImageNet. Higher is better. The first and second best results are respectively in \textbf{bold} and \underline{underlined}.}
    \label{tab:metrics_imagenet}
\end{table}

Table~\ref{tab:metrics_imagenet} reports the \fidelity~(\muf), \consistency~(\reco) and \representativity~(\mege) scores obtained for the ResNet-50 predictors trained without degradation on ImageNet. 
We can exploit a selection criterion from the differences in scores.
First of all, we notice that the two methods obtaining a good fidelity score are RISE and Grad-CAM, they reflect well the predictor functioning. 
Their high fidelity score acts as a confidence bound on the \mege~and \reco~metrics: by correctly transcribing the functioning of the predictor, we obtain at the same time the \representativity~and the \consistency~of the explanations.
This score can then be used as a criterion to separate RISE from Grad-CAM. In view of the differences, RISE method seems preferable.

Concerning the \representativity~score, it is important to note that two methods tested here involve the element-wise product of the explanation with the input: Integrated Gradients and Gradient Input. This operation could eliminate the attribution score on a part of the image, thus reducing the distance between the two explanations. The result is an artificially better \mege~score which is in fact due to the dominance of input in the element-wise product.

It can be observed that the change of predictor has an effect on this ranking, and that a good method of explainability must be chosen according to a context: predictor and data set. However, even considering these effects, the experiments carried out suggest $3$ methods that give faithful, representative and consistent explanations: Grad-CAM, Grad-CAM++ and RISE (for more results on Cifar-10, EuroSAT and Fashion MNIST).

\paragraph{Towards predictors with better explanations.} In an attempt to find predictors that give better explanations, we extend the experience on the Cifar-10 dataset by adding a family of 1-Lipschitz networks. Indeed different works mention the Lipschitz constrained networks as particularly robust~\cite{usama2018robust, scaman2019lipschitz, pauli2020training, louislip} and have good generalizability. As a reminder, a $\f$ function is called $L$-Lipschitz, with $L \in \mathbb{R}^+$ if 
$| \f(\vx) - \f(\v{z}) | \leq L |\vx - \v{z}|$
For every pair $(\vx, \v{z}) \in \sx^2$. The smallest of these $L$ is called the Lipschitz constant of $\f$. This constant certifies that the input gradients ($\grad_{\vx} \f(\vx)$) of the function represented by the deep neural network are bounded and that this bound is known. This robustness certificate also comes with new generalization bounds that critically rely on the Lipschitz constant of the neural network~\cite{von2004distance, neyshabur2017exploring, bartlett2017spectrallynormalized}.

The predictors were trained using the Deel-Lip library~\cite{deelLip}. All the predictors, including the 1-Lipschitz, have comparable accuracy ($78 \pm 4\%$). To our knowledge, no previous work has made the link between Lipschitz networks and the chosen explainability methods.

\begin{figure}[ht]
    \centering
    \includegraphics[width=0.85\textwidth]{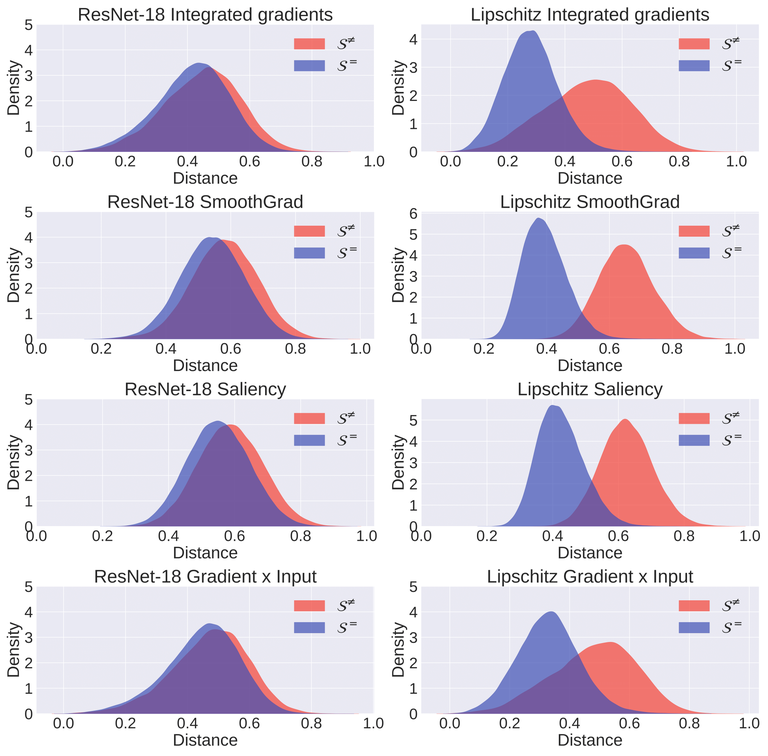}
    \caption{Lipschitz predictors (right column) on Cifar10.
    As explained in this paper, a clear separation between the $\Setm$ and $\Setp$ histograms is a sign of consistent explanations.
    }
    \label{fig:lipVsNormalDistrib}
\end{figure}

\begin{table}[ht]
    \centering
    \scalebox{1.0}{
        \begin{tabular}{l lllllll}
        \toprule
        \mege & IG & SG & SA & GI & GC & G+ & RI \\
        \midrule
        ResNet-18    & 0.58 & 0.46 & 0.45 & 0.55 & 0.72 & \textbf{0.83} & 0.57 \\
        1-Lipschitz  & \textbf{0.72} & \textbf{0.60} & \textbf{0.58} & \textbf{0.67} & \textbf{0.75} & 0.54 & \textbf{0.85} \\
        \bottomrule \\
        \end{tabular}
    }
    \caption{\mege~scores obtained by 1-Lipschitz models and ResNet-18 models on Cifar10. Higher is better. For almost all methods, the \representativity~of explanations increases significantly on 1-Lipschitz models.}
    \label{tab:lip_mege}
     \vspace{-1mm}
\end{table}

\begin{table}[ht]
    \centering
    \scalebox{1.0}{
        \begin{tabular}{l lllllll}
        \toprule
        \reco & IG & SG & SA & GI & GC & G+ & RI \\
        \midrule
        ResNet-18     & 0.11 & 0.15 & 0.15 & 0.09 & 0.64 & \textbf{0.49} & 0.52 \\
        1-Lipschitz   & \textbf{0.60} & \textbf{0.90} & \textbf{0.81} & \textbf{0.50} & \textbf{0.67} & 0.24 & \textbf{0.84} \\
        \bottomrule \\
        \end{tabular}
    }
    \caption{\reco~scores obtained by 1-Lipschitz models and ResNet-18 models on Cifar10. Higher is better. For almost all methods, the \consistency~of explanations increases significantly on 1-Lipschitz models.}
    \label{tab:lip_reco}
     \vspace{-1mm}
\end{table}

The Fig.~\ref{fig:lipVsNormalDistrib} shows the difference in $\Setp$ and $\Setm$ between ResNet and 1-Lipschitz predictors. In the left column, the results come from ResNet-18 predictors trained on Cifar-10 while the right column is dedicated to 1-Lipschitz predictors. We observe a clear improvement of the consistency and generalization of the explanations respectively as a result of better separation of the histograms and a smaller expectation of $\Setm$. SmoothGrad is the method that obtains the most consistent explanations as indicated in the table~\ref{tab:lip_reco}, in front of RISE and Saliency.

Concerning \mege, the results reported in Table~\ref{tab:lip_mege} show an improvement in the \representativity~of the explanations for the 1-Lipschitz predictors. Indeed, the \representativity~score has increased compared to the ResNet predictors for all tested methods, except Grad-CAM++.

Like \mege, the results in Table~\ref{tab:lip_reco} show an improvement for the 1-Lipschitz predictors in the \consistency~of the explanations for all the methods tested except for Grad-CAM++, reflecting the more marked separation between the two histograms of $\Setm$ and $\Setp$ in Fig.~\ref{fig:lipVsNormalDistrib}.

In general, the experiments carried out allow us to observe a clear improvement in the quality of explanations from the 1-Lipschitz predictor.
These encouraging results show that there is a close link between the methods used and predictor architectures, as well as the usefulness of Lipschitz networks for explainability.
Furthermore, it underlines the fact that the search for new methods is not the only path to explainability: the search for predictors with better explanations is another under-exploited avenue. 

\subsection{Conclusion}

We introduced a procedure to derive two new measures to characterize important properties of a good explanation: \representativity~and~\consistency~using Algorithmic Stability inspired procedure.
We highlight the fact that current \fidelity~metrics are intended to ensure that the explanations correctly reflect the underlying strategies of the model, regardless of whether these strategies are general or consistent.
We conducted several experimental sanity checks to ensure the proposed measures capture the notion of \representativity~and \consistency.
In addition, we showed that it is possible to use these measures as criteria for selecting an explanation method in conjunction with the fidelity metric.
Finally, as a case in point, we presented a novel analysis using 1-Lipschitz networks. We  used our measures to quantify the consistency of their explanations and showed that this class of networks gives much more stable and trustworthy explanations compared to standard neural networks. The code for reproducing and computing the proposed metrics is available in \xplique.

\clearpage

\section{Global sensitivity for Explainable AI with Sobol' indices}
\label{sec:attributions:sobol}
\newcommand{\sob}{\mathcal{S}}
\newcommand{\Vemp}{\hat{\text{V}}}
\newcommand{\perturbation}{\v{\pi}}
\newcommand{\estA}{\rm{A}}
\newcommand{\estB}{\rm{B}}
\newcommand{\estAB}{\rm{C}}

In this section, we address a challenge faced by black-box attribution methods by presenting a new approach grounded in Sensitivity Analysis and utilizing Sobol indices. These indices offer a streamlined method to not only model the individual contributions of different parts of an image but also to capture complex interactions among these parts and their impact on a neural network's prediction, as seen through variations in output.
Our method involves efficiently computing these indices for high-dimensional problems, such as those posed by images, by employing perturbation masks along with efficient estimators. This strategy effectively handles the large number of dimensions.
Crucially, we demonstrate that our proposed method achieves favorable performance on standard benchmarks for both vision and language models, while significantly reducing computational time compared to other black-box methods. Remarkably, it even surpasses the accuracy of state-of-the-art white-box methods, which rely on access to internal representations of the model.

We will begin by briefly reviewing relevant prior work, then introduce our method based on random perturbation and Sobol indices. We will propose an efficient estimator and finally, demonstrate through experiments that the Sobol attribution method not only outperforms previous methods in faithfulness but also offers faster computation and enables the discovery of intricate interactions within the model.

\begin{figure*}[ht]
    \centering
    \includegraphics[width=0.99\linewidth]{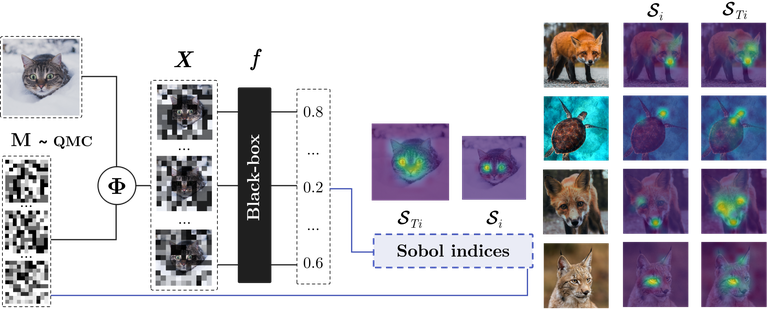}
    \caption{\textbf{(Left)} \textbf{Sobol Attribution Method overview.}
    Our method aims to explain the prediction of a black-box model for a given image. We first sample a set of real-valued masks $\rm{M}$ drawn from a Quasi-Monte Carlo (QMC) sequence.
    We apply these masks to the input image through a perturbation function $\perturbation(\cdot)$ (here the \textit{Inpainting} function) to form perturbed inputs $\rm{X}$ that we forward to the black box $\f$ to obtain prediction scores.
    Using the masks $\rm{M}$ and the associated prediction scores, we finally produce an explanation $\sob_{T_i}$ which characterizes the importance of each region by estimating the total order Sobol indices.
    While $\sob_{T_i}$ encompasses the effects of first and all higher-order non-linear interactions between pixel regions, we can also produce the first-order Sobol indices $\sob_i$ that reflect the importance of a region in isolation (e.g., the eyes of the cats).
    \textbf{(Right)} \textbf{Sample explanations for ResNet50V2.}
    Comparing explanations produced with $\sob_i$ and $\sob_{T_i}$ helps highlight the importance of individual image regions in isolation vs. jointly (e.g., the lynx tips are important but conditioned on the presence of the presence of an eye).
    }
    \vspace{-0.5cm}
    \label{fig:sobol}
\end{figure*}

\subsection{Background}
\label{sobol:sec:related_work}

\paragraph{Attribution methods for black-box models.} Most similar to our approach are attribution methods that can be used to explain the predictions of truly black-box models.
These methods probe a neural network's responses to perturbations over image regions and combine the resulting predictions into an influence score for each individual pixel or group of pixels.
The simplest method, ``Occlusion''~\cite{zeiler2014visualizing}, masks individual image regions -- one at a time -- with an occluding mask set to a baseline value and assigns the corresponding prediction scores to all pixels within the occluded region. Then the explanation is given by these prediction scores and can be easily interpreted.
However, occlusion fails to account for the joint (higher-order) interactions between multiple image regions. For instance, occluding two image regions -- one at a time -- may only decrease the model's prediction minimally (say a single eye or mouth component on a face)  while occluding these two  regions together may yield a substantial change in the model's prediction if these two regions interact non-linearly as is expected for a deep neural network.

This work, together with related methods such as LIME~\cite{ribeiro2016lime} and RISE~\cite{petsiuk2018rise}, addresses this problem by randomly perturbating the input image in multiple regions at a time. Obviously, perturbating multiple image locations simultaneously leads to a combinatorial explosion in the number of combinations and methods have been proposed to make these approaches more tractable.
For instance, a popular method, LIME~\cite{ribeiro2016lime}, takes superpixels as regions to perturbate instead of individual pixels.
An influence score is then computed for a set of connected pixel patches indicating how strongly a patch is correlated to the model predictions.

RISE~\cite{petsiuk2018rise} relies on Monte Carlo sampling to generate a set of binary masks, each value in the masks representing a pixel region.
By probing the model with  randomly masked versions of the input, RISE~\cite{petsiuk2018rise} produces a importance map by considering the average of the masks weighted by their associated prediction scores.
Instead of using binary masks, our method considers a continuous range of perturbations which allows for a finer exploration of the model's response.
Our method can still use the same perturbations as used in Occlusion~\cite{zeiler2014visualizing}, LIME~\cite{ribeiro2016lime} and RISE~\cite{petsiuk2018rise}, but it also enables the use of more advanced perturbation functions that take continuous inputs.

More importantly, the aforementioned methods lack a rigorous framework. Here, we introduce a theoretical framework that decomposes the influence score of each individual region between multiple orders of influence.
The first-order approximates Occlusion~\cite{zeiler2014visualizing} by considering the influence of one region at a time, while the second-order considers two regions at a time, etc. The decomposition also includes higher-orders.

\paragraph{Variance-based sensitivity analysis.} Our attribution method builds on the variance-based sensitivity analysis framework.
The approach was introduced in the 70s~\cite{cukier1973study} and reached a cornerstone with the Sobol indices~\cite{sobol1993sensitivity}.
Sobol indices are currently used in many fields (including those that are said to be safety-critical), especially for the analysis of physical phenomena~\cite{iooss2015}. More recently, connections have been successfully made between these indices and existing metrics of fairness~\cite{benesse2021fairness}.

They are used to identify the input dimensions that have the highest influence on the output of a model or a mathematical system.
Several statistical estimators to compute these indices are available~\cite{saltelli2010variance, marrel2009calculations, janon2014asymptotic, owen2013better, tarantola2006random} and have asymptotic guarantees~\cite{janon2014asymptotic, da2013efficient, tissot2012bias}.
We build on this literature by adapting these Sobol indices in the context of black-box models to compute the influence of regions of an image on the output predictions using perturbation masks.

\subsection{Sobol attribution method}
\label{sobol:sec:method}

In this work, we formulate the feature attribution problem as quantifying the contribution of a collection of $d$ real-valued variables $\v{x} = (x_1, ..., x_d)$  with respect to a model decision. Specifically, we consider a black-box decision function $\f : \sx \to \sy$ whose internal states and analytical form are unknown (for instance, $\f$ can score the probability for the input to belong to a specific class). Our goal is to quantify the importance of each feature to the decision score $\f(\vx)$, not just individually but also collectively. To capture these higher-order interactions, our method consists in estimating the Sobol indices of the variables $\vx$ by randomly perturbating them and evaluating the impact of these perturbations on the prediction of the black-box model (Fig.~\ref{fig:sobol}).

Considering variations of $\f(\vx)$ in response to meaningful perturbations of the input $\vx$ is a natural way to interpret the local behavior of the decision function around $\vx$. Several methods build on this idea, e.g., by removing one or a group of input variables~\cite{zeiler2014visualizing, ribeiro2016lime, fong2017perturbation, petsiuk2018rise, fong2019extremal} or by back-propagating the gradient to the input space through the model~\cite{simonyan2014deep, sundararajan2017axiomatic, smilkov2017smoothgrad, Selvaraju_2019}. Most of these methods use the model's internal representations and/or require computing the gradient w.r.t. the input, which makes them unusable in a black-box setting. Moreover, these methods focus on estimating the intrinsic contribution of each feature, neglecting the combinatorial components. Our method applies perturbations directly on the input in order to deal with a black-box scenario, and allows us to estimate higher-order interactions between the variables.

\paragraph{Random Perturbation}
\label{sobol:sec:perturbation_masks}

Formally, let us define a probability space $(\Omega,\sx,\P)$ of possible input perturbations of $\vx$
There are several ways to define random perturbations corresponding to different coverage of the data manifold around $\vx$. For instance, we can consider the perturbation mask operator $\perturbation: \sx \times \s{M} \to \sx$ which combines a stochastic mask $\rv{m} = (\r{m}_1, ..., \r{m}_d) \in \s{M}$ (i.e., an i.i.d sequence of real-valued random variables on $[0, 1]^d$) with the original input $\vx$. This formulation encompasses \textit{Inpainting} perturbations: $\perturbation(\vx, \rv{m}) = \vx \odot \rv{m} + (\v{1} - \rv{m}) \mu$ with $\mu \in \mathbb{R}$ a baseline value, and $\odot$ the Hadamard product. This consists in linearly varying the pixel intensities towards a baseline intensity such as a pure black with a value of zero~\cite{fong2017perturbation,ribeiro2016lime,zeiler2014visualizing,petsiuk2018rise}. Similarly, \textit{Blurring} consists of applying a blur operator with various intensities to certain regions of the image~\cite{fong2017perturbation}.
Different perturbation domains can be considered for other types of data such as textual or tabular data that we discuss further in the experimental section. In the next section, we explain how we adapt the Sobol-based sensitivity analysis using a class of perturbations to explain the predictions of a black-box model.

\paragraph{Sensitivity analysis using Sobol indices}
\label{sec:sobol_indices}

We first briefly review the classical Sobol-Hoeffding decomposition from~\cite{hoeffding1948} and introduce the Sobol indices. Let $(\r{x}_1,...,\r{x}_d)$ be independent variables and assume that $\f$ belongs to $\mathbb{L}^2(\sx,\P)$. Moreover we denote the set $\mathcal{U} =\{1, ..., d\}$, $\v{u}$ a subset of $\s{U}$, its complementary ${\complementary}\v{u}$ and $\E(\cdot)$ the expectation over the perturbation space. The Hoeffding decomposition allows us to express the function $\f$ into summands of increasing dimension, denoting $\f_{\v{u}}$ the partial contribution of variables $\rvx_{\v{u}} = (\r{x}_i)_{i \in \v{u}}$ to the score $\f(\rvx)$:

\begin{equation}
    \label{eq:anova}
    \begin{aligned}
    \f(\rvx) &= \f_{\emptyset} + \sum_i^d \f_i(\rx_i)
    + \sum_{1 \leqslant i < j \leqslant d} \f_{i,j}(\rx_i, \rx_j)
    + \cdots + \f_{1,...,d}(\rx_1, ..., \rx_d) \\
    &= \sum_{\substack{\v{u} \subseteq \mathcal{U}}} \f_{\v{u}}(\rvx_{\v{u}})
    \end{aligned}
\end{equation}

Eq.~\ref{eq:anova} consists of $2^d$ terms and is unique under the following orthogonality constraint:

\begin{equation}
    \label{eq:anova_ortho}
    \begin{aligned}
    \forall (\v{u},\v{v}) \subseteq \mathcal{U}^2 \; s.t. \;  \v{u} \neq \v{v}, \;\; \E\big(\f_{\v{u}}(\rvx_{\v{u}}) \f_{\v{v}}(\rvx_{\v{v}})\big) = 0
    \end{aligned}
\end{equation}

Furthermore, orthogonality yields the characterization $\f_{\v{u}}(\rvx) = \E(\f(\rvx)|\rvx_{\v{u}}) - \sum_{\bm{v}\subset \v{u}}\f_{\v{v}}(\rvx)$ and allows us to decompose the model variance as:
\begin{equation}
    \label{eq:var_decomposition}
    \begin{aligned}
        \V(\f(\rvx)) &= \sum_i^d \V(\f_i(\rx_i)) +
        \sum_{1 \leqslant i < j \leqslant d} \V(\f_{i,j}(\rx_i, \rx_j)) +
        ... + \V(\f_{1,...,d}(\rx_1, ..., \rx_d)) \\
        &=\sum_{\substack{\rv{u} \subseteq \mathcal{U}}} \V(\f_{\v{u}}(\rvx_{\v{u}}))
        \end{aligned}
\end{equation}
Building from Eq.~\ref{eq:var_decomposition}, it is natural to characterize the influence of any input subset $\v{u}$ as its own variance w.r.t. the total variance. This yields, after normalization by $\V(\f(\rvx))$, the general definition of Sobol indices.
\begin{definition}[Sobol indices~\cite{sobol1993sensitivity}]
\label{def:sobol_indice}
The sensitivity index $\sob_{\v{u}}$ which measures the contribution of the variable set $\rvx_{\v{u}}$ to the model response $\f(\rvx)$ in terms of fluctuation is given by:
\begin{equation}
    \label{eq:sobol_indice}
    \sob_{\v{u}}  = \frac{ \V(\f_{\v{u}}(\rvx_{\v{u}})) }{ \V(\f(\rvx)) }
    = \frac{ \V(\E(\f(\rvx) | \rvx_{\v{u}})) - \sum_{\bm{v}\subset \v{u}}\V(\E(\f(\rvx) | \rvx_{\bm{v}} ))}{ \V(\f(\rvx)) }
\end{equation}
\end{definition}
Sobol indices give a quantification of the importance of any subset of features with respect to the model decision, in the form of a normalized measure of the model output deviation from $\f(\rvx)$. Thus, Sobol indices sum to one : $\sum_{\v{u} \subseteq \mathcal{U}} \sob_{\v{u}} = 1$.

For each subset of variables $\rvx_{\v{u}}$, the associated Sobol index $\sob_{\v{u}}$ describes the proportion of the model's output variance explained by this subset. In particular, the first-order Sobol indices $\sob_i$ capture the intrinsic share of total variance explained by a particular variable, without taking into account its interactions.
Many attribution methods construct such intrinsic importance estimator. However, the framework of Sobol indices enables us to capture higher-order interactions between features. In this view, we define the Total Sobol indices.
\begin{definition}[Total Sobol indices~\cite{homma1996importance}]
\label{def:total_sobol_indice}
The total Sobol index $\sob_{T_i}$ which measures the contribution of the variable $\rx_i$ as well as its interactions of any order with any other input variables to the model output variance  is given by:
\begin{equation}
    \label{eq:sobol_total}
    \sob_{T_i}
    = \sum_{\substack{\v{u} \subseteq \mathcal{U} \\ i \in \v{u} }} \sob_{\v{u} }
    = 1 - \frac{\V_{ \rvx_{\complementary i} }(\E_{ \rx_i }(\f(\rvx) | \rvx_{\complementary i})) }{ \V(\f(\rvx))}
    = \frac{ \E_{\rvx_{\complementary i}}( \V_{\rx_i} ( \f(\rvx) | \rvx_{\complementary i} )) }{ \V(\f(\rvx)) }
\end{equation}
\end{definition}
Where $\E_{\rvx{\complementary i}}( \V_{\rx_i} ( \f(\rvx) | \rvx_{\complementary i}))$ is the expected variance that would be left if all variables but $\rx_i$ were to be fixed. $\sob_{T_i}$ is the sum of the Sobol indices for the all the possible groups of variables where $i$ appears, i.e. first and higher order interactions of variable $\rx_i$.

Since the total interaction index contains the first order index, it is natural that it is greater than or equal to the first order index. We thus note the property which can easily be deduced: $\forall i, 0 \leq \sob_i \leq \sob_{T_i} \leq 1$. We remind that naturally the score is bounded between 0 and 1 as it represents a (relative) part of the model's variance.
We will now see why these two indices and the difference between them make them relevant for the explainability of a black-box model.

These statistics quantify the intrinsic (first-order indices) and relational (total indices) impact of each variable to the model output.
A variable with a low total Sobol index is therefore not important to explain the model decision. Also, a variable has a weak interaction with other variables when $\sob_{T_i} \approx \sob_i$, while it has a strong interaction when the difference between its two indices is high $\sob_{T_i} \gg \sob_i$. A strong interaction means that the effect of one variable on the variation of the model output depends on other variables.
Thus, using Sobol indices allows to describe fine grained interactions between inputs which leads to the model decision.
We next present an efficient method to estimate these indices.

\begin{figure*}[ht]
  \centering
  \includegraphics[width=0.85\textwidth]{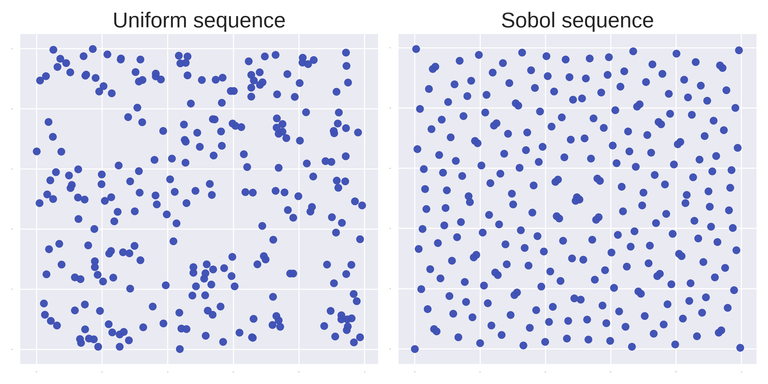}
  \caption{\textbf{Quasi-Monte Carlo vs. Crude Monte Carlo Sampling.} Comparison of Sampling Techniques in a Two-Dimensional Unit Cube: On the left, we observe a crude Monte Carlo sampling where points are distributed according to a uniform random sequence, exhibiting clusters and gaps due to the stochastic nature of the sampling process. On the right, a quasi-Monte Carlo sampling using a Sobol sequence is displayed, demonstrating a more uniform and structured distribution of points across the space, which is characteristic of low-discrepancy sequences that aim for even coverage in the unit cube $[0,1]^2$. This illustrates the advantage of quasi-Monte Carlo methods in achieving a more evenly distributed set of sample points, which can lead to more efficient numerical integration and optimization within multidimensional domains.}
  \label{fig:sobol:qmc}
\end{figure*}

\paragraph{Efficient estimator.} As models are becoming more and more complex, the proposed estimator must take into account the computational cost of model evaluation. Many efficient estimators have been proposed in the literature~\cite{iooss2015}. In this work, we use the Jansen~\cite{jansen1999} estimator which is often considered as one of the most efficient~\cite{puy2020comprehensive}.
Jansen is typically used with a Monte Carlo sampling strategy.
We improve over Monte Carlo by using a Quasi-Monte Carlo (QMC) sampling strategy which generates low-discrepancy sample sequences allowing a faster and more stable convergence rate~\cite{gerber2015}, see Figure~\ref{fig:sobol:qmc}.
Interestingly, QMC samling allow us to add posteriori points to refine the result.
We will now describe the procedure to implement these estimators.

We start by drawing two independent matrices of size $N \times d$ of $N$ perturbation masks from a Sobol low discrepancy $LP_{\tau}$ sequences.
$N$ will be our number of designs and we recall that $d$ is our dimensions (e.g, $d=121$ for $11$ by $11$ mask).
Once the perturbation operator is applied to our input $\perturbation(\vx, \rm{M})$ with these masks, we obtain two matrices $\estA$ and $\estB$ of the same size as the perturbed inputs (i.e., partially masked images). We note $\estA_{ji}$ and $\estB_{ji}$ the elements of the matrices such that $i = 1, ..., d$ the number of variables studied and $j = 1, ..., N$ the number of samples in each matrix.
We form the new matrix $\estAB^{(i)}$ in the same way as $\estA$ except for the fact that the column corresponding to the variable $i$ is now replaced by the column of $\estB$.
We denote $f_{\emptyset} = \frac{1}{N} \sum_{j=0}^N \f(\estA_j)$ and the empirical variance $\hat{V} = \frac{1}{N-1} \sum_{j=0}^N (\f(\estA_j) - f_{\emptyset})^2 $. The empirical estimators for first ($\hat{\sob}_i$) and total order ($\hat{\sob}_{T_i}$) can be formulated as:
\vspace{-1mm}
\begin{equation}
\label{eq:jansen_estimator}
        \hat{\sob}_i = \frac
        { \Vemp - \frac{1}{2N} \sum_{j=1}^N (\f(\estB_j) - \f(\estAB_j^{(i)}))^2 }
        { \Vemp }
        \text{ }\text{ }\text{ }\text{ }\text{ }\text{ }\text{ }\text{ }
        \hat{\sob}_{T_i} = \frac
        { \frac{1}{2N} \sum_{j=1}^N ( \f(\estA_j) - \f(\estAB_j^{(i)}) )^2  }
        { \Vemp } \\
\end{equation}
Hence, to compute the set of first order and total indices, it is necessary to perform $N(d+2)$ forwards of the model. We study in section \ref{sec:sobol:efficient} how to choose a sufficient number of forwards ($N$).
To ease understanding and demonstrate that these estimators can be easily implemented, we show in Algorithm~\ref{alg:sobol:sti} a minimal pythonic implementation of the total order estimator that outputs $\hat{\sob}_{T_i}$ indices. The input $Y$ contains the prediction scores of the $N \times (d+2)$ forwards. The scores are ordered following the same QMC sampling ordering of their associated mask. The output \texttt{STis} contains $d$ importance scores, one for each dimension of the mask. In the case of images, we obtain our final explanation map by applying a bilinear upsampling to match the dimensions of the input image.

\begin{figure}
\begin{lstlisting}
def total_order_estimator(Y, N=32, d=11*11):
    fA, fB = Y[:N], Y[N:N*2]
    fC = [Y[N*2+N*i:N*2+N*(i+1)] for i in range(d)]
    f0 = mean(fA)
    V = sum([(val - f0)**2 for val in fA]) / (len(fA) - 1)
    STis = [sum((fA - fC[i])**2) / (2 * N) / V for i in range(d)]
    return STis
\end{lstlisting}
\caption{\textbf{Pythonic implementation of the estimator.} We just need to have access to the output of the model for the corresponding $\rm{A},\rm{B}$ and $\rm{C}$ matrix. Using only the output, we can efficiently estimate the total Sobol indices $\sob_T$.}
\label{alg:sobol:sti}
\end{figure}

\paragraph{Signed estimator}
Although the proposed Sobol-based attribution method allows us to determine the impact of any variables for a given prediction and thus to identify diagnostic ones, it lacks the ability to highlight the type of contributions made, whether positive or negative. Simple methods such as ``Occlusion'' typically include this information. Hence, we propose a variant that combines the importance scores of the total Sobol indices with the sign of the occlusion. We compute the difference in score between the prediction on the original input $\vx$ and a partial version $\vx_{[x_i = 0]}$ with the variable $x_i$ occluded. Intuitively, this provides an estimate of the direction of the variations generated by the variables studied with respect to a reference state.
\vspace{-3mm}\begin{equation}
    \label{eq:sobol_signed}
    \hat{\sob}_{T_i}^{\Delta} = \hat{\sob}_{T_i} \times \text{sign}( \f(\vx) - \f(\vx_{[x_i = 0]})
\end{equation}

\subsection{Experiments}

To evaluate the benefits and the reliability of the Sobol attribution method, we performed multiple systematic experiments on vision and natural language models using common explainability metrics.

For our vision experiments, we compared the plausibility of the explanations produced on the Pointing Game~\cite{zhang2018top} benchmark. We evaluate the fidelity of our explanations using the Deletion metric for $4$ representative models commonly used in explainability studies: ResNet50V2~\cite{he2016deep} , VGG16~\cite{simonyan2014deep}, EfficientNet~\cite{tan2019efficientnet} and MobileNetV2~\cite{sandler2018mobilenetv2} trained on  ILSVRC-2012~\cite{imagenet_cvpr09}.
In addition, we also compared the speed of convergence of the proposed estimator with that of the leading approach, RISE~\cite{petsiuk2018rise}, on the same models.
For our NLP experiments, we fine-tuned a Bert model and trained a bi-LSTM on the IMDB sentiment analysis dataset~\cite{maas2011} before comparing  fidelity scores using word-deletion for representative methods.

Throughout this work, explanations were generated using the Sobol total estimator $\hat{\sob}_{T_i}$ on the target class output.
In the supplementary material, we demonstrate the effectiveness of modeling higher-order interactions between image regions by comparing $\hat{\sob}_{T_i}$ against $\hat{\sob}_{i}$ which only models the main effects.
For the experiments involving images, the masks were generated at a resolution of $d' = 11 \times 11$ pixels, then upsampled with a nearest-neighbor interpolation method before being applied with the \textit{Inpainting} perturbation function. Finally,  $N$ was set to $32$ which is equivalent to $3,936$ forward passes, half the number of forward used by RISE (see Section~\ref{sec:sobol:efficient} for details).
For $\smash{\hat{\sob}_{T_i}^\Delta}$, an occlusion using the same resolution as the masks was used to sign $\smash{\hat{\sob}_{T_i}}$, with zero as baseline.
For RISE~\cite{petsiuk2018rise}, we have followed the recommendations of the original paper with $8,000$ forward passes for all models.

\begin{table*}[t]
\centering
\begin{tabular}{c lccc}
\toprule
 & & Pointing Game & Deletion & \textit{Time (s)} \\
\midrule
& Baseline Center &  27.8 & 0.235 & - \\
\midrule
\multirow{7}{*}{\rotatebox[origin=c]{90}{{\footnotesize White box}}}
& Saliency~\cite{simonyan2014deep} & 37.7 & 0.174 & 0.031 \\
& Guided-Backprop.~\cite{springenberg2014striving} & 39.1 & 0.142 & 0.051 \\
& MWP~\cite{zhang2018top}  &  39.8 & - & 0.039 \\
& cMWP~\cite{zhang2018top} &  49.7 & - & 0.040 \\
& Integ.-Grad. ~\cite{shrikumar2017learning} &  49.7 & \underline{0.123} & 0.040 \\
& GradCAM~\cite{Selvaraju_2019} & \underline{54.2} & 0.141 & 0.015 \\
& ExtremalPerturbation~\cite{fong2019extremal} & 51.5 & - & 26.48\\
\midrule
\multirow{3}{*}{\rotatebox[origin=c]{90}{{\footnotesize Black box}}}
& Occlusion & 35.6 & 0.350 & 1.134 \\
& RISE~\cite{petsiuk2018rise} & 50.8 & 0.127 & 13.19 \\
& Sobol ($\hat{\sob}_{T_i}$) (ours) & \textbf{54.6} & \textbf{0.121} & 6.381 \\
\bottomrule
\end{tabular}
\caption{\textbf{Pointing game.} Accuracy over the full test set and a subset of difficult images (defined in~\cite{zhang2018top}).
The first and second best results are  \textbf{bolded} and \underline{underlined}.
Results are based on PyTorch re-implementations using the TorchRay package.
The reported execution time is an average over 100 runs on ResNet50 using an Nvidia Tesla P100 on Google Colab and a batch size of 64. Lower execution time can be reached with higher batch size. \vspace{-3mm}
}\label{tab:pointing_game}
\end{table*}

\subsubsection{Pointing game}

Different evaluation methods have been proposed to compare attribution methods and their explanations~\cite{samek2016evaluating, hooker2018benchmark, aggregating2020, fel2020representativity}.
The first common approach consists in measuring the plausibility of an explanation as the correlation between attribution maps and human-provided semantic annotations. Here, we focused on the Pointing Game used in~\cite{zhang2018top, fong2017perturbation, fong2019extremal, petsiuk2018rise}. For each attribution method, we compute a contribution score for each pixel of a given class of objects, e.g., bike or car.
We then calculated the percentage of times the pixel with the highest score is included in the bounding box surrounding the object of interest. In this benchmark, a good attribution method should point to the most important evidence of the object appearance in accordance with a human user.

In Table~\ref{tab:pointing_game}, a report results for the Pascal VOC~\cite{everingham2010pascal} and MS COCO~\cite{coco} datasets using VGG16~\cite{simonyan2014deep} and ResNet50\cite{he2016deep}.
In the last column we report the computation times for each method averaged over $100$ MS COCO samples for the ResNet50 model.
We subdivided explanation methods into two categories: white-box methods which require the use of backpropagation, such as Gradient~\cite{zeiler2014visualizing} and Extremal Perturbation~\cite{fong2019extremal}, and/or access to the internal states of the model, such as GradCAM~\cite{Selvaraju_2019} versus black-box methods such as Occlusion~\cite{zeiler2014visualizing}, RISE~\cite{petsiuk2018rise}, or the proposed Sobol method $\sob_{T_i}$ which only require the final model predictions. The proposed method outperforms RISE~\cite{petsiuk2018rise} on all of the tested cases, while reducing the number of forward passes by half.
Surprisingly, white-box methods do not always lead to higher scores, and indeed $\hat{\sob}_{T_i}^\Delta$ is the leading method for Pascal VOC / VGG16 and our two estimators $\hat{\sob}_{T_i}^\Delta, \hat{\sob}_{T_i}$ prevail on COCO / VGG16.
Also note that our signed version of the estimator obtains higher scores overall.
This might be due to the fact that images from VOC and COCO often feature several types of objects. Thus, the maximum variance in the output is not always induced by the object of interest but can be due to the masking of another object in the image. This result suggests that our signed version $\hat{\sob}_{T_i}^\Delta$ should be used on multi-label datasets, while $\hat{\sob}_{T_i}$ should be used on multi-class datasets. We indeed confirm this in the next set of experiments on a multi-class dataset.

\subsubsection{Fidelity}
There is a broad consensus that measuring the plausibility of an explanation alone is insufficient~\cite{adebayo2018sanity, ghorbani2017interpretation}. Indeed, if an explanation is used to make a critical decision, users expect an explanation to reflect the true underlying decision process of the model and not just a consensus with humans. Failures to do so could have disastrous consequences.
A first major limitation of current evaluation methods based on human-provided groundtruth such as the pointing game is that they do not work when a model prediction is wrong. In this case, an explanation method can be penalized for not pointing to the correct evidence even though explaining prediction errors is a critical use case for explanation methods. Another limitation of these evaluation methods is that they make the implicit assumption that the models should be relying on the same image regions than humans for recognition~\cite{ullman2016atoms, linsley2018learning}, which is likely to be an incorrect assumption. We thus use the fidelity metric as a complementary type of evaluation. This metric assumes that the more faithful an explanation is, the quicker the prediction score should drop when pixels that are considered important are reset to a baseline value (e.g., gray values).

In Table~\ref{tab:deletion}, we report results for the Deletion Metric~\cite{petsiuk2018rise} (or $1 - AOPC$~\cite{samek2016evaluating}) for 4 different pre-trained models: ResNet50~\cite{he2016deep} , VGG16~\cite{simonyan2014deep}, EfficientNet~\cite{tan2019efficientnet} and MobileNet~\cite{sandler2018mobilenetv2} on 2,000 images sampled from the ImageNet validation set. TensorFlow~\cite{tensorflow2015} and the Keras~\cite{chollet2015keras} API were used to run the models.
Several baseline values can be used~\cite{sturmfels2020visualizing}, but we chose the standard approach with gray values.
We observe that the proposed Sobol $\hat{\sob}{T_i}$ is the most faithful black-box methods with the lowest deletion scores across all models.
Overall $\hat{\sob}_{T_i}$ is able to match the scores of the most faithful white-box method, namely Integrated Gradients~\cite{sundararajan2017axiomatic}, and gets the lowest score on ResNet50V2 with 0.121 against 0.123 (lower is better).
We also report that our signed version $\hat{\sob}_{T_i}^\Delta$ is less faithful that the standard Sobol $\hat{\sob}{T_i}$. This can be explained by the fact that ImageNet images contains only one object and therefore the main variance area generally coincides with the class to be explained.
This confirms our observation on the previous pointing game benchmark that $\hat{\sob}{T_i}$ should be preferred in  a multi-class setup and $\hat{\sob}_{T_i}^\Delta$ in a multi-label setup.

\begin{table*}[t]
\vspace{10mm}
\centering
\scalebox{0.85}{\begin{tabular}{c lcccc}
\toprule
 & Method & \textit{ResNet50V2} & \textit{VGG16} & \textit{EfficientNet} & \textit{MobileNetV2} \\
\midrule
& Baseline Random (ours) & 0.235 & 0.168 & 0.124 & 0.137 \\
\midrule
\multirow{7}{*}{\rotatebox[origin=c]{90}{{\footnotesize White box}}}
& Saliency~\cite{simonyan2014deep} & 0.174 & 0.134 & 0.105 & 0.125 \\
& Guided-Backprop.~\cite{springenberg2014striving} & 0.142 & 0.138 & 0.105 & 0.102 \\
& DeconvNet~\cite{zeiler2014visualizing} & 0.159 & 0.146 & 0.105 & 0.111 \\
& Grad.-Input~\cite{shrikumar2017learning} & 0.140 & \underline{0.096} & \underline{0.093} & 0.103 \\
& Integ.-Grad.~\cite{sundararajan2017axiomatic} & \textbf{0.123} & \textbf{0.095} & \textbf{0.091} & \textbf{0.093} \\
& SmoothGrad~\cite{smilkov2017smoothgrad} & \underline{0.130} & 0.106 & 0.094 & \underline{0.098} \\
& GradCAM~\cite{Selvaraju_2019} & 0.141 & 0.118 & 0.130 & 0.122 \\

\midrule
\multirow{4}{*}{\rotatebox[origin=c]{90}{{\footnotesize Black box}}}
& Occlusion~\cite{zeiler2014visualizing} & 0.350 & 0.357 & 0.252 & 0.357 \\
& RISE~\cite{petsiuk2018rise} & \underline{0.127} & 0.121 & \underline{0.119} & \underline{0.114} \\
& Sobol ($\hat{\sob}_{T_i}$) (ours) & \textbf{0.121} & \textbf{0.109} & \textbf{0.104} & \textbf{0.107} \\
& Sobol signed ($\hat{\sob}^{\Delta}_{T_i}$) (ours) & 0.145 & \underline{0.114} & 0.147 & 0.141 \\
\bottomrule
\end{tabular}}
\vspace{0mm}\caption{\textbf{Deletion} scores obtained on 2,000 ImageNet validation set images. Lower is better.
Random consists in removing  pixels at each step at random.
The first and second best results are \textbf{bolded} and \underline{underlined}. \vspace{-0.5cm}
}\label{tab:deletion}
\end{table*}
Another metric called Insertion has been proposed by the authors of RISE~\cite{petsiuk2018rise}.
Instead of deleting pixels in the original image like with Deletion, Insertion consists in adding pixels on a baseline image, e.g. one gray image, starting with pixels that are associated with the highest importance scores for a given explanation method.
An issue with Insertion is that the score computed along the insertion path is highly influenced by the first inserted pixels which contributes disproportionately. A good score on this metric therefore requires exploring a region very far from the original image and closer to the baseline. For this reason, we rather preferred to focus our study on Deletion than Insertion. However, we also report results on Insertion in the supplementary material using the same hyperparameters as used in Deletion.

\subsubsection{Efficiency}
\label{sec:sobol:efficient}

The black-box methods presented so far compete with white-box methods that do not require access to the internal representation of the model at the cost of a large number of forward passes, e.g., around $8,000$ for RISE~\cite{petsiuk2018rise}. This weakness leads us to take a more serious look at the performance of the proposed method.
It seems critical for the deployment of black-box methods to lower the amount of compute required to produce correct explanations.
We describe an experiment to show that beyond producing higher quality explanations, our estimator converges quickly.
We first generate an explanation with a high number of forward passes that is large enough to reach convergence, e.g. $10,000$ forward passes. Then we compare this explanation that ``converged'' to other explanations obtained with lower numbers of forward passes. It allows us to measure the stability and rate of convergence towards this explanation that ``converged'', but more practically to find the proper trade-off between the amount of compute and the quality of explanations.
This procedure requires defining a measure of similarity between two explanations.
Since the proper interpretation method is to rank the features most sensitive to the model's decision, it seems natural to consider the  Spearman rank correlation~\cite{spearman1904measure} to compare the similarity between explanations (see the Appendix \autoref{ap:mege:distances}). Moreover, prior work has provided theoretical and experimental arguments in line with this choice~\cite{ ghorbani2017interpretation, adebayo2018sanity, tomsett2019sanity}.

\begin{figure*}
  \centering
    \includegraphics[width=0.98\linewidth]{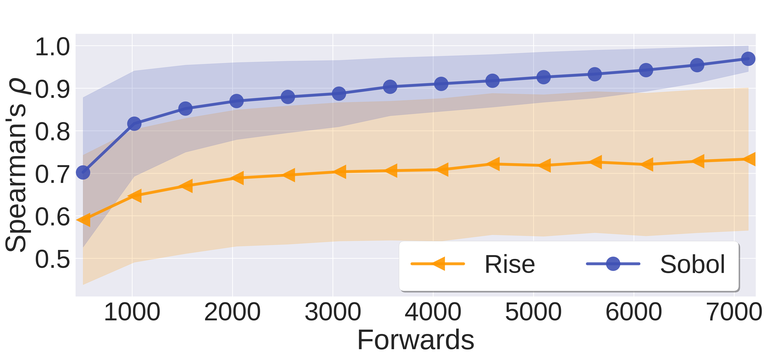}
  \caption{
  Comparison between the rate and stability of convergence of Sobol $\hat{\sob}_{T_i}$ and RISE~\cite{petsiuk2018rise}.
  A high Spearman correlation rank corresponds to producing an explanation that is similar to the explanation that ``converged'', i.e. with $10,000$ forward passes.
  We report mean and variance computed over $500$ images from the ImageNet dataset using EfficientNet.
  } \label{fig:convergence_efficient}
\end{figure*}

In Fig.~\ref{fig:convergence_efficient}, we compare the proposed Sobol attribution method $\hat{\sob}_{T_i}$ against RISE~\cite{petsiuk2018rise}, which is the current state-of-the-art for black-box methods.
We use their respective gold explanation generated after $10,000$ forwards. To allow a fair comparison, both methods use masks generated in $7\times7$ dimensions (as recommended by RISE~\cite{petsiuk2018rise}).
We report the average results and variance over $500$ images from the ImageNet validation set using EfficientNet, a convolutional neural network optimized for fast forward computing times.
We observe that our method exhibit higher convergence rate by getting higher Spearman's rank correlation of 0.8 after only 1,000 forwards against 0.65 for RISE, and consistently obtain higher scores until reaching 0.97 with 7,000 forwards against 0.73 for RISE. Additionally, we observe that Sobol has a more stable convergence by getting an overall lower variance than RISE. This implies that the number of forward passes used in Sobol can be greatly reduced to accommodate computational resources constraints compared to RISE. Indeed, RISE proposes to use $8,000$ forwards, but our method is faster and reaches better results with half the number of passes.
Finally, we perform an ablation study of Sobol to show the impact of lowering the number of forwards on the Deletion benchmark. We report competitive scores with 16 times fewer number of forwards than RISE by reaching 0.151 in Deletion score with 492 forwards. For reference, Sobol was reaching state-of-the-art results of 0.121 with 3,936 forwards.

\paragraph{Word deletion}

\begin{table*}[t]
\centering
\scalebox{0.95}{\begin{tabular}{l ccccccc}
\toprule
 & Saliency & Grad-Input & SmoothGrad & Integ-Grad & Occlusion & $\hat{\sob}_{T_i}$ & $\hat{\sob}^{\Delta}_{T_i}$  \\
\midrule
BERT & 0.684 & 0.682 & 0.682 & 0.689&  \textbf{0.531} & 0.662& \underline{0.598}   \\
LSTM & 0.541 & 0.529 & 0.541 & 0.538 & \textbf{0.440}& 0.523 & \underline{0.461} \\
\bottomrule
\end{tabular}}
\caption{\textbf{Word deletion} scores, obtained on 1,000 sentences. Delete up to 20 words per sentence accordingly to their relevance and track the impact on the classification performance. Lower is better.
The first and second best results are \textbf{bolded} and \underline{underlined}. \vspace{-0.2cm}
}\label{tab:word_deletion}
\vspace{-0.5cm}
\end{table*}

For NLP,  black-box methods require the use of perturbations that can be applied to the space of characters, words or sentences. For instance, a common perturbation consists in simply removing one word of the sentence to be explained. Therefore, the \textit{Inpainting} perturbation that we used with Sobol to reduce the intensity of pixels in a continuous manner cannot be directly applied in this context. Instead, we adapt it by binarizing the masks such that $\perturbation(\vx, \rm{M}) = \vx \odot \lceil\rm{M} - 0.5\rceil $, i.e., if the value is greater than 0.5 the word is kept, otherwise it is removed. We then verify that our Sobol method can be used to identify words that support a specific decision of a text classifier. Inspired by previous work~\cite{arras2017relevant, arras2017explaining, bach2015pixel}, we introduce an experimental benchmark on the IMDB Review dataset~\cite{imdb2011}. It is similar to the previous Deletion benchmark for images in that it focuses on assessing the faithfulness of the explanation and does not require specific human annotations.
More precisely, we first trained two models: a bi-LSTM from scratch and a BERT model fine-tuned for the task. We generate explanations on $1,000$ sentences from the validation dataset. An explanation associates an importance score to each word. Similar to Deletion, we use these scores to successively remove the most relevant word of the sentence and measure the corresponding drop in the prediction score of the model.

In Table~\ref{tab:word_deletion}, we report results for explanation methods that are commonly used in NLP.
Both our two Sobol methods have a better faithfulness than all the tested gradient-based white-box methods including Saliency, Grad-Input, SmoothGrad, and Integr-Grad. For instance, Sobol $\hat{\sob}^{\Delta}_{T_i}$ even reaches a low Word deletion scores of 0.461 (lower is better) for the bi-LSTM compared to 0.529 for the best white-box approach.
However, the proposed methods are only the second and third most faithful methods. Occlusion (often called Omit-1 in NLP) reaches the lowest score of 0.440 for LSTM and 0.531 for BERT, against 0.461 and 0.598 for Sobol $\hat{\sob}^{\Delta}_{T_i}$. This is due to the fact that the default distribution of masks used in Sobol is centered around $0.5$, which corresponds to removing on average half of the words as opposed to a single word for Occlusion.
In IMDB, this causes the frequent removal of critical words that support the model decision.
Indeed, we report comparable results with Occlusion (e.g., $0.527$ for BERT) for a lower threshold of $0.05$ to remove far fewer words. Since Sobol can model higher-order interactions between words, we believe that it could successfully be used for NLP tasks that are more complex than sentiment classification on IMDB.

\subsection{Conclusion}

We have presented a novel explainability method to study and understand the predictions of a black-box model. This new approach tackle important challenges, namely the efficiency of current black-box method, by proposing an efficient method grounded within the theoretical framework of sensitivity analysis using Sobol indices.
A non-trivial contribution of this work  was to make the approach tractable and efficient for high-dimensional data such as images. For this purpose, we have introduced a method using perturbation masks coupled, a Quasi-Monte Carlo sampling coupled with efficient estimators from the sensitivity analysis literature. One additional benefit of the approach is that it provides a way to study the importance of not just the main effects of input variables but also higher-order interactions between them.
We showed that our method can be efficiently used to explain the decisions of image classifiers.
It reaches performance on par with or better than the current best black-box methods while being twice as fast. It even reaches comparable results to the best white-box methods without requiring access to internal states.
We also showed that our method could be applied to language models and reported initial competitive results, and we hope that further links will be made with the field of sensitivity analysis and Attribution methods.

\clearpage

\section{Application to FRSign}
\label{sec:attribution:frsign}
In this section, we examine the application of attribution methods to models trained on the FRSign dataset~\cite{2020frsign}, and use our recently introduced Sobol method. This dataset, containing images of French railway signals, serves as a practical case for assessing our attribution technique's effectiveness in making models more transparent. Detailed setup is documented in~\autoref{sec:intro:frsign}. Our analysis primarily features results from ResNet50, yet findings are applicable to VGG and ViT models.

For this application, our focus will be twofold: firstly, we will analyze fidelity scores to determine which attribution methods is more faithful; secondly, we aim to understand the model's strategies for the classes under study. We will observe that for most classes, the model appears to use plausible features. However, for one class, the attributions are somewhat mysterious. To have deeper understanding, we will employ feature visualization to formulate a diagnosis and hypotheses.

\subsection{Fidelity Scores}

We begin with a fidelity measure to identify which attribution methods best transcribe the model's behavior. \autoref{tab:frsign:fidelity} displays the results computed from 100 randomly selected images from the test dataset\footnote{It should be noted that the question of whether it is relevant to apply explainability to the training set remains open. Up to my knowledge, I see no a priori issues with it, but out of an abundance of caution and to ensure that nothing is overlooked, we will exclusively conduct explainability analyses on the test set.}.

\begin{table}[h]
\centering
\begin{tabular}{l c c}
\textbf{Attribution Method}&\textbf{Deletion Score}&\textbf{Insertion Score}\\
\hline
Sobol&\textbf{0.329}&0.377\\
RISE&0.348&\textbf{0.396}\\
Saliency&0.402&0.325\\
Integrated Gradient&0.396&0.348\\
Grad-CAM&0.419&0.372\\
SmoothGrad&\underline{0.338}&0.363\\
Occlusion&0.345&\underline{0.380}\\
\\
\hline
\end{tabular}
\caption{\textbf{Insertion and Deletion Scores for Seven Attribution Methods on the FRSign Dataset.} This table presents the scores for each attribution method according to the fidelity metrics of insertion and deletion. It's important to remember that a lower deletion score is preferable, and a higher insertion score is considered better. The best method is highlighted in \textbf{bold}, while the second best is \underline{underlined}. The methods that appear to be the most effective are RISE, Sobol, and SmoothGrad.}
\label{tab:frsign:fidelity}
\end{table}

In \autoref{tab:frsign:fidelity}, we observe that the method we previously introduced also achieves favorable deletion scores. This finding is reassuring as it suggests that Sobol's performance generalizes beyond the datasets studied earlier. Additionally, RISE and SmoothGrad both exhibit strong performance across both metrics. Therefore, for the remainder of our study, we will primarily focus on these three methods to draw our conclusions.

\begin{figure}[ht!]
\centering
\includegraphics[width=0.9\textwidth]{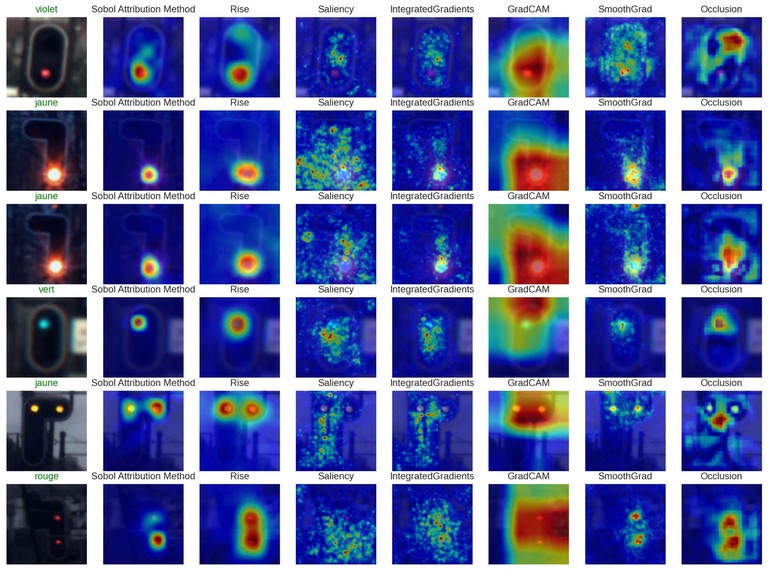}
\includegraphics[width=0.9\textwidth]{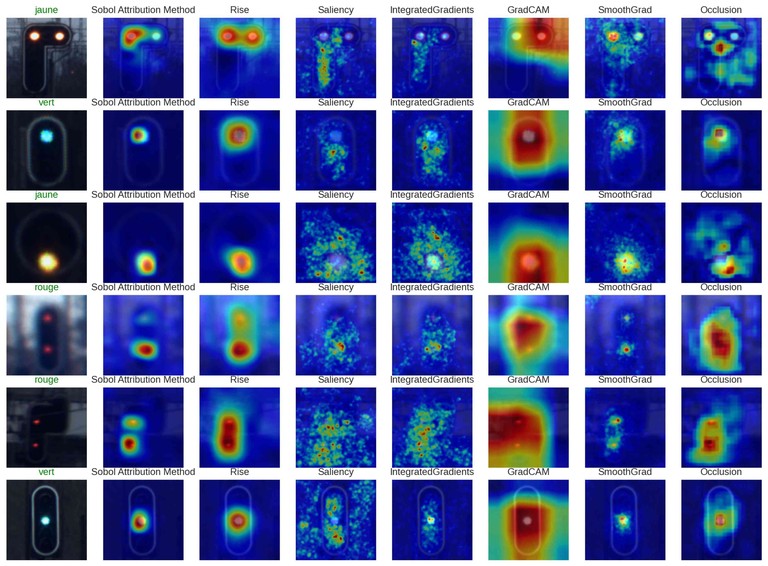}
\caption{\textbf{Comparative Visual Analysis of Attribution Methods.} We applied seven attribution methods to our model trained on the FRSign dataset. According to~\autoref{tab:frsign:fidelity}, the most faithful methods are Sobol, RISE, and SmoothGrad. Remarkably, the model tends to focus on the areas it should, specifically the traffic lights (or light) for the target class, which is reassuring.}
\label{fig:frsign:good_attributions}
\end{figure}

\subsection{Comparative Visual Analysis}

After computing the fidelity scores, we have a clearer understanding of which attribution methods more accurately reflect the model's decisions. This allows us to place greater trust in certain methods over others based on these initial tests. Sobol, RISE, and SmoothGrad emerge as the top three methods. However, we will continue to consider all methods to comprehensively assess our results. An interesting observation is that when all methods achieve good fidelity scores but highlight different areas of importance, this could be interpreted as indicating multiple ways to explain the model's reasoning. This is an intriguing aspect to explore~\footnote{Some preliminary remarks have been done on this topic in~\cite{aggregating2020}, but the ``diversity'' of explanation and the capability to aggregate them is still an interesting open questions.}. Nonetheless, it is important to remember that methods with lower fidelity scores should be approached with caution.

Figure~\autoref{fig:frsign:good_attributions} displays examples of attributions for each class that appear to be accurate. For critical signals such as violet, red, and yellow lights, the model seems to focus on the specific light or lights it is supposed to, which could increase our confidence in the model's decision-making for these types of signals.

These examples focus solely on instances where the model's predictions are correct. Next, we will apply our attribution methods to investigate failure cases, that is, instances where the model has made incorrect predictions.

\subsection{Explaining Failure Cases}

\begin{figure}[ht!]
\centering
\includegraphics[width=0.95\textwidth]{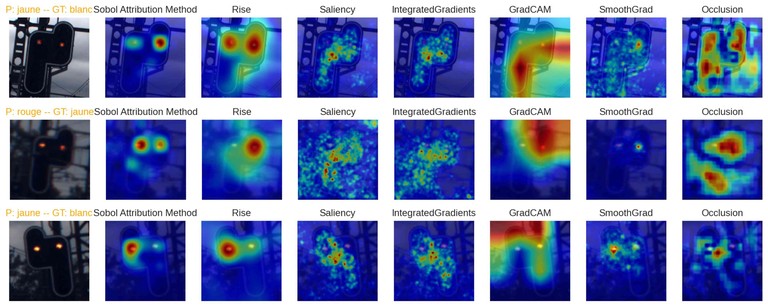}
\caption{\textbf{Attribution Methods on wrongly classified points.} We applied the same seven attribution methods to points where the model's predictions were incorrect. ``P'' denotes the model's prediction, $\f(\vx)$, and ``GT'' for the label $y$. Upon review, the human eye tends to agree with the model's predictions, which might lead us to suspect incorrect labeling. However, for some images, the labeling is indeed accurate, and it is light aberrations or capture problem that obscure the correct ground truth from view.}
\label{fig:frsign:bad_attributions}
\end{figure}

Despite the ResNet-50 model being our most performant, with an accuracy above 90\%, it is not without its share of incorrect predictions. Figure \autoref{fig:frsign:bad_attributions} presents several examples of explanations for misclassified points.

Upon further analysis, it appears that a portion of the data points were indeed incorrectly labeled, while a significant number are correctly labeled, although human observation alone may not always accurately identify the correct label due to noise, errors, or anomalies in the image capture process. This leads to an intriguing question that extends beyond the scope of this thesis: whether the model or the label is at fault. In other words, if in reality a signal was violet but appears red in our images, should we expect the model to perceive it as humans do, with all associated biases, or should it interpret the data optimally for the task at hand, potentially employing mechanisms or perceptions different from those of humans? These considerations open up a broader discourse, yet there is one final observation to be made before concluding this section.

The analysis of failure cases does not encompass the entirety of our observations. There remains one particularly perplexing scenario, observed post-analysis: the case of the white signals.

\subsection{White Signal}

\begin{figure}[ht!]
\centering
\includegraphics[width=0.95\textwidth]{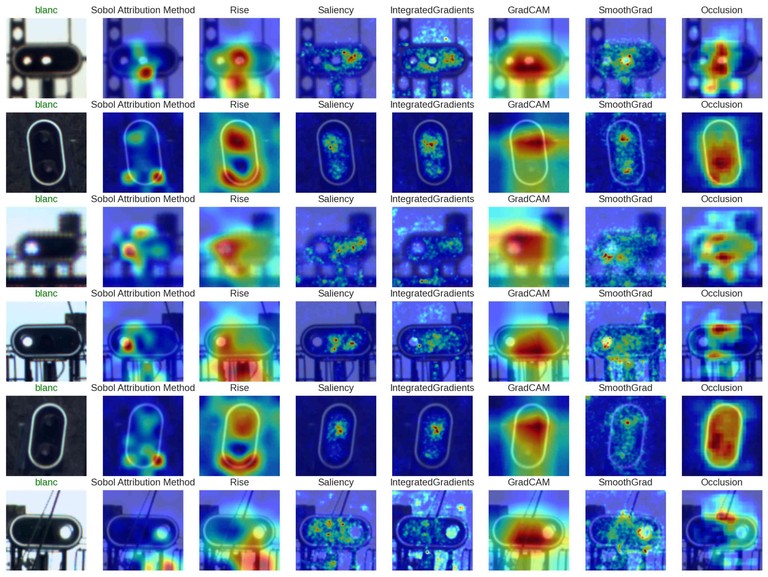}
\caption{\textbf{Attribution Methods on White Signals.} The set of explainability methods applied to images correctly predicted as white signals is concerning. The model appears to focus on areas other than the traffic lights; however, it is unclear what specifically garners the model's attention.}
\label{fig:frsign:white_signals}
\end{figure}

The interpretability of white signals poses a challenge, as the focal points of the model remain unclear. This is illustrated in Figure \autoref{fig:frsign:white_signals}.

The areas of attention for white signals appear cryptic and are not consistently focused on the lights. Furthermore, the focus does not always seem to be located in the same manner, which prevents a clear understanding of what the model is observing or relying upon for its decisions. We will now employ feature visualizations to delve deeper into this issue.

\begin{figure}[ht]
\centering
\includegraphics[width=0.95\textwidth]{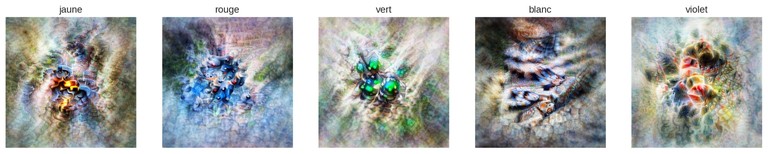}
\includegraphics[width=0.95\textwidth]{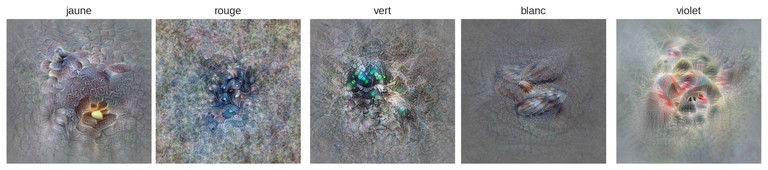}
\caption{\textbf{Feature Visualization for the Logits of $\f$.} The images represent the results of two settings of feature visualization (in Fourier space) for the image that maximizes the logits for the classes yellow, red, green, white, and violet.}
\label{fig:frsign:fviz}
\end{figure}

\paragraph{Feature Visualization.} To gain a better understanding of the potential strategies employed by our model, we utilized feature visualization. The results are shown in \autoref{fig:frsign:fviz}.

For crucial traffic lights such as red, yellow, and violet, the feature visualizations seem to make sense, which is reassuring. However, for white, the interpretations remain somewhat cryptic. Nonetheless, we can hypothesize that the model focuses on the frame (contour of the light) as indicated in the top feature visualization for the white light in \autoref{fig:frsign:fviz}.

\subsection{Conclusion}

Attribution methods serve as a valuable tool for understanding the model and verifying that it relies on plausible features. They provide reassurance in most cases by ensuring that the areas most important to the model are also those containing information meaningful to humans.

However, two main issues arise. Firstly, we wish to extend our methods to offer stronger guarantees; that is, to establish confidence bounds around our explanations to ensure the model's reliance on these interpretations. Secondly, in some instances, the features the model focuses on remain ambiguous, such as with the case of white signals. This observation suggests that further research is necessary to make attribution methods both safer and more informative.

\clearpage

\section{Guarantee for Explainable AI with Verified perturbation Analysis}
\label{sec:attributions:eva}
\newcommand{\lowerf}{\f_{\text{min}}}
\newcommand{\upperf}{\f_{\text{max}}}
\newcommand{\evaEmp}{\eva\textsubscript{emp}}
\newcommand{\evaH}{\eva\textsubscript{hybrid}}
\newcommand{\adv}{\textit{adversarial overlap}}
\newcommand{\AO}{\textit{AO}}
\newcommand{\AOup}{\overline{\textit{AO}}}
\newcommand{\AOemp}{\hat{\textit{AO}}}
\newcommand{\Adv}{\textit{Adversarial overlap}}

\newcommand{\ballu}{\ball_{\bm{u}}}

\newcommand{\rsr}{{Robustness\text{-}S\textsubscript{r}}}

In this section, we tackle the challenge of generating attributions maps with strong formal guarantee. 

We first remark that among the  plethora of attribution methods have recently been developed to explain deep neural networks, many methods use different classes of perturbations (e.g, occlusion, blurring, masking, etc.) to estimate the importance of individual image pixels to drive a model's decision.
Nevertheless, the space of possible perturbations is vast and current attribution methods typically require significant computation time to accurately sample the space in order to achieve high-quality explanations. To say it otherwise, the actual methods cannot ``scan'' the entire space of perturbation, and the ability to do so would enable us to derive strong guarantee.

In this work, we introduce EVA (Explaining using Verified Perturbation Analysis) -- the first explainability method which comes with guarantees that an entire set of possible perturbations has been exhaustively searched. We leverage recent progress in verified perturbation analysis methods to directly propagate bounds through a neural network to exhaustively probe a -- potentially infinite-size --  set of perturbations in a single forward pass. Our approach takes advantage of the beneficial properties of verified perturbation analysis, i.e., time efficiency and guaranteed complete -- sampling agnostic -- coverage of the perturbation space -- to identify image pixels that drive a model's decision. 
We  evaluate EVA systematically and demonstrate state-of-the-art results on multiple benchmarks.

\begin{figure}[t!]
  \centering
  \includegraphics[width=0.9\textwidth]{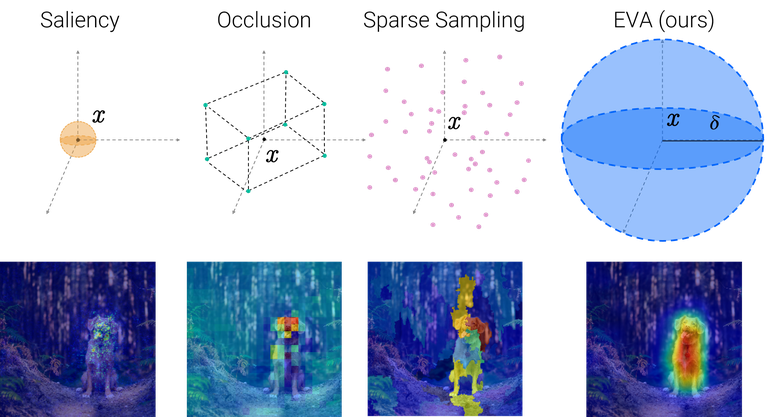}
  \caption{
  \textbf{Manifold exploration of current attribution methods.}
  Current methods assign an importance score to individual pixels using perturbations around a given input image $\vx$. Saliency~\cite{simonyan2013deep} uses infinitesimal perturbations around $\vx$, Occlusion~\cite{zeiler2013visualizing} switches individual pixel intensities on/off. More recent approaches~\cite{ribeiro2016lime, lundberg2017unified, petsiuk2018rise, fel2021sobol, novello2022making} use (Quasi-) random sampling methods in specific perturbation spaces (occlusion of segments of pixels, blurring, ...). However, the choice of the perturbation space undoubtedly biases the results -- potentially even introducing serious artifacts~\cite{sturmfels2020visualizing,hsieh2020evaluations,haug2021baselines,kindermans2019reliability}.
  We propose to use verified perturbation analysis to efficiently perform a complete coverage of a perturbation space around $\vx$ to produce reliable and faithful explanations.
  }
  \label{fig:eva:big_picture}
\end{figure}

\subsection{Background}

The applications of Attributions methods these methods are multiple -- from helping to improve or debug their decisions to helping instill confidence in the reliability of their decisions~\cite{doshivelez2017rigorous}, as explained in our  \autoref{sec:intro:motivation}. 
Unfortunately, a severe limitation of these approaches is that they are subject to a confirmation bias: while they appear to offer useful explanations to a human experimenter, they may produce incorrect explanations~\cite{adebayo2018sanity, ghorbani2017interpretation, slack2021counterfactual}.
In other words, just because the explanations make sense to humans does not mean that they actually convey what is actually happening within the model.
Therefore, the community is actively seeking for better benchmarks involving humans~\cite{hsieh2020evaluations,nguyen2021effectiveness,fel2021cannot,kim2021hive}.

In the meantime, it has been shown that some of our current and commonly used benchmarks are biased and that explainability methods reflect these biases -- ultimately providing the wrong explanation for the behavior of the model~\cite{sturmfels2020visualizing,hsieh2020evaluations,hase2021out}.
For example, some of the current fidelity metrics \cite{petsiuk2018rise, aggregating2020,jacovi2020towards,hedstrom2022quantus,fel2022xplique} mask one or a few of the input variables (with a fixed value such as a gray mask) in order to assess how much they contribute to the output of the system. Trivially, if these variables are already set to the mask value in a given image (e.g., gray), masking these variables will not yield any effect on the model's output and the importance of these variables is poised to be underestimated. 
Finally, these methods rely on sampling a space of perturbations that is far too vast to be fully explored -- e.g., LIME on a image divided in $64$ segments image would need more than $10^{19}$ samples to test all possible perturbations. 
As a result, current attribution methods may be subject to bias and are potentially not entirely reliable.

\paragraph{Explainability through the lens of Robustness.} To try to address the aforementioned limitations, several groups~\cite{ignatiev2019abduction, ignatiev2019relating, slack2021reliable, hsieh2020evaluations, boopathy2020proper, lin2019explanations, fel2020representativity} have focused on the development of a new set of robustness-based evaluation metrics for trustworthy explanations. 
These new metrics are in contrast with the previous ones, which consisted in removing the pixels considered important in an explanation by substituting them with a fixed baseline -- which inevitably introduces bias and artifacts~\cite{hsieh2020evaluations,sturmfels2020visualizing,haug2021baselines,kindermans2019reliability,hase2021out}. 
Key to these new metrics is the assumption that when the important pixels are in their nominal (fixed) state, then perturbations applied to the complementary pixels -- deemed unimportant -- should not affect the model's decision to any great extent. The corollary that follows is that perturbations limited to the pixels considered important should easily influence the model's decision~\cite{lin2019explanations,hsieh2020evaluations}.
Going further along the path of robustness, abductive reasoning was used in~\cite{ignatiev2019abduction} to compute optimal subsets with guarantees.  The challenge consists  in looking for the subset with the smallest possible  cardinality -- to guarantee the decision of the model. This work constituted one of the early successes of formal methods for explainability, but the approach was limited to low-dimensional problems and shallow neural networks. It was later extended to relax the subset minimum explanation by either providing multiple explanations, aggregating pixels in bundles~\cite{bassan2022towards} or by using local surrogates~\cite{boumazouza2021asteryx}.

Some heuristics-oriented works also propose to optimize these new robustness based criteria and design new methods using a generative model~\cite{o2020generative} or adversarial attacks~\cite{hsieh2020evaluations}.
The latter approach requires searching for the existence or lack of an adversarial example for a multitude of $\ell_p$ balls around the input of interest. As a result, the induced computational cost is quite high as the authors used more than $50000$  computations of adversarial examples to generate a single explanation. 

More importantly, a failure to find an adversarial perturbation for a given radius does not guarantee that none exists. In fact, it is not uncommon for adversarial attacks to fail to converge --  or fail to find an adversarial example -- which will result in a failure to output an importance score.
Our method addresses these issues while drastically reducing the computation cost.
An added benefit of our approach is that verified perturbation analysis provides additional guarantees and hence opens the doors of certification which is a necessity for safety-critical applications.

\paragraph{Verified Perturbation Analysis.} This growing field of research focuses on the development of methods that outer-approximate neural network outputs given some input perturbations. 
Simply put, for a given input $\vx$ and a bounded perturbation $\v{\delta}$, verification methods yield minimum $\lowerf(\vx)$ and maximum $\upperf(\vx)$ bounds on the output of a model. Formally $\forall~ \v{\delta} ~s.t~ ||\v{\delta}||_p \leq \varepsilon$:

$$
\lowerf(\vx) \leq \pred(\vx + \v{\delta}) \leq \upperf(\vx). 
$$
This allows us to explore the whole perturbation space without having to explicitly sample points in that space.

Early works focused on computing reachable lower and upper bounds based on satisfiability modulo theories~\cite{katz2017reluplex, ehlers2017formal}, and mixed-integer linear programming problems~\cite{tjeng2017verifying}. While these early results were encouraging, the proposed methods struggled even for small networks and image datasets. More recent work has led to the independent development of methods for computing looser certified lower and upper bounds more efficiently thanks to convex linear relaxations either in the primal or dual space~\cite{salman2019convex}. 
While looser, those bounds remain tight enough to yield non-ubiquitous robustness properties on medium size neural networks. CROWN (hereafter called Backward) uses Linear Relaxation-based Perturbation Analysis (LiRPA) and achieves the tightest bound for efficient single neuron linear relaxation~\cite{zhang2018efficient, singh2019abstract, wang2021beta}. 
In addition, linear relaxation methods offer a wide range of possibilities with a vast trade-off between ``tigthness'' of the bounds and efficiency. 
These methods form two broad classes: `forward' methods which propagate constant bounds (more generally affine relaxations from the input to the output of the network) also called Interval Bound Propagation (IBP, Forward, IBP+Forward) vs. `backward' methods which bound the output of the network by affine relaxations given the internal layers of the network, starting from the output to the input. Note that these methods can be combined, e.g. (CROWN + IBP + Forward).
For a thorough description of the LiRPA framework and theoretical analysis of the worst-case complexities of each variant, see~\cite{xu2020automatic}.
In this work, we remain purposefully agnostic to the verification method used and opt for the most accurate LiRPA method applicable to the predictor. Our approach is based on the formal verification framework DecoMon, based on Keras~\cite{ducoffe2021decomon}.

In this work, we propose to address this limitation by introducing \eva~(Explaining using Verified perturbation Analysis), a new explainability method based on robustness analysis. Verified perturbation analysis is a rapidly growing toolkit of methods to derive bounds on the outputs of neural networks in the presence of input perturbations. In contrast to current attributions methods based on gradient estimation or sampling, verified perturbation analysis allows the full exploration of the perturbation space, see Fig.~\ref{fig:eva:big_picture}. We use a tractable certified upper bound of robustness confidence to derive a new estimator to help quantify the importance of input variables (i.e., those that matter the most). That is, the variables most likely to change the predictor's decision.

\subsection{Explainability with Verified Perturbation Analysis}
\label{sec:eva:method}

\paragraph{Notation.} We still consider a standard supervised machine-learning classification setting with input space $\mathcal{X} \subseteq \mathbb{R}^d$, an output space $\mathcal{Y} \subseteq \mathbb{R}^c$, and a predictor function $\pred : \mathcal{X} \to \mathcal{Y}$ that maps an input vector $\vx~=~(x_1,~\ldots{},~x_d)$ to an output $\pred(\vx)~=~\left(\pred^{(1)}(\vx),~\ldots{},~\pred^{(c)}(\vx)\right)$.
We denote $\ball~=~\{\v{\delta} \in \mathbb{R}^d ~:~  || \v{\delta}||_{p} \leq \radius \}$ 
the perturbation ball with radius $\radius > 0$,  with 
$p \in \{1, 2, \infty\}$.
For any subset of indices $\v{u}\subseteq\{1,~\ldots{},~d\}$, we denote 
$\ballu$ the ball without perturbation on the variables in $\v{u}$:
$\ballu = \{ \v{\delta} ~:~ \v{\delta} \in \ball, ~ \v{\delta}_{\v{u}} = 0 \}$ and $\ball(\bm{x})$ the perturbation ball centered on $\bm{x}$. We denote the lower (resp. upper) bounds obtained with verification perturbation analysis as: 

$$
\lowerf(\vx, \ball) = \left(\lowerf^{(1)}(\vx, \ball),~\ldots{},~\lowerf^{(c)}(\vx, \ball)\right)
$$
$$
\upperf(\vx, \ball) = \left(\upperf^{(1)}(\vx, \ball),~\ldots{},~\upperf^{(c)}(\vx, \ball)\right).
$$
Intuitively, these bounds delimit the output prediction for any perturbed sample in $\ball(\vx)$, such that:
$$
\forall \v{\delta} \in \ball, \lowerf(\vx, \ball) \leq \f(\vx + \v{\delta}) \leq \upperf(\vx, \ball).
$$

\subsubsection{The importance of setting the importance right}

Different attribution methods implicitly assume different definitions of the notion of importance for input variables based either on game theory~\cite{lundberg2017unified}, the notion of conditional expectation of the score logits~\cite{petsiuk2018rise}, their variance~\cite{fel2021sobol} or on some  measure of statistical dependency between different areas of an input image and the output of the model~\cite{novello2022making}.
For this work, we build on robustness-based explainability methods~\cite{hsieh2020evaluations} which assume that a variable is important if small perturbations of this variable lead to large changes in the model decision.
Conversely, a variable is said to be unimportant if changes to this variable only yield small changes in the model decision.
From this intuitive assertion, we construct an estimator that we call \Adv.

\subsubsection{Adversarial overlap}

We go one step beyond previous work and propose to compute importance by taking into account not only the ability of individual variables to change the network's decision but also its confidence in the prediction.
\Adv ~ measures the extent to which a modification on a group of pixels can generate overlap between classes, i.e. generate a point close to $\vx$ such that the attainable maximum of an unfavorable class $c'$ can match the minimum of the initially predicted class $c$.

Indeed, if a modification of a pixel -- or group of pixels -- allows generating a new image that changes the decision of $\pred$, this variable must be considered important. 
Conversely, if the decision does not change regardless of the value of the pixel, then the pixel can be left at its nominal value and should be considered unimportant. 

Among the set of possible variable perturbations $\v{\delta}$ around a point $\vx$, we, therefore, look for points that can modify the decision\footnote{Throughout this section, when $c$ is not specified, it is assumed that $c = \argmax \pred(\vx)$.} with the most confidence.
Hence our scoring criterion can be formulated as follows:

\begin{equation}\label{eq:eva:adv_surface}
    \AO(\vx, \ball) = 
    \max_{\substack{\v{\delta} \in \ball, c'\neq c}} \f^{(c')}(\vx + \v{\delta}) - \f^{(c)}(\vx + \v{\delta}).
\end{equation}

Intuitively, this score represents the confidence of the ``best'' adversarial perturbation that can be found in the perturbation ball $\ball$ around $\vx$.

In order to estimate this criterion, a naive strategy could be to use adversarial attacks to search within $\ball$. 
However, when they converge - which is not ensured, such methods only explore certain points of the considered space, thus giving no guarantee regarding the optimality of the solution. 
Moreover, adversarial methods have no guarantee of success and therefore cannot ensure a valid score under every circumstance.
Finally, the large dimensions of the current datasets make exhaustive searches impossible.

To overcome these issues, we take advantage of one of the main results from verified perturbation analysis to derive a guaranteed upper bound on the criterion introduced in Eq.~\ref{eq:eva:adv_surface}. 
We can upper bound the \adv{} criterion as follows: 

$$ 
\AO(\vx, \ball) \leq \AOup(\vx, \ball) = \max\limits_{c'\neq c} \upperf^{(c')}(\vx, \ball) - \lowerf^{(c)}(\vx, \ball). 
$$
The computation of this upper bound becomes tractable using any verified perturbation analysis method.

\begin{figure*}[t!]
  \includegraphics[width=0.99\textwidth]{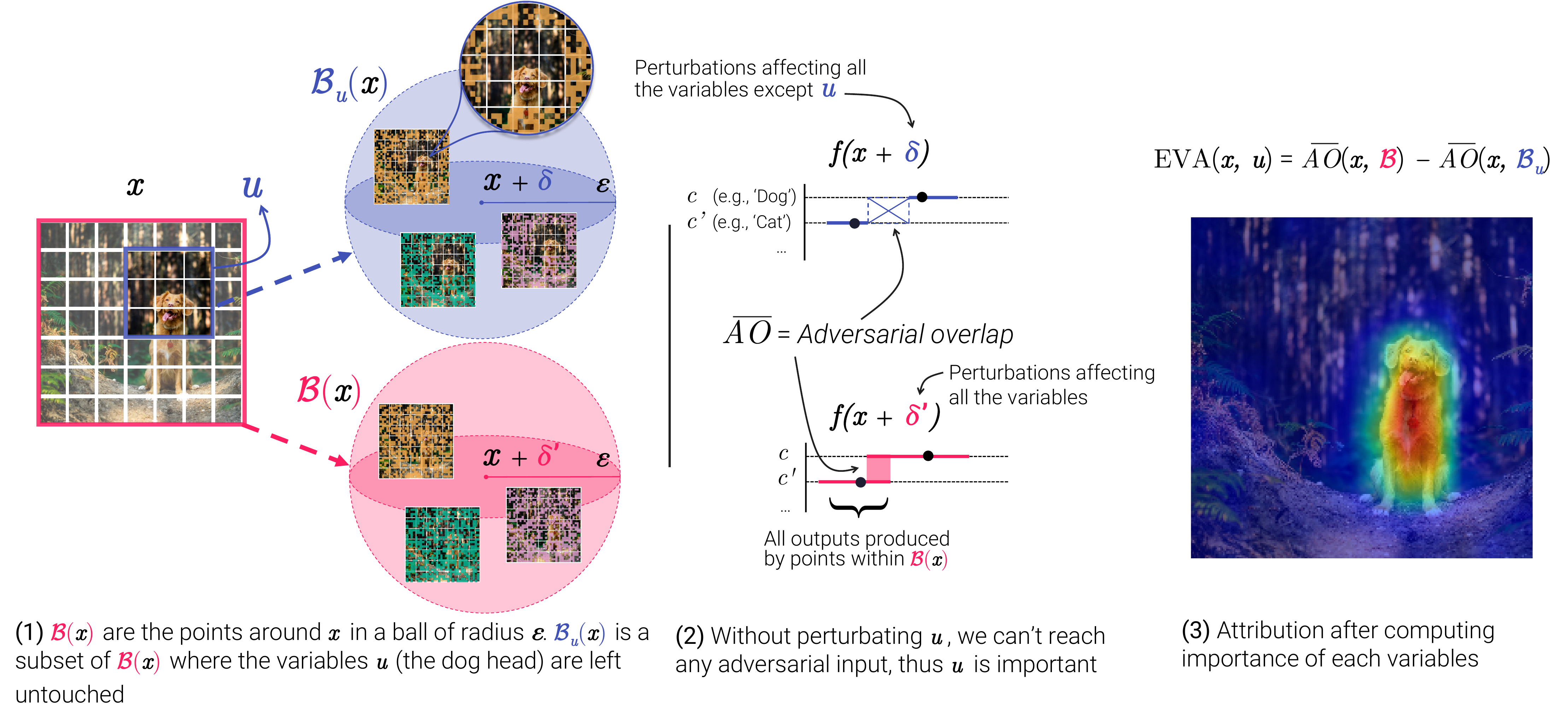}
  \caption{
  \textbf{\eva~attribution method.} In order to compute the importance for a group of variables $\v{u}$ -- for instance the dog's head -- the first step (1) consists in designing the perturbation ball $\textcolor{indigo}{\ballu}(\vx)$. This ball is centered in $\vx$ and contain all the possible images perturbed by $\textcolor{indigo}{\v{\delta}} ~s.t~ ||\textcolor{indigo}{\v{\delta}}||_{p} \leq \varepsilon, ||\textcolor{indigo}{\v{\delta}}_{\v{u}}||_p = 0$ which do not perturb the variables $\textcolor{indigo}{\v{u}}$. Using verified perturbation analysis, we then compute the \adv~ $\AOup(\vx, \textcolor{indigo}{\ballu})$ which corresponds to the overlapping between the class $c$ -- here dog -- and $c'$, the maximum among the other classes. Finally, the importance score for the variable $\v{u}$ corresponds to the drop in \adv~ when $\v{u}$ cannot be perturbed, thus the difference between $\AOup(\vx, \textcolor{pink}{\ball})$ and $\AOup(\vx, \textcolor{indigo}{\ballu})$. 
  Specifically, this measures how important the variables $\v{u}$ are for changing the model's decision.
  }
  \label{fig:eva:method}
\end{figure*}

For example, $\AOup(\vx, \ball) \leq 0$  guarantees that no adversarial perturbation is possible in the perturbation space.\footnote{Note that with adversarial attacks, failure to find an adversarial example does not guarantee that it does not exist.}
Our upper bound $\AOup(\vx, \ball)$ corresponds to the difference between the verified lower bound of the class of interest $c$ and the maximum over the verified upper bounds among the other classes.
Thus, when important variables are modified (e.g the head of the dog in Fig.~\ref{fig:eva:method}, using $ \textcolor{pink}{\ball} $), the lower bound for the class of interest will get smaller than the upper bound of the adversary class. On the other hand, this overlap is not possible when important variables are fixed (e.g in Fig.~\ref{fig:eva:method} when the head of the dog is fixed, using $ \textcolor{indigo}{\ballu} $).
We now demonstrate how to leverage this score to derive an efficient estimator of variable importance.

\subsubsection{\eva}

We are willing to assign a higher importance score for a variable allowing (1) a change in a decision, (2) a greater adversarial -- thus a solid change of decision. Modifying all variables gives us an idea of the robustness of the model.
In the same way, the modification of all variables without the subset $\v{u}$ allows quantifying the change of the strongest adversarial perturbation and thus quantifies the importance of the variables $\v{u}$. Intuitively, if an important variable $\v{u}$ is discarded, then it will be more difficult, if not impossible, to succeed in finding any adversarial perturbation. Specifically, removing the possibility to modify $\vx_{\v{u}}$ allows us to reveal its importance by taking into account its possible interactions.

The complexity of current models means that the variables are not only treated individually in neural network models, but collectively. In order to capture these higher-order interactions, our method consists in measuring the \adv~ allowed by all the variables together $\AOup(\vx, \ball)$ -- thus taking into account their interactions -- and then forbidding to play on a group of variables $\AOup(\vx, \ballu)$ to estimate the importance of $\v{u}$. Making the interactions of $\v{u}$ disappear reveals their importance. Note that several works have mentioned the importance of taking into account the interactions of the variables when calculating the importance~\cite{petsiuk2018rise,fel2021sobol, ferrettini2021coalitional,idrissi2023coalitional}. Formally:

\begin{definition}[\eva]
We introduce \eva~(Explainability using Verified perturbation Analysis) that measure the drop in \adv~ when we fixed the variables $\v{u}$:

\begin{equation}
    \label{eq:eva:tod_estimator}
    \bm{\eva}(\vx, \v{u}, \ball) \defas \AOup(\vx, \ball) - \AOup(\vx, \ballu).
\end{equation}

\end{definition}

As explained in Fig.~\ref{fig:eva:method}, the estimator requires two passes of the perturbation analysis method; one for $\AOup(\ball)$, and the other for $\AOup(\ballu)$: the first term consists in measuring the \adv~ by modifying all the variables, the second term measures the adversarial surface when fixing the variables of interest $\v{u}$.
In other words, \eva~measures the \adv~that would be left if the variables $\v{u}$ were to be fixed.

From a theoretical point of view, we notice that \eva~- under reasonable assumptions - yield the optimal subset of variables to minimize the theoretical \rsr.

\begin{theorem}\textbf{\eva~provide the optimal set from step $|\v{u}|$ to the last step.}
\label{thm:eva:rsr}
With $\v{u}$ the essential variables of $\v{\delta}^*$, \eva~will rank the $\v{u}$ variables first and provide the optimal set from the step $|\v{u}|$ to the last step. 
\end{theorem}

Proof in Appendix \autoref{app:eva:rsr}. Moreover, we note that the explanation stability can be easily bounded by the model Lipschitz constant.

\begin{theorem}\textbf{\eva~ has bounded Stability}
\label{thm:eva:stab}
Given a $L$-lipschitz predictor $\f$, $\radius$ the radius of $\ball$ and $\radiusbis$ the Stability radius, then
$$
\textit{Stability}(\vx, \eva) \leq 4L(\radius + \radiusbis)
$$
\end{theorem}
Proof in Appendix \autoref{app:eva:stab}.
From a computational point of view, we can note that the first term of the \adv~$\AOup(\vx, \ball)$ -- as it does not depend on $\v{u}$ -- can be calculated once and re-used to evaluate the importance of any other variables considered. 
Moreover, contrary to an iterative process method~\cite{Fong_2017, hsieh2020evaluations, ignatiev2019abduction}, each importance can be evaluated independently and thus benefit from the parallelization of modern neural networks. Finally, the experiments in Section~\ref{sec:eva:experiments} show that even with two calls to $\AOup$~ per variables, our method remains much faster than the one based on sampling or on adversarial attacks (such as Greedy-AS or Greedy-AO, see appendix \ref{ap:eva:benchmarks}). %

In this work, the verified perturbation-based analysis considered is not always adapted to high dimensional models, especially those running on ImageNet~\cite{imagenet_cvpr09}. We are confident that the verification methods will progress towards more scalability in the near future, enabling the original version of \eva~ on deeper models. 

In the meantime, we introduce an empirical method that allows to scale \eva~to high dimensional models. This method sacrifices theoretical guarantees, but the results section reveals that it may be a good compromise.

\begin{figure*}[t!]
  \centering
  \includegraphics[width=0.95\textwidth]{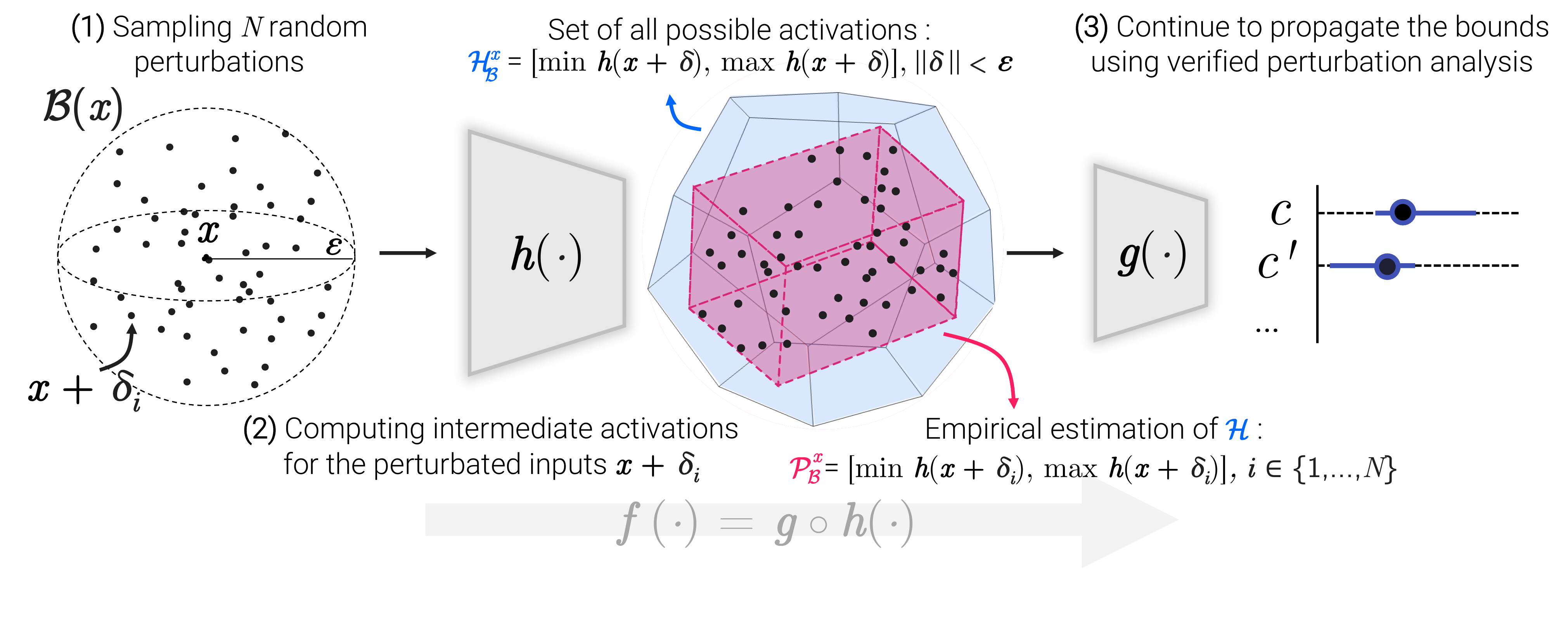}
  \caption{\textbf{Scaling strategy.} 
  In order to scale to very large models, we propose to estimate the bounds of an intermediate layer's activations empirically by (1) Sampling $N$ input perturbations and (2) calculating empirical bounds on the resulting activations for the layer $\bm{h}(\cdot)$. 
  We can then form the set $\textcolor{pink}{\mathcal{P}_{\ball}^{\vx}}$ which is a subset of the true bounds $\textcolor{indigo}{\mathcal{H}_{\ball}^{\vx}}$ since the sampling is never exhaustive. We can then plug this set into a verified perturbation analysis method (3) and continue the forward propagation of the inputs through the rest of the network.
  }
  
  \label{fig:eva:eva_hybrid}
\end{figure*}

\subsubsection{Scaling to larger models}
\label{sec:eva:scaling}

We propose a second version of \eva, 
which is a combination of sampling and verification perturbation analysis. 
The aim of this hybrid method is twofold: (\textit{\textbf{i}}) take advantage of sampling to approach the bounds of an intermediate layer in a potentially very large model, (\textit{\textbf{ii}}) then complete only the rest of the propagations with verified perturbation analysis and thus move towards the native \eva~method which benefits from theoretical guarantees.
Note that, combining verification methods with empirical methods (a.k.a adversarial training) has notably been proposed in~\cite{balunovic2019adversarial} for robust training.

Specifically, our technique consists of splitting the model into two parts, and (\textit{\textbf{i}}) estimating the bounds of an intermediate layer using sampling, (\textit{\textbf{ii}}) propagating these empirical intermediate bounds onto the second part of the model with verified perturbation analysis methods.

For the first step (\textit{\textbf{i}}) we consider the original predictor $\pred$ as a composition of functions 
$\pred(\vx) = (\bm{g} \circ \bm{h})(\vx)$. 
For deep neural networks, $\bm{h}(\cdot)$ is a function that maps input to an intermediate feature space
and $\bm{g}(\cdot)$ is a function that maps this same feature space to the classification.

We propose to empirically estimate bounds 
$( \underline{\bm{h}}_{\ball}^{\vx}, \overline{\bm{h}}_{\ball}^{\vx} )$ for the intermediate activations $ \bm{h}(\cdot) \in \mathbb{R}^{d'}$ using Monte-Carlo sampling on the perturbation $\v{\delta} \in \ball$. Formally:

\begin{equation*}
\begin{split}
    \forall j \in \{0, \ldots, d'\}, ~
& \underline{\bm{h}}_{\ball}^{\vx}[j] = 
  \min\limits_{\v{\delta}_1,\ldots, \v{\delta}_n \overset{\text{iid}}{\sim} \P(\ball)}
  \bm{h}(\vx+\v{\delta}_i)[j]\\
& \overline{\bm{h}}_{\ball}^{\vx}[j] =  
  \max\limits_{\v{\delta}_1,\ldots,\v{\delta}_n  \overset{\text{iid}}{\sim} \P(\ball)}
  \bm{h}(\vx+\v{\delta}_i)[j].
\end{split}
\end{equation*}

With $\P(\ball)$ the uniform distribution over $\ball$. Obviously, since the sampling is never exhaustive, the bounds obtained underestimate the true maximum $ \overline{\bm{h}}_{\ball}^{\vx} \leq \max \bm{h}(\vx + \v{\delta})$ and overestimates the true minimum $ \underline{\bm{h}}_{\ball}^{\vx} \geq \min \bm{h}(\vx + \v{\delta})$ as illustrated in the Fig.~\ref{fig:eva:eva_hybrid}.
In a similar way, we define $\underline{\bm{h}}_{\ballu}^{\vx}$ and $\overline{\bm{h}}_{\ballu}^{\vx}$ when  $\v{\delta} \in \ballu$. 
Once the empirical bounds are estimated, we may proceed to the second step and use the obtained bounds to form the new perturbation set 

$$ 
\mathcal{P}^{\vx}_{\ball} = 
    [\underline{\bm{h}}_{\ball}^{\vx} - \bm{h}(\vx), 
    \overline{\bm{h}}_{\ball}^{\vx} - \bm{h}(\vx)].
$$ 
Intuitively, this set bounds the intermediate activations obtained empirically and can then be fed to a verified perturbation verification method.

We then carry out the end of the bounds propagation in the usual way, using verified perturbation analysis. This amounts to computing bounds for the outputs of the network for all possible activations contained in our empirical bounds. The only change is that we no longer operate in the pixel space $\vx$ with the ball $\ball$, but in the activation space $\bm{h}(\cdot)$  with the perturbations set $\mathcal{P}^{\vx}_{\ball}$. The importance score of a set of variables $\v{u}$ is then : 

$$ \evaH(\vx, \v{u}, \ball) \defas \eva(\bm{h}(\vx), \v{u}, \mathcal{P}^{\vx}_{\ball}).$$

This hybrid approach allows us to use \eva~ on state-of-the-art models and thus to benefit from our method while remaining tractable. We believe this extension to be a promising step towards robust explanations on deeper networks.

\subsection{Experiments}
\label{sec:eva:experiments}

\begin{table*}[t]
    \centering
    \scalebox{0.75}{
        \begin{tabular}{l C{0mm} P{0mm} P{5mm}P{5mm}P{5mm}P{5mm}P{5mm} P{1mm} P{5mm}P{5mm}P{9mm}P{5mm}P{5mm} P{1mm} P{5mm}P{5mm}P{5mm}P{5mm}P{5mm}}
        \toprule
        &&& \multicolumn{5}{c}{MNIST} &&  \multicolumn{5}{c}{Cifar-10} &&  \multicolumn{5}{c}{ImageNet}  \\
        \cmidrule(lr){4-8} \cmidrule(lr){10-14} \cmidrule(lr){16-20}
        
        &&& Del.$\downarrow$ & Ins.$\uparrow$ & Fid.$\uparrow$ & Rob.$\downarrow$ & Time 
        && Del.$\downarrow$ & Ins.$\uparrow$ & Fid.$\uparrow$ & Rob.$\downarrow$ & Time 
        && Del.$\downarrow$ & Ins.$\uparrow$ & Fid.$\uparrow$ & Rob.$\downarrow$ & Time
        \\
        
        \midrule
        Saliency &   && .193 & \underline{.633} & \underline{.378} & .071 & 0.04
                 && \underline{.171} & .172 & -.021 & .026 & 0.16
                 && \underline{.057} & .126 & .035 & .769 & 0.36 \\
        GradInput &   && .222 & .611 & .107 & .074 & 0.04 
                 && .200 & .143 & -.018 & .095 & 0.17
                 && \underline{.057} & .050 & .023 & .814 & 0.36  \\
        SmoothGrad &   && .185 & .621 & .331  & .070 & 1.91
                 && .174 & .181 & .092 & .048 & 9.07
                 && \textbf{.051} & .069 & .019 & .809 & 9.63 \\
        VarGrad &   && .207   & .555   & .216   & .077   & 1.76 
                 && .183 & .211 & -.012 & .193 & 9.07
                 && .098 & .201 & .021 & .787 &  9.62 \\
        InteGrad &   && .209 & .615 & .108 & .074 & 1.77
                 && .194 & .171 & -.016 & .154 & 7.19
                 && .058 & .052 & .023 & .813 & 8.39 \\
        Occlusion &   && .247  & .545  & .137  & .082  & 0.04 
                 && .217 & \underline{.290} & .105 & .232 & 1.13
                 && .100 & .266 & .026 & .821 & 4.97 \\
        GradCAM &   && n/a &  n/a  &  n/a  &  n/a  &  n/a  
                 && .297 & .282 & .056 & .195 & 0.39
                 && .073 & .232 & .036 & .817 & 0.18 \\
        GradCAM++ &   && n/a  & n/a & n/a & n/a & n/a
                 && .270 & \textbf{.326} & .102 & .094 & 0.39
                 && .074 & \underline{.285} & \underline{.054} & .800 & 0.19 \\
        RISE &   && .248  & .558 & .133  & .093  & 2.26 
                 && .196 & .273  & \underline{.157} & .385 & 20.5
                 && .074 & .276 & \textbf{.154} & .818 & 1215 \\
        Greedy-AS &   && .260  & .497  & .110  & \textbf{.061}  & 335
                 && .205 & .264 & -.003 & \textbf{.013} & 4618
                 && .088 & .047 & .023 & \textbf{.612} & 180056  \\
        \midrule 
        \textbf{\eva}~(ours)     & && \textbf{.089} & \textbf{.736} & \textbf{.428} & \underline{.069} & 1.29 
                 && \textbf{.164} & \underline{.290} & \textbf{.352} & \underline{.025} & 12.7 
                 && .070 & \textbf{.289} & .048 & \underline{.758} & 6454
                 \\
        
        \bottomrule \\
        \end{tabular}
    }
    \caption{Results on Deletion (Del.), Insertion (Ins.), $\mu$Fidelity (Fid.) and \rsr~ (Rob.) metrics. 
Time in seconds corresponds to the generation of 500 (MNIST/CIFAR-10) and 100 (ImageNet) explanations on an Nvidia P100.
Note that \eva~is the only method with guarantees that the entire set of possible perturbations has been exhaustively searched.
Verified perturbation analysis with IBP + Forward + Backward is used for MNIST, with Forward only for  CIFAR-10 and with our hybrid strategy described in Section.\ref{sec:eva:scaling} for ImageNet. 
Grad-CAM and Grad-CAM++ are not calculated on the MNIST dataset since the network  only has dense layers. 
The first and second best results are  in \textbf{bold} and \underline{underlined}, respectively. 
}
    \label{tab:eva:cifar_mnist_metrics}
    \vspace{-3mm}
\end{table*}

To evaluate the benefits and reliability of our explainability method, we performed several experiments on a standard dataset, using a set of common explainability metrics against \eva.
In order to test the fidelity of the explanations produced by our method, we compare them to that of 10 other explainability methods using the (1) Deletion, (2) Insertion, and (3) MuFidelity metrics. As it has been shown that these metrics can exhibit biases, we completed the benchmark by adding the (4) \rsr metric. Each score is averaged over 500 samples.

We evaluated these 4 metrics on 3 image classification datasets, namely MNIST~\cite{lecun2010mnist}, CIFAR-10~\cite{krizhevsky2009learning} and ImageNet~\cite{imagenet_cvpr09}.

Through these experiments, the explanations were generated using \eva~estimator introduced in Equation~\ref{eq:eva:tod_estimator}. The importance scores were not evaluated pixel-wise but on each cell of the image after having cut it into a grid of 12 sides (see Fig.~\ref{fig:eva:method}). For MNIST and Cifar-10, we used 
$\varepsilon = 0.5$, whereas for ImageNet $\varepsilon = 5$. Concerning the verified perturbation analysis method, we used (IBP+Forward+Backward) for MNIST, and (IBP+Forward) on Cifar-10 and $p=\infty$. For computational purposes, we used the hybrid approach introduce in Section~\ref{sec:eva:scaling} for ImageNet using the penultimate layer (FC-4096) as the intermediate layer $\bm{h}(\cdot)$. We give in Appendix the complete set of hyperparameters used for the other explainability methods, metrics considered as well as the architecture of the models used on MNIST and Cifar-10.

\subsubsection{Comparison with the state of the art}

There is a general consensus that fidelity is a crucial criterion for an explanation method. That is, if an explanation is used to make a critical decision, then users are expecting it to reflect the true decision-making process underlying the model and not just a consensus with humans. Failure to do so could have disastrous consequences. Pragmatically, these metrics assume that the more faithful an explanation is, the faster the prediction score should drop when pixels considered important are changed.
In Table~\ref{tab:eva:cifar_mnist_metrics}, we present the results of the Deletion~\cite{petsiuk2018rise} (or $1 - AOPC$~\cite{samek2016evaluating}) metric for the MNIST and Cifar-10 datasets on 500 images sampled from the test set. TensorFlow~\cite{tensorflow2015} and the Keras API~\cite{chollet2015keras} were used to run the models and Xplique~\cite{fel2022xplique} for the explainability methods.
In order to evaluate the methods, the metrics require a baseline and several were proposed~\cite{sturmfels2020visualizing, hsieh2020evaluations}, but we chose to keep the choice of~\cite{hsieh2020evaluations} using their random baseline.
 
We observe that \eva~is the explainability method getting the best Deletion, Insertion, and $\mu$Fidelity scores on MNIST, and is just behind Greedy-AS on \rsr. This can be explained by the fact that the Robustness metric uses the adversarial attack PGD~\cite{madry2017pgd}, which is the same one used to generate Greedy-AS, thus biasing the adversarial search. Indeed, if PGD does not find an adversarial perturbation using a subset $\v{u}$ does not give a guarantee of the robustness of the model, just that the adversarial perturbation could be difficult to reach with PGD.

For Cifar-10, \eva~remains overall the most faithful method according to Deletion and $\mu$Fidelity, and obtains the second score in Insertion behind Grad-Cam++~\cite{chattopadhay2018grad}. 
Finally, we notice that if Greedy-AS~\cite{hsieh2020evaluations} allows us to obtain a good \rsr~score, but this comes with a considerable computation time, which is not the case of \eva~which is much more efficient. Eventually, EVA is a very good compromise for its relevance to commonly accepted explainability metrics and more recent robustness metrics.

\paragraph{ImageNet}

After having demonstrated the potential of the method on vision datasets of limited size, we consider the case of ImageNet which has a significantly higher level of dimension.
The use of verified perturbation analysis methods other than IBP is not easily scalable on these datasets. We, therefore, used the hybrid method introduced in Section ~\ref{sec:eva:scaling} in order to estimate the bounds in a latent space  and then plug those bounds into the perturbation analysis to get the final \adv~ score.

Table~\ref{tab:eva:cifar_mnist_metrics} shows the results obtained with the empirical method proposed in Section~\ref{sec:eva:scaling}. We observe that even with this relaxed estimation, \eva~is able to score high on all the metrics. Indeed, \eva~obtains the best score on the Insertion metric and ranks second on $\mu$Fidelity and 
\rsr.
Greedy-AS ranks first on \rsr~at the expense of the other scores where it performs poorly. Finally, both RISE and SmoothGrad perform  well on all the fidelity metrics but collapse on the robustness metric. Extending results with ablations of \eva, including Greedy-AO, are available in Table~\ref{tab:eva:ablation_ao}.

Qualitatively, Fig.~\ref{fig:eva:imagenet_explanations} shows examples of explanations produced on the ImageNet VGG-16 model. The explanations produced by \eva~ are more localized than Grad-CAM or RISE, while being less noisy than the gradient-based or Greedy-AS methods.

In addition, as the literature on verified perturbation analysis is evolving rapidly we can conjecture that the advances will benefit the proposed explainability method.
Indeed, \eva~proved to be the most effective on the benchmark when an accurate formal method was used.
After demonstrating the performance of the proposed method, we study its ability to generate class explanations specific. 

\subsubsection{Tighter bounds lead to improved explanations}

\begin{table}[h]
    \centering
    \scalebox{0.9}{
        \begin{tabular}{l c P{5mm}P{5mm}P{5mm}P{5mm}}
        \toprule
        & Tightness$\downarrow$ & Del.$\downarrow$ & Ins.$\uparrow$ & Fid.$\uparrow$ & Rob.$\downarrow$ \\
        \midrule
        IBP & 4.58 & .148 & .588 & .222 & .077 \\
        Forward & 2.66 & .150 & .580 & .209 & .078 \\
        Backward & \underline{2.36} &  \underline{.115} & \underline{.607} & \underline{.274} & \underline{.074} \\
        IBP + Fo. + Ba. & \textbf{1.55} & \textbf{.089} & \textbf{.736} & \textbf{.428} & \textbf{.069} \\
        \bottomrule
        \vspace{-4mm}
        \end{tabular}
    }
    \caption{\textbf{Impact of the verified perturbation analysis method on EVA.}
    Results of \eva on Tightness, Deletion (Del.), Insertion (Ins.), Fidelity (Fid.) and $\rsr$ (Rob.) metrics obtained on MNIST. The Tightness score corresponds to the average adversarial surface. A lower Tightness score indicates that the method is more precise: it reaches tighter bound, resulting in better explanations and superior scores on the other metrics. The first and second best results are respectively in \textbf{bold} and \underline{underlined}.
    \vspace{-4mm}
    }
    \label{tab:eva:ablation_vpa}
\end{table}

The choice of the verified perturbation analysis method is a hyperparameter of EVA.
Hence, it is interesting to see the effect of the choice of this hyperparameter on the previous benchmark. 
We recall that only the MNIST dataset could benefit from the (IBP+Forward+Backward) combo. Table~\ref{tab:eva:ablation_vpa}  reports the results of the fidelity metrics using other verified perturbation analysis methods. We also report a tightness score which corresponds to the average of the \adv~: 
$\mathbb{E}_{\vx \sim \mathcal{X}}(\AOup(\vx, \ball))$.
Specifically, a low score indicates that the verification method is precise, meaning that the over-approximation is closer to the actual value. It should be noted that the true value is intractable, but remains the same across all three tested cases. We observe that the tighter the bounds, the higher the scores. This allows us to conjecture that the more scalable the formal methods will become, the better the quality of the generated explanations will be.
We perform additional experiments to ensure that the certified component of \eva~score is significant by comparing \eva~ to 
a sampling-based version of \eva. The details of these experiments are available in Appendix~\ref{ap:eva:benchmarks}.

\subsubsection{Targeted Explanations}
\label{sec:eva:targeted_explanations}

\begin{figure}[t!]
  \centering
  \includegraphics[width=0.9\textwidth]{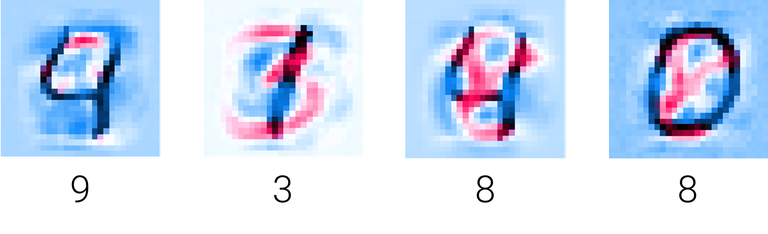}
  \caption{\textbf{Targeted explanations.} Generated explanations for a decision other than the one predicted by the model. The class explained is indicated at the bottom of each sample, e.g., the first sample is a `4' and the explanation is for the class `9'. As indicated in section~\ref{sec:eva:targeted_explanations}, the red areas indicate that a black line should be added and the blue areas that it should be removed. More examples are available in the Appendix.
  }
  \label{fig:eva:targeted_explanations}
\end{figure}

In some cases, it is instructive to look at the explanations for unpredicted classes in order to get information about the internal mechanisms of the models studied. Such explanations allow us to highlight contrastive features: elements that should be changed or whose absence is critical. 
Our method allows us to obtain such explanations: for a given input, we are then exclusively interested in the class we are trying to explain, without looking at the other decisions. Formally, for a given targeted class $c'$
the \adv~ (Equation~\ref{eq:eva:adv_surface}) become $\AO(\vx, \ball) = \max_{\substack{\v{\delta} \in \ball}} \f^{(c')}(\vx + \v{\delta}) - \f^{(c)}(\vx + \v{\delta})$. Moreover, by splitting the perturbation ball into a positive one $\ball^{(+)} = \big\{ \v{\delta} \in \ball ~:~ \v{\delta}_i \geq 0,~ \forall i \in \{1, ..., d \} \big\}$ and a negative one $\ball^{(-)} = \big\{ \v{\delta} \in \ball ~:~ \v{\delta}_i \leq 0,~ \forall i \in \{1, ..., d \} \big\}$, one can deduce which direction -- adding or removing the black line in the case of gray-scaled images -- will impact the most the model decision.

We generate targeted explanations on the MNIST dataset using (IBP+Forward+Backward). For several inputs, we generate the explanation for the 10 classes. Fig.~\ref{fig:eva:ap_targeted} shows 4 examples of targeted explanations, the target class $c'$ is indicated at the bottom. The red areas indicate that adding a black line increases the \adv~ with the target class. Conversely, the blue areas indicate where the increase of the score requires removing black lines.
All other results can be found in the Appendix.
In addition to favorable results on the fidelity metrics and guarantees provided by the verification methods, \eva~can provide targeted explanations that are easily understandable by humans, which are two qualities that make them a candidate of choice to meet the recent General Data Protection Regulation (GDPR) adopted in Europe~\cite{kaminski2021right}. More examples are available in the Appendix~\ref{ap:eva:targeted}.

\subsection{Conclusion}

In this work, we presented the first explainability method that uses verification perturbation analysis that exhaustively explores the perturbation space to generate explanations. We presented an efficient estimator that yields explanations that are state-of-the-art on current metrics. We also described a simple strategy to scale up perturbation verification methods to complex models. Finally, we showed that this estimator can be used to form easily interpretable targeted explanations.

We hope that this work will %
for searching for safer and more efficient explanation methods for neural networks -- and that it will inspire further synergies with the field of formal verification.

\clearpage

\section{How useful are attributions method ? A Meta-predictor perspective.}
\label{sec:attributions:metapred}

\newcommand{\metapred}{\bm{\psi}}
\newcommand{\metabase}{\bm{\psi}^{(0)}}
\newcommand{\method}{\explainer}
\newcommand{\complexity}{\textit{Complexity}}

\renewcommand{\mp}{Meta-predictor}

\newcommand{\expsa}{Saliency}
\newcommand{\expgi}{Gradient $\odot$ Input}
\newcommand{\expig}{Integrated Gradients}
\newcommand{\expsg}{SmoothGrad}
\newcommand{\expgc}{Grad-CAM}
\newcommand{\expoc}{Occlusion}

As we have seen so far, a multitude of explainability methods has been described to try to help users better understand how modern AI systems make decisions. However, most performance metrics developed and used in this manuscript to evaluate these methods have remained largely theoretical -- without much consideration for the human end-user. In particular, it is not yet clear (1) how useful current explainability methods are in real-world scenarios; and (2) whether current performance metrics accurately reflect the usefulness of explanation methods for the end user. To fill this gap, we conducted psychophysics experiments at scale ($n=1,150$) to evaluate the usefulness of representative attribution methods in three real-world scenarios. Our results demonstrate that the degree to which individual attribution methods help human participants better understand an AI system varies widely across these scenarios. This suggests the need to move beyond quantitative improvements of current attribution methods, towards the development of complementary approaches that provide qualitatively different sources of information to human end-users.

\subsection{Background.}

There is now broad consensus that modern AI systems might not be safe to be deployed in the real world~\cite{deel2021whitepaper} despite their exhibiting very high levels of accuracy on held-out data because these systems have been shown to exploit dataset biases and other statistical shortcuts~\cite{geirhos2020shortcut, d2020underspecification, shahamatdar2022deceptivelearning, fel2022aligning,moayeri2021sample,moayeri2022comprehensive,moayeri2022hard,moayeriexplicit}. A growing body of research thus focuses on the development of explainability methods to help better interpret these systems' predictions~\cite{ribeiro2016lime, sundararajan2017axiomatic, smilkov2017smoothgrad, petsiuk2018rise, selvaraju2017gradcam, linsley2018learning, fel2021sobol, eva2, novello2022making} to make them more trustworthy. The application of these explainability methods will find broad societal uses, like easing the debugging of self-driving vehicles~\cite{zablocki2021explainability} 
and helping to fulfill the ``right to explanation'' that European laws guarantee to its citizens~\cite{goodman2017european}.

In this work, we restrict our research on the most commonly used methods in eXplainable AI (XAI): attribution methods~\cite{simonyan2014deep, zeiler2014visualizing, ribeiro2016lime, selvaraju2017gradcam, sundararajan2017axiomatic, ancona2017better, smilkov2017smoothgrad,  petsiuk2018rise, fel2021sobol}. Despite a large array of methods, assessing the quality and reliability of these methods remains an open problem. So far the community has mostly focused on evaluating these methods using surrogate measures defined axiomatically such as fidelity or faithfulness measures.

\paragraph{Evaluations based on faithfulness measures.} Common approaches~\cite{samek2015evaluating, petsiuk2018rise} measure the faithfulness of an explanation through the change in the classification score when the most important pixels are progressively removed. The bigger the drop, the more faithful is the explanation method.
To ensure that the drop in score does not come from a change in the distribution of the perturbed images, the ROAR~\cite{hooker2018benchmark} methods include an additional step whereby the image classifier is re-trained between each removal step.
Because these methods do not require ground-truth annotations (i.e. object masks or bounding boxes), they are quite popular in computer vision~\cite{samek2015evaluating, petsiuk2018rise, fel2021sobol, fong2017perturbation, fong2019extremal, kapishnikov2019xrai} and natural language processing~\cite{arras2017explaining, arras2017relevant, fel2021sobol}.

Nevertheless, {\em faithfulness} measures have recently been criticized as they all rely on a baseline for removing important areas, a baseline that will obviously give better scores to methods relying internally on the same baseline~\cite{hsieh2020evaluations}. More importantly, they do not consider humans at any time in the evaluation. As a result, it is unclear if the most faithful attribution method is practically useful to humans.

\paragraph{Evaluations based on humans}
A second class of approaches consists in evaluating the ability of humans to leverage explanations for different purposes~\cite{ribeiro2016lime, selvaraju2017gradcam, mac2018teaching, chandrasekaran2018explanations, alufaisan2020does, biessmann2021quality, nguyen2021effectiveness, shen2020useful,nguyen2018comparing,hase2020evaluating,nguyen2022visual}.
\cite{ribeiro2016lime} were the first to evaluate the usefulness of explanations. Their work focused on the use case of bias detection: they trained a classifier on a biased dataset of wolves and huskies and found that the model consistently used the background to classify. They asked participants if they trusted the model before and after seeing the explanation for the model's predictions, and found that explanations helped detect bias here. We use a similar dataset to reproduce those results, but our evaluation differs greatly from theirs as we do not ask if participants trust the model but instead measure directly if they understand it. 

Closest to this work are~\cite{nguyen2018comparing,shen2020useful,hase2020evaluating, kim2021hive, sixt2022users}.~\cite{nguyen2018comparing,shen2020useful,hase2020evaluating} design their evaluation around the notion of \textit{simulatability}~\cite{kim2016examples, doshivelez2017rigorous}. They introduce different experimental procedures to measure if humans can learn from the explanations how to copy the model prediction on unseen data. Some provide the explanations at test time~\cite{nguyen2018comparing, shen2020useful}. Similar to us but for tabular data, \cite{hase2020evaluating} proposes to hide explanations at test time, this forces the participants to learn the rules driving the model's decision at training time where the explanations are shown. There are two limitations to their work: (1) they provide ground-truth labels associated with input images during training, (2) the participants see the same set of images without explanations, and then with explanations, always in that order. This creates learning effects that can heavily bias their results. We differ from their work by: (1) removing ground-truth labels from our framework as they serve no purpose and can bias participants, and (2) we have different participants go through the different conditions. This removes any learning effect, and more importantly, new explainability methods can be evaluated independently and still be compared to the previously evaluated methods. 
A recent study~\cite{kim2021hive} evaluated how AI systems may be able to assist human decisions by asking participants to identify the correct prediction out of four prediction-explanations pairs shown simultaneously. This measure reflects how well explanations help users tell apart correct from incorrect predictions. While the approach was useful to evaluate explanations in this specific scenario, it is not clear how this framework could be used to evaluate explainability methods more generally. Furthermore, when comparing different types of methods, they adapt the complexity of certain explainability methods to ease the task for participants. We argue that the complexity of explanations is an important property of explanations and that abstracting it away from the evaluation lead to unfair comparisons between methods.
In contrast, we propose a more general evaluation framework that can be used for any kind of explainability method without the need to adapt them for the evaluation procedure -- hence allowing for an unbiased and scalable comparison between methods.
Finally, \cite{sixt2022users} proposes to evaluate if users are able to identify important features biasing the predictions of a model using a synthetic dataset. By controlling the generation process of the dataset, they have access to the ground-truth attributes biasing the classifier, and can measure the accuracy of users at identifying these features. They evaluate if concept-based or counterfactual explanations help users improve over a baseline accuracy when no explanations are provided, and find no explanation tested to be useful. While both works highlights the importance of human evaluation, they differ in: the metrics employed (identifying relevant features for the model \textit{vs.} meta-prediction), the type of dataset used (synthetic \textit{vs.} real-world scenarios), and the type of methods evaluated (counterfactual and concept-based methods \textit{vs.} attribution methods).

The main contributions of this paper are as follows:
\begin{itemize}[leftmargin=*]
    \item We propose a novel human-centered explainability performance measure together with associated psychophysics methods to experimentally evaluate the practical usefulness of modern explainability methods in real-world scenarios.
    \item Large-scale psychophysics experiments $(n=1,150)$ revealed that SmoothGrad~\cite{smilkov2017smoothgrad} is the most useful attribution method amongst all tested and that none of the faithfulness performance metrics appear to predict if and when attribution methods will be of practical use to a human end-user. 
    \item Perceptual scores derived from attribution maps, characterizing either the complexity of an explanation or the challenge associated with identifying ``what'' features drive the system's decision, appear to predict failure cases of explainability methods better than faithfulness metrics.
\end{itemize}

\begin{figure}[t!] %
  \centerline{\includegraphics[width=0.99\textwidth]{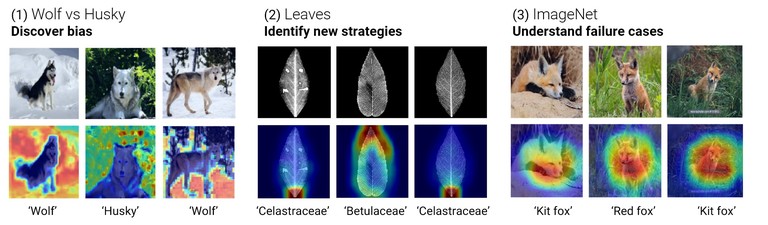}}
  \caption{
  We study the practical usefulness of recent explainability methods in three real-world scenarios, each corresponding to different use cases for XAI. 
  The first dataset is Husky vs. Wolf where the goal of the explanations is to help the user to identify a source of bias in a model (classification is based on the background (snow, grass) as opposed to the animal). 
  The second dataset corresponds to a real-world leaf classification problem which is complex for non-experts. The goal of the explanations is to help the end-user identify the strategy discovered by the vision system. 
  Finally, the third dataset is a subset of ImageNet, which consists of a collection of images where half have been misclassified by the system. The goal of the explanations here is to help the end-user understand the failure sources of a high performing model.   
  }
  \label{fig:metapred:big_picture}
\end{figure}

\begin{figure*}[t!]
  \includegraphics[width=0.99\textwidth]{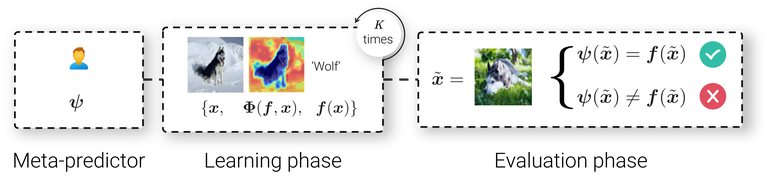}

  \caption{
  We describe a human-centered framework to evaluate explainability methods borrowing the concept of \mp. 
  The framework requires a black box model $\pred$ (the predictor), an explanation method $\method$ and a human subject $\metapred$ which will try to predict the predictor, hence, the name \mp.
  The first step is the learning phase where the \mp~is training using $K$ samples $\bm{x}$, together with the associated model predictions $\pred(\bm{x})$ and explanations $\method(\pred, \bm{x})$. 
  The goal of this learning phase is for the \mp~to uncover the rules driving the decisions of the model from the triplets $(\bm{x}, \method(\pred, \bm{x}), \pred(\bm{x}))$.
  Then, the second step is the evaluation phase where we test the Meta-predictor's ability to correctly predict the model's outputs on new samples $\bm{\tilde{x}}$ by comparing its predictions $\metapred(\bm{\tilde{x}})$ to those of $\pred(\bm{\tilde{x}})$.
  The \metric~score of the explanation method is then computed as the relative accuracy improvement of \mp~trained with vs. without explanations.
  }
  \label{fig:metapred:metapred}
\end{figure*}

\subsection{A \mp~ perspective}

Before proposing a rigorous definition of interpretability, let us motivate our approach with an example: a linear classifier is often considered to be readily interpretable because its inner working is sufficiently intuitive that it can be comprehended by a human user. A user can in turn build a mental model of the classifier -- predicting the classifier's output for arbitrary inputs. 
In essence, we suggest that the model is interpretable because the output can be predicted -- i.e, we say we understand the rules used by a model, if we can use those inferred rules to correctly predict its output.
This concept of predicting the classifier's output is central to our approach and we conceptualize the human user as a \textit{\mp}~of the machine learning model. This notion of \mp~is also closely related to the notion of \textit{simulatability}~\cite{doshivelez2017rigorous, hase2020evaluating, pruthi2020evaluating, kim2016examples, fong2017perturbation}. 
We will now define the term more formally. 

\label{sec:metapred:framework}

We consider a standard supervised learning setting where $\pred$ is a black-box predictor that maps an input $\bm{x} \in \mathcal{X}$ (e.g., an image) to an output $\pred(\bm{x}) \in \mathcal{Y}$ (e.g., a class label).
One of the main goals of eXplainable AI is to yield useful rules to understand the inner-working of a model $\pred$ such that it is possible to infer its behavior on unseen data points.
To correctly infer those rules, the usual approach consists in studying explanations (from Attribution Map, Concept Activation Vectors, Feature Visualization, \textit{etc..}) for several predictions.
Formally, $\method$ is any explanation functional which, given a predictor $\pred$ and a point $\bm{x}$, provides an information $\method(\pred, \bm{x})$ about the prediction of the predictor. In our experiments, $\method$ is an attribution method but we would like to remind that the framework is naturally adaptable to other explainability methods such as concept-based methods or feature visualization.

\paragraph{The understandability-completeness trade-off} 

Different attribution methods will typically produce different heatmaps -- potentially highlighting different image regions and/or presenting the same information in a different format. The quality of an explanation can thus be affected by two factors:  {\em faithfulness} of the explanation (i.e., how many pixels or input dimensions deemed important effectively drive the classifier's prediction) and the understandability of the explanation for an end-user (i.e., how much of the pattern highlighted by the explanation is grasped by the user).

At one extreme, an explanation can be entirely {\em faithful} and provide all the information necessary to predict how a classifier will assign a class label to an arbitrary image (i.e., by giving all the parameters of the classifiers). However, such information will obviously be too complex to be understood by a user and hence it is not {\em understandable}. 
Conversely, an explanation that overly simplifies the model might offer an approximation of the rule used by the model that will be more easily grasped by the user --a more {\em understandable} explanation-- but this approximation might ultimately mislead the user if it is not {\em faithful}. That is to say, just because a human agrees with the evidence pointed out by an explanation does not necessarily mean that it reflects how the model works.

Overall, this means that there is a trade-off between the amount of information provided by an explanation and its comprehensibility to humans. The most useful explanations should lie somewhere in the middle of this trade-off.

\paragraph{The {\em usefulness} metric} 

We describe a new human-centered measure that incorporates this trade-off into a single {\em usefulness} measure by empirically evaluating the ability of human participants to learn to ``predict the predictor'', i.e.,  to be an accurate \mp.
Indeed, if an explanation allows users to infer precise rules for the functioning of the predictor on past data, the correct application of these same rules should allow the user to correctly anticipate the model's decisions on future data. 
Scrutable but inaccurate explanations will result in an inaccurate \mp~ -- just like accurate inscrutable ones.
This \mp~framework avoids current pitfalls such as confirmation bias - just because a user likes the explanation does not mean they will be a better \mp~- or prediction leakage on the explanation - in simulatability experiments, as the explanation is available during the test phase, any explanation that leaks the prediction would have a perfect score, without giving us any additional information about the model. We will now formally describe the metric build using this framework.

We assume a dataset\footnote{We note that, in this paper, we only considered binary dataset --Class 1 vs Class 2-- because having the participants classify more than 2 classes would increase their cognitive load and bring unnecessary difficulty to the task. Nonetheless, any dataset could have been used as classification problems with more than 2 classes can always be trivially reformulated as Target class vs. Other / binary classification problems, instead of Class 1 vs Class 2, without lack of generality.} $\mathcal{D} = \{(\bm{x}_i, \pred(\bm{x}_i), \method(\pred, \bm{x}_i) \}_{i=1}^K$ used to train human participants to learn to predict a classifier's output $\pred$ from $K$ samples made of an input image $\bm{x}_i$, the associated  predictions $\pred(\bm{x}_i)$ and explanations $\method(\pred, \bm{x}_i)$. 
We denote $\metapred^{(K)}$ a human \mp~after being trained on the dataset $\mathcal{D}$ (see Fig.~\ref{fig:metapred:metapred}) using explanations. In addition, let $\metabase$ be the human \mp~after participants were trained on the same dataset but without explanations to offer baseline accuracy scores. 
We can now define the usefulness of an explainability method $\method$ after training participants on $K$ samples through the accuracy score of the \mp~normalized by the baseline \mp~accuracy:

\begin{equation}
    \label{eq:metapred:metric_k}
    \metrick = \frac{
    \mathbb{P}(\metapred^{(K)}(\bm{x}) = \pred(\bm{x}))
    }{
    \mathbb{P}(\metabase(\bm{x}) = \pred(\bm{x}))
    }
\end{equation}

with $\mathbb{P}(\cdot)$ the probability over a test set. Thus, \metrick~score measures the improvement in accuracy that the explanation has brought. It is important to emphasize that this \metric~measure only depends on the classifier prediction and not on the ground-truth label as recommended by~\cite{jacovi2020towards}. After fixing the number of training samples $K$, we compare the normalized accuracy of different Meta-predictors. The \mp~with the highest score is then the one whose explanations were the most useful as measures compared to a no-explanation baseline. 

\paragraph{\metric~metric} In practice, we propose to vary the number of observations $K \in \{ K_0, ..., K_n \}$ and to report an aggregated $\metric$~score by computing the area under the curve (AUC) of the \metrick. 
The higher the AUC the better the corresponding explanation method is. Formally, given a curve represented by a set of $n$ points $\mathcal{C} = \{ (K_0, \metrick_0), ...,  (K_n, \metrick_n) \}$ where $K_{i-1} < K_i$ we define the metric as $\metric = AUC(\mathcal{C})$. 

\subsection{Experimental design}
\label{sec:metapred:exp_design}

We first describe how participants were enrolled in the study, then our general experimental design (See SI for more informations).

\paragraph{Participants} Behavioral data were gathered from $n=1,150$ participants using Amazon
Mechanical Turk (AMT) ({\url{www.mturk.com}}). 
All participants provided informed consent electronically and were compensated $\$1.4$ for their time ($\sim 5 - 8$ min). The protocol was approved by the University IRB and was carried out in accordance with the provisions of the World Medical Association Declaration of Helsinki. For each of the three tested datasets, we ensured that there was a sufficient number of participants after filtering out uncooperative participants ($n = 240$ participants, 30 per condition, 8 conditions) to guarantee sufficient statistical power (See SI for details).
Overall, the cost of evaluating one method using our benchmark is relatively modest (\$50 per test scenario). 

\paragraph{General study design}

It included 3 conditions: an experimental condition where an explanation is provided to human participants during their training phase (see Fig.~\ref{fig:metapred:metapred}), a baseline condition where no explanation was provided to the human participants, and a control condition where a bottom-up saliency map~\cite{itti2005bottomup} was provided as a non-informative explanation. This last control is critical, and indeed lacking from previous work~\cite{hase2020evaluating, ribeiro2016lime}, because it  provides a control for the possibility that providing explanations along with training images simply increases participants' engagement in the task. As we will show in Sec.~\ref{sec:metapred:results}, such non-informative explanations actually led to a decrease in participants' ability to predict the classifier's decisions -- suggesting that giving a wrong explanation is worse than giving no explanations at all.

Each participant was only tested on a single condition to avoid possible experimental confounds. 
The main experiment was divided into 3 training sessions (with 5 training samples in each) each followed by a brief test. In each individual trial, an image was presented with the associated prediction of the model, either alone for the baseline condition or together with an explanation for the experimental and control condition. After a brief training phase (5 samples), participants' ability to predict the classifier's output was evaluated on 7 new samples during a test phase. During the test phase, no explanation was provided to limit confounding effects: one possible effect is if the explanation leaks information about the class label.\footnote{Imagine an attribution method that would solely encode the classifiers' prediction. Participants would be able to guess the classifier's prediction perfectly from the explanation but the explanation per se would not help participants understand how the classifier works.}
We also propose to use a reservoir that subjects can refer to during the testing phase to minimize memory load as a confounding factor which was reported in~\cite{hase2020evaluating} (see SI for an illustration).

\paragraph{Datasets and models} 

We performed three distinct experiments in total -- using a variety of neural network architectures and $6$ representative attributions methods. Each of these experiments aimed at testing the usefulness of the explanation in a different context.

\begin{table}[!t]
    \centering
    \resizebox{1.00\textwidth}{!}{%
        \begin{tabular}{|c|cccc|cccc|cccc|}
        \toprule
         Method & \multicolumn{4}{c|}{Husky vs. Wolf} & \multicolumn{4}{c|}{\textit{Leaves}} & \multicolumn{4}{c|}{\textit{ImageNet}} \\
         \midrule
        Session~n$^{\circ}$               & 1 & 2 & 3 & \metric & 1 & 2 & 3 & \metric & 1 & 2 & 3 & \metric \\
         \midrule
        Baseline                    & 55.7 & 66.2 & 62.9 &  &       70.1 & 76.8 & 78.6 &  &       58.8 & 62.2 & 58.8 &  \\
        Control                             & 53.3 & 61.0 & 61.4 & 0.95 &       72.0 & 78.0 & 80.2 & 1.02 &     60.7 & 59.2 & 48.5 & 0.94 \\
        \midrule
        Saliency~\cite{simonyan2014deep}                  & 53.9 & 69.6 & 73.3 & 1.06  &        83.2 & 88.7 & 82.4 & \textbf{1.13} &      61.7 & 60.2 & 58.2 & 1.00 \\ 
        Integ.-Grad.~\cite{sundararajan2017axiomatic}     & 67.4 & 72.8 & 73.2 & 1.15 &      82.5 & 82.5 & 85.3 & \textbf{1.11} &       59.4 & 58.3 & 58.3 & 0.98\\
        SmoothGrad~\cite{smilkov2017smoothgrad}           & 68.7 & 75.3 & 78.0 & \textbf{1.20} &    83.0 & 85.7 & 86.3 & \textbf{1.13} &       50.3 & 55.0 & 61.4 & 0.93 \\
        GradCAM~\cite{selvaraju2017gradcam}               & 77.6 & 85.7 & 84.1 & \textbf{1.34} &       81.9 & 83.5 & 82.4 & 1.10 &      54.4 & 52.5 & 54.1 & 0.90 \\
        Occlusion~\cite{zeiler2014visualizing}            & 71.0 & 75.7 & 78.1 & \textbf{1.22} &       78.8 & 86.1 & 82.9 & 1.10 &     51.0 & 60.2 & 55.1 & 0.92 \\
        Grad.-Input~\cite{shrikumar2017learning}               & 65.8 & 63.3 & 67.9 & 1.06 &      76.5 & 82.9 & 79.5 & 1.05 &      50.0 & 57.6 & 62.6 & 0.95 \\
        \bottomrule
        \end{tabular}%
    }
    \caption{\textbf{\metrick~and \metric~scores}. \metrick scores across the 3 sessions for each attribution method, for each of the 3 datasets considered, followed by the \metric~scores. Higher is better. The \metric~scores of attribution methods that are statistically significant are \textbf{bolded}.}
    \label{tab:metapred:big_tab}
\end{table}

Our first scenario focuses on the detection of biases in AI systems using the popular Wolf vs. Husky dataset from~\cite{ribeiro2016lime} where an evaluation measure was already proposed around the usefulness of explanations for humans to detect biases. This makes it a good control experiment to measure the effectiveness of the framework proposed in Sec.~\ref{sec:metapred:framework}.
For this first experiment, we used the same model as in the original paper: InceptionV1~\cite{szegedy2015going}, and a similar dataset of Husky and Wolf images to bias the model. In this situation where prior knowledge of subjects can affect their \mp~score, we balance data correctness ($50\%$ of correct/incorrect examples shown). Therefore, a subject relying only on their prior knowledge will  end up as a bad \mp~of the model. For this experiment, the results come from $n=242$ subjects who all passed our screening process.

In our second scenario, we focus on a representative challenging image categorization task which would be hard to solve by a non-expert untrained human participant and the goal is for the end-user to understand the strategy that was discovered by the AI system. Here, we chose the leaf dataset described in~\cite{wilf2016computer}. We selected 2 classes from this dataset (Betulaceae and Celastracea) that could not be classified by shape to reduce the chances that participants will discover the solution on their own -- forcing them instead to rely on non-trivial features highlighted by the explanations (veins, leaf margin, etc). This scenario is far from being artificial as it reflects a genuine problem for the paleobotanist~\cite{spagnuolo2022decoding}. Can explainability methods help non-specialists discover the strategies discovered by an AI system? As participants are lay people from Amazon Mechanical Turk we do not expect them to be experts in botany, therefore we did not explicitly try to control for prior knowledge. In this experiment, $n=240$ subjects passed all our screening and other filtering processes. 

Finally, our last scenario focuses on identifying cases where an AI system fails\footnote{We acknowledge the existence of some overlap between the scenario 1 and scenario 3 as bias detection is a special case of a failure case. The reason we still use scenario 1 is because of the work previously done on it, allowing us  to validate our framework.} using ImageNet~\cite{imagenet_cvpr09}, also used in previous explainability work~\cite{fong2017perturbation,elliott2021perceptualball,hooker2018benchmark,fel2021sobol, shen2020useful,nguyen2021effectiveness}. We used this dataset because we expect it to be representative of real-world scenarios where it is difficult to understand what the model relies on for classification which makes it very difficult to understand these failure cases. 
Moreover, previous work has pointed out that attribution methods are not useful on this dataset ~\cite{shen2020useful}, we have thus chosen to extend our analysis to this particular case.
We use a ResNet50~\cite{he2016deep} pretrained on this dataset as predictor. Because prior knowledge is a major confounding factor on ImageNet, we select a pair of classes that was heavily miss-classified by the model, to be able to show subjects 50\% of correct/incorrect predictions: the pair Kit Fox and Red Fox fits this requirement.
In this experiment, we analyzed data from $n=241$ participants who passed our screening and filtering processes.

For all experiments, we compared $6$ representative attribution methods: \expsa~(SA)~\cite{simonyan2014deep}, 
\expgi~(GI)~\cite{ancona2017better}, 
\expig~(IG)~\cite{sundararajan2017axiomatic},
\expoc~(OC)~\cite{zeiler2014visualizing},
\expsg~(SG)~\cite{smilkov2017smoothgrad} and 
\expgc~(GC)~\cite{selvaraju2017gradcam}.
Further information on these methods can be found in SI.
Table \ref{tab:metapred:big_tab} summarizes all the results from our psychophysics experiments. 

\begin{figure*}[t!]
    \includegraphics[width=0.48\textwidth]{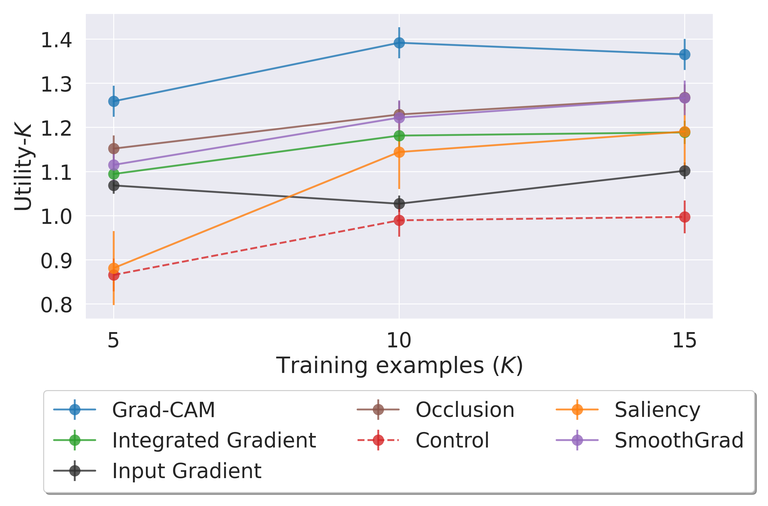}
    \includegraphics[width=0.48\textwidth]{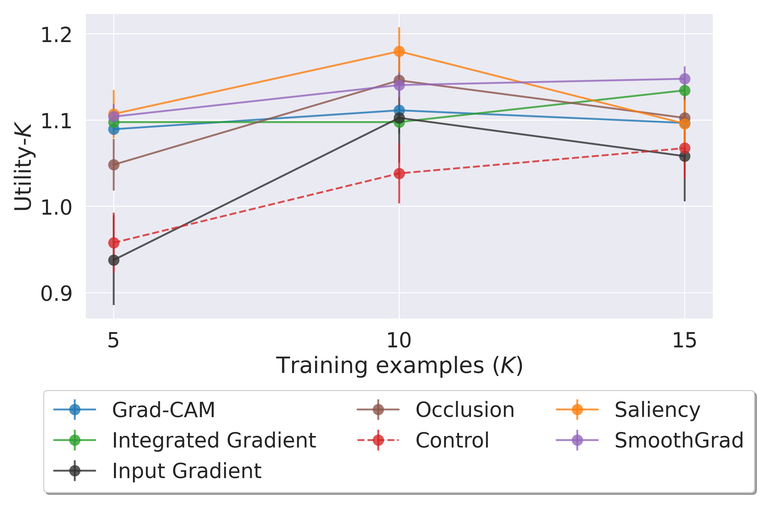}
    \caption{\textbf{\metrick~for both Husky vs. Wolf (left) and the Leaves (right) dataset.} 
    The \metrick~of the explanation, or the accuracy of the human \mp~after training, is measured after each training session (3 in total) for the scenario (1) of bias detection (on the left) and the scenario (2) concerning the identification of new strategies. 
    Concerning the first scenario, all methods have a positive effect on the score obtained - they improve the subjects' ability to predict the model - and are thus useful to better understand the model. \expgc, \expoc~and \expsg~are particularly useful for bias detection. On the Leaves dataset~\cite{wilf2016computer}, explanations are also useful, but specifically \expsa, \expsg~and \expig.
    }
    \label{fig:metapred:utility}
\end{figure*}

\subsection{Results}
\label{sec:metapred:results}

\paragraph{Scenario (1): Bias detection}

Fig.~\ref{fig:metapred:utility} shows the \metrick~scores for each method after different numbers of training samples were used to train participants for the biased dataset of Husky vs. Wolf.
The \metric~score encodes the quality of the explanations provided by a method, the higher the score, the better the method, with the baseline score being 1 (every score is divided by the baseline score corresponding to human accuracy after training without explanations).

A first observation is that the explanations have a positive effect on the \metrick~score: the explanation allows participants to better predict the model's decision (as the \metric~scores are above 1). These results are consistent with those reported in ~\cite{ribeiro2016lime}. This is confirmed with an Analysis of Variance (ANOVA) for which we found a significant main effect, with a medium effect size ($F(7, 234) = 9.19,\ p < .001,\ \eta^2 = 0.089$). 
Moreover, the only score below the baseline is that of the control explanation, which do not make use of the model. 
We further explore our results by performing pairwise comparisons using Tukey’s Honestly Significant Difference~\cite{tukey1949comparing} to compare the different explanations against the baseline. We found 3 explainability methods to be significantly better than the baseline: \expgc~($p<0.001$), \expoc~($p = 0.01$) and \expsg~($p = 0.034$).
Thus, participants who received the \expgc, ~\expoc~ or ~\expsg~  explanations performed much better than those who did not receive them. 

\paragraph{Scenario (2): Identifying an expert strategy}

In Fig.~\ref{fig:metapred:utility}, we show results on the Leaves dataset. An ANOVA analysis across all conditions revealed a significant main effect, albeit small ($F(7, 232) = 4.29,\ p < .001,\ \eta^2 = 0.042$).
This implies that explanation also had a positive effect resulting in better \mp~ in this use case.
A Tukey’s Honestly Significant Difference test suggests that the best explanations are \expsa~, \expsg~and \expig~as they are the only ones to be significantly better than our baseline (WE) ($p = .004$, $p = .007$ and $p = .03$ respectively). An interesting result is that \expsg~seems to be consistently useful across both use cases where explanations are indeed practically useful.
A more surprising result is that \expsa~which was one of the worst explanations for bias detection, is now the best explanation on this use case (We discuss possible reasons in SI).

\paragraph{Scenario (3): Understanding failure cases}

Table~\ref{tab:metapred:big_tab} shows that, on the ImageNet dataset, none of the methods tested exceeded baseline accuracy.  Indeed, the experiment carried out, even with an improved experimental design, led us to the same conclusion as previous works~\cite{shen2020useful}: none of the tested attribution methods are useful (ANOVA: $F(7, 233) = 1.26,\ p>.05$). 
In the use case of understanding failure cases on ImageNet, \textbf{no attributions methods} seem to be useful.

\subsection*{Why do attribution methods fail?}
After studying the usefulness of attribution methods across 3 real-world scenarios for eXplainable AI, we found that attribution methods help, sometimes, but not always. We are interested in better understanding why sometimes attribution methods fail to help. Because this question has yet to be properly studied, there is no consensus if we can still make attribution methods work on those cases with incremental quantitative improvements. In the follow-up sections we explore 3 hypothesis to answer that question. 

\paragraph{Faithfulness as a proxy for Utility?} 
\begin{figure}[ht]
  \centering
  \begin{center}
    \includegraphics[width=0.8\textwidth]{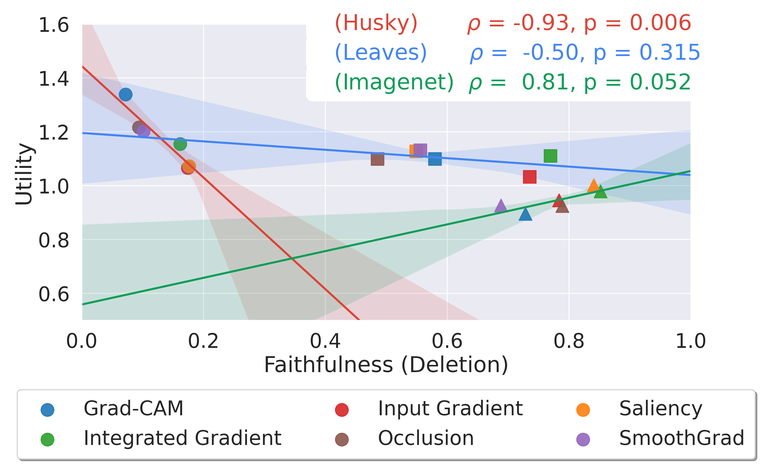}
  \end{center}

  \caption{\textbf{\metric~vs Faithfulness correlation.}
        The results suggest that current faithfulness metrics are poor predictors of the end-goal usefulness of explanation methods. 
        Concerning the ImageNet dataset (triangle marker), the \metric~scores are insignificant since none of the methods improves the baseline.
        }
    \label{fig:metapred:correlations_plot}
\end{figure}

Faithfulness is often described as one of the key desiderata for a good explanation~\cite{aggregating2020,yeh2019infidelity,fel2020representativity}. If an explanation fails to be sufficiently faithful, the rules it highlights won't allow a user to understand the inner-working of the model. Thus, a lack of faithfulness on ImageNet could explain our results. 
To test this hypothesis, we use the  {\em faithfulness} metrics: Deletion\cite{samek2015evaluating, petsiuk2018rise},  commonly used to compare attribution methods~\cite{samek2015evaluating, petsiuk2018rise, fel2021sobol, fong2017perturbation, fong2019extremal, kapishnikov2019xrai}. A low Deletion score indicates a good {\em faithfulness}, thus for ease of reading we report the  {\em faithfulness} score as $1 -$ Deletion such that a higher {\em faithfulness} score is better. 

Fig \ref{fig:metapred:correlations_plot} shows the linear relationship between our Utility metrics and the faithfulness scores computed for every attribution method across all 3 datasets. We observe two main trends: 1) \textbf{There does not appear to be any specific pattern regarding faithfulness  that could explain why attribution methods are not useful for ImageNet}, and 2) the least useful attribution methods for both use cases for which methods help (Bias and Leaves) are some of the leading methods in the field measured by the faithfulness metric.
We also found a weak, if maybe anti-correlated, relation between faithfulness and usefulness: just focusing on making attribution methods more faithful does not translate to having methods with higher practical usefulness for end-users. And, in fact, focusing too heavily on faithfulness seem to come at the expense of usefulness, resulting in explanations that are counter-intuitively less useful. This second observation may seems rather alarming for the field given that the faithfulness measure is one of the driving benchmarks.

\paragraph{Are explanations too complex?} 

Using the trade-off between completeness and understandability previously discussed in Section~\ref{sec:metapred:framework}, we formulate another hypothesis: some explanations may be faithful  but too complex and therefore cannot be understood by humans. In that view, an explanation with low complexity would tend to be more useful.

\begin{figure}[ht]
  \begin{center}
    \includegraphics[width=0.8\textwidth]{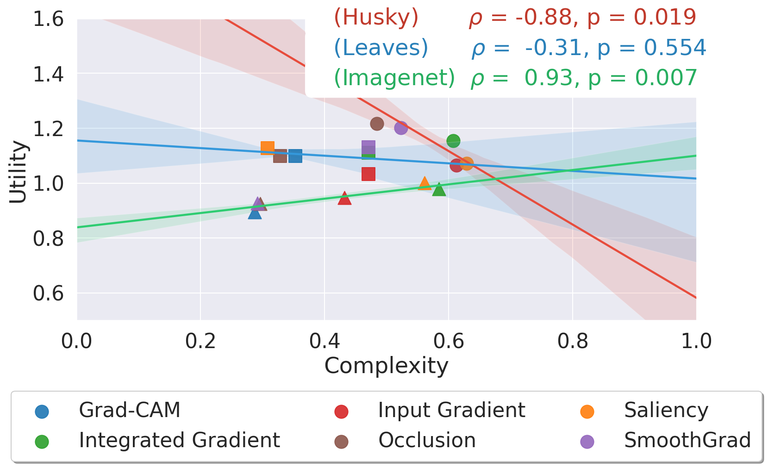}
  \end{center}
  \caption{\textbf{\metric~vs Complexity correlation.} 
  This suggest a weak but possibly existing link between \complexity~ and \metric~ scores. 
    }
    \label{fig:metapred:complexity}
\end{figure}

As a simple measure of the complexity of visual explanations, it would be ideal to be able to compute the Kolmogorov complexity~\cite{li2004similarity} of each explanation. It was shown in previous work~\cite{da2011image} to correlate well with human-derived ratings for the complexity of natural images~\cite{forsythe2008confounds,forsythe2009visual}. As suggested by~\cite{li2004similarity,de2006approximating} we used a standard compression technique (JPEG) to approximate the Kolmogorov complexity.
Fig.~\ref{fig:metapred:complexity} shows the \metric~vs complexity score of attribution methods for each dataset. For one of the datasets where attribution methods help, the results suggest the presence of a strong correlation between \textit{usefulness} and \textit{complexity}: the least complex method is the most useful to end-users. For the other datasets, the results are either not conclusive (Leaves), or are not relevant as methods are not useful (ImageNet). 
Overall, \textbf{across datasets there is no significant difference in the complexity of explanations} that can explain why attribution methods do not help on ImageNet.
This could be because the Kolmogorov Complexity does not perfectly reflect human visual complexity, or because this is not the key element to explain failure cases of attribution methods.

\paragraph{An intrinsic limitation of Attribution methods?} 

The role of attribution methods is to  help identify ``where'' to look in an image to understand the basis for a system's prediction. However, attribution methods do not tell us ``what'' in those image regions is driving decisions. For categorization problems which involve perceptually similar classes (such as when discriminating between different breeds of dogs) and fine-grained categorization problems more generally, simply looking at diagnostic image regions tells the user very little about the specific shape property being relevant. For instance, knowing that the ear shape is being used for recognition does not say what specific shape feature is being encoded (e.g., pointed vs. round or narrow vs. broad base, etc). Our main hypothesis is that such a lack of explicit ``what'' information is precisely what is driving the failure of attribution methods on our ImageNet use-case.
\begin{figure}[ht]
    \centering
    \includegraphics[width=0.8\textwidth]{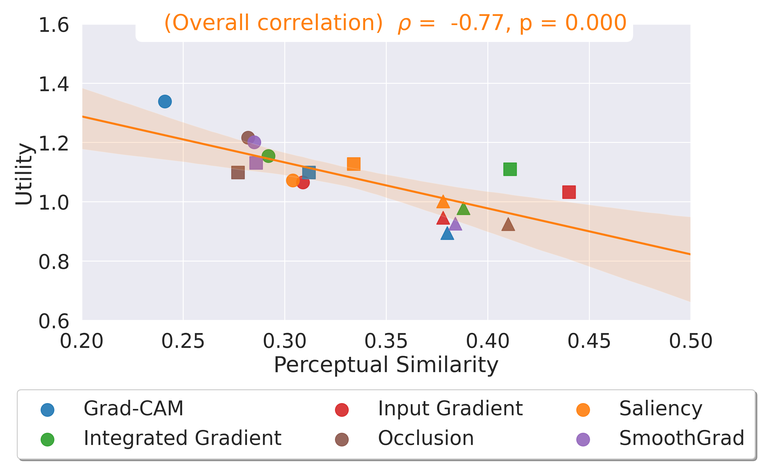}
    \caption{\textbf{\textit{Perceptual Similarity} scores vs. \metric.} The perceptual similarity of highlighted regions by a given attribution method for both classes is measured, for each method, for each dataset. We observe a strong correlation between the perceptual similarity of features highlighted for both classes and the practical usefulness of methods.
    }
    \label{fig:metapred:similarity}
\end{figure}

To test this hypothesis, we estimated the perceptual similarity between classes measured within diagnostic regions (see SI for more details) using the Learned Perceptual Image Patch Similarity (LPIPS) metric~\cite{zhang2018unreasonable} as it has been shown to approximate human perceptual similarity judgments well~\cite{zhang2018unreasonable, nanda2021exploring}. We report the perceptual similarity score as 1 - LPIPS score so that a high score means a high similarity.
Fig.~\ref{fig:metapred:similarity} shows the correlation between the perceptual similarity scores vs. our \metric~scores on all methods and datasets studied. Our results suggest a strong correlation  between perceptual similarity and practical usefulness: the more perceptually similar discriminative features of both classes are, the less useful attribution methods become. More importantly, the results across datasets show that on ImageNet, where attribution methods do not help, every method has a high similarity score. This result suggests that \textbf{after a certain threshold of perceptual similarity, attribution methods might no longer be useful, no matter how faithful or low in complexity the explanation is.} 
Overall, the results suggest that the perceptual similarity of discriminative features could explain why attribution methods fail on ImageNet.

\subsection{Discussion \& Hypothesis}

In summary, we conducted a large-scale human psychophysics experiment to test the utility of explainability methods in real-world scenarios. 
Our work shows that in two of the three tested scenarios (bias detection and identification of new strategies), explainability methods have indeed progressed and they provide meaningful assistance to human end-users. 
Nevertheless, we identified a scenario (understanding failure case) for which none of the tested attribution methods were helpful. This result is consistent with previous work~\cite{shen2020useful} and highlights a fundamental challenge for XAI.

Further analysis of associated faithfulness performance metrics driving the development of explainability methods revealed that they did not correlate with our empirical measure of utility -- suggesting that they might not be suited anymore to move the field forward.
We also investigated the possibility that the complexity of individual explanations may play a role in explaining human failures to learn to leverage those explanations to understand the model and, while we found a weak correlation between complexity and our empirical measure of utility, this correlation appears too low to explain the failure of these methods.

Finally, because attribution methods appear to be just as faithful and low in complexity whether they are useful or not, we explored the possibility that their failure lies, not in the quality of their explanations, but in the intrinsic limitations of attribution methods. If fully grasping the strategy of a model requires understanding, not just ``where'' to look (as revealed by attribution maps) but also ``what'' to look at, something not currently revealed by these methods, attribution methods will not help. Our assumption is that the need for finer ``what'' information should arise when diagnostic image locations across classes look perceptually very similar and potentially semantically related for certain classification problems (e.g., looking at the ears or the snout to discriminate between breeds of cats and dogs) and one needs to identify what visual features are driving decisions. We computed a perceptual score for classification problems by estimating the perceptual similarity between diagnostic image regions (as predicted by attribution methods) and found that, indeed, when this score predicts a certain level of perceptual similarity between classes, attribution methods fail to contribute useful information to human users, regardless of the faithfulness or complexity of the explanations. This suggests that explainability methods may need to communicate additional information to the end user beyond attribution maps.

\clearpage

\section{Conclusion}

This chapter has provided a detailed examination of attribution methods in explainability, covering gradient-based, internal, and black-box approaches. 
Initially, we explored a novel metric inspired by Algorithmic Stability to assess the quality of explanations for a given model. This investigation revealed that fidelity metrics are not enough and that robust models tend to offer more general and consistent explanations.
Subsequently, we introduced the Sobol method, an efficient black-box attribution technique grounded in global sensitivity analysis. This method identify significant pixel regions via perturbation and use quasi-Monte Carlo sampling, marking a notable advancement in computational efficiency while maintaining strong theoretical grounding in Global Sensitivity Analysis.
Additionally, we discussed Explainability with Verified Perturbation Analysis (\eva), which introduces formal guarantees to importance estimation, thereby enhancing trust in the insights derived from models.
To finish, the chapter evaluated the practical utility of attribution in real-life scenarios and highlighted a fundamental shortcoming of current methods: their tendency to falter in complex situations. This shortfall underscores a need for further development in our understanding of models.

\begin{figure}[ht]
    \centering
    \includegraphics[width=1.0\textwidth]{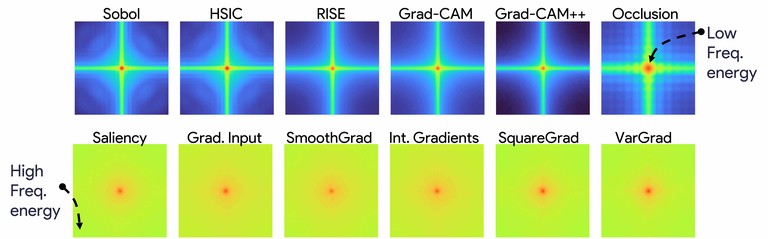}
    \caption{\textbf{Fourier footprint of attribution methods, adapted from \cite{muzellec2023gradient}.} We show on the top row the Fourier spectrum of prediction-based attribution methods and of the gradient-based methods on the bottom row, computed with a ResNet50. The two families can be distinguished by methods but also by their signature in the Fourier domain. The former method has magnitudes largely concentrated in the low frequencies, while the latter is more spread out: it features non-trivial magnitudes almost everywhere, including in high frequencies.}
    \label{fig:attribution:spectrum}
\end{figure}

\paragraph{Additional Remarks.} Through this research, several noteworthy observations about attribution methods were made:

\begin{itemize}
    \item \textbf{Explainable AI need to adapt GSA tools.} Global Sensitivity Analysis (GSA) is already a mature field and a fertile research area that \textit{could} significantly contribute to Explainable AI (XAI), especially in attribution methods. Our work with Sobol is merely a starting point, and the active exploration of GSA in XAI, including recent advancements using Hilbert Schmidt Independence Criterion (HSIC)~\cite{novello2022making}, promises further reductions in computational time while improving interpretability scores. The integration of kernels or novel tools~\cite{da2015global,sarazin2023new} from this research domain could potentially fuel an entire thesis. 

    \item \textbf{Attribution methods are not always consistent.} On a more practical note, after inspection of thousands of heatmaps, certain methods appear to cluster together, such as gradient-based methods (e.g., Saliency and SmoothGrad) on one hand, and black-box methods (e.g., RISE, Occlusion) on the other, with Grad-CAM and Grad-CAM++ forming another cluster. Each cluster tends to offer similar explanations. During analysis, prompting a diversified approach when examining results should be recommended, ideally combining methods from different clusters, e.g. SmoothGrad, Sobol, and Grad-CAM.

    \item \textbf{A frequency perspective on Attribution is promising.} Another observation that could be made is the significant variance in the frequency spectrum of methods -- as showcased in \autoref{fig:attribution:spectrum}. The impact of high-frequency energy on interpretability remains unclear. An initial investigation in \autoref{sec:attributions:metapred} with the complexity measure is a starting point, as well as the recent work of \cite{muzellec2023gradient}. Analyzing methods from a frequency perspective could offer valuable insights into attribution methods.

\end{itemize}

As we conclude this chapter, it's evident that attribution methods have not fully resolved the spectrum of use-cases in explainability. To move forward effectively, we propose utilizing the conclusion of the meta-prediction metric and hypothesize the reasons behind the shortcomings of attribution methods, and then suggest solutions that verify these hypotheses.

\paragraph{Hypotheses on the Shortcomings of Attribution Methods.} We posit two hypotheses for the failures of attribution methods, attributing them either to the model itself or to the explanation method -- implying that solutions may involve changing the model or the explainability method.

The first hypothesis questions the necessity of altering models. Despite the goal of explainability to elucidate any system, we could easily admit that among models achieving the same accuracy on benchmarks like ImageNet, they might vary in interpretability score (such as the \metric~score proposed). This is the idea behind \textit{Rashomon set}~\cite{xin2022exploring}: among the set of predictor with the same accuracy level $\kappa$ defined as $\{ \f \in \fspace : \P_{\rvx, \ry}(\f(\rvx) = \ry) = \kappa \}$ some of them may use strategies that are more aligned with humans (or easier to \textit{meta-predict}). This hypothesis leads us to ask, \textit{``Among high-performing models, how can we identify the most explainable ones?''} We will explore potential strategies to address this model alignment hypothesis in the \autoref{chap:alignment}.

\begin{customhypothesis}{Hypothesis 1: Model Alignment.}
\label{hyp:alignment}
Attribution methods fail in various scenarios due to the model's fault. The model might employ processes and strategies too divergent from human reasoning. Aligning the model with human understanding could make these processes more transparent.
\end{customhypothesis}

Our second hypothesis defends the notion that attribution methods are inherently limited. A commonly discussed limitation is the difference between the "\where" and the "\what"—that is, attribution methods show \where~the model focuses but not \what~ it perceives. Future research could aim to develop methods that explain the very nature of the features influencing the model's decisions. This approach has begun to be explored, and we dedicate \autoref{chap:concepts} to this investigation, formally stating this hypothesis:

\begin{customhypothesis}{Hypothesis 2: The Need to Go Beyond Attributions.}
\label{hyp:what}
Attribution methods fail in various scenarios because they do not convey sufficient information. They only show where the model looks, leaving humans to guess the type of features seen by the model. This can lead to ambiguities and misalignment between what humans think the model uses and what it actually does.
\end{customhypothesis}

Moving forward, we will propose methods to address these two hypotheses in \autoref{chap:alignment} and \autoref{chap:concepts}, respectively.

\clearpage

\chapter{Alignment}
\label{chap:alignment}

\begin{chapterabstract}
\textit{
This chapter confronts a problem highlighted in the preceding chapter: Can we build models that align more closely with human cognition to amplify our understanding of their mechanisms ? 
Our investigation commences in \autoref{sec:harmonization}, where we study the Alignment issue through the prism of explainability. Our methodology is inspired by \cite{linsley2018learning}'s seminal work, which amassed a large dataset via psychophysical experiments. Utilizing explainability techniques, we craft a metric to gauge the similarity between model and human explanations. Initially, our metric reveals an unexpected pattern: there is a discernible trend where higher model performance correlates with decreased alignment with human in terms of explanation. In simpler terms, the more performant the models, the less they align with human explanations, evidenced by a divergence from human-generated attribution maps or heatmaps.
In response, we introduce a novel training paradigm designed to synchronize machine learning models' explanations mechanisms with human heatmaps called ``Click-maps''. We leverage attribution methods -- particularly their differentiable characteristic -- to directly steer the model's focus towards alignment with human. 
We observe that this apparent performance-alignment dichotomy can be effectively addressed with our ``harmonization'' training approach. This technique not only increase alignment between models and human explanations but also enhances model accuracy.
In \autoref{sec:lipschitz}, we explore an alternative thesis pathway: the study of robust models, with a focus on 1-Lipschitz networks. Remarkably, we discover that these networks inherently exhibit a greater alignment with human explanations, bypassing the need for specialized training routines.
}
\end{chapterabstract}

The work in this chapter has led to the publication of the following conference papers:
{\small{
\begin{itemize}

    \item \textbf{Thomas Fel}\equal, Ivan F Rodriguez\equal, Drew Linsley\equal, Thomas Serre, (2022). \textit{``Harmonizing the object recognition strategies of deep neural networks with humans''.} In: \textit{Advances in Neural Information Processing Systems}  (\textcolor{confcolor}{NeurIPS})

    \item Mathieu Serrurier, Franck Mamalet, \textbf{Thomas Fel}, Louis Béthune, Thibaut Boissin, (2023). \textit{``On the explainable properties of 1-Lipschitz Neural Networks: An Optimal Transport Perspective''.} In: \textit{Advances in Neural Information Processing Systems} (\textcolor{confcolor}{NeurIPS})

    \item Drew Linsley, Ivan F Rodriguez, \textbf{Thomas Fel}, Michael Arcaro, Saloni Sharma, Margaret Livingstone, Thomas Serre, (2023). \textit{``Performance-optimized deep neural networks are evolving into worse models of inferotemporal visual cortex''.} In: \textit{Advances in Neural Information Processing Systems} (\textcolor{confcolor}{NeurIPS}).

\end{itemize}
}}

\minitoc
\clearpage

\section{Introduction}

\textit{Are there shared strategies between neural networks and humans?} This question has long motivated neuroscientists, cognitive scientists, and machine learning researchers. It is fascinating to observe that although today's neural networks outperform humans in many tasks, certain anomalies, such as adversarial examples~\cite{goodfellow2014explaining}, strongly suggest that the underlying mechanisms of these networks are not entirely the same as those of humans; they are not \textit{aligned}.

In the previous chapter, we hypothesized that a model more closely aligned with human understanding would be more interpretable. What remains to be clarified is our definition of alignment. This research field draws from neuroscience, cognitive science, and machine learning and has thus many definitions, depending on the domain. However, a common point is that Alignment aims to (1) measure how closely the internal representations of two systems match and (2) if possible, to correct any differences between them. Recently, numerous methods have been proposed, many of which are discussed in this chapter. For an excellent review of the state of the art in alignment, see~\cite{sucholutsky2023getting}.

In addressing the challenge of aligning models with human cognition, we identify two primary obstacles. The first is the difficulty in accurately capturing human cognitive processes or judgments, a task that is inherently complex and resource-intensive. Human cognition encompasses a vast array of processes, including perception, decision-making, and problem-solving, each influenced by subjective experiences and external contexts. Quantifying such a multifaceted construct requires sophisticated methodologies that often entail significant time and financial resources.

The second challenge lies in integrating these human cognitive metrics into the training of machine learning models. This integration demands innovative strategies, such as incorporating supplementary types of data, designing novel loss functions, or embedding specific biases within model architectures. The objective is to ensure that the models not only process information but also interpret and act upon it in a manner that aligns with human reasoning and judgment.

Recent advancements, particularly in the field of large language models (LLMs), have highlighted the potential of alignment through methods like Reinforcement Learning from Human Feedback (RLHF)~\cite{ouyang2022training}. RLHF integrates human judgments directly into the fine-tuning process, thereby enhancing the model's performance and alignment with human expectations. However, the deployment of such techniques is often hampered by the high cost associated with gathering and processing the requisite human feedback data. This underscores the need for efficient and targeted alignment strategies that judiciously select and utilize data and methodologies to align models effectively with human cognition.

In this chapter, we propose to take an explainability perspective on Alignment, leveraging what we have previously observed with attribution methods and aligning not explicitly the representations themselves but rather implicitly through the attribution explanations of the models. This approach guides the models toward more human-like explanations. In \autoref{sec:harmonization} we will craft a metric as well as a routine to build more aligned models, and in \autoref{sec:lipschitz} we will briefly see an alternative to the Harmonization procedure using robust models.

\clearpage

\section{Harmonizing Human and Machine Explanations}
\label{sec:harmonization}
\definecolor{CNN}{RGB}{99, 143, 238}
\definecolor{CNN_data}{RGB}{213, 149, 101}
\definecolor{transformer}{RGB}{241,194,70}
\definecolor{selfsup}{RGB}{217, 88, 73}
\definecolor{robust}{RGB}{103, 174, 108}
\definecolor{meta}{RGB}{236, 178, 46}

Richard Sutton's \textit{bitter lesson} articulates that seven decades of AI research have taught us that "general methods leveraging computational power overwhelmingly outperform more specialized approaches"~\cite{Sutton2019-vf}. This insight has been underscored by the advent of deep learning, particularly following the groundbreaking success of AlexNet~\cite{krizhevsky2012imagenet} on the ImageNet challenge~\cite{imagenet_cvpr09} over ten years ago. Deep neural networks (DNNs) have since seen continuous advancements, further validating Sutton's observation as these networks now match or even surpass human capabilities on the benchmark, primarily through the sheer scale of computational resources: significantly expanding the network's parameters and the volume of training images far beyond what was utilized for AlexNet~\cite{Liu2022-es,Zhai2021-al,Kaplan2020-zx}. The triumphs of these "scaling laws" are incontrovertible. However, this relentless pursuit of performance has often overlooked a critical inquiry vital for the advancement of brain sciences and the practical deployment of object recognition models: \textit{Do the visual strategies developed by DNNs mirror those employed by humans?}

The visual strategies that mediate object recognition in humans can be decomposed into two related but distinct processes: identifying \textit{where} the important features for object recognition are in a scene, and determining \textit{how} to integrate the selected features into a categorical decision~\cite{DiCarlo2012-nx, ullman2016atoms}. It has been known for nearly a century~\cite{Buswell1935-uu, Yarbus_undated-cq, Posner1980-hh, Mannan2009-xq} that different humans attend to similar locations when asked to find and recognize objects. After selecting these important features, human observers are also consistent in how they use those features to categorize objects -- the inclusion of a few pixels in an image can be the difference between recognizing an object or not~\cite{ullman2016atoms, Gruber2021-uq}.

Has the past decade of DNN development produced any models that are aligned with these human visual strategies for object recognition? Such a model could transform cognitive science by supporting a better mechanistic understanding of how vision works. More human-like models of object recognition would also resolve the problems with predictablity and interpretablity of DNNs~\cite{fel2021cannot}, and control their alarming tendency to rely on ``shortcuts'' and dataset biases to perform well on tasks~\cite{geirhos2020shortcut}. In this work, we perform the first large-scale and systematic comparison of the visual strategies of DNNs and humans for object recognition on ImageNet. 

\paragraph{Contributions.}
In order to compare human and DNN visual strategies, we first turn to the human feature importance maps collected by Linsley et al.~\cite{linsley2018learning, Lin2017-mj}. Their datasets, \textit{ClickMe} and \textit{Clicktionary}, contain maps of nearly 200,000 unique images in ImageNet that highlight the visual features humans believe are important for recognizing them. These datasets amount to a reverse inference on \textit{where} important visual features are in ImageNet images (Fig.~\ref{fig:harmonization:intro}). We complement these datasets with new psychophysics experiments that directly test \text{how} important visual features are used for object recognition (Fig.~\ref{fig:harmonization:intro}). \textbf{As DNN performance has increased on ImageNet, their alignment with human visual strategies captured in these datasets has worsened.} This trade-off is found over 84 different DNNs representing all popular model classes -- from those trained for adversarial robustness to those pushing the scaling laws in network capacity and training data. To summarize our findings:
\begin{itemize}[leftmargin=*]
    \item The trade-off between DNN object recognition accuracy and alignment with human visual strategies replicates across three unique datasets: \textit{ClickMe}~\cite{Linsley2019-ew}, \textit{Clicktionary}~\cite{Lin2017-mj}, and our psychophysics experiments.
    \item We shift this trade-off with our neural harmonizer, a novel drop-in module for co-training any DNN to align with human visual strategies while also achieving high task accuracy. Harmonized DNNs learn visual strategies that are significantly more aligned with humans \textit{than any other DNN we tested}.
    \item We release our data and code at \url{https://serre-lab.github.io/Harmonization/} to help the field tackle the growing misalignment between DNNs and humans.
\end{itemize}

\subsection{Background}

\begin{figure}[ht]
  \centering
    \includegraphics[width=0.9\textwidth]{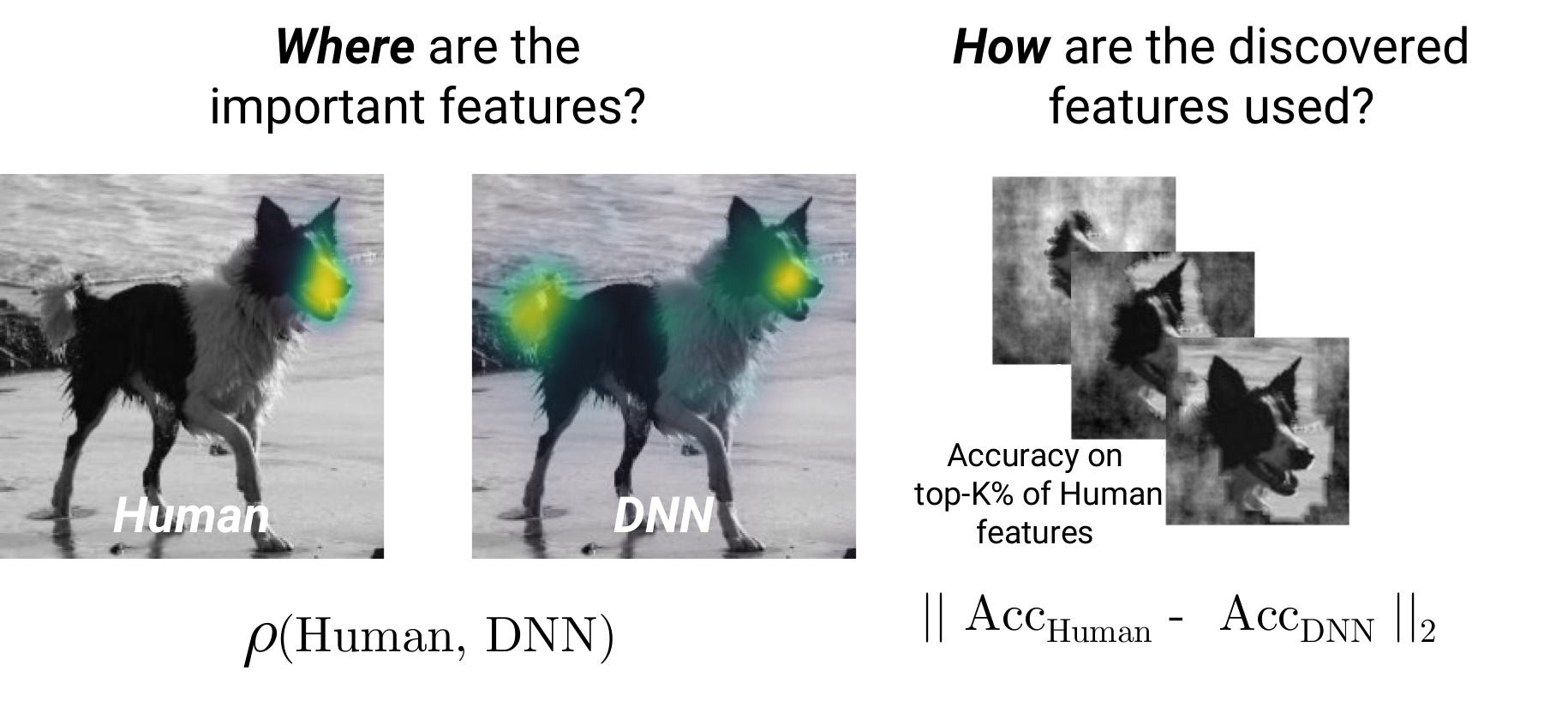}
  \caption{\textbf{Visual strategies of object recognition}. We investigate the alignment of human and DNN visual strategies in object categorization. We decompose human visual strategies into descriptions of \textit{where} important features are~\cite{Linsley2017-qe,Linsley2019-ew}, and \textit{how} those features are integrated into visual decisions.}\label{fig:harmonization:intro}
\end{figure}
\paragraph{Do DNNs explain human visual perception?} Despite the continued success of DNNs on computer vision benchmarks, there are conflicting accounts on their ability to explain human vision. On the one hand, there is evidence that DNNs are improving as models of human visual perception on challenging tasks, such as recognizing objects obscured by noise~\cite{Geirhos2021-rr}. On the other hand, there is also evidence that DNNs struggle to explain perceptual phenomena in human vision like contextual illusions~\cite{Linsley2020-en}, perceptual grouping~\cite{Kim2020-yw,Linsley2021-vx,Geirhos2020-nl}, and categorical prototypes~\cite{Golan2020-zw}. Others have found differences between human attention data and DNN models of visual attention~\cite{Linsley2019-ew,Langlois2021-ns}. Moreover, DNNs have stopped improving as models of the ventral visual system in humans and primates over recent years. While the original theory was that model explanations of object-evoked neural activity patterns improved alongside model categorization accuracy~\cite{Yamins2014-ba}, recent large-scale DNNs are worse at explaining neural data than older ones with lower ImageNet accuracy~\cite{Schrimpf2020-wp}.

\paragraph{What are the visual strategies underlying human object recognition?} Ever since its inception, a goal of vision science has been to characterize the neural processes supporting object recognition in humans. It has been discovered that object recognition can be decomposed into different processing stages that emerge over time~\cite{Fabre-Thorpe2011-js,Roelfsema2000-op,DiCarlo2012-nx,Serre2007-hq,Kietzmann2019-xy,Jagadeesh2022-df,Berrios2022-qm}, where the earliest stage is associated with processing through feedforward connections in the visual system, and the later stage is associated with processing through feedback connections. Since the DNNs used today mostly rely on feedforward connections, it is likely that they are better models for that rapid feedforward phase of processing than the subsequent feedback phase~\cite{Serre2019-bb, Serre2007-hq}. To maximize the likelihood that the visual strategies learned by DNNs align with those used by humans, our experiments focus on the visual strategies of rapid feedforward object recognition in humans.

Most closely related to our work, are studies of ``top-down'' image saliency and \textit{where} category diagnostic visual features are in images. These studies typically involve asking participants to search for an object in an image, or find visual features that are diagnostic for an object's category or identity~\cite{Linsley2017-qe, Linsley2019-ew, Koehler2014-li, Buswell1935-uu, Yarbus_undated-cq, Posner1980-hh, Mannan2009-xq}. In our work, we complement these descriptions of \textit{where} important features are in images with psychophysics testing \textit{how} those features are used to categorize objects.

\paragraph{Comparing visual strategies of humans and machines.} As methods in explainable artificial intelligence have developed over the past decade, they have opened up opportunities for comparing the visual regions selected by humans and DNNs when solving tasks. Many of these comparisons have focused on human image saliency measurements captured by eye tracking or mouse clicks during passive or active viewing~\cite{Linsley2017-qe,Linsley2019-ew,Jiang2015-vl,Peterson2018-pu,Lai2019-ln,Ebrahimpour2019-dc}. Others have compared categorical representation distances~\cite{Peterson2018-pu,Roads2020-gd} or combined those distances with measures of human attention~\cite{Langlois2021-ns}. The most direct comparisons between human and DNN visual strategies involved analyzing the minimal image patches needed to recognize objects~\cite{Ullman2016-wy,Funke2018-ft,Srivastava2019-jg}. However, these studies were limited and compared humans with older DNNs on tens of images. To the best of our knowledge, the largest-scale evaluation of human and DNN visual strategies relied on the \textit{ClickMe} dataset to compare visual regions preferred by humans and attention models trained for object recognition~\cite{Linsley2019-ew}. What is noticeably missing from each of these studies is an large-scale analysis spanning many images and models of how human and DNN alignment has changed as a function of model performance.

\paragraph{Improving the correspondence between humans and machines.} Inconsistencies between human and DNN representations can be resolved by directly training models to act more like humans. DNNs have been trained to have more human-like attention, or human-like representational distances in their output layers~\cite{Peterson2018-pu,Roads2020-gd,Linsley2019-ew,Boyd2021-mh,Bomatter2021-zs}. Here, we add to these successes with the neural harmonizer, a training routine that automatically aligns the visual strategies (Fig.~\ref{fig:harmonization:intro}) of any two observers by minimizing the dissimilarity of their decision explanations.

\subsection{Methods}\label{sec:methods}
\paragraph{Human feature importance datasets.} We focused on the ImageNet dataset to compare the visual strategies of humans and DNNs for object recognition at scale. We relied on the two significant efforts for gathering feature importance data from humans on ImageNet: the \textit{Clicktionary}~\cite{Linsley2017-qe} and \textit{ClickMe}~\cite{Linsley2019-ew} games, which use slightly different methods to collect their data. Both games begin with the same basic setup: two players work together to locate features in an object image that they believe are important for categorizing it. As one of the players selects important image regions, those regions are filled into a blank canvas for the other observer to see and categorize the image as quickly as possible. In \textit{Clicktionary}~\cite{Linsley2017-qe}, both players are humans, whereas in \textit{ClickMe}~\cite{Linsley2019-ew}, the player selecting features is a human and the player recognizing images is a DNN (VGG16~\cite{Simonyan2014-jd}). For both games, feature importance maps depicting the average object category diagnosticity of every pixel was computed as the probability of it being clicked by a participant. In total, \textit{Clicktionary}~\cite{Linsley2017-qe} contained feature importance maps for 200 images from the ImageNet validation set, whereas \textit{ClickMe}~\cite{Linsley2019-ew} contained feature importance maps for a non-overlapping set of 196,499 images from ImageNet training and validation sets. Thus, \textit{ClickMe} has far more data than \textit{Clicktionary}, but \textit{Clicktionary} data has more reliable human feature importance data than \textit{ClickMe}. Our experiments measure the alignment between human and DNN visual strategies using \textit{ClickMe} and \textit{Clicktionary} feature importance maps captured on the ImageNet validation set. As we describe in \textsection{\ref{sec:meta_pred}}, \textit{ClickMe} feature importance maps from the ImageNet training set are used to implement our neural harmonizer.

\paragraph{Psychophysics participants and dataset.} We complemented the feature importance maps from \textit{Clicktionary} and \textit{ClickMe} with psychophysics experiments on rapid visual categorization. We recruited 199 participants from Amazon Mechanical Turk (\url{mturk.com}) to complete the experiments. Participants viewed a psychophysics dataset consisting of the 100 animal and 100 non-animal images in the Clicktionary game taken from the ImageNet validation set~\cite{Linsley2017-qe}. We used the feature importance maps for each image as masks for the object images, allowing us to control the proportion of important features observers were shown when asked to recognize objects (Fig.~\ref{fig:harmonization:psychophysics}a). We generated versions of each image that reveal anywhere between 1\% to 100\% (at log-scale spaced intervals) of the important object pixels against a phase scrambled noise background (see Appendix \textsection{1} for details on mask generation). The total number of revealed pixels was equal for every image at a given level of image masking, and the revealed pixels were centered against the noise background. Each participant saw only one masked version of each object image.

\paragraph{Psychophysics experiment.} Participants were instructed to categorize images in the psychophysics dataset as animals or non-animals as quickly and accurately as possible. Each experimental trial consisted of the following sequence of events overlaid onto a white background (SI Fig. 1): (\textit{i}) a fixation cross displayed for a variable time (1,100–1,600ms); (\textit{ii}) an image for 400ms; (\textit{iii}) an additional 150ms of response time. In other words, the experiment forced participants to perform rapid object categorization. They were given a total of 550ms to view an image and press a button to indicate its category (feedback was provided on trials in which responses were not provided within this time limit). Images were sized at 256 x 256 pixel resolution, which is equivalent to a stimulus size approximately between 5 -- 11 degrees of visual angle across a likely range of possible display and seating setups we expect participants used for the experiment. Similar paradigms and timing parameters have been shown to capture pre-attentive visual system processing~\cite{Eberhardt2016-cw, Kirchner2006-xc, Fabre-Thorpe2011-js, Muriel2007-co}. Participants provided informed consent electronically and were compensated \$3.00 for their time ($\sim$ 10--15 min; approximately \$15.00/hr).

\paragraph{Models.}
We compared humans with 84 different DNNs representing the variety of approaches used in the field today: 50 CNNs trained on ImageNet~\cite{Chen2021-is,Tan2019-uh,Radosavovic2020-cs,Howard2019-cr,Simonyan2014-jd,Huang2018-yt,He2015-lm,Zhang2020-my,Gao2021-er,Kolesnikov2019-gg,Sandler2018-lh,Liu2022-es,Szegedy2016-fd,Szegedy2015-pr,Chollet2016-np,Radford2021-km,Xie2019-ju,Xie2016-ol,Szegedy2015-pr,Brendel2019-mw,Mehta2020-ad,Chen2017-wp,Wang2019-jm,Tan2018-zk}, 6 {\color{CNN}{CNNs}} trained on other datasets in addition to ImageNet (which we refer to as ``{\color{CNN_data}{CNN extra data}}'')~\cite{Xie2019-rp,Radford2021-km,Liu2022-es}, 10 {\color{transformer}{vision transformers}}~\cite{DAscoli2021-xw,Touvron2020-fo,Tolstikhin2021-hw,Dosovitskiy2020-if,Steiner2021-pl}, 6 CNNs trained with {\color{selfsup}{self-supervision}}~\cite{Chen2020-lw,Zeki_Yalniz2019-yo}, and 13 models trained for {\color{robust}{robustness}} to noise or adversarial examples~\cite{Geirhos2018-ag,Salman2020-lo}. We used pretrained weights for each of these models supplied by their authors, with a variety of licenses (detailed in SI~\textsection{2}), implemented in Tensorflow 2.0, Keras, or PyTorch.

\begin{figure}[!t]
  \centering
    \includegraphics[width=0.99\textwidth]{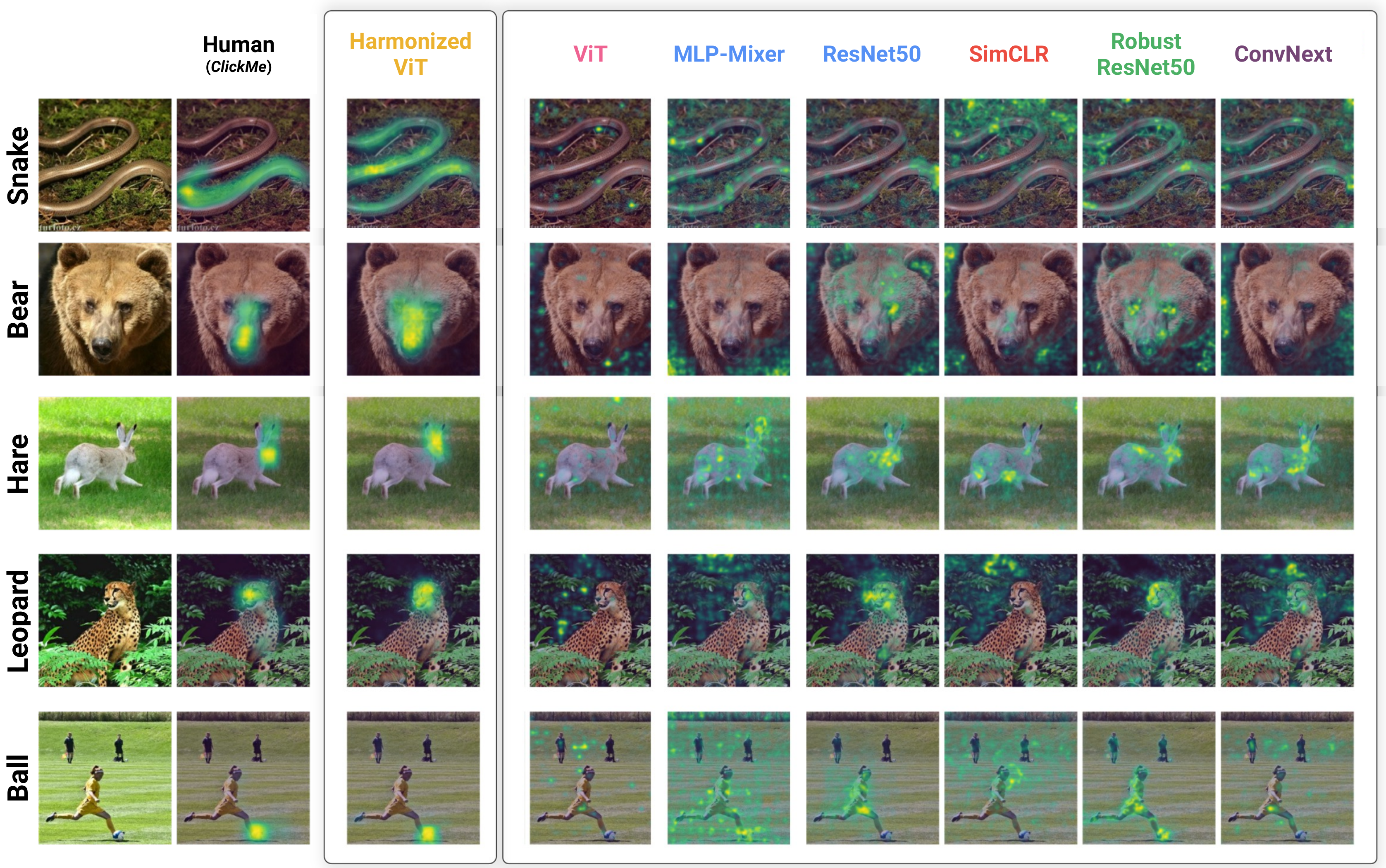}
  \caption{\textbf{Human and DNNs rely on different features to recognize objects.} In contrast, our neural harmonizer aligns DNN feature importance with humans. We smooth feature importance maps from humans (\textit{ClickMe}) and DNNs with a Gaussian kernel for visualization.}
\label{fig:harmonization:clickme_qualitative}
\end{figure}

\subsection{Results}\label{sec:meta_pred}
\subsubsection{\textit{Where} are diagnostic object features for humans and DNNs?} To systematically compare the visual strategies of object recognition for humans and DNNs on ImageNet, we first turned to the \textit{ClickMe} dataset of feature importance maps~\cite{Linsley2019-ew}. In order to derive comparable feature importance maps for DNNs, we needed a method that could be efficiently and consistently applied to each of the 84 DNNs we tested without any idiosyncratic hyperparameters. This led us to choose a classic method for explainable artificial intelligence, image feature saliency~\cite{Simonyan2013-ln}. We prepared human feature importance maps from \textit{ClickMe} by taking the average importance map produced by humans for every image that also appeared in ImageNet validation. We then used Spearman's rank-correlation to measure the similarity between human feature maps and DNN feature maps for each image~\cite{Eberhardt2016-cw}. We also computed the inter-rater alignment of human feature importance maps as the mean split-half correlation across 1000 random splits of the participant pool ($\rho=0.66$). We then normalized each human-DNN correlation by this score~\cite{Linsley2019-ew}.

\begin{figure}[ht!]
\begin{center}
   \includegraphics[width=.99\linewidth]{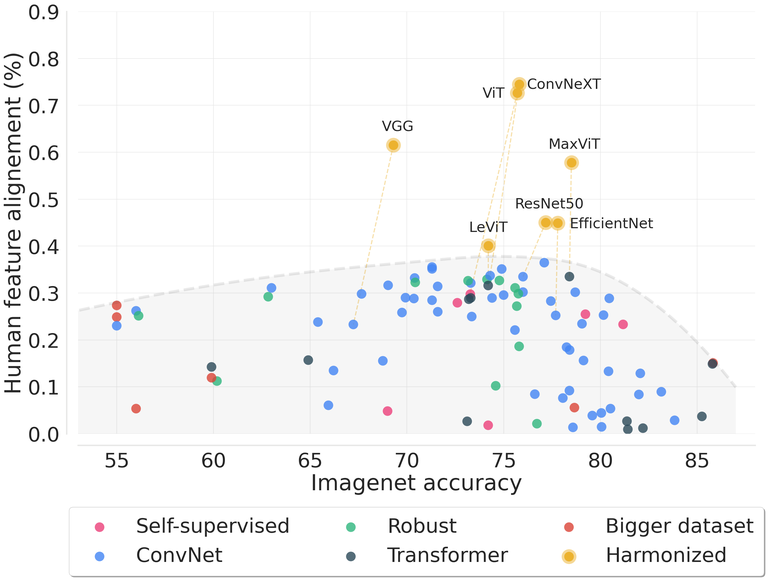}
\end{center}
   \caption{\textbf{The trade-off between DNN performance and alignment with human feature importance from \textit{ClickMe}\cite{Linsley2019-ew}}. Human feature alignment is the mean Spearman correlation between human and DNN feature importance maps, normalized by the average inter-rater alignment of humans. The shaded region denotes the pareto frontier of the trade-offs between ImageNet accuracy and human feature alignment for unharmonized models.  {\color{meta}{Harmonized}} models (VGG16, ResNet50, ViT, and EfficientNetB0) are more accurate and aligned than versions of those models trained only for categorization. Error bars are bootstrapped standard deviations over feature alignment. Arrows show a shift in performance after training with the {\color{meta}{neural harmonizer}}. The feature alignment of an average of \textit{ClickMe} maps with held-out maps is denoted by $\dagger$.}
\label{fig:harmonization:clickme_results}
\end{figure}

There were dramatic qualitative differences between the features selected by humans and DNNs on ImageNet. In general, humans selected less context and focused more on object parts: for animals, parts of their faces; for non-animals, parts that enable their usage, like the spade of a shovel (see Fig.~\ref{fig:harmonization:clickme_qualitative} and SI Fig.~5. The DNN that was most aligned with humans, the DenseNet121, was still only 38\% aligned with humans (Fig.~\ref{fig:harmonization:clickme_results}).

Plotting the relationship between DNNs' top-1 accuracy on ImageNet with their human alignment revealed a striking trade-off: as the accuracy of DNNs has improved beyond DenseNet121, their alignment with humans has worsened (Fig.~\ref{fig:harmonization:clickme_results}). For example, consider the ConvNext~\cite{Liu2022-es}, which achieved the best top-1 accuracy in our experiments (85.8\%), was only 22\% aligned with humans -- equivalent to the alignment of the BagNet33~\cite{Brendel2019-mw} (63\% top-1 accuracy). As an additional control, we computed the similarity between the average \textit{ClickMe} map, which exhibits a center bias~\cite{Deza2020-fq,Wang2017-dp} (SI Fig.~5), and each individual \textit{ClickMe} map. This center-bias control was only outperformed by 42/84 CNNs we tested ($\dagger$ in Fig.~\ref{fig:harmonization:clickme_results}). Overall, we observe that human and DNN alignment has considerably worsened since the introduction of these two models.

\paragraph{The neural harmonizer.} While scaling DNNs has immensely helped performance on popular benchmark tasks, there are still fundamental differences in the architectures of DNNs and the human visual system~\cite{Serre2019-bb} which could part of the reason to blame for poor alignment. While introducing biological constraints into DNNs could help this problem, there is plenty of evidence that doing so would hurt benchmark performance and require bespoke development for every different architecture~\cite{Tang2018-vg,Kubilius2019-qr,Schrimpf2020-em}. \textit{Is it possible to align a DNN's visual strategies with humans without hurting its performance?}

Such a general-purpose method for aligning human and DNN visual strategies should satisfy the following criteria: (\textbf{\textit{i}}) The method should work with any fully-differentiable network architecture. (\textbf{\textit{ii}}) It should not present optimization issues that interfere with learning to solve a task, and the task-accuracy of a model trained with the method should not be worse than a model trained without the method. We created the neural harmonizer to satisfy these criteria.

We propose to recall some notations before formally introducing our loss. Still within the standard supervised learning framework, we define an input space, $\sx$, and an output space, $\sy$, alongside a parameterized predictor function $\f : \sx \to \sy$. This function maps an input vector $\vx \in \sx$ to an output $\f(\vx;\parameters)$. We introduce an explanation functional, $\explainer : \fspace \times \s{X} \to \s{X}$, that produces a feature importance map $\explainer(\f, \vx)$. 

Our goal is to synchronize the model's explanation, denoted as $\explanation$, with a human-provided explanation, denoted as $\explanation^\star_{\vx}$, here a \textit{Clickmap}, without compromising the model's accuracy. We aim to achieve this alignment without sacrificing the model's accuracy. A simplicist approach to this alignment might be simply to force the model's explanations to match the human explanations as closely as possible and add a cross-entropy loss:

$$
\s{L}_{\text{naive}} = \s{L}_{\text{cross-entropy}}(\f, \vx, \vy) + \lambda \norm{\explainer(\f, \vx) - \explanation^\star_{\vx}}_2^2
$$

Interestingly, under mild assumption, one can show that \textit{aligning the explanations implicitly align the predictions} (up to a constant), a concept we encapsulate in a theorem formalizing this relationship.

\begin{definition}[$\explainer$-Aligned predictors.]
Let $\fspace : \sx \to \sy$, with $\sx = (0, 1]^d$, $\sy \subseteq \Real$, and given an explanation functionnal $\explainer : \fspace \times \sx \to \sx$. For any couple of predictors $(\f, \fbis) \in \fspace^2$ we say that the two predictor are $\explainer$-Aligned if and only if:
$$
\forall ~ \vx \in \sx ~~~~
\explainer(\f, \vx) = \explainer(\v{\psi}, \vx)
$$

\end{definition}

In other terms, this denotes that two predictors are aligned if, for each point in the input space, they yield identical explanations. An interesting property of many popular attribution methods, such as Saliency, Gradient-Input, Integrated-Gradients, or Occlusion, is that if two predictors are $\explainer$-Aligned, their predictions are also aligned (up to a constant):

\begin{theorem}[$\explainer$-Aligned imply Aligned predictions.]
Let $(\f, \fbis)$ be two $\explainer$-Aligned predictors. For any explanation functionnal $\explainer \in \{ \explainer_{\text{Sa}}, \explainer_{\text{GI}}, \explainer_{\text{IG}}, \explainer_{\text{OC}} \}$, aligning the explanations implies aligning the predictions (up to a constant):

$$
\forall \vx \in \sx ~~ \f(\vx) = \v{\psi}(\vx) + \kappa
$$

With $\kappa$ a constant independent of $\vx$.

\end{theorem}

Proofs in \autoref{app:harmonization:thm}. This suggests that theoretically, by learning the explanation of a model, one implicitly learns its decision function. Surprisingly, this is a commonality among attribution methods with apparent diverse definitions.

While this observation and theorem are insightful, the current loss present two main issues: first, the human attributions, $\explanation^\star$, are not pixel-perfect, meaning we don't have a pixel-resolution explanation for the \textit{Clickmaps}. Second, the range of values of these attributions isn't well-defined, meaning we prefer to align with the model's gradient value range to avoid overly penalizing the model. This ensures our routine can adapt to a wide range of models. To address these issues, we refine our loss function further.

\begin{itemize}
    \item To tackle the ``not pixel-perfect'' issue, we employ a multi-scale alignment strategy. This means we do not insist on a perfect pixel-wise match between explanations. Instead, we seek an approximate alignment within a reasonable delta. We utilize a Gaussian pyramid representation to demand alignment at lower-resolution versions of the explanation, thus accommodating for the lack of pixel perfection. Formally, we employ a Gaussian pyramid representation, $\pyramid_i(\cdot)$, to rescale the feature importance map $\explanation$ over $n$ levels, where $i \in \{1, \ldots, n\}$. This is accomplished by iteratively downsampling the map with a Gaussian kernel, starting from $\pyramid_1(\explanation) = \explanation$. Our objective is to minimize $\sum_{i}^n || \pyramid_i(\explainer(\f, \vx)) - \pyramid_i(\explanation^\star_{\vx}) ||^2$, ensuring the alignment of DNNs' feature importance maps with those of humans across each pyramid level.

    \item Regarding the issue of value range, we propose to stabilize the loss by standardizing both heatmaps and gradients. This standardization ensures that each explanations operate within their respective value ranges. Importantly, it allows for the most critical image regions to match, regardless of their exact values. Formally, we define the standardization function as $\v{z}(\cdot)$, which normalizes the explanation $\explanation$ so that $\v{z}(\explanation)$ has an average value of zero and a standard deviation of one. To emphasize alignment on the most critical image regions, we only consider the positive part of the standardized explanation, $\v{z}(\explanation)^{+}$.
\end{itemize}

\begin{figure}[t]
  \centering
    \includegraphics[width=0.9\textwidth]{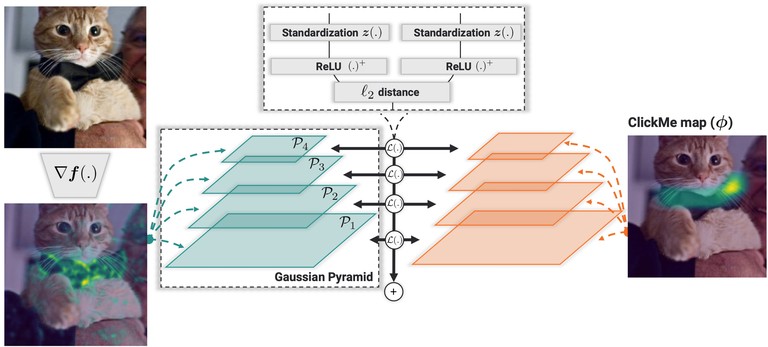}
  \caption{\textbf{Harmonization loss}. This figure illustrates the proposed harmonization loss, aiming to align the model's decision gradient (explanation) with the click-map. Instead of merely computing an $\ell_2$ distance between the two explanations, we introduce a multi-scale representation to compensate for the click-maps not being pixel-perfect. Additionally, we standardize both explanations to ensure that the ground truth explanations are adjusted to match the model gradient's value range.}\label{fig:harmonization:loss}
\end{figure}

This leads to the the complete neural harmonization loss illustrated in \autoref{fig:harmonization:loss}:

\begin{align}
    \mathcal{L}_{\text{Harmonization}} =&
    ~\mathcal{L}_{CCE}(\f, \vx, \vy) + \beta ||\parameters||_2^2 + \\
    &\lambda \sum_{i}^n || \big( (\v{z} \circ \pyramid_i \circ \explainer)(\f, \vx) \big)^+ ~ - ~ \big( (\v{z} \circ \pyramid_i)(\explanation^\star_{\vx}) \big)^+ ||_2^2 
\end{align}

\begin{figure}[ht]
\begin{center}
   \includegraphics[width=.99\linewidth]{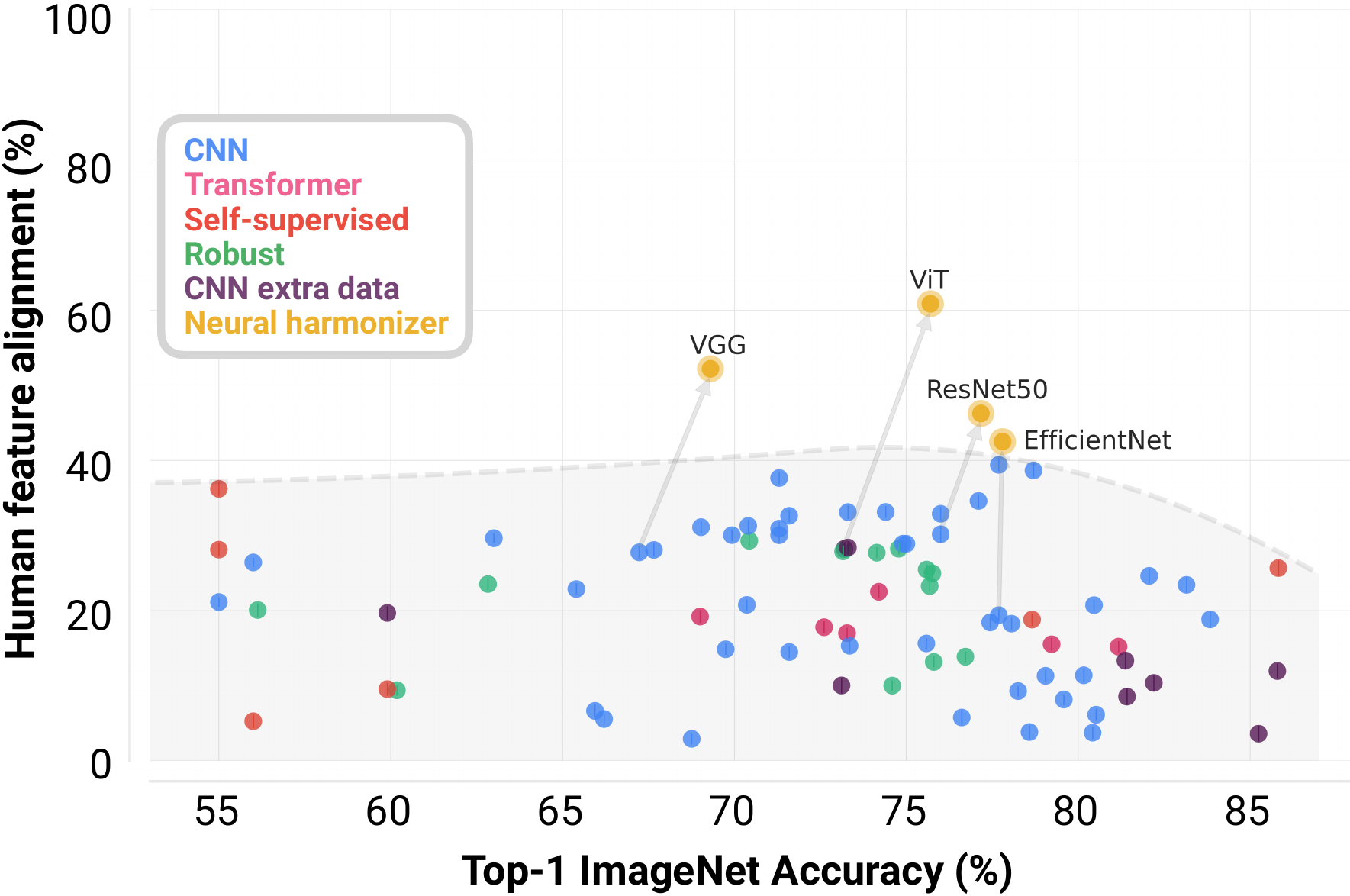}
\end{center}
   \caption{\textbf{The trade-off between DNN performance and alignment with human feature importance from \textit{Clicktionary}\cite{Linsley2017-qe}}. Human feature alignment is the mean Spearman correlation between human and DNN feature importance maps, normalized by the average inter-rater alignment of humans. The shaded region denotes the pareto frontier of the trade-offs between ImageNet accuracy and human feature alignment for unharmonized models.  {\color{meta}{Harmonized}} models (VGG16, ResNet50, MobileNetV1, and EfficientNetB0) are more accurate and aligned than versions of those models trained only for categorization. Error bars are bootstrapped standard deviations over feature alignment. Arrows denote a shift in performance after training with the {\color{meta}{neural harmonizer}}.}
\label{fig:harmonization:clicktionary_results}
\end{figure}

\paragraph{Training.} We trained four different DNNs with the neural harmonizer: VGG16, ViT, ResNet50, and EfficientNetB0. These models were selected because they are popular convolutional and transformer networks with open-source architectures that are straightforward to train and also sit near the boundary of the trade-off between DNN performance and alignment with humans. Models were trained using the neural harmonizer to optimize categorization performance on ImageNet and feature importance map alignment with human data from \textit{ClickMe}. We trained models on all images in the ImageNet training set, but because \textit{ClickMe} only contains human feature importance maps for a portion of those images, we computed the categorization loss but not the neural harmonizer loss for images without importance maps. Models were trained using 8 cores V4 TPUs on the Google Cloud Platform, and training lasted approximately one day. Models were trained with an augmented ResNet training recipe (built from \url{https://github.com/tensorflow/tpu/}). Models were optimized with SGD and momentum over batches of 512 images, a learning rate of $0.3$, and label smoothing~\cite{Muller2019-td}. Images were augmented with random left-right flips and mixup~\cite{Zhang2017-hw}. The learning rate was adjusted over the course of training with a schedule that began with an initial warm-up period of 5 epochs and then  decaying according to a cosine function over 90 epochs, with decay at step 30, 50 and 80. We validated that a ResNet50 and VGG16 trained with these hyperparameters and schedule using standard cross-entropy (but not the neural harmonizer) matched published performance.

\begin{figure}[t!]
\begin{center}
   \includegraphics[width=1\linewidth]{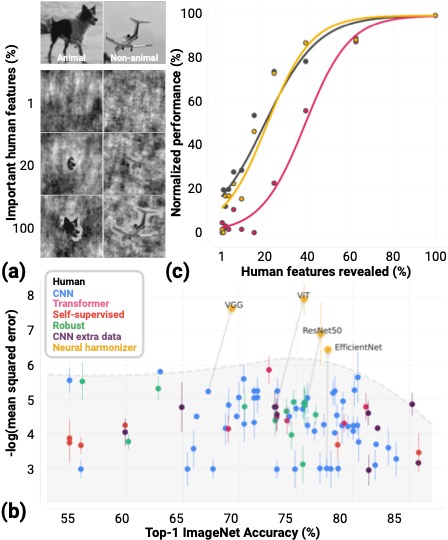}
\end{center}
   \caption{\textbf{Comparing \textit{how} humans and DNNs use visual features during object recognition}. \textbf{(a)} Humans and DNNs categorized ImageNet validation images as animals or non-animals. The images revealed only a portion of the most important visual features according to the \textit{Clicktionary} game~\cite{Linsley2017-zt}. \textbf{(b)} There was a trade-off between DNN top-1 accuracy on ImageNet and alignment with human visual decision making. The shaded region denotes the pareto frontier of the trade-off between ImageNet accuracy and human feature alignment for unharmonized models. Arrows denote a shift in performance after training with the {\color{meta}{neural harmonizer}}. Error bars are bootstrapped standard deviations over decision-making alignment. \textbf{(c)} A state-of-the-art DNN like the ViT learned a different strategy for integrating visual features into decisions than humans or a harmonized ViT.}
 \label{fig:harmonization:psychophysics}
\end{figure}

\paragraph{The neural harmonizer aligns human and DNN visual strategies.} We found that harmonized models broke the trade-off between ImageNet accuracy and model alignment with \textit{ClickMe} human feature importance maps (Fig.~\ref{fig:harmonization:clickme_results}). Harmonized models were significantly more aligned with feature importance maps and also performed better on ImageNet. The changes in \textit{where} harmonized models find important features in images were dramatic: a harmonized ViT had feature importance maps that are far less reliant on context (Fig.~\ref{fig:harmonization:clickme_qualitative}) and approximately 150\% more aligned with humans (Fig.~\ref{fig:harmonization:clickme_results}; ViT goes from 28.7\% to 72.6\% alignment after harmonization). The same model also performed 4\% better in top-1 accuracy without any changes to its architecture. Similar improvements were found for the harmonized VGG16 and ResNet50. While the EfficientNetB0 had only a minimal improvement in accuracy, it too exhibited a large boost in human feature alignment.

\paragraph{Clicktionary.} To test if the trade-off between DNN ImageNet accuracy and alignment with humans is a general phenomenon we next turned to \textit{Clicktionary}~\cite{Linsley2017-qe}. Indeed, we observed a similar trade-off on this dataset as we found for \textit{ClickMe}: alignment with human feature importance from \textit{Clicktionary} has worsened as DNN accuracy has improved on ImageNet (Fig.~\ref{fig:harmonization:clicktionary_results}). As with \textit{ClickMe}, harmonized DNNs shift the accuracy-alignment trade-off on this dataset.

\subsubsection{\textit{How} do humans and DNNs integrate diagnostic object features into decisions?}
The trade-off we discovered between DNN accuracy on ImageNet and alignment with human visual feature importance suggests that the two use different visual strategies for object classification. However, there is potential for an even deeper problem. Even if two observers deem the same regions of an image as important for recognizing it, there is no guarantee that they use the selected features in the same way to render their decisions. We posit that if two observers have aligned visual strategies, the will agree on both \textit{where} important features are in an image and \textit{how} they use those features for decisions.

We developed a psychophysics experiment to measure how different humans use features in ImageNet images to recognize objects. Participants viewed versions of these images where only a proportion of the features that were deemed most important in the \textit{Clicktionary} game were visible (Fig.~\ref{fig:harmonization:psychophysics}a). Participants had to accurately detect whether or not the image contained an animal within 550ms, which forced them to rely on feedforward processing as much as possible~\cite{Serre2007-hq}. Each of the 200 images we used were shown to a single participant only once. We accumulated responses from all participants to construct decision curves that showed how accurately the average human converted any given proportion of image features into an object decision. We performed the same experiment on DNNs as we did on humans, recording animal \textit{vs} non-animal decisions according to whether or not the most probable category in the model's 1000-category output was an animal. Because the experiment was speeded, humans did not achieve perfect accuracy. Thus, we normalized performance for humans and DNNs to compare the rate at which each integrated features into accurate decisions.

We discovered a similar trade-off between ImageNet accuracy and alignment with human visual decision making in this experiment as we did in \textit{ClickMe} and \textit{Clicktionary} (Fig.~\ref{fig:harmonization:psychophysics}b). Indeed, the model that was most aligned with human decision-making -- the BagNet33~\cite{Brendel2019-mw} -- only achieved 63.0\% accuracy on ImageNet. Surprisingly, harmonized models broke this trend, particularly the harmonized ViT (Fig.~\ref{fig:harmonization:psychophysics}b, top-right), despite no explicit constraints in that procedure which forced consistent decision-making with humans. In contrast, an unharmonized ViT integrates visual information into accurate decisions less efficiently than humans or harmonized models (Fig.~\ref{fig:harmonization:psychophysics}c).

\subsection{Conclusion}
Models that reliably categorize objects like humans do would shift the paradigms of the cognitive sciences and artificial intelligence. But despite continuous progress over the past decade on the ImageNet benchmark, DNNs are becoming \textit{worse} models of human vision, less aligned. Our solution to this problem, the neural harmonizer, can be applied to any DNN to align their visual strategies with humans and even improve performance.

We observed the greatest benefit of harmonization on the visual transformer, the ViT. This finding is particularly surprising given that transformers eschew the locality bias of convolutional neural networks that has helped them become the new standard for modeling human vision and cognition~\cite{Serre2019-bb}. Thus, we suspect that the neural harmonizer is especially well-suited for large-scale training on low-inductive bias models, like transformers. We also hypothesize that the improvements in human alignment provided by the neural harmonizer will yield a variety of downstream benefits for a model like the ViT, including better predictions of perceptual similarity, stimulus-evoked neural responses, and even performance on visual reasoning tasks. We leave these analyses for future work.

The field of computer vision today is following Sutton's prescient lesson: benchmark tasks can be scaling architectural capacity and the size of training data. However, as we have demonstrated here, these scaling laws are exchanging performance for alignment with human perception. We encourage the field to re-analyze the costs and benefits of this exchange, particularly in light of the growing concerns about DNNs leveraging shortcuts and dataset biases to achieve high performance~\cite{Geirhos2020-nl}. Alignment with human vision need not be exchanged with performance if DNNs are harmonized. Our codebase (\url{https://serre-lab.github.io/Harmonization/}) can be used to incorporate the neural harmonizer into any DNN created and measure its alignment with humans on the datasets we describe in this paper.

\paragraph{Limitations.} One possible explanation for the misalignment between DNNs and humans that we observe is that recent DNNs have achieved superhuman accuracy on ImageNet. Superhuman DNNs have been described in biomedical applications~\cite{Linsley2021-tb,Lee2017-ip} where there is definitive biological ground-truth labels, but ImageNet labels are noisy, making it unclear if such an achievement is laudable. Thus, an equally likely explanation is that the continued improvements of DNNs at least partially reflect their exploitation of shortcuts in ImageNet~\cite{Geirhos2020-nl}. 

The scope of our work is also limited in that it focuses on object recognition in ImageNet. It is possible that models trained on other tasks, such as segmentation, may be more aligned with humans.

Finally, our modeling efforts were hamstrung for the largest-scale models in existence. Our work does not answer how much harmonization would help a model like CLIP because of the massive investment needed to train it. The neural harmonizer can be applied to CLIP but it is possible that more \textit{ClickMe} human feature importance maps are needed for successful harmonization.

\clearpage

\section{On the Intriguing Effect of Robustness Towards Alignment}
\label{sec:lipschitz}
\begin{figure*}[ht]
  \centering
  \includegraphics[width=0.99\linewidth]{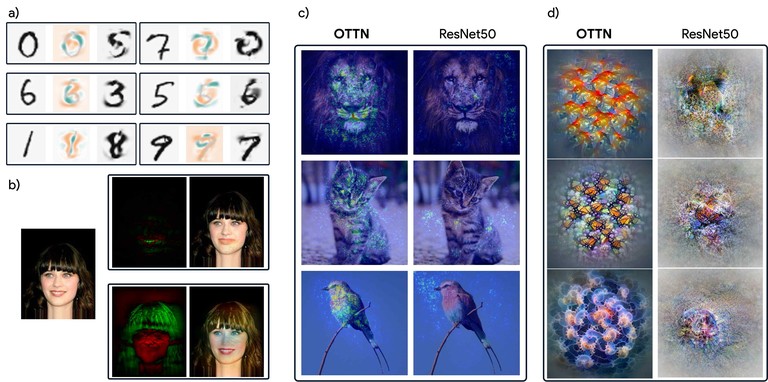}
\caption{
\textbf{Illustration of the beneficial properties of $\losshkr$ gradients.} Examples \textbf{a)} and \textbf{b)} show that the gradients naturally provide a direction that enables the generation of adversarial images - a theoretical justification based on optimal transport is provided in the \autoref{sec:lipschitz:theory}. By applying the gradient $\rvx' = \rvx  -\alpha \nabla_{\rvx}  \f(\rvx)$ to the original image $\rvx$ (on the left), any digit from MNIST can be transformed into its counterfactual $\rvx'$ (e.g., turning a 0 into a 5). 
In \textbf{b)}, we illustrate that this approach can be applied to larger datasets, such as Celeb-A, by creating two counterfactual examples for the closed-mouth and blonde  classes. In \textbf{c)}, we compare the Saliency Map of a classical model with those of $\losshkr$ gradients, which are more focused on relevant elements. Finally, in \textbf{d)}, we show that following the gradients of $\losshkr$ could generate convincing feature visualizations that ease the understanding of the model's features. 
}
\label{fig:lipschitz:big_picture}
\end{figure*}

The method outlined in the previous section utilizes a routine and tailored data to regularize models during training, aligning them with human attention. However, collecting such data can be laborious and sometimes impossible. In this section, we pivot to an alternative approach that shifts \textbf{from regularizing to constraining the model}. We employ 1-Lipschitz networks, trained with a transport loss. In \autoref{sec:attributions:mege}, we used our metric of algorithmic stability to demonstrate that 1-Lipschitz models provide more general explanations. Here, we will illustrate that the gradient of these models has a compelling interpretation: it points towards the counterfactual, meaning the closest real point belonging to a different class (see Figure~\ref{fig:lipschitz:big_picture}). We will show that these models, originally designed for robustness against adversarial attacks, are \textit{also} naturally aligned.

\subsection{Background}

Let us consider a probability space $(\Omega, \mathcal{F}, \P)$, where $\Omega$ represents the set of outcomes, $\mathcal{F}$ a $\sigma$-algebra of events, and $\P$ a probability measure. The space of all probability measures on a metric space $(\sx, \norm{\cdot})$ is denoted as $\s{P}(\sx)$. Here, $\sx \subseteq \Real^d$ signifies the input space, and $\sy = \{\pm 1\}$ the output space. The input data $\rvx : \Omega \rightarrow \sx$ and target label $\ry : \Omega \rightarrow \sy$ are modeled as random variables with distributions $\P_{\rvx}$ and $\P_{\ry}$, respectively, with $\P_{\rvx,\ry}$ representing their joint distribution over $\sx \times \sy$.

The Wasserstein distance, inspired by the theory of optimal transport~\cite{villani2009optimal}, measures the minimal cost required to transform one probability distribution into another. It roots back to the work of Gaspard Monge in the 18th century~\cite{monge1781memoire}, and was originally defined as:

\begin{equation}
    \wasserstein_1(\mu, \nu) = \inf_{\pi \in \Pi(\mu, \nu)} \int_{\sx \times \sx} \norm{\rvx - \rv{z}} \,d\pi(\rvx, \rv{z}),
\end{equation}

Where $\Pi(\mu, \nu)$ is the set of all couplings of $\mu$ and $\nu$, $\wasserstein_1$ denote the 1-Wasserstein distance, also known as the Earth-Mover's distance, between two probability measures $\mu$ and $\nu$ over $\sx$.
Moreover, it can be shown that dual representation of $\wasserstein_1$ is a special case of the duality theorem of Kantorovich and Rubinstein (\cite{kantorovich1960mathematical}) and is defined as:

\begin{align}\label{eq:lip:kantorovich}
    \wasserstein_1(\mu, \nu) &= \sup_{\f \in \lip_1(\sx)} \left( \int_{\sx} \f(\rvx) \,d\mu(\rvx) - \int_{\sx} \f(\rvx) \,d\nu(\rvx) \right) \\
    & = \sup_{\f \in \lip_1(\sx)} ~ ~ \underset{\rvx \sim \mu}{\E}(\f(\rvx)) ~ - ~ \underset{\rvx \sim \nu}{\E}(\f(\rvx)).
\end{align}

Where $\lip_1(\sx)$ denotes the space of 1-Lipschitz functions on $\sx$. For reference, a function is considered L-Lipschitz if for all pairs $(\rvx, \rv{z}) \in \sx^2$, the norm of the difference between $\rvx$ and $\rv{z}$ is less than or equal to $L$ times the norm of the difference between $\f(\rvx)$ and $\f(\rv{z})$ :

$$
\forall (\rvx, \rv{z}) \in \sx^2, \norm{\f(\rvx) - \f(\rv{z})} \leq L \norm{\rvx - \rv{z}}.
$$

This formulation in Equation~\ref{eq:lip:kantorovich} is particularly intriguing, as it renders the computation of the Wasserstein distance tractable if one can correctly parametrize to optimize over the space of 1-Lipschitz functions.
Recent works have proposed to use deep neural network to parametrize the function $\f(\cdot, \parameters)$ and have found various ways to constraint the function space such that $\f \in \lip_1$ at every step in the training process. We refer the reader to \cite{serrurier2022adversarial,hein_formal_2017,Sokolic_2017,tsipras2018robustness,salimans2016weight,miyato2018spectral} for more information.

\paragraph{\hkr: Robust Classification via Transport-Based Loss Function}

Building on these foundations, the \hkr~Loss introduced in~\cite{serrurier2021achieving}~incorporates a hinge regularization term to the Kantorovich-Rubinstein optimization objective, aiming to enhance binary classification performance. It is formulated as:
\begin{equation}
    \losshkr(\f) = 
    \underset{\rvx \sim \mu}{\E}(\f(\rvx)) ~ - ~ \underset{\rvx \sim \nu}{\E}(\f(\rvx)) 
    ~ +  \underset{(\rvx, \ry) \sim \P_{\rvx, \ry}}{\lambda ~ \E}\big(\margin - \ry \f(\rvx)\big)^+
\end{equation}
With $\margin > 0$, the margin introduces a significant contribution to the model's robustness and interpretability by promoting separation between the distributions of positive and negative classes. This loss has been thoroughly analyzed in~\cite{bethune2022pay}, providing insights into its interpretation, limitations, and advantages, especially in controlling the Lipschitz constant. Moreover, the HKR loss has been applied in computing SDF functions~\cite{bethune2023robust} and in DP-training~\cite{bethune2023dp}. In practice, the model is trained using the DeelLip\footnote{https://github.com/deel-ai/deel-lip} library (\cite{deelLip}).

\subsection{An optimal transport perspective of Saliency}
\label{sec:lipschitz:theory}

Models trained with the previously introduced \hkr~loss exhibit interesting properties from a transport perspective: the gradient points towards a point of the opposite class, a counterfactual. We will revisit these propositions and interpret the significance of this gradient, then explore how this translates into terms of alignment.

We note $\optiplan$ the optimal transport plan corresponding to the minimizer of the \hkr loss. In the most general setting, $\optiplan$ is a joint distribution over $\mu,\nu$ pairs. However, when $\mu$ and $\nu$ admit a density function~\cite{peyre2018computational} with respect to Lebesgue measure, then the joint density describes a deterministic mapping, i.e. a Monge map. Given $\rvx \sim \mu$ 
(resp. $\nu$) we note $\rv{z} = \transport(\rvx) \in \nu$ (resp. $\mu$) the image of $\rvx$ with respect to $\optiplan$. When $\optiplan$ is not deterministic (on real datasets that are defined as a discrete collection of Diracs), we take $\transport(\rvx)$ as the point of maximal mass with respect to $\optiplan$.

\begin{theorem}[Transportation plan direction~\cite{serrurier2024explainable}]\label{th:gradient_transport_plan}
Let $\f^\star$ an optimal solution minimizing the $\losshkr$. Given $\rvx \sim \mu$ (resp. $\nu$)  and  $\rv{z} = \transport(\rvx)$, then $\exists \alpha \geq 0$ (resp. $\alpha \leq 0$) such that $\transport(\rvx) = \rvx  -\alpha \nabla_{\rvx}  \f^\star(\rvx)$ almost surely. 
\end{theorem}

This proposition also holds for the Kantorovich-Rubinstein dual problem without hinge regularization, demonstrating that for $\rvx \sim \P_{\rvx,\ry}$, the gradient $\nabla_{\rvx} \f^{\star}(\rvx)$ indicates the direction in the transportation plan almost surely.

\begin{theorem}[Decision boundary~\cite{serrurier2024explainable}]\label{boundary_distance}
Let $\mu$ and $\nu$ two distributions with disjoint supports with minimal distance~$\xi$ and  $\f^\star$ an optimal solution minimizing the $\losshkr$~with~$\delta < 2\xi$. Given $\rvx \sim \P_{\rvx,\ry}$, $\rvx_\delta = \rvx -\f^\star(\rvx) \nabla_{\rvx} \f^\star(\rvx) \in \boundary $
where $\boundary = \left\{\rvx' \in \sx | \f^\star(\rvx') = 0 \right\}$  is the decision boundary (i.e. the 0 level set of $\f^\star$).
\end{theorem}

Experiments perform in \cite{serrurier2024explainable} suggest this probably remains true when the supports of $\mu$ and $\nu$  are not disjoint. 

\begin{corollary}[\cite{serrurier2024explainable}]\label{fx_grad_adversarial}
Let $\mu$ and $\nu$ two separable distributions with minimal distance $\xi$ and  $\f^\star$ an optimal solution minimizing the $\losshkr$ with $\delta <2\xi$, given $\rvx \sim \P_{\rvx, \ry}$, 
$adv(\f^\star,\rvx) = \rvx_{\delta}$ 
almost surely, where $\rvx_\delta = \rvx -\f^\star(\rvx) \nabla_{\rvx} \f^\star(\rvx)$.
\end{corollary}

\begin{figure*}[ht]
    \centering
    \includegraphics[width=.9\textwidth]{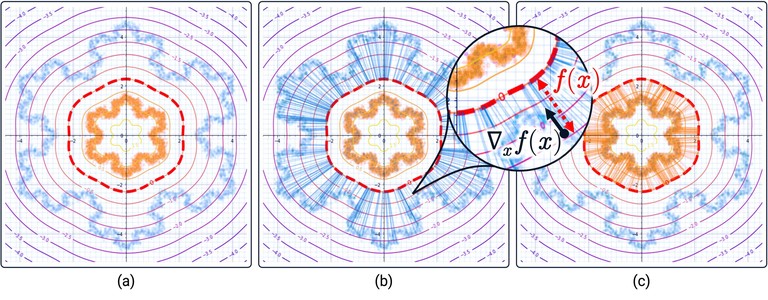}
    \caption{Level sets of an 1-Lipschitz classifier train with $\losshkr$ for two concentric Koch snowflakes \textbf{(a)}.  The decision boundary (denoted $\boundary$, also called the 0-level set) is the red dashed line. Figure \textbf{(b)} (resp. \textbf{(c)}) represents the translation of the form $\rvx'= \rvx -\f(\rvx)\nabla_{\rvx} \f(\rvx)$ of each point $\rvx$ of the first class (resp second class). 
    $(\rvx,\rvx')$ pairs are represented by blue (resp. orange)  segments.}
    \label{fig:lipschitz:koch}
\end{figure*}

This corollary is of great interest as it shows that adversarial examples are precisely identified for the classifier based on $\losshkr$: the direction is given by the gradient $\nabla_{\rvx} \f^\star(\rvx)$ and the distance by $\norm{\f^\star(\rvx)}$. In this scenario, the optimal adversarial attacks align with the gradient direction.

To illustrate these propositions, we learned a dense binary classifier with $\losshkr$ to separate two complex distributions, following two concentric Koch snowflakes. Figure~\ref{fig:lipschitz:koch}) \textbf{(a)} shows the two distributions (blue and orange snowflakes), the learned boundary ($0$-level set) (red dashed line). In the same figure, \textbf{(b)} and \textbf{(c)} show, for random samples $\rvx$ from the two distributions, the segments $[\rvx,\rvx_\delta]$ where $\rvx_\delta$ is defined in Proposition.~\ref{boundary_distance}.

\paragraph{Alignment induced by $\losshkr$.}

Thus, the learning process of those models induces a strong constraint on the gradients of the neural network, aligning them to the optimal transport plan. We claim that is the reason why the simple Saliency Maps have very good properties for those networks.

\begin{figure*}[ht]
\centering
\includegraphics[width=.99\textwidth]{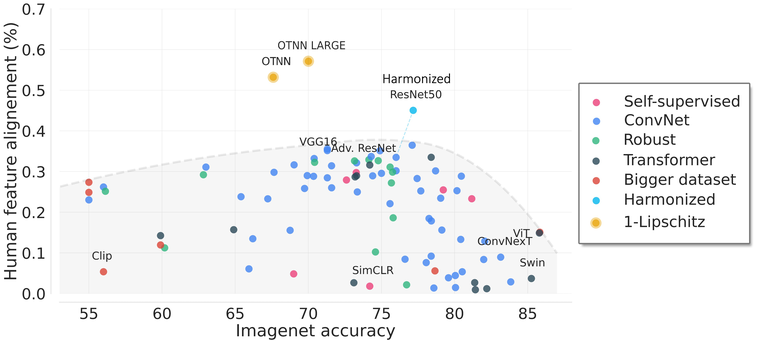}
\caption{\textbf{$\losshkr$~ naturally align gradients with Human attention.} Our study shows that the Saliency Map of $\losshkr$ model (denoted OTNN, for Optimal Transport Neural Network) is highly aligned with human attention. The degree of alignment between human and DNN saliency is measured using the mean Spearman correlation, normalized by the average inter-rater alignment of humans. 
}
\label{fig:lipschitz:human_alignement}
\end{figure*}

By adopting the metric we have proposed in~\autoref{sec:harmonization}, we have computed the human feature alignment of $\losshkr$ Saliency Maps and compare with the others models tested in ~\cite{fel2022aligning}-- more than 100 recent deep neural networks. In Figure ~\ref{fig:lipschitz:human_alignement}, we demonstrate that those model's Saliency Maps, do not only carry strong theoretical interpretation as the direction of the transport plan, it is also more aligned with human attention than any other tested models and significantly surpasses the Pareto front discovered previously. Perhaps the most surprising is that no clickmap or any specific routine like the harmonization one was used: the OTNN model is even more aligned than a ResNet50 model trained with the specific alignment objective proposed in~\autoref{sec:harmonization}.
The implications of these results are crucial for both cognitive science and industrial applications. A model that more closely aligns with human attention and visual strategies can provide a more comprehensive understanding of how vision operates for humans, and also enhance the predictability, interpretability, and performance of object recognition models in industry settings. 
Furthermore, the drop in alignment observed in recent models highlights the necessity of considering the alignment of model visual strategies with human attention while developing object recognition models to reduce the reliance on spurious correlations and ensure that our models are accurate for the right reasons.

\clearpage

\section{Conclusion}

In this chapter, we have explored Hypothesis~\ref{hyp:alignment}, stating that alignment represents a valuable avenue for enhancing our understanding of neural networks. This approach led us to focus on the alignment between models and humans, particularly through explanations. Specifically, we have concentrated on training models to share explanations with humans. This effort addresses a critical need in the field of Deep Learning to narrow the divide between machine learning models and human interpretability. We propose two promising directions: a training routine paired with an innovative metric for assessing alignment, grounded in explainable AI, and an analysis of model robustness as a potential facilitator of alignment.

Our research was bifurcated into distinct yet complementary approaches. The first one revolved around the proposition of a new loss function to encourage regularization. The second propose to directly constrain the model architecture to be robust by design -- more specifically, to optimize over the $\lip_1$ function spaces. Both avenues, through preliminary findings, suggest seems promising for achieving more human-aligned models.

\paragraph{Perspective.} The potential for alignment goes well beyond these first steps. Tasks that more accurately reflect human cognitive processes could open up new dimensions of alignment. Additionally, the influence of diverse types of data -- such as video -- deserves in-depth exploration. Investigating these aspects could reveal how various data modalities and complexities affect the path to model alignment.

I believe that a promising direction would be to have a \textit{holistic} approach, to truly have substantial progress. This would involve integrating more accurate human \textbf{data}, using more biologically realistic \textbf{architectures}, and focusing on more \textbf{human-like tasks}.

\begin{itemize}

    \item From a data perspective, our investigation has so far focused on explanations through heatmaps, while emerging studies highlight the advantages of integrating human preferences~\cite{muttenthaler2024improving}. Many other approaches could be considered, and undoubtedly, data is likely to be a pivotal factor.
    
    \item Regarding architecture, we've explored robust designs like 1-Lipschitz networks, but there are also more biologically plausible architectures~\cite{serre2006learning} available. Notably, modeling recurrent connections found in the visual cortex offers a promising direction for aligning internal mechanisms more closely with human processes. Seminal works~\cite{linsley2020stable,chalvidal2020go} demonstrate the potential benefits of incorporating biological realism into artificial systems, suggesting these architectures could lead to computational models that align more closely with human cognition.

    \item Finally, the deep learning field's focus on classification tasks might not fully represent the complexity of human cognition. Classification is fundamental but captures just a narrow slice of human cognitive skills, which include problem-solving, learning from minimal examples, nuanced context understanding or more interestingly learning to learn~\cite{chalvidal2022meta}. 

\end{itemize}

Pursuing a broader array of actionable components to enhance model alignment, coupled with adopting this holistic viewpoint, lays the groundwork towards models that are not only more interpretable but also more deeply aligned with human thinking and learning processes.

\clearpage

\chapter{From Pixels to Features: Towards Deeper Explainability with Concepts}
\chaptermark{\protect\parbox{.5\textwidth}{From Pixels to Features}}
\label{chap:concepts}

\begin{chapterabstract}
\textit{
In this chapter, we address a challenge identified in \autoref{chap:attributions}: Is it possible to transcend attributions methods to forge methods that do more than just spotlight where a model directs its attention -- \where~the model is looking -- but also clarify \what~exactly it perceives? Essentially, existing methods primarily disclose the ``\where'' in terms of the model's focus, rather than elucidating the "\what" it discerns, in terms of feature. The question then becomes, how can we define and characterize this ``\what''? This is the subject of this chapter that aims to extend beyond attribution methods to lay a more robust foundation for a deeper and more precise Explainability. \\
Our exploration begins in \autoref{sec:concepts:craft}, which propose a significant advancement in concept-based explainability by introducing an automated method, \craft, for extracting a model's learned concepts. We demonstrate that it is feasible to easily assess the significance of these derived concepts using Sobol indices presented in \autoref{sec:attributions:sobol}. The findings from this work substantially improve upon the benchmarks established in \autoref{sec:attributions:metapred}, and offer new avenues for addressing complex scenarios requiring in-depth explainability.
Progressing to \autoref{sec:concepts:holistic}, the cornerstone of this chapter, we show \tbi{i} how \craft~and related research fit within a broader framework of dictionary learning. We propose a unified framework for concept extraction, paving the way for new methodologies. Further, \tbi{ii} we establish a link between concept importance estimation and traditional attribution methods, demonstrating that concept importance estimation methods can be viewed as attribution methods recontextualized within the concept space for evaluative purposes.
With this framework in place, we find it possible to derive insightful answers to literature questions such as ``where should concept decomposition be performed?'' or ``which importance method to choose''. Furthermore, we delve into the importance measure of concepts, revealing that this information can be utilized to address a significant open problem in Explainability: ``how to identify points classified for similar reasons'', by proposing the strategic clustering plot.
The final section of this chapter, \autoref{sec:concepts:maco}, is dedicated to scaling feature visualization through a reformulation of the optimization problem within the Fourier space, by constraining magnitude. This new module allows for the use of feature visualization to create prototypes of the concepts extracted with \craft.
In conclusion, we will showcase the powerful synergies this new framework offers with \Lens, a demo that enables the visualization of the concepts used by a ResNet50 model for the 1000 ImageNet classes.
In sum, this chapter not only tackles foundational questions within the domain of machine learning explainability, but also sets forth a comprehensive framework that integrates advanced methodologies for concept extraction and importance estimation.
}
\end{chapterabstract}

The work in this chapter has led to the publication of the following conference papers:
{\small{
\begin{itemize}

    \item \textbf{Thomas Fel}\equal, Agustin Picard\equal, Louis Bethune\equal, Thibaut Boissin\equal, David Vigouroux, Julien Colin, Rémi Cadène, Thomas Serre, (2023). \textit{``CRAFT: Concept Recursive Activation FacTorization for Explainability''.} In: \textit{IEEE/CVF Conference on Computer Vision and Pattern Recognition} (\textcolor{confcolor}{CVPR})
    
    \item \textbf{Thomas Fel}\equal, Victor Boutin\equal, Mazda Moayeri, Rémi Cadène, Louis Bethune, Mathieu Chalvidal, Thomas Serre (2023). \textit{``A Holistic Approach to Unifying Automatic Concept Extraction and Concept Importance Estimation''.} In: \textit{Advances in Neural Information Processing Systems}  (\textcolor{confcolor}{NeurIPS})
    
    \item \textbf{Thomas Fel}\equal, Thibaut Boissin\equal, Victor Boutin\equal, Agustin Picard\equal, Paul Novello\equal, Julien Colin, Drew Linsley, Tom Rousseau, Rémi Cadène, Lore Goetschalckx, Thomas Serre (2024). \textit{``Unlocking feature visualization for deep network with MAgnitude constrained optimization''.} In: \textit{Advances in Neural Information Processing Systems}  (\textcolor{confcolor}{NeurIPS})

\end{itemize}
}}

\minitoc
\clearpage

\section{Introduction}
This chapter is a direct response to the critical question raised at the conclusion of \autoref{chap:attributions}: why are existing attribution methods not enough to properly understand deep learning models? Our ambition is to investigate Hypothesis~\ref{hyp:what} by extending the methods of explainability beyond  attribution, aiming to probe into the deeper, more intricate aspects of internal features of deep neural network.

This investigation is not a novel expedition. Prior research has ventured beyond simple attribution methods, among which we can identify two strong candidates: Feature visualization (see~\autoref{def:intro:feature_viz}) and concept-based analysis (also briefly presented in~\autoref{def:intro:cav}). Yet, these approaches, while rich with potential, are fraught with challenges due to their early stages of development. In this chapter, we will develop new tools and methodologies within this emerging field.

To frame our investigation, we introduce several critical challenges that must be addressed to advance the state of those new approaches. Concerning the Concept-based approach, we identify 2 main problems in the literature:

\begin{customchallenge}{Challenge 1: Automatically discover concepts used by the model.}
\end{customchallenge}

One of the most pressing challenges is developing methodologies capable of automatically and efficiently uncovering the concepts that models inherently use in decision-making processes. Current methodologies are  focused on testing pre-defined concepts~\cite{kim2018interpretability}. This approach is limited, as models may develop and rely on unexpected features or "shortcuts"~\cite{geirhos2020shortcut} for decision-making, which can surprise researchers. Therefore, there's a significant need for methods that not only test for anticipated concepts but also uncover and interpret the full range of strategies a model might employ, including those unanticipated by developers. This capability would mark a substantial advancement in our understanding of how models process information, offering insights into their internal logic and potentially highlighting biases or unintended behaviors. 

\begin{customchallenge}{Challenge 2: Theoretical framework deficiency.}
\end{customchallenge}

The field's reliance on empirical research has led to a significant gap in theoretical foundations, leaving concept-based methods without solid ground for evaluating the relevance and significance of identified concepts.
Concerning Feature visualization, they offer profound insights yet struggles with scalability and adaptability to the increasing complexity of contemporary models.

\begin{customchallenge}{Challenge 3: Scalability of Feature Visualization.}
\end{customchallenge}

The scalability of feature visualization techniques is limited, often resulting in noisy and less interpretable images on advanced models, highlighting the need for methodological renewal and innovation.
Finally, a more global issue is the lack of clear link between Attributions methods, Concepts and Feature Visuzaliation. 

\begin{customchallenge}{Challenge 4: Lack of synergy across methods}
\end{customchallenge}

Despite underlying conceptual connections, attribution, feature visualization, and concept-based methods have evolved in isolation, lacking integration and synergy.

Addressing these challenges, this chapter proposes a unified theoretical framework aimed at not just incrementally improving model explainability but trying to lay a ground for a more robust and deeper understanding of deep neural network. 

\clearpage

\section{CRAFT : Concept Recursive Activation FacTorization}
\label{sec:concepts:craft}
\definecolor{metalgreen}{RGB}{51, 157, 144}
\definecolor{metalorange}{RGB}{244, 161, 97}

We propose to directly start with our first work, \craft. As we have seen in \autoref{chap:attributions}, Attribution methods employ heatmaps to identify the most influential regions of an image that impact model decisions, and those methods have gained widespread popularity as a type of explainability method.
However, they only reveal~\where~the model looks, failing to elucidate \what~the model sees in those areas.
In this section, we will try to fill in this gap with \craft -- a novel approach to identify both ``\what'' and ``\where'' by generating concept-based explanations.
We introduce 3 new ingredients to the automatic concept extraction literature: (\textbf{i}) a recursive strategy to detect and decompose concepts across layers, (\textbf{ii}) a novel method for a more faithful estimation of concept importance using Sobol indices, and (\textbf{iii}) the use of implicit differentiation to unlock Concept Attribution Maps.

We conduct both human and computer vision experiments, specifically the one proposed in \autoref{sec:attributions:metapred}, to demonstrate the benefits of the proposed approach. We show that the proposed concept importance estimation technique -- based on Sobol indices -- is more faithful to the model than previous methods. Moreover, we have open-sourced our code at
\href{https://github.com/deel-ai/Craft}{\nolinkurl{github.com/deel-ai/Craft}}, and also in the \href{https://github.com/deel-ai/xplique}{Xplique} library.

\begin{figure}[ht]\centering
\includegraphics[width=0.99\textwidth]{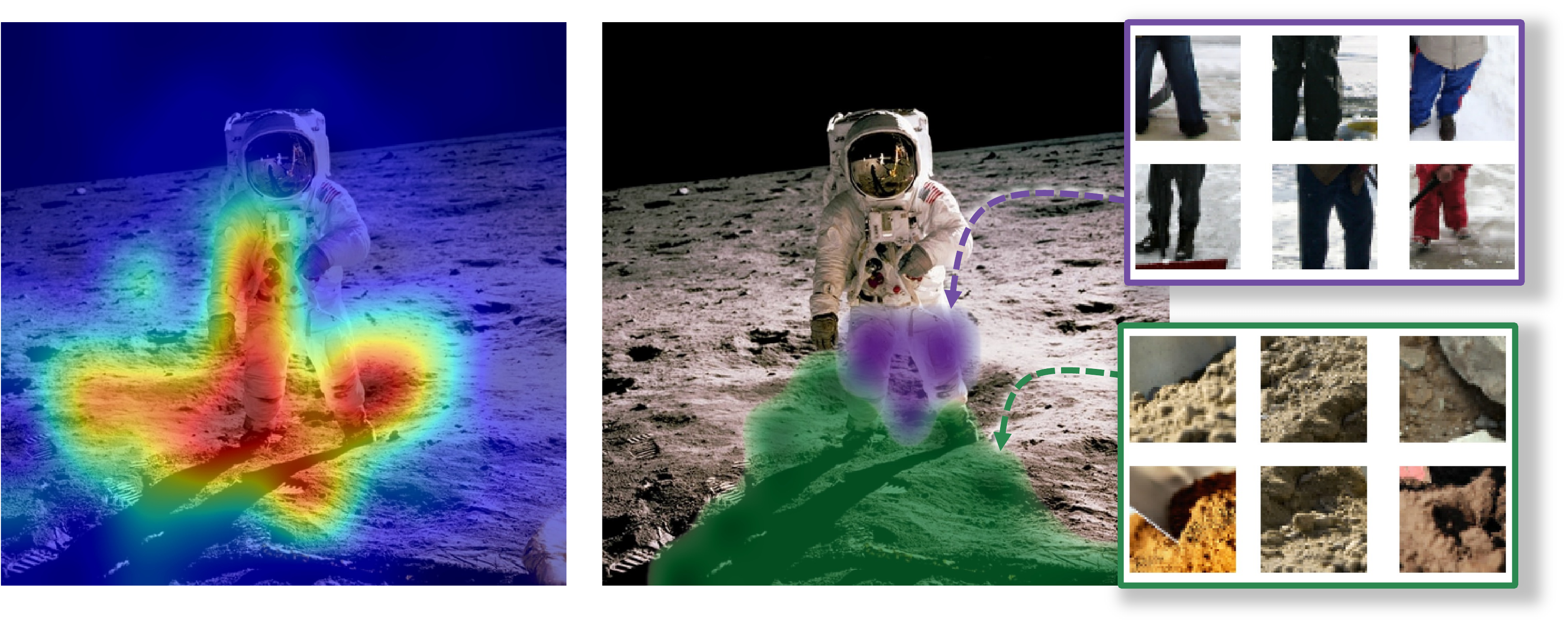}
\caption{\textbf{The ``Man on the Moon'' incorrectly classified as a ``shovel'' by an ImageNet-trained ResNet50.} Heatmap generated by a classic attribution method~\cite{petsiuk2018rise} (left) vs.  \textit{concept attribution maps} generated with the proposed \craft~approach (right) which highlights the two most influential concepts that drove the ResNet50's decision along with their corresponding locations. 
\craft~suggests that the neural net arrived at its decision because it identified the concept of ``dirt'' \textcolor{green}{$\bullet$} commonly found in members of the image class ``shovel'' and the concept of ``ski pants'' \textcolor{violet}{$\bullet$} typically worn by people clearing snow from their driveway with a shovel instead the correct concept of astronaut's pants (which was probably never seen during training).
}
\label{fig:craft:shovel}
\end{figure}

\subsection{Background}

\begin{figure*}[t!]\centering
\includegraphics[width=0.99\textwidth]{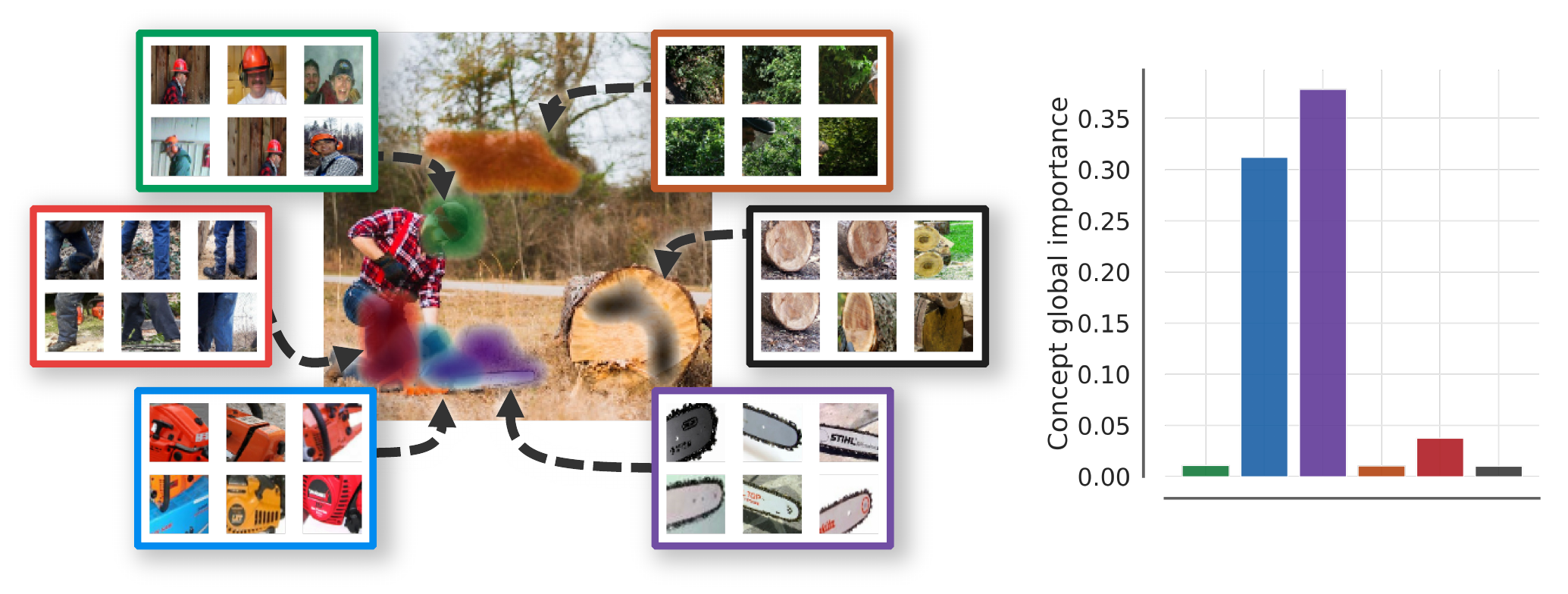}
\caption{\textbf{\craft~results for the prediction ``chain saw''.} 
First, our method uses Non-Negative Matrix Factorization (NMF) to extract the most relevant concepts used by the network (ResNet50V2) from the train set (ILSVRC2012~\cite{imagenet_cvpr09}). The global influence of these concepts on the predictions is then measured using Sobol indices (right panel). Finally, the method provides local explanations through \textit{concept attribution maps} (heatmaps associated with a concept, and computed using grad-CAM by backpropagating through the NMF concept values with implicit differentiation).
Besides, concepts can be interpreted by looking at crops that maximize the NMF coefficients. For the class ``chain saw'', the detected concepts seem to be:
\textcolor{blue}{$\bullet$} the chainsaw engine, 
\textcolor{purple}{$\bullet$} the saw blade, 
\textcolor{green}{$\bullet$} the human head, 
\textcolor{orange}{$\bullet$} the vegetation, 
\textcolor{red}{$\bullet$} the jeans and
\textcolor{dark}{$\bullet$} the tree trunk.
}
\label{fig:craft:craft_demo}
\end{figure*}

\paragraph{Attribution methods}

Attribution methods are widely used as post-hoc explainability techniques to determine the input variables that contribute to a model's prediction by generating importance maps, such as the ones shown in Fig.\ref{fig:craft:shovel}. The first attribution method, Saliency, introduced in~\cite{zeiler2014visualizing}, generates a heatmap by utilizing the gradient of a given classification score with respect to the pixels. This method was later improved upon in the context of deep convolutional networks for classification in subsequent studies, such as~\cite{zeiler2013visualizing, springenberg2014striving, sundararajan2017axiomatic, smilkov2017smoothgrad}.

Unfortunately, a severe limitation of these approaches -- apart from the fact that they only show the ``\where'' -- is that they are subject to confirmation bias: while they may appear to offer useful explanations to a user, sometimes these explanations are actually incorrect~\cite{adebayo2018sanity, ghorbani2017interpretation,slack2020fooling}.
These limitations raise questions about their usefulness, as recent research has shown by using human-centered experiments to evaluate the utility of attribution~\cite{hase2020evaluating,nguyen2021effectiveness,fel2021cannot,kim2021hive,shen2020useful}.

In particular, in our previous \autoref{sec:attributions:metapred}, we have proposed a protocol to measure the usefulness of explanations, corresponding to how much they help users identify rules driving a model's predictions (correct or incorrect) that transfer to unseen data -- using the concept of meta-predictor (also called simulatability)~\cite{kim2016examples,doshivelez2017rigorous,fong2017meaningful}. 
The main idea is to train users to predict the output of the system using a small set of images along with associated model predictions and corresponding explanations. 
A method that performs well on this this benchmark is said useful, as it help users better predict the output of the model by providing meaningful information about the internal functioning of the model.
This framework being agnostic to the type of explainability method, we have chosen to use it in Section~\ref{sec:craft:exp} in order to compare \craft~with attribution methods.

\paragraph{Concepts-based methods}
\cite{kim2018interpretability} introduced a method aimed at providing explanations that go beyond attribution-based approaches by measuring the impact of pre-selected concepts on a model's outputs. Although this method appears more interpretable to human users than standard attribution techniques, it requires a database of images describing the relevant concepts to be manually curated.
Ghorbani et al.~\cite{ghorbani2019towards} further extended the approach to extract concepts  without the need for human supervision. The approach, called ACE~\cite{ghorbani2019towards}, uses a segmentation scheme on images, that belong to an image class of interest. %
The authors leveraged the intermediate activations of a neural network for specific image segments. These segments were resized to the appropriate input size and filled with a baseline value. The resulting activations were clustered to produce prototypes, which they referred to as "concepts". However, some concepts contained background segments, leading to the inclusion of uninteresting and outlier concepts. To address this, the authors implemented a postprocessing cleanup step to remove these concepts, including those that were present in only one image of the class and were not representative. While this improved the interpretability of their explanations to human subjects, the use of a baseline value filled around the segments could introduce biases in the explanations~\cite{hsieh2020evaluations,sturmfels2020visualizing,haug2021baselines,kindermans2019reliability}.

Zhang et al.~\cite{zhang2021invertible} developed a solution to the unsupervised concept discovery problem by using matrix factorizations in the latent spaces of neural networks. However, one major drawback of this method is that it operates at the level of convolutional kernels, leading to the discovery of localized concepts. For example, the concept of "grass" at the bottom of the image is considered distinct from the concept of "grass" at the top of the image.

\begin{figure*}[t!]
\centering\includegraphics[width=1.0\textwidth]{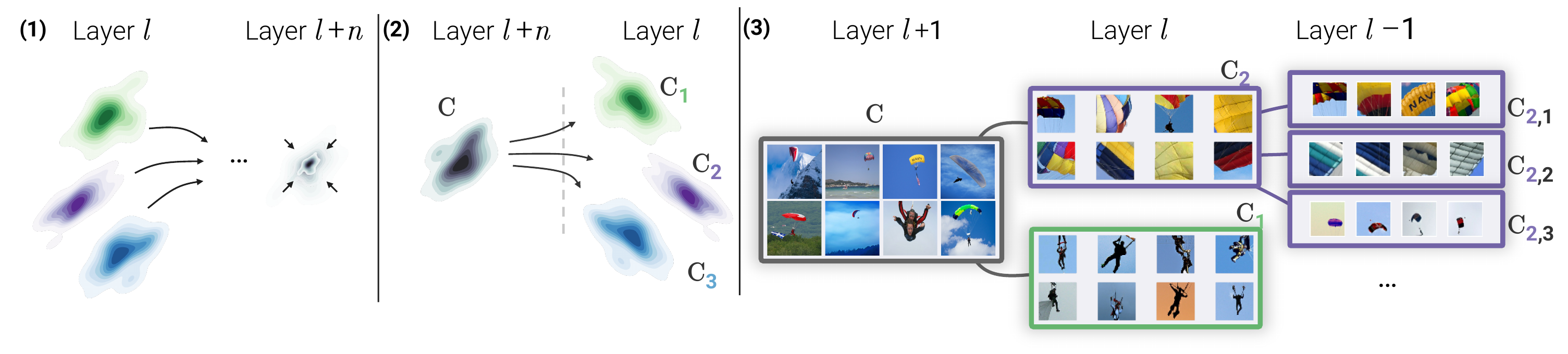}
\caption{ 
\textbf{(1) Neural collapse (amalgamation).}
A classifier needs to be able to linearly separate classes by the final layer. It is commonly assumed that in order to achieve this, image activations from the same class get progressively ``merged'' such that these image activations converge to a one-hot vector associated with the class at the level of the logits layer~\cite{paypan2020collapse}. 
In practice, this means that different concepts get ultimately blended together along the way. 
\textbf{(2) Recursive process.} When a concept is not understood (e.g., $\mathcal{C}$), we propose to decompose it into multiple sub-concepts (e.g., $\mathcal{C}_{\textcolor{green}{1}}, \mathcal{C}_{\textcolor{purple}{2}}, \mathcal{C}_{\textcolor{blue}{3}}$) using the activations from an earlier layer to overcome the aforementioned neural collapse issue.
\textbf{(3) Example of recursive concept decomposition} using \craft~on the ImageNet class ``parachute''.
}
\label{fig:craft:collapse}
\end{figure*}

Here, we try to fill these gaps with a novel method called \craft~which uses Non-Negative Matrix Factorization (NMF)~\cite{lee1999learning} for concept discovery. In contrast to other concept-based explanation methods, our approach provides an explicit link between their global and local explanations (Fig.~\ref{fig:craft:craft_demo}) and identifies the relevant layer(s) to use to represent individual concepts (Fig.~\ref{fig:craft:collapse}). Our main contributions can be described as follows:

{\textbf{(i)}} A novel approach for the automated extraction of high-level concepts learned by deep neural networks. We validate its practical utility to users with human psychophysics experiments.

{\textbf{(ii)}} A recursive procedure to automatically identify concepts and sub-concepts at the right level of granularity -- starting with our decomposition at the top of the model and working our way upstream. We validate the benefit of this approach with human psychophysics experiments showing that (i) the decomposition of a concept yields more coherent sub-concepts and (ii) that the groups of points formed by these sub-concepts are more refined and appear meaningful to humans.

{\textbf{(iii)}} A novel technique to quantify the importance of individual concepts for a model's prediction using Sobol indices~\cite{sobol1993sensitivity,da2013efficient,sobol2001,sobol2005global,saltelli2002} -- a technique borrowed from Sensitivity Analysis.

{\textbf{(iv)}} The first concept-based explainability method which produces  \textit{concept attribution maps} by backpropagating  concept scores  into the pixel space by leveraging the implicit function theorem in order to localize the pixels associated with the concept of a given input image. This effectively opens up the toolbox of both white-box~\cite{smilkov2017smoothgrad, zeiler2014visualizing, sundararajan2017axiomatic, selvaraju2017gradcam, springenberg2014striving, eva2} and black-box~\cite{ribeiro2016lime, lundberg2017unified, petsiuk2018rise, fel2021sobol} explainability methods to derive concept-wise attribution maps.

\subsection{Overview of the method} \label{sec:craft:method}

In this section, we first describe our concept activations factorization method. Below we highlight the main differences with related work.
We then proceed to introduce the three novel ingredients that make up \craft: %
(1) a method to recursively decompose concepts into sub-concepts, 
(2) a method to better estimate the importance of extracted concepts, and 
(3) a method to use any attribution method to create \textit{concept attribution maps}, using implicit differentiation~\cite{krantz2002implicit,griewank2008evaluating,blondel2021implicitdiff}.%

\paragraph{Notations}
In this work, we consider a general supervised learning setting, where $(\vx_1, ..., \vx_n) \in \sx^n \subseteq \Real^{n \times d}$ are $n$ inputs images and $(y_1, ..., y_n) \in \sy^n$ their associated labels. 
We are given a (machine-learnt) black-box predictor $\f : \sx \to \sy$, which at some test input $\vx$ predicts the output $\f(\vx)$.
Without loss of generality, we establish that $\f$ is a neural network that can be decomposed into two distinct components. The first component is a function $\v{g}$ that maps the input to an intermediate state, and the second component is $\v{h}$, which takes this intermediate state to the output, such that $\f(\vx) = (\v{h} \circ \v{g})(\vx)$. In this context, $\v{g}(\vx) \subseteq \Real^p$ represents the intermediate activations of $\vx$ within the network.
Further, we will assume non-negative activations: $ \v{g}(\vx) \geq 0$. In particular, this assumption is verified by any architecture that utilizes \textit{ReLU}, but any non-negative activation function works.

\subsubsection{Concept activation factorization.}\label{subsec:caf}

\begin{figure*}[t!]
\centering
\centering
\includegraphics[width=1.0\textwidth]{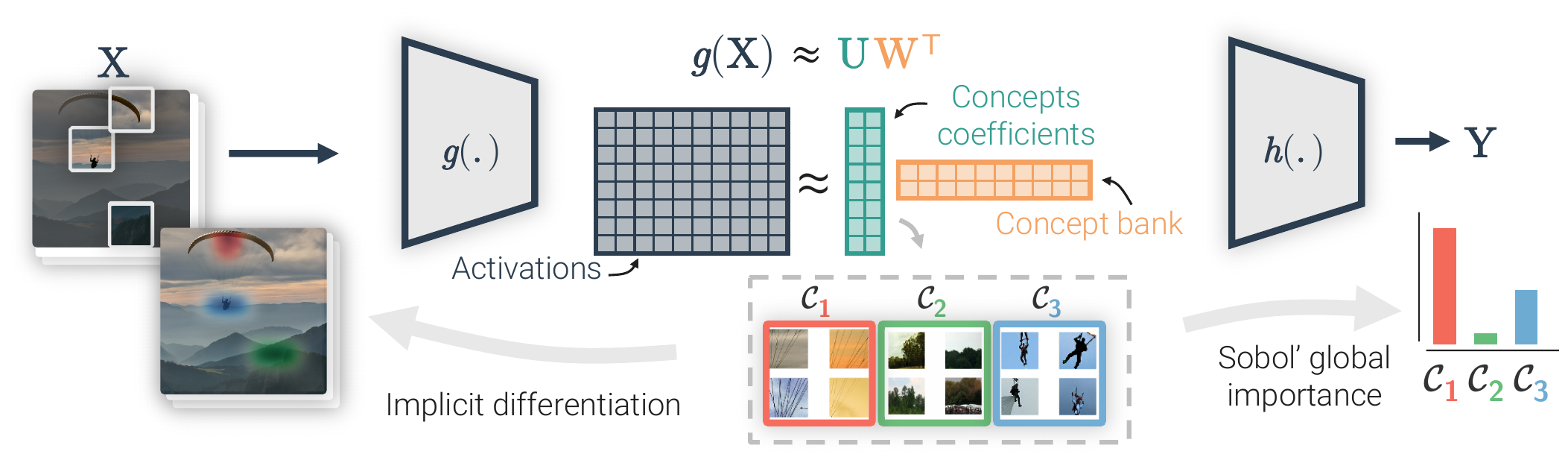}

\caption{
\textbf{Overview of \craft.}
Starting from a set of crops $\vx$ containing a concept $\mathcal{C}$ (e.g., crops images of the class ``parachute''), we compute activations $\v{g}(\vx)$ corresponding to an intermediate layer from a neural network for random image crops. 
We then factorize these activations into two lower-rank matrices, $(\textcolor{metalgreen}{\m{U}}, \textcolor{metalorange}{\m{W}})$. $\textcolor{metalorange}{\m{W}}$ is what we call a ``concept bank'' and is a new basis used to express the activations, while $\textcolor{metalgreen}{\m{U}}$ corresponds to the corresponding coefficients in this new basis.
We then extend the method with 3 new ingredients: (1) recursivity -- by proposing to re-decompose a concept (e.g., take a new set of images containing $\mathcal{C}_{\textcolor{red}{1}}$) at an earlier layer, (2) a better importance estimation using Sobol indices and (3) an approach to leverage implicit differentiation to generate \textit{concept attribution maps} to localize concepts in an image.
}
\label{fig:craft:craft}
\end{figure*}

We use Non-negative matrix factorization to identify a basis for concepts based on a network's activations (Fig.\ref{fig:craft:craft}). Inspired by the approach taken in ACE~\cite{ghorbani2019towards}, we will use image sub-regions to try to identify coherent concepts. 

The first step involves gathering a set of images that one wishes to explain, such as the dataset, in order to generate associated concepts. In our examples, to explain a specific class $y \in \sy$, we selected the set of points $\mathcal{C}$ from the dataset for which the model's predictions matched a specific class $\mathcal{C} = \{ \vx_i : \f(\vx_i) = y, 1 \leq i \leq n \}$.
It is important to emphasize that this choice is significant. The goal is not to understand how humans labeled the data, but rather to comprehend the model itself. By only selecting correctly classified images, important biases and failure cases may be missed, preventing a complete understanding of our model.

Now that we have defined our set of images, we will proceed with selecting sub-regions of those images to identify specific concepts within a localized context. It has been observed that the implementation of segmentation masks suggested in ACE can lead to the introduction of artifacts due to the associated inpainting with a baseline value.
In contrast, our proposed method takes advantage of the prevalent use of modern data augmentation techniques such as randaugment, mixup, and cutmix during the training of current models.
These techniques involve the current practice of models being trained on image crops, which enables us to leverage a straightforward crop and resize function denoted by $\bm{\pi}(\cdot)$ to create sub-regions (illustrated in Fig.\ref{fig:craft:craft}). By applying $\bm{\pi}$ function to each image in the set $\mathcal{C}$, we obtain an auxiliary dataset $\mx \in \Real^{n \times d}$ such that each entries $\mx_i = \bm{\pi}(\vx_i)$ is an image crop.

To discover the concept basis, we start by obtaining the activations for the random crops $\m{A} = \v{g}(\mx) \in \Real^{n \times p}$.
In the case where $\f$ is a convolutional neural network, a global average pooling is applied to the activations.

We are now ready to apply Non-negative Matrix Factorization (NMF) to decompose  positive activations $\m{A}$ into a product of non-negative, low-rank matrices $\m{U} \in \Real^{n \times r}$ and $\m{W} \in \Real^{p \times r}$ by solving:

\begin{equation}
\label{eq:craft:nmf}
(\m{U}, \m{W}) = \argmin_{\m{U} \geq 0, \m{W} \geq 0} ~ \frac{1}{2}\|\m{A} - \m{U}\m{W}^\tr \|^2_{F}, %
\end{equation}  
where $||\cdot||_F$ denotes the Frobenius norm.

This decomposition of our activations $\m{A}$ yields two matrices: $\m{W}$ containing our Concept Activation Vectors (CAVs) and $\m{U}$ that redefines the data points in our dataset according to this new basis. Moreover, this decomposition in this new basis has some interesting properties that go beyond the simple low-rank factorization -- since $r \ll \min(n,p)$.
First, NMF can be understood as the joint learning of a dictionary of Concept Activation Vectors -- called a ``concept bank'' in Fig.~\ref{fig:craft:craft} -- that maps a $\Real^p$ basis onto $\Real^r$, and $\m{U}$ the coefficients of the vectors $\m{A}$ expressed in this new basis. 
The minimization of the reconstruction error $\frac{1}{2}\|\m{A} - \m{U}\m{W}\|^2_F$ ensures that the new basis contains (mostly) relevant concepts. Intuitively, the non-negativity constraints $\m{U} \geq 0, \m{W} \geq 0$ encourage (\textbf{\textit{i}}) $\m{W}$ to be sparse (useful for creating disentangled concepts), (\textbf{\textit{ii}})  $\m{U}$ to be sparse (convenient for selecting a minimal set of useful concepts)  and (\textbf{\textit{iii}})  missing data to be imputed~\cite{ren2020using}, which corresponds to the sparsity pattern of \textit{post-ReLU} activations $\m{A}$. 

It is worth noting that each input $\mx_i$ can be expressed as a linear combination of concepts denoted as $\m{A}_i = \sum_{j=1}^r U_{i,j} \m{W}_j^\tr$.  This approach is advantageous because it allows us to interpret each input as a composition of the underlying concepts. Furthermore, the strict positivity of each term -- NMF is working over the anti-negative semiring, -- enhances the interpretability of the decomposition. Another interesting interpretation could be that each input is represented as a superposition of concepts~\cite{elhage2022superposition}.

While other methods in the literature solve a similar problem (such as low-rank factorization using SVD or ICA), the NMF is both fast and effective and is known to yield concepts that are meaningful to humans~\cite{fu2019nonnegative,zhang2021invertible}. Finally, once the concept bank $\m{W}$ has been precomputed, we can associate the concept coefficients $\bm{u} \in \Real^r$ to any new input $\vx$ (e.g., a full image) by solving the underlying Non-Negative Least Squares (NNLS) problem $\min_{\bm{u} \geq 0} ~ \frac{1}{2}\|\v{g}(\vx) - \bm{u}\m{W}^\tr\|^2_{F}$, and therefore recover its decomposition in the concept basis.

In essence, the core of our method can be summarized as follows: using a set of images, the idea is to re-interpret their embedding at a given layer as a composition of concepts that humans can easily understand. 
In the next section, we show how one can recursively apply concept activation factorizations to preceding layer for an image containing a previously computed concept.

\subsubsection{Ingredient 1: A pinch of recursivity}\label{subsec:rec}

One of the most apparent issues in previous work~\cite{ghorbani2019towards,zhang2021invertible} is the need for choosing a priori a layer at which the activation maps are computed. This choice will critically affect the concepts that are identified  because certain concepts get amalgamated~\cite{paypan2020collapse} into one at different layers of the neural network, resulting in incoherent and indecipherable clusters, as illustrated in Fig.~\ref{fig:craft:collapse}. We posit that this can be solved by iteratively applying our decomposition at different layer depths, and for the concepts that remain difficult to understand, by looking for their sub-concepts in earlier layers by isolating the images that contain them. This allows us to build hierarchies of concepts for each class.

We offer a simple solution consisting of reapplying our method to a concept by performing a second step of concept activation factorization on a set of images that contain the concept $\mathcal{C}$ in order to refine it and create sub-concepts (e.g., decompose $\mathcal{C}$ into $\{ \mathcal{C}_1,\mathcal{C}_2,\mathcal{C}_3 \}$) see Fig.~\ref{fig:craft:collapse} for an illustrative example. 
Note that we generalize current methods in the sense that taking images $(\vx_1, ..., \vx_n)$ that are clustered in the logits layer (belonging to the same class) and decomposing them in a previous layer -- as done in \cite{ghorbani2019towards, zhang2021invertible} -- is a valid recursive step.
For a more general case, let us assume that a set of images that contain a common concept is obtained using the first step of concept activation factorization. 

We will then take a subset of the auxiliary dataset points to refine any concept $j$. To do this, we select the subset of points that contain the concept $\mathcal{C}_j = \{\vx_i : U_{i,j} > \lambda_j, 1 \leq i \leq n \}$, where $\lambda_j$ is the 90th percentile of the values of the concept $\m{U}_{:,j}$ across the $n$ points. In other words, the 10\% of images that activate the concept $j$ the most are selected for further refinement into sub-concepts.
Given this new set of points, we can then re-apply the Concept Matrix Factorization method to an earlier layer to obtain the sub-concepts decomposition from the initial concept -- as illustrated in Fig.\ref{fig:craft:collapse}.

\subsubsection{Ingredient 2: A dash of sensitivity analysis}\label{subsec:sobol}

A major concern with concept extraction methods is that concepts that makes sense to humans are not necessarily the same as those being used by a model to classify images.
In order to prevent such confirmation bias during our concept analysis phase, a faithful estimate the overall importance of the extracted concepts is crucial. 
Kim et al.~\cite{kim2018interpretability} proposed an importance estimator based on directional derivatives: the partial derivative of the model output with respect to the vector of concepts. 
While this measure is theoretically grounded, it relies on the same principle as gradient-based methods, and thus, suffers from the same pitfalls: neural network models have noisy gradients~\cite{smilkov2017smoothgrad,sundararajan2017axiomatic}. Hence, the farther the chosen layer is from the output, the noisier the directional derivative score will be.

Since we essentially want to know which concept has the greatest effect on the output of the model, it is natural to consider the field of sensitivity analysis~\cite{sobol2005global, sobol1993sensitivity, sobol2001, cukier1973study,idrissi2021developments}.
In this section, we briefly recall the classic ``total Sobol indices'' on wich we based our previous method in the \autoref{sec:attributions:sobol}, and how to apply them to our problem. The complete derivation of the Sobol-Hoeffding decomposition for concepts is presented in Section~\ref{apdx:sobol} of the supplementary materials.
Formally, a natural way to estimate the importance of a concept $i$ is to measure the fluctuations of the model's output $\v{h}(\m{U} \m{W}^\tr)$ in response to meaningful perturbations of the concept coefficient $\m{U}_{:,i}$ across the $n$ points.
Concretely, we will use perturbation masks $\rm{M}  = (\r{M}_1, ..., \r{M}_r) \sim \mathcal{U}([0, 1]^r)$, here an i.i.d sequence of real-valued random variables, we introduce a concept fluctuation to generate a perturbed activation $\rm{A} = (\m{U} \odot \rm{M})\m{W}^\tr$ where $\odot$ denote the Hadamard product (e.g., the masks can be used to remove a concept by setting its value to zero). We can then propagate this perturbed activation to the model output and get the associated random output $\rm{Y} = \v{h}(\rm{A})$.
Simply put, removing or applying perturbation of an important concept will result in a substantial variation in the output, whereas an unused concept will have minimal effect on the output.

Finally, we can capture the importance that a concept might have as a main effect -- along with its interactions with other concepts -- on the model's output by calculating the expected variance that would remain if all the concepts except the $i$ were to be fixed. This yields the general definition of the total Sobol indices.

\begin{definition}[\textbf{Total Sobol indices for Concept}]
\textit{The total Sobol index $\mathcal{S}^T_i$, which measures the contribution of a concept $i$ as well as its interactions of any order with any other concepts to the model output variance, is given by:}

\begin{align}
\label{eq:craft:total_sobol}
\mathcal{S}^T_i 
& = \frac{ \E_{\rm{M}_{\sim i}}( \V_{M_i} ( \rm{Y} | \rm{M}_{\sim i} )) }{ \V(\rm{Y}) } \\
& = \frac{ \E_{\bm{M}_{\sim i}}( \V_{M_i} ( \v{h}((\m{U} \odot \rm{M})\m{W}^\tr) | \rm{M}_{\sim i} )) }{ \V( \v{h}((\m{U} \odot \rm{M})\m{W}^\tr)) }.
\end{align}
\end{definition}

In practice, this index can be calculated very efficiently~\cite{saltelli2010variance, marrel2009calculations, janon2014asymptotic, owen2013better, tarantola2006random}, more details on the Quasi-Monte Carlo sampling and the estimator used are left in appendix~\ref{apdx:sobol}.

\begin{figure*}[ht]
\centering
\includegraphics[width=0.95\textwidth]{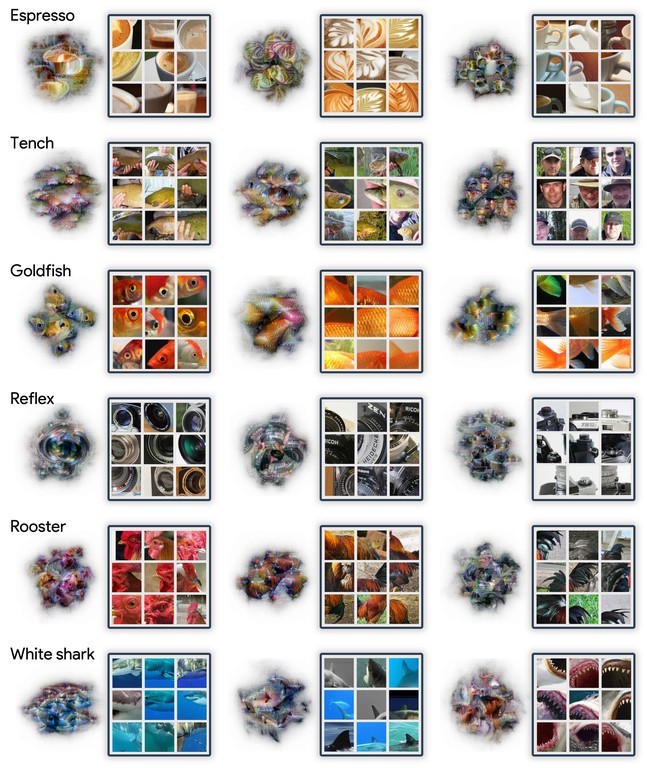}
\caption{
\textbf{Qualitative Results:} \craft~results on 6 classes of ILSVRC2012~\cite{imagenet_cvpr09} for a trained ResNet50V2. The results showcase the top 3 most important concepts for each class. This is done by displaying crop images that activate the concept the most (using $\m{U}$) and also feature visualization~\cite{olah2017feature} of the associated CAVs (using $\m{W}$). 
}
\label{fig:craft:qualitative}
\end{figure*}

\subsubsection{Ingredient 3: A smidgen of implicit differentiation}\label{subsec:cam}

Attribution methods are useful for determining the regions deemed important by a model for its decision, but they lack information about what exactly triggered it.
We have seen that we can already extract this information from the matrices $\m{U}$ and $\m{W}$, but as it is, we do not know in what part of an image a given concept is represented.
In this section, we will show how we can leverage attribution methods (forward and backward modes) to find where a concept is located in the input image (see Fig.~\ref{fig:craft:craft_demo}). Forward attribution methods do not rely on any gradient computation as they only use inference processes, whereas backward methods require back-propagating through a network's layers. By application of the chain rule, computing $\partial \m{U} / \partial \mx$ requires access to $\partial \m{U} /\partial \m{A}$.

To do so, one could be tempted to solve the linear system $\m{U}\m{W}^\tr=\m{A}$. 
However, this problem is ill-posed since $\m{W}^\tr$ is low rank. A standard approach is to calculate the Moore-Penrose pseudo-inverse $(\m{W}^\tr)^\pinv$, which solves rank deficient systems by looking at the minimum norm solution~\cite{barata2012moore}. In practice, $(\m{W}^\tr)^{\dagger}$ is computed with the Singular Value Decomposition (SVD) of $\m{W}^\tr$. Unfortunately, SVD is also the solution to the \textit{unstructured minimization} of $\frac{1}{2}\|\m{A}-\m{U}\m{W}^\tr\|^2_F$ by the Eckart-\-Young-\-Mirsky theorem~\cite{eckart1936approximation}. Hence, the non-negativity constraints of the NMF are ignored, which prevents such approaches from succeeding. Other issues stem from the fact that the $\m{U},\m{W}$ decomposition is generally not unique.

Our third contribution consists of tackling this problem to allow the use of attribution methods, i.e., \textit{concept attribution maps}, by proposing a strategy to differentiate through the NMF block.

\paragraph{Implicit differentiation of NMF block}

The NMF problem~\ref{eq:craft:nmf} is NP-hard~\cite{vavasis2010complexity}, and it is not convex with respect to the input pair $(\m{U},\m{W})$. However, fixing the value of one of the two factors and optimizing the other turns the NMF formulation into a pair of Non-Negative Least Squares (NNLS) problems, which are convex. This ensures that alternating minimization (a standard approach for NMF) of $(\m{U},\m{W})$ factors will eventually reach a local minimum.
Each of this alternating NNLS problems fulfills the Karush-–Kuhn-–Tucker (KKT) conditions~\cite{karush1939minima,kuhn1951nonlinear}, which can be encoded in the so-called \textit{optimality function} $\implicit$ from \cite{blondel2021implicitdiff}, see Eq.~\ref{apeq:craft:optimality_fun} Appendix~\ref{app:craft:implicit}. The implicit function theorem~\cite{griewank2008evaluating} allows us to use implicit differentiation~\cite{krantz2002implicit,griewank2008evaluating,bell2008algorithmic} to efficiently compute the Jacobians $\partial \m{U}/ \partial \m{A}$ and $\partial \m{W} / \partial \m{A}$ without requiring to back-propagate through each of the iterations of the NMF solver:

Let the optimality function $\implicit$, as introduced in Blondel et al. (2021) and based on the Karush-Kuhn-Tucker (KKT) conditions (Karush, 1939; Kuhn and Tucker, 1951), encapsulate the optimality conditions of the Non-negative Matrix Factorization (NMF) problem as formulated in Equation \ref{eq:craft:nmf}. The function $\implicit$ is defined for a given matrix $\m{A}$ and the tuple of matrices $(\m{U},\m{W},\bar{\m{U}},\bar{\m{W}})$ as follows:

\begin{theorem}[Implicit differentiation of NMF.] Let the optimality function $\implicit$ as introduced in~\cite{blondel2021implicitdiff} adapted for the Karush-Kuhn-Tucker (KKT) conditions~\cite{karush1939minima,kuhn1951nonlinear} capturing the optimality conditions of the problem~\ref{eq:craft:nmf} reads:

\begin{equation}
    \implicit((\m{U},\m{W},\bar{\m{U}},\bar{\m{W}}),\m{A})=
    \begin{cases}
    (\m{U}\m{W}^T-\m{A})\m{W}-\bar{\m{U}}    ,& \\ 
    (\m{W}\m{U}^T-\m{A}^T)\m{U}-\bar{\m{W}}  ,& \\ 
    \bar{\m{U}} \odot \m{U}   ,& \\ 
    \bar{\m{W}} \odot \m{W}   .& \\
    \end{cases}
\end{equation}

Given the optimal tuple $(\m{U},\m{W},\bar{\m{U}},\bar{\m{W}})$ that constitutes a root of $\implicit$ which is a root of $\implicit$, then, the implicit differentiation yields:

\begin{equation}
\frac{\partial (\m{U},\m{W},\bar{\m{U}},\bar{\m{W}})}{\partial \m{A}}=-(\partial_1 \implicit)^{-1}\partial_2 \implicit.
\end{equation}
\end{theorem}

See \autoref{app:craft:implicit} for full derivation. In particular, this requires the dual variables $\bar{\m{U}}$ and $\bar{\m{W}}$, which are not computed in scikit-learn's~\cite{pedregosa2011scikit} popular implementation\footnote{Scikit-learn uses a block coordinate descent algorithm~\cite{cichocki2009fast,fevotte2011algorithms}, with a randomized SVD initialization.}. Consequently, we leverage the work of~\cite{huang2016flexible} and we re-implement our own solver with Jaxopt~\cite{blondel2021implicitdiff} based on ADMM~\cite{boyd2011distributed}, a GPU friendly algorithm (see Appendix~\ref{app:craft:implicit}).

\begin{table*}[t]
    \centering
    \resizebox{0.9\textwidth}{!}{%
        \begin{tabular}{c l cccc cccc cccc}
        \toprule
          & & \multicolumn{4}{c}{\textit{Husky vs. Wolf}} & \multicolumn{4}{c}{\textit{Leaves}} & \multicolumn{4}{c}{\textit{``Kit Fox'' vs ``Red Fox''}} \\
         \midrule
       & Session~n$^{\circ}$               & 1 & 2 & 3 & \metric & 1 & 2 & 3 & \metric & 1 & 2 & 3 & \metric \\
         \midrule
        & Baseline                    & 55.7 & 66.2 & 62.9 &  &       70.1 & 76.8 & 78.6 &  &       58.8 & 62.2 & 58.8 &  \\
        & Control                             & 53.3 & 61.0 & 61.4 & 0.95 &       72.0 & 78.0 & 80.2 & 1.02 &     60.7 & 59.2 & 48.5 & 0.94 \\
        \midrule
        \multirow{6}{*}{\rotatebox[origin=c]{90}{{\footnotesize Attributions}}}
        & Saliency                  & 53.9 & 69.6 & 73.3 & 1.06  &        83.2 & 88.7 & 82.4 & \underline{1.13} &      61.7 & 60.2 & 58.2 & 1.00 \\ 
        & Integ.-Grad.     & 67.4 & 72.8 & 73.2 & 1.15 &      82.5 & 82.5 & 85.3 & 1.11 &       59.4 & 58.3 & 58.3 & 0.98\\
        & SmoothGrad           & 68.7 & 75.3 & 78.0 & 1.20 &    83.0 & 85.7 & 86.3 & \underline{1.13} &       50.3 & 55.0 & 61.4 & 0.93 \\
        & GradCAM               & 77.6 & 85.7 & 84.1 & 1.34 &       81.9 & 83.5 & 82.4 & 1.10 &      54.4 & 52.5 & 54.1 & 0.90 \\
        & Occlusion            & 71.0 & 75.7 & 78.1 & 1.22 &       78.8 & 86.1 & 82.9 & 1.10 &     51.0 & 60.2 & 55.1 & 0.92 \\
        & Grad.-Input               & 65.8 & 63.3 & 67.9 & 1.06 &      76.5 & 82.9 & 79.5 & 1.05 &      50.0 & 57.6 & 62.6 & 0.95 \\

        \midrule
        
        \multirow{3}{*}{\rotatebox[origin=c]{90}{{\footnotesize Concepts}}}
        
        & ACE & 68.8 & 71.4 & 72.7 & 1.15 &   79.8 & 73.8 & 82.1 & 1.05 &    48.4 & 46.5 & 46.1 & 0.78 \\

        & CRAFTCO (ours) & 82.4 & 87.0 & 85.1 & \underline{1.38} &  78.8 & 85.5 & 89.4 & 1.12 &    55.5 & 49.5 & 53.3 & 0.88 \\
        & CRAFT (ours) & 90.6 & 97.3 & 95.5 & \textbf{1.53} &   86.2 & 86.6 & 85.5 & \textbf{1.15} &    56.5 & 50.6 & 49.4 & 0.87 \\
        \bottomrule
        \end{tabular}%
    }
    \caption{\textbf{\metric~scores on 3 datasets from 
    \cite{fel2021cannot}} (presented in \autoref{sec:meta_pred}). \metric~benchmark evaluates how well explanations help users identify general rules driving classifications that readily transfer to unseen instances. At training time, users are asked to infer rules driving the decisions of the model given a set of images, and their associated predictions and explanations. At test time, the \metric~metric measures the accuracy of users at predicting the model decision on novel images averaged over 3 sessions, and normalized by the baseline accuracy of users trained without explanations.
    The higher the \metric~score, the more useful the explanation, and the more crucial the information provided is for understanding --and thus predicting the model's output-- on novel samples. %
    CRAFTCO stands for ``CRAFT Concept Only'' and designates an experimental condition where only global concepts are given to users, without local explanations (i.e., the concept attribution maps).
    The first and second best results above the baseline are in \textbf{bold} and \underline{underlined}, respectively. 
    }
    \label{tab:utility}
\end{table*}

Concretely, given our concepts bank $\m{W}$, the concept attribution maps of a new input $\vx$ are calculated by solving the NNLS problem $\min_{\m{U} \geq 0} \frac{1}{2}\|\v{g}(\vx)-\m{U}\m{W}^\tr\|^2_F$. The implicit differentiation of the NMF block $\partial \m{U} / \partial \m{A}$ is integrated into the classic back-propagation to obtain $\partial \m{U} / \partial \vx$. Most interestingly, this technical advance enables the use of all white-box explainability methods~\cite{smilkov2017smoothgrad, zeiler2014visualizing, sundararajan2017axiomatic, selvaraju2017gradcam, springenberg2014striving} to generate concept-wise attribution maps and trace the part of an image that triggered the detection of the concept by the network. Additionally, it is even possible to employ black-box methods~\cite{ribeiro2016lime, petsiuk2018rise, lundberg2017unified, fel2021sobol} since it only amounts to solving an NNLS problem. %

\subsection{Experimental evaluation}\label{sec:craft:exp}

In order to evaluate the interest and the benefits brought by \craft, we start in Section~\ref{sec:craft:expUtility} by assessing the practical utility of the method on a human-centered benchmark composed of 3 XAI scenarios and presented in~\autoref{sec:attributions:metapred}.

After demonstrating the usefulness of the method using these human experiments, we independently validate the 3 proposed ingredients.
First, we provide evidence that recursivity allows refining concepts, making them more meaningful to humans using two additional human experiments in Section \ref{sec:craft:expRecursivity}.
Next, we evaluate our new Sobol estimator and show quantitatively that it provides a more faithful assessment of concept importance in Section~\ref{sec:craft:expSobol}.
Finally, we run an ablation experiment that measures the interest of local explanations based on concept attribution maps coupled with global explanations.
Additional experiments, including a sanity check and an example of deep dreams applied on the concept bank, as well as many other examples of local explanations for randomly picked images from ILSVRC2012, are included in Section~\ref{apx:craft:more-craft} of the supplementary materials.
We leave the discussion on the limitations of this method and on the broader impact in appendix~\ref{apx:craft:limitations}.

\subsubsection{Utility Evaluation}
\label{sec:craft:expUtility}

As emphasized by Doshi-Velez et al.~\cite{doshivelez2017rigorous}, the goal of XAI should be to develop methods that help a user better understand the behavior of deep neural network models. An instantiation of this idea was proposed in \autoref{sec:attributions:metapred} where we described an experimental framework to quantitatively measure the practical usefulness of explainability methods in real-world scenarios. In the initial setup, we recruited  $n=1,150$ online participants (evaluated over 8 unique conditions and 3 AI scenarios) -- making it the largest benchmark to date in XAI. Here, we extend our framework to allow for the robust evaluation of the utility of our proposed \craft~method and the related ACE.
The 3 representative real-world scenarios are: (1) identifying bias in an AI system (using Husky vs Wolf dataset from~\cite{ribeiro2016lime}), (2) characterizing the visual strategy that are too difficult for an untrained non-expert human observer (using  the Paleobotanical dataset from \cite{wilf2016computer}), (3) understanding complex failure cases (using ImageNet ``Red fox'' vs ``Kit fox'' binary classification).
Using this benchmark, we evaluate \craft, ACE, as well as \craft~with only the global concepts (\craft CO) to allow for a fair comparison with ACE.
To the best of our knowledge, we are the first to systematically evaluate concept-based methods against attribution methods.

Results are shown in Table~\ref{tab:utility} and demonstrate the benefit of \craft, which achieves higher scores than all of the attribution methods tested as well as ACE in the first two scenarios. To date, no method appears to exceed the baseline on the third scenario suggesting that additional work is required.
We also note that, in the first two scenarios, \craft CO is one of the best-performing methods and it always outperforms ACE -- meaning that even without the local explanation of the concept attribution maps, \craft~largely outperforms ACE. Examples of concepts produced by \craft~are shown in the Appendix~\ref{app:craft:utility}.

\subsubsection{Validation of Recursivity}
\label{sec:craft:expRecursivity}

\begin{table}
\centering
\begin{tabular}{lll}
\toprule
& Experts ($n=36$) & Laymen ($n=37$)\\
\cmidrule[0.1pt](lr){2-3}
\textit{Intruder}  \\
\cmidrule[0.1pt](lr){1-3}
Acc. Concept     & 70.19\%  & 61.08\%     \\
Acc. Sub-Concept & 74.81\% ($p = 0.18$)  & \textbf{67.03}\% ($p = 0.043$)      \\
\midrule
\textit{Binary choice} \\
\cmidrule[0.1pt](lr){1-3}
Sub-Concept & \textbf{76.1}\% ($p < 0.001$) & \textbf{74.95}\% ($p < 0.001$)\\
Odds Ratios & $3.53$ & $2.99$\\
\bottomrule
\end{tabular}
\caption{\textbf{Results from the psychophysics experiments to validate the recursivity ingredient. }}\label{tab:results}
\end{table}

To evaluate the meaningfulness of the extracted high-level concepts, we performed psychophysics experiments with human subjects, whom we asked to answer a survey in two phases. Furthermore, we distinguished two different audiences: on the one hand, experts in machine learning, and on the other hand, people with no particular knowledge of computer vision. Both groups of participants were volunteers and did not receive any monetary compensation. Some examples of the developed interface are available the appendix~\ref{app:craft:human-exp}. It is important to note that this experiment was carried out independently from the utility evaluation and thus it was setup differently.
\newline\textbf{Intruder detection experiment} First, we ask users to identify the intruder out of a series of five image crops belonging to a certain class, with the odd one being taken from a different concept but still from the same class. Then, we compare the results of this intruder detection with another intruder detection, this time, using a concept (e.g., $\mathcal{C}_1$) coming from a layer $l$ and one of its sub-concepts (e.g., $\mathcal{C}_{12}$ in Fig.\ref{fig:craft:collapse}) extracted using our recursive method. If the concept (or sub-concept) is coherent, then it should be easy for the users to find the intruder.
Table~\ref{tab:results} summarizes our results, showing that indeed both concepts and sub-concepts are coherent, and that recursivity can lead to a slightly higher understanding of the generated concepts (significant for non-experts, but not for experts) and might suggest a way to make concepts more interpretable.
\newline\textbf{Binary choice experiment} In order to test the improvement of coherence of the sub-concept generated by recursivity with respect to the larger parent concept, we showed participants an image crop belonging to both a subcluster and a parent cluster (e.g.,  $\bm{\pi}(\vx) \in \mathcal{C}_{11} \subset \mathcal{C}_1$) and asked them which of the two clusters (i.e., $\mathcal{C}_{11}$ or $\mathcal{C}_{1}$) seemed to accommodate the image the best. If our hypothesis is correct, then the concept refinement brought by recursivity should help form more coherent clusters.
The results in Table~\ref{tab:results} are satisfying since in both the expert and non-expert groups, the participants chose the sub-cluster more than 74\% of the time. We measure the significance of our results by fitting a binomial logistic regression to our data, and we find that both groups are more likely to choose the sub-concept cluster (at a $p < 0.001$).

\subsubsection{Fidelity analysis} \label{sec:craft:expSobol}

\begin{figure}[ht]
\includegraphics[width=\linewidth]{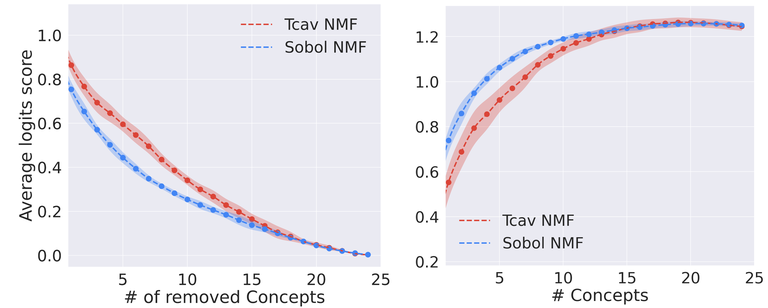}
\caption{
\textbf{(Left)} Deletion curves (lower is better). \textbf{(Right)} Insertion curves (higher is better). 
For both the deletion or insertion metrics, Sobol indices lead to better estimates (calculated on >100K images) of important concepts. %
}
\label{fig:craft:deletion}
\end{figure}

We propose to simultaneously verify that identified concepts are faithful to the model and that the concept importance estimator performs better than that used in TCAV~\cite{kim2018interpretability} by using the fidelity metrics introduced in \cite{ghorbani2019towards, zhang2021invertible}. These metrics are similar to the ones used for attribution methods, which consist of studying the change of the logit score when removing/adding pixels considered important. Here, we do not introduce these perturbations in the pixel space but in the concept space: once $\m{U}$ and $\m{W}$ are computed, we reconstruct the matrix $\m{A}\approx \m{U}\m{W}^\tr$ using only the most important concept (or removing the most important concept for deletion) and compute the resulting change in the output of the model.  As can be seen from Fig.~\ref{fig:craft:deletion}%
, ranking the extracted concepts using Sobol's importance score results in steeper curves than when they are sorted by their TCAV scores. %
We confirm that these results generalize with other matrix factorization techniques (PCA, ICA, RCA) in Section~\ref{app:craft:fidelity} of the Appendix.

\subsection{Conclusion}

In this first section, we introduced \craft, a method for automatically extracting human-interpretable concepts from deep networks. Our method aims to explain a pre-trained model's decisions both on a per-class and per-image basis by highlighting both ``\what'' the model saw and ``\where'' it saw it  -- with complementary benefits. The approach relies on 3 novel ingredients: \tbi{i} a recursive formulation of concept discovery to identify the correct level of granularity for which individual concepts are understandable; \tbi{ii} a novel method for measuring concept importance through Sobol indices to more accurately identify which concepts influence a model's decision for a given class; and \tbi{iii} the use of implicit differentiation methods to backpropagate through non-negative matrix factorization (NMF) blocks to allow the generation of concept-wise local explanations or \textit{concept attribution maps} independently of the attribution method used. Using our previously introduced human-centered utility benchmark, we conducted psychophysics experiments to confirm the validity of the approach: and that the concepts identified by \craft~are useful and meaningful to human experimenters. 

\clearpage

\section{Application: FRSign}
\label{sec:concepts:frsign}
In this comprehensive examination, we extend our investigation into the utility of concept-based methods applied to models trained on the FRSign dataset~\cite{2020frsign}, aiming to delve deeper than conventional attribution methods allow. This inquiry builds upon our previous work outlined in~\autoref{sec:attribution:frsign}, where we expressed reservations about the strategies the model employs, particularly concerning the interpretation of white signals.

In this section, our exploration is structured in three parts: initially, we conduct a review of classes for which we hypothesize the model's behavior aligns closely with expectations. Subsequently, we direct our focus towards the more enigmatic white signal. Finally, we venture further by examining secondary concepts, leading us to propose a hypothesis we term ``support concepts''.

\begin{figure}[ht]
\centering
\includegraphics[width=0.9\textwidth]{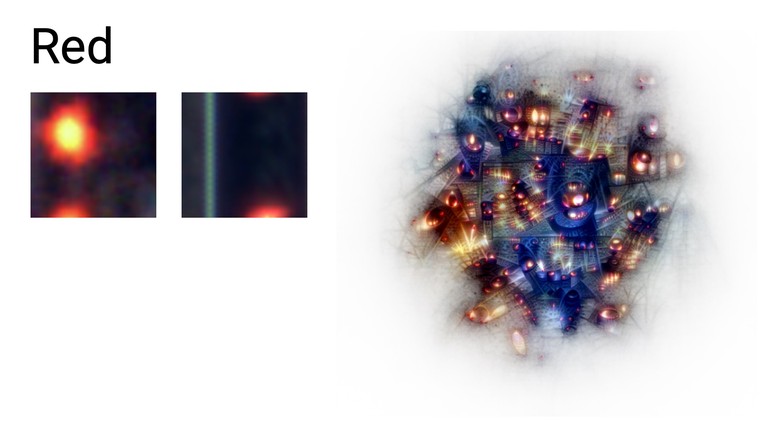}
\includegraphics[width=0.9\textwidth]{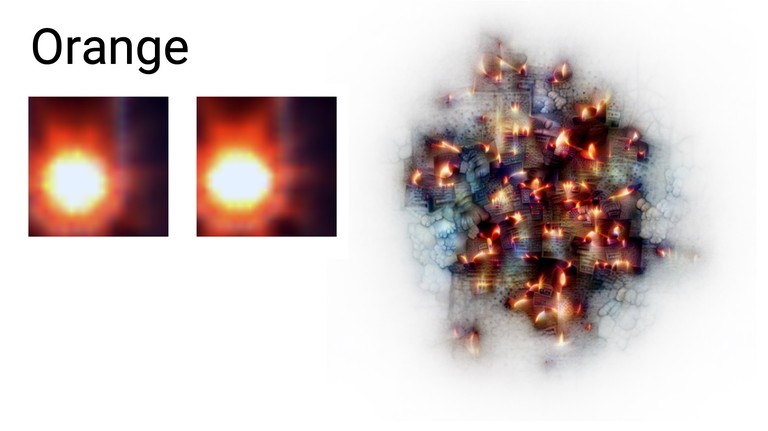}
\caption{\textbf{Most Important Concepts for Red and Orange Class.} Applying~\craft method, we extracted the most significant concepts for the ResNet50 model trained on the FRSign dataset. Consistent with findings from~\autoref{sec:attribution:frsign}, the key concepts appear reasonable and are focused on the traffic light itself, reaffirming the model's attention to relevant features.}
\label{fig:frsign:good_concepts}
\end{figure}

\subsection{Visual inspection using concepts}

We commence with a visual inspection of the model's behavior across various classes using our concept-based method, \craft. We visualize the most important concepts that the ResNet50 model -- as detailed in~\autoref{sec:intro:frsign} --  leverages. 

The~\autoref{fig:frsign:good_concepts} illustrates these concepts which, as hypothesized in~\autoref{sec:attribution:frsign}, appear aligned and plausible. In this instance, the concepts do not seem to offer substantial new insights at first glance. We will now proceed to address the challenging case highlighted in the previous section: the interpretation of white signals.

\begin{figure}[ht]
\centering
\includegraphics[width=0.48\textwidth]{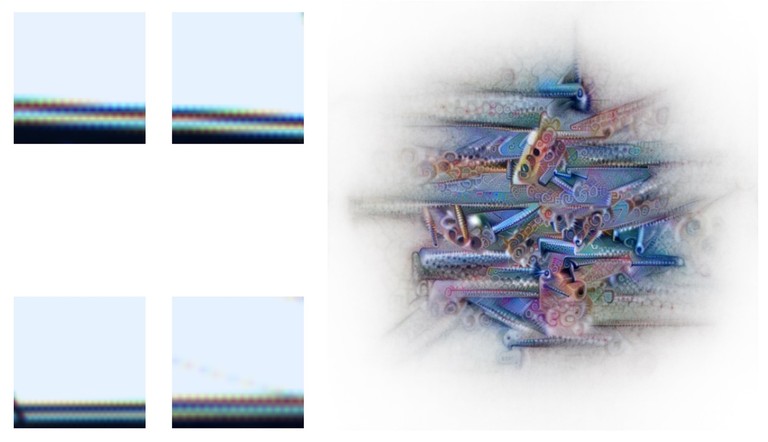}
\includegraphics[width=0.48\textwidth]{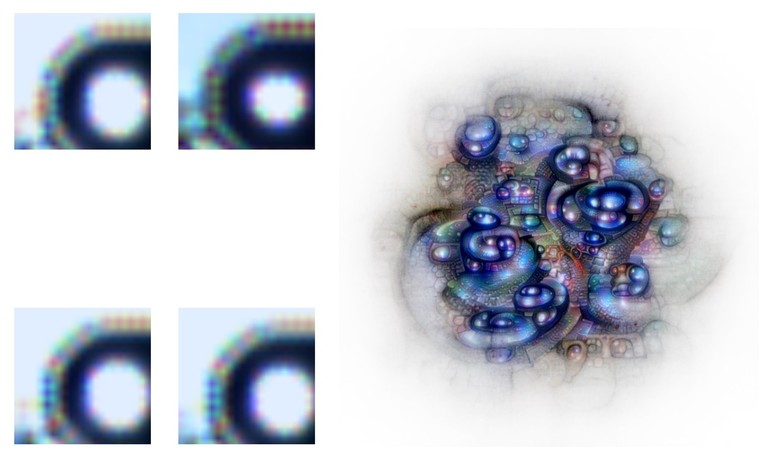}
\caption{\textbf{Most and second most important concepts for White class.} The predominant concept for the white signal class seems to be the shear effect on image borders. This finding not only confirms our earlier hypothesis but also sheds additional light on the significance of the frame's edge in the model's decision-making process.}
\label{fig:frsign:white_concepts}
\end{figure}

\subsection{Understanding the White Signal Case}

Our analysis takes a deeper dive into the peculiar case of the white signal, where the primary concept identified appears to be the shear effect on images around the edges of the frame (\autoref{fig:frsign:white_concepts}). This observation supports and further illuminates our previous hypothesis from~\autoref{sec:attribution:frsign}, suggesting that the frame's edge plays a crucial role in the model's interpretation.

\begin{figure}[ht]
\centering
\includegraphics[width=0.45\textwidth]{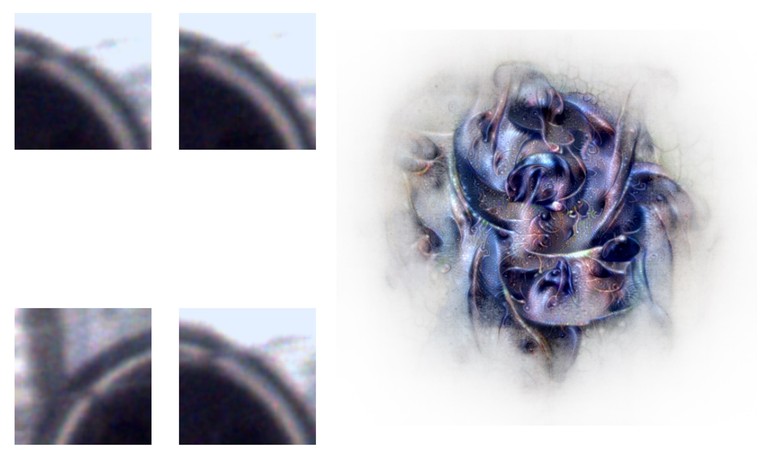}
\includegraphics[width=0.45\textwidth]{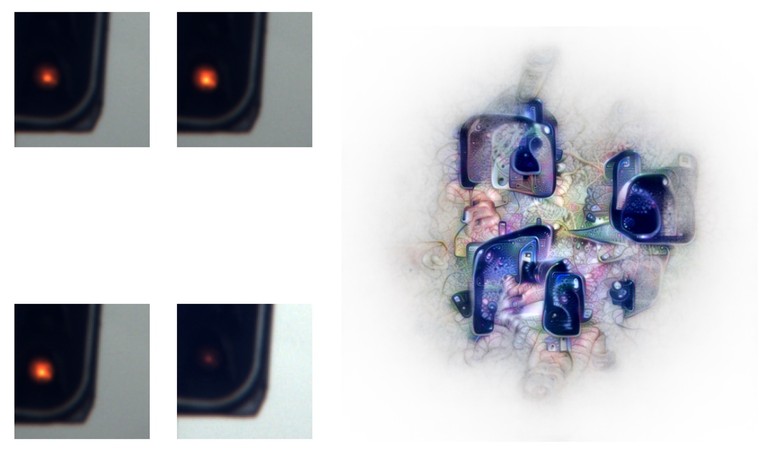}
\includegraphics[width=0.45\textwidth]{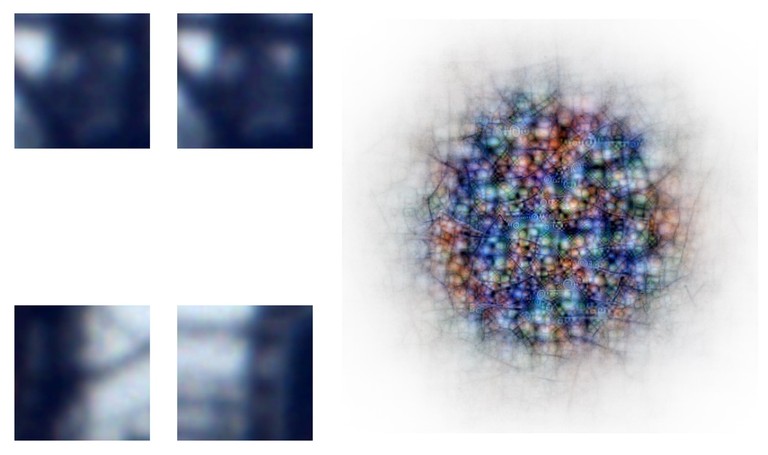}
\includegraphics[width=0.45\textwidth]{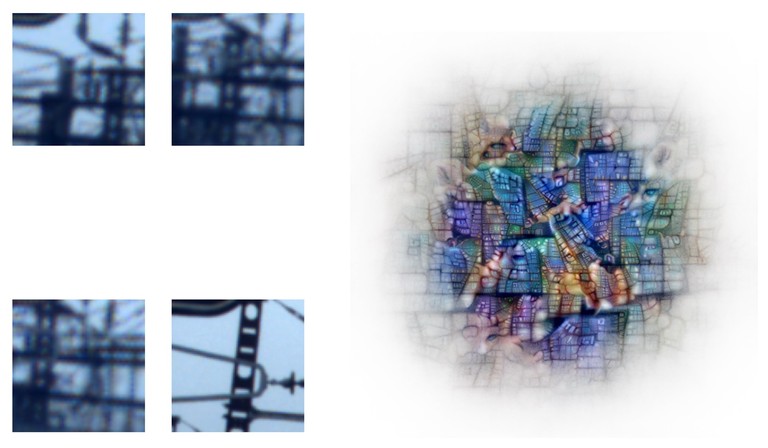}
\caption{\textbf{Examples of Four Secondary, Yet Significant, Concepts.} These concepts are used for the orange, red, yellow, and violet logits, respectively. Although they are not the top-1 concepts, this does not mean they do not contribute to the logits. A portion of the logits is influenced by these concepts, underscoring their importance in the model's decision-making process.}
\label{fig:frsign:support}
\end{figure}

\subsection{Hypothesis: Support Concepts}

In an effort to further our understanding, we investigate secondary, yet influential concepts, which we refer to as "support concepts." These concepts, while not being the top-1 most important for a given class, still significantly contribute to the model's logits for various signals, as illustrated in \autoref{fig:frsign:support}.

We can conjecture that, more problematically, attribution methods that highlight the most important pixels or areas may overlook several features that also drive decision-making and could represent shortcuts. Thus, these concepts could be "hidden" by the attribution maps but still present internally. We introduce the idea of "support concepts" as an avenue for future work, suggesting that a deeper exploration into these underlying influences could unveil additional layers of model reasoning not immediately apparent through conventional attribution techniques.

\subsection{Conclusion}

The application of concept-based explanations has provided us with a more granular understanding of the model's behavior, particularly elucidating the case of the white signal. It appears that biases, possibly inherent in the dataset, necessitate a broader collection of images to mitigate such issues. Alarmingly, our analysis confirms that "support concepts," while not paramount for a class, play a critical role in achieving high performance levels. This revelation affirms the pervasive nature of biases and shortcuts in model training. Consequently, concept-based explainability holds promising potential for unveiling these complexities in model interpretations, offering a path towards more transparent and interpretable machine learning models.

\clearpage

\section{Unifying Automatic Concept Extraction and Concept Importance Estimation}
\label{sec:concepts:holistic}
\definecolor{color1}{RGB}{67, 160, 71}
\definecolor{color2}{RGB}{53, 183, 121}
\definecolor{color3}{RGB}{62, 74, 137}

\newcommand{\ACE}{\textbf{\textcolor{color1}{ACE}}}
\newcommand{\ICE}{\textbf{\textcolor{color2}{ICE}}}

\newcommand{\fa}{\bm{g}}
\newcommand{\fb}{\bm{h}}

In the first section (\autoref{sec:concepts:craft}), we have introduced a first framework able to automatically extract concept and estimate their importance. Recently, other approaches have been proposed, either for concept extraction or concept importance estimation. However no proper metric, benchmark or theoretical framework have been proposed. In this section, we start by noticing that all current concept-based approaches seek discover intelligible visual ``concepts'' buried within the complex patterns of activations using two key steps: (1) concept extraction followed by (2) importance estimation. Again, while these two steps are shared across methods, they all differ in their specific implementations.

Starting from that, we introduce a unifying theoretical framework that recast the first step -- concept extraction problem -- as a special case of \textbf{dictionary learning}, and we formalize the second step -- concept importance estimation -- as a more general form of \textbf{attribution method}.
This framework offers several advantages as it allows us: \tbi{i} to propose new evaluation metrics for comparing different concept extraction approaches; \tbi{ii} to leverage modern attribution methods and evaluation metrics to extend and systematically evaluate state-of-the-art concept-based approaches and importance estimation techniques; \tbi{iii}  to derive theoretical guarantees regarding the optimality of such methods.

We further leverage our framework to try to tackle a crucial question in explainability: how to \textit{efficiently} identify clusters of data points that are classified based on a similar shared strategy.
To illustrate these findings and to highlight the main strategies of a model, we introduce a visual representation called the strategic cluster graph.

\subsection{Introduction}

One promising set of explainability methods to address the issue posed in~\autoref{hyp:what} includes concept-based explainability methods, which are methods that aim to identify high-level concepts within the activation space of ANNs~\cite{kim2018interpretability}. These methods have recently gained renewed interest due to their success in providing human-interpretable explanations~\cite{ghorbani2019towards, zhang2021invertible, fel2023craft, graziani2023concept}. However, concept-based explainability methods are still in the early stages, and progress relies largely on researchers' intuitions rather than well-established theoretical foundations.  A key challenge lies in formalizing the notion of concept itself~\cite{genone2012concept}. 
Researchers have proposed desiderata such as meaningfulness, coherence, and importance~\cite{ghorbani2019towards} but the lack of formalism in concept definition has hindered the derivation of appropriate metrics for comparing different methods.

\begin{figure}[ht]
\begin{center}
   \includegraphics[width=.99\textwidth]{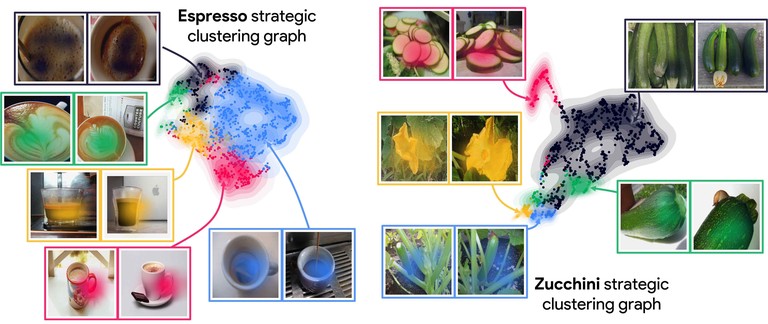}
\end{center}
   \caption{\textbf{Strategic cluster graphs for the espresso and zucchini classes.}
    The framework presented in this section provides a comprehensive approach to uncover local importance using any attribution methods. 
    Consequently, it allow us to estimate the critical concepts influencing the model's decision for each image.
    As a results, we introduced the Strategic cluster graph, which offers a visual representation of the main strategies employed by the model in recognizing an entire object class.
    For espresso (left), the main strategies for classification appear to be: \textcolor{anthracite}{$\bullet$} bubbles and foam on the coffee, \textcolor{green}{$\bullet$} Latte art, \textcolor{yellow}{$\bullet$} transparent cups with foam and black liquid, \textcolor{red}{$\bullet$} the handle of the coffee cup, and finally \textcolor{blue}{$\bullet$} the coffee in the cup, which appears to be the predominant strategy.
    As for zucchini, the strategies are: \textcolor{blue}{$\bullet$} a zucchini in a vegetable garden, \textcolor{yellow}{$\bullet$} the corolla of the zucchini flower, \textcolor{red}{$\bullet$} sliced zucchini, \textcolor{green}{$\bullet$} the spotted pattern on the zucchini skin and \textcolor{anthracite}{$\bullet$} stacked zucchini.
   }
\label{fig:holistic:clustering_graph}
\end{figure}

This section presents a theoretical framework to unify and characterize current concept-based explainability methods. Our approach builds on the fundamental observation that all concept-based explainability methods share two key steps: (1) concepts are extracted, and (2) importance scores are assigned to these concepts based on their contribution to the model's decision~\cite{ghorbani2019towards}. Here, we show how the first extraction step can be formulated as a dictionary learning problem while the second importance scoring step can be formulated as an attribution problem in the concept space. To summarize, our contributions are as follows:

\setlist[itemize]{leftmargin=5.5mm}
\begin{itemize}
    \item We describe a novel framework that unifies all modern concept-based explainability methods and we borrow metrics from different fields (such as sparsity, reconstruction, stability, FID, or OOD scores) to evaluate the effectiveness of those methods. 
    \item We leverage modern attribution methods to derive seven novel concept importance estimation methods and provide theoretical guarantees regarding their optimality. 
    Additionally, we show how standard faithfulness evaluation metrics used to evaluate attribution methods (i.e., Insertion, Deletion~\cite{petsiuk2018rise}, and $\mu$Fidelity~\cite{aggregating2020}) can be adapted to serve as benchmarks for concept importance scoring.
    In particular, we demonstrate that Integrated Gradients, Gradient Input, RISE, and Occlusion achieve the highest theoretical scores for 3 faithfulness metrics when the concept decomposition is on the penultimate layer. 
    \item We introduce the notion of local concept importance to address a significant challenge in explainability: the identification of image clusters that reflect a shared strategy by the model (see Figure~\ref{fig:holistic:clustering_graph}). We show how the corresponding cluster plots can be used as visualization tools to help with the identification of the main visual strategies used by a model to help explain false positive classifications. 
\end{itemize}

\subsection{A Unifying perspective}

\paragraph{Notations.} Throughout, $||\cdot||_2$ and $||\cdot||_F$ represent the $\ell_2$ and Frobenius norm, respectively. 
We consider a general supervised learning setting, where a classifier $\f : \sx \to \sy$ maps inputs from an input space $\mathcal{X} \subseteq \Real^d$ to an output space $\sy \subseteq \Real^c$. 
For any matrix $\mx \in \Real^{n \times d}$, $\vx_i$ denotes the $i^{th}$ row of $\mx$, where $i \in \{1, \ldots, n \}$ and $\vx_i \in \Real^{d}$.
Without loss of generality, we assume that $\f$ admits an intermediate space $\s{I} \subseteq \Real^p$. In this setup, $\fa : \sx \to \s{I}$ maps inputs to the intermediate space, and $\fb : \s{I} \to \sy$ takes the intermediate space to the output. Consequently, $\f(\vx) = (\fb \circ \fa)(\vx)$. Additionally, let $\bm{a} = \fa(\vx) \in \s{I}$ represent the activations of $\vx$ in this intermediate space.
We also abuse notation slightly: $\f(\mx) = (\fb \circ \fa)(\mx)$ denotes the vectorized application of $\f$ on each element $\vx$ of $\mx$, resulting in $(\f(\vx_1),\ldots, \f(\vx_n))$.

\paragraph{2 Fundamental steps.} Prior methods for concept extraction, namely \ACE~\cite{ghorbani2019towards}, \ICE~\cite{zhang2021invertible}~and \CRAFT~\cite{fel2023craft}, can be distilled into two fundamental steps:

\begin{enumerate}[label=(\textit{\textbf{\roman*}}), labelindent=0pt,leftmargin=5mm]

\item {\bf Concept extraction:} A set of images $\mx \in \Real^{n\times d}$ belonging to the same class is sent to the intermediate space giving activations $\m{A} = \fa(\mx) \in \Real^{n \times p}$.
These activations are used to extract a set of $k$ CAVs using K-Means~\cite{ghorbani2019towards}, PCA (or SVD)~\cite{zhang2021invertible, graziani2023concept} or NMF~\cite{zhang2021invertible, fel2023craft}. Each CAV is denoted $\cav_i$ and $\m{V} = (\cav_1, \ldots, \cav_k) \in \Real^{p \times k}$ forms the dictionary of concepts.

\item {\bf Concept importance scoring:} 
It involves calculating a set of $k$ global scores, which provides an importance measure of each concept $\cav_i$ to the class as a whole. Specifically, it quantifies the influence of each concept $\cav_i$ on the final classifier prediction for the given set of points  $\mx$. Prominent measures for concept importance include TCAV~\cite{kim2018interpretability} and the Sobol indices~\cite{fel2023craft}. 

\end{enumerate}

The two-step process described above is repeated for all classes. In the following subsections, we theoretically demonstrate that the concept extraction step \tbi{i} could be recast as a dictionary learning problem (see~\ref{sec:dico_learning}). It allows us to reformulate and generalize the concept importance step \tbi{ii} using attribution methods (see~\ref{sec:importance}). 

\subsubsection{Concept Extraction}

\paragraph{A dictionary learning perspective.}
\label{sec:dico_learning} 
The purpose of this section is to redefine all current concept extraction methods as a problem within the framework of dictionary learning. Given the necessity for clearer formalization and metrics in the field of concept extraction, integrating concept extraction with dictionary learning enables us to employ a comprehensive set of metrics and obtain valuable theoretical insights from a well-established and extensively researched domain. 

The goal of concept extraction is to find a small set of interpretable CAVs (i.e., $\m{V}$) that allows us to faithfully interpret the activation $\m{A}$. By preserving a linear relationship during the reconstruction, from $\m{U}$ to $\m{A}$ (and not necessarily from $\m{A}$ to $\m{U}$), we facilitate the understanding and interpretability of the learned concepts~\cite{kim2018interpretability, elhage2022superposition}. Therefore, we look for a coefficient matrix $\m{U} \in \Real^{n \times k}$ (also called loading matrix) and a set of CAVs $\m{V}$, so that $\m{A} \approx \m{U} \m{V}^\tr$.
In this approximation of $\m{A}$ using the two low-rank matrices $(\m{U}, \m{V})$,
$\m{V}$ represents the concept basis used to reinterpret our samples, and $\m{U}$ are the coordinates of the activation in this new basis. Interestingly, such a formulation allows a recast of the concept extraction problem as an instance of dictionary learning problem~\cite{mairal2014sparse} %
in which all known concept-based explainability methods fall:%

\begin{numcases}{(\m{U}^\star, \m{V}^\star) = \argmin_{\m{U},\m{V}} || \m{A} - \m{U} \m{V}^\tr ||^2_F ~~s.t~~}
 \forall ~ i, \v{u}_i \in \{ \e_1, \ldots, \e_k \} ~~ \text{\small\cite{ghorbani2019towards})}, \label{eq:holistic:dico_kmeans}\nonumber\\
  \m{V}^\tr \m{V} = \mathbf{I} ~~~ \text{\small(\cite{graziani2023concept,zhang2021invertible})}, \label{eq:holistic:dico_pca}\nonumber\\
 \m{U} \geq 0, \m{V} \geq 0 ~~~ \text{\CRAFT} \nonumber\\
 \m{U} = \bm{\psi}(\m{A}), ||\m{U}||_0 \leq \kappa  ~~\text{\small \cite{makhzani2013k}} \label{eq:holistic:dico_nmf}\nonumber
\end{numcases}

with $\bm{e}_i$ the $i$-th element of the canonical basis, $\mathbf{I}$ the identity matrix and $\bm{\psi}$ any neural network. 
In this context, $\m{V}$ is the \emph{dictionary} and $\m{U}$ the \emph{representation} of $\m{A}$ with the atoms of $\m{V}$. $\v{u}_i$ denote the $i$-th row of $\m{U}$. 
These methods extract the concept banks $\m{V}$ differently, thereby necessitating different interpretations\footnote{Concept extractions are typically overcomplete dictionaries, meaning that if the dictionary for each class is combined, $k >> p$, as noted in our previous section. The collapse problem in \autoref{sec:concepts:craft}, and a more detailed work \cite{bricken2023monosemanticity} suggest that overcomplete dictionaries are serious candidates to the superposition problem~\cite{elhage2022superposition}.}. 

In \ACE, the CAVs are defined as the centroids of the clusters found by the K-means algorithm.
Specifically, a concept vector $\cav_i$ in the matrix $\m{V}$ indicates a dense concentration of points associated with the corresponding concept, implying a repeated activation pattern. 
The main benefit of ACE comes from its reconstruction process, involving projecting activations onto the nearest centroid, which ensures that the representation will lie within the observed distribution (no out-of-distribution instances). %
However, its limitation lies in its lack of expressivity, as each activation representation is restricted to a single concept ($||\v{u}||_{0}=1$). As a result, it cannot capture compositions of concepts, leading to sub-optimal representations that fail to fully grasp the richness of the underlying data distribution.

On the other hand, the PCA benefits from superior reconstruction performance due to its lower constraints, as stated by the Eckart-Young-Mirsky~\cite{eckart1936approximation} theorem. %
The CAVs are the eigenvector of the covariance matrix: they indicate the direction in which the data variance is maximal. %
An inherent limitation is that the PCA will not be able to properly capture stable concepts that do not contribute to the sample variability (e.g. the dog-head concept might not be considered important by the PCA to explain the dog class if it is present across all examples).
Neural networks are known to cluster together the points belonging to the same category in the last layer to achieve linear separability (\cite{paypan2020collapse, fel2023craft}). Thus, the orthogonality constraint in the PCA might not be suitable to correctly interpret the manifold of the deep layer induced by points from the same class (it is interesting to note that this limitation can be of interest when studying all classes at once).
Also, unlike K-means, which produces strictly positive clusters if all points are positive (e.g., the output of ReLU), PCA has no sign constraint and can undesirably reconstruct out-of-distribution (OOD) activations, including negative values after ReLU. %

In contrast to K-Means, which induces extremely sparse representations, and PCA, which generates dense representations, the NMF (used in \CRAFT~and \ICE) strikes a harmonious balance as it provides moderately sparse representation. This is due to NMF relaxing the constraints imposed by the K-means algorithm (adding an orthogonality constraint on $\m{V}$ such that $\m{V} \m{V}^\tr = \mathbf{I}$ would yield an equivalent solution to K-means clustering~\cite{ding2005equivalence}). This sparsity facilitates the encoding of compositional representations that are particularly valuable when an image encompasses multiple concepts. Moreover, by allowing only additive linear combinations of components with non-negative coefficients, %
NMF inherently fosters a parts-based representation. This distinguishes NMF from PCA, which offers a holistic representation model. Interestingly, the NMF is known to yield representations that are interpretable by humans~\cite{zhang2021invertible, fel2023craft}.
Finally, the non-orthogonality of these concepts presents an advantage as it accommodates the phenomenon of superposition~\cite{elhage2022superposition}, wherein neurons within a layer may contribute to multiple distinct concepts simultaneously.

\begin{figure}[t]
\begin{center}
   \includegraphics[width=.99\textwidth]{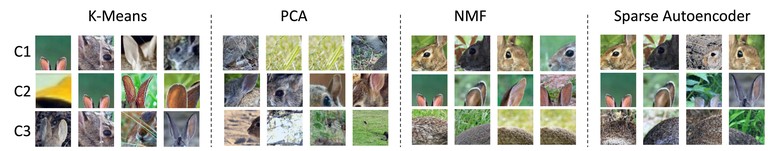}
\end{center}
   \caption{\textbf{Most important concepts extracted for the studied methods.} This qualitative example shows the three most important concepts extracted for the 'rabbit' class using a ResNet50 trained on ImageNet. The crops correspond to those maximizing each concepts $i$ (i.e., $\vx$ where $\m{U}(\vx)_i$ is maximal). As demonstrated in previous works \cite{zhang2021invertible,fel2023craft,parekh2022listen}, NMF (requiring positive activations) produces particularly interpretable concepts despite poorer reconstruction than PCA and being less sparse than K-Means. Details for the sparse Autoencoder architecture are provided in the appendix.}
\label{fig:holistic:qualitative_comparison}
\end{figure}

To summarize, we have explored three approaches to concept extraction, each necessitating a unique interpretation of the resulting Concept Activation Vectors (CAVs). Among these methods, NMF (used in \CRAFT~ and \ICE) emerges as a promising middle ground between PCA and K-means. Leveraging its capacity to capture intricate patterns, along with its ability to facilitate compositional representations and intuitive parts-based interpretations (as demonstrated in Figure~\ref{fig:holistic:qualitative_comparison}), NMF stands out as a compelling choice for extracting meaningful concepts from high-dimensional data. These advantages have been underscored by our human studies, and also evidenced by works such as~\cite{zhang2021invertible}.

\begin{table*}[h!]
\centering
\scalebox{0.76}{\begin{tabular}{l c c c c c}
\toprule
& \multicolumn{1}{c}{Relative $\ell_2$ ($\downarrow$)} 
& \multicolumn{1}{c}{Sparsity ($\uparrow$)} 
& \multicolumn{1}{c}{Stability ($\downarrow$)} 
& \multicolumn{1}{c}{FID ($\downarrow$)} 
& \multicolumn{1}{c}{OOD ($\downarrow$)} \\
 
\cmidrule(lr){2-2}
\cmidrule(lr){3-3}
\cmidrule(lr){4-4}
\cmidrule(lr){5-5}
\cmidrule(lr){6-6}

& 
Eff / R50 / Mob &
Eff / R50 / Mob &
Eff / R50 / Mob &
Eff / R50 / Mob &
Eff / R50 / Mob 
\\

\midrule
PCA 
   & 0.60 / 0.54 / 0.73 
   & 0.00 / 0.00 / 0.0
   & 0.41 / 0.38 / 0.43
   & 0.47 / 0.17 / 0.24
   & 2.44 / 0.36 / 0.16
\\
KMeans 
   & 0.72 / 0.66 / 0.84 
   & 0.95 / 0.95 / 0.95
   & 0.07 / 0.08 / 0.04
   & 0.46 / 0.21 / 0.33
   & 1.76 / 0.29 / 0.15
\\
NMF 
   & 0.63 / 0.57 / 0.75 
   & 0.68 / 0.44 / 0.64
   & 0.17 / 0.14 / 0.16
   & 0.38 / 0.21 / 0.24
   & 1.98 / 0.29 / 0.15
\\
\bottomrule
\end{tabular}}
\caption{\textbf{Concept extraction comparison.} Eff, R50 and Mob denote EfficientNetV2~\cite{zhang2018efficient}, ResNet50~\cite{he2016deep}, MobileNetV2~\cite{sandler2018mobilenetv2}. The concept extraction methods are applied on the last layer of the networks. Each results is averaged across 10 classes of ImageNet and obtained from a set of 16k images for each class.
}\label{tab:quantitative_comparison}
\vspace{-6mm}
\end{table*}

\paragraph{Evaluation of concept extraction}
Following the theoretical discussion of the various concept extraction methods, we conduct an empirical investigation of the previously discussed properties to gain deeper insights into their distinctions and advantages. In our experiment, we apply the PCA, K-Means, and NMF concept extraction methods on the penultimate layer of three state-of-the-art models. We subsequently evaluate the concepts using five different metrics (see Table \ref{tab:quantitative_comparison}). 
All five metrics are connected with the desired characteristics of a dictionary learning method. They include achieving a high-quality reconstruction (Relative l2), sparse encoding of concepts (Sparsity), ensuring the stability of the concept base in relation to $\m{A}$ (Stability), performing reconstructions within the intended domain (avoiding OOD), and maintaining the overall distribution during the reconstruction process (FID).
All the results come from 10 classes of ImageNet (the one used in Imagenette \cite{imagenette}), and are obtained using $n=16k$ images for each class.

We begin our empirical investigation by using a set of standard metrics derived from the dictionary learning literature, namely Relative $l_2$ and Sparsity. 
Concerning the Relative $\ell_2$, PCA achieves the highest score among the three considered methods, confirming the theoretical expectations based on the Eckart–Young–Mirsky theorem~\cite{eckart1936approximation}, followed by NMF.
Concerning the sparsity of the underlying representation $\v{u}$, we compute the proportion of non-zero elements $||\v{u}||_0 / k$. Since K-means inherently has a sparsity of $1 / k$ (as induced by equation \ref{eq:holistic:dico_kmeans}), it naturally performs better in terms of sparsity, followed by NMF.

We deepen our investigation by proposing three additional metrics that offer complementary insights into the extracted concepts. Those metrics are the Stability, the FID, and the OOD score.
The Stability (as it can be seen as a loose approximation of algorithmic stability~\cite{bousquet2002stability}) measures how consistent concepts remain when they are extracted from different subsets of the data.
To evaluate Stability, we perform the concept extraction methods $N$ times on $K$-fold subsets of the data. Then, we map the extracted concepts together using a Hungarian loss function and measure the cosine similarity of the CAVs. If a method is stable, it should yield the same concepts (up to permutation) across each $K$-fold, where each fold consists of $1000$ images.
K-Means and NMF demonstrate the highest stability, while PCA appears to be highly unstable, which can be problematic for interpreting the results and may undermine confidence in the extracted concepts.

The last two metrics, FID and OOD, are complementary in that they measure: (i) how faithful the representations extracted are w.r.t the original distribution, and (ii) the ability of the method to generate points lying in the data distribution (non-OOD).
Formally, the FID quantifies the 1-Wasserstein distance~\cite{villani2009optimal} $\mathcal{W}_1$ between the empirical distribution of activation $\m{A}$, denoted $\mu_{\bm{a}}$, and the empirical distribution of the reconstructed activation $\m{U}\m{V}^\tr$ denoted $\mu_{\bm{u}}$. Thus, FID is calculated as $\text{FID} = \mathcal{W}_1(\mu_{\bm{a}}, \mu_{\bm{u}})$.
On the other hand, the OOD score measures the plausibility of the reconstruction by leveraging Deep-KNN~\cite{sun2022out}, a recent state-of-the-art OOD metric. More specifically,  we use the Deep-KNN score to evaluate the deviation of a reconstructed point from the closest original point. In summary, a good reconstruction method is capable of accurately representing the original distribution (as indicated by FID) while ensuring that the generated points remain within the model's domain (non-OOD). 
K-means leads to the best OOD scores because each instance is reconstructed as a centroid, resulting in proximity to in-distribution (ID) instances. However, this approach collapses the distribution to a limited set of points, resulting in low FID. On the other hand, PCA may suffer from mapping to negative values, which can adversely affect the OOD score. Nevertheless, PCA is specifically optimized to achieve the best average reconstructions. NMF, with fewer stringent constraints, strikes a balance by providing in-distribution reconstructions at both the sample and population levels.

In conclusion, the results clearly demonstrate NMF as a method that strikes a balance between the two approaches as NMF demonstrates promising performance across all tested metrics. Henceforth, we will use the NMF to extract concepts without mentioning it.

\begin{figure}[ht]
\begin{center}
   \includegraphics[width=.99\textwidth]{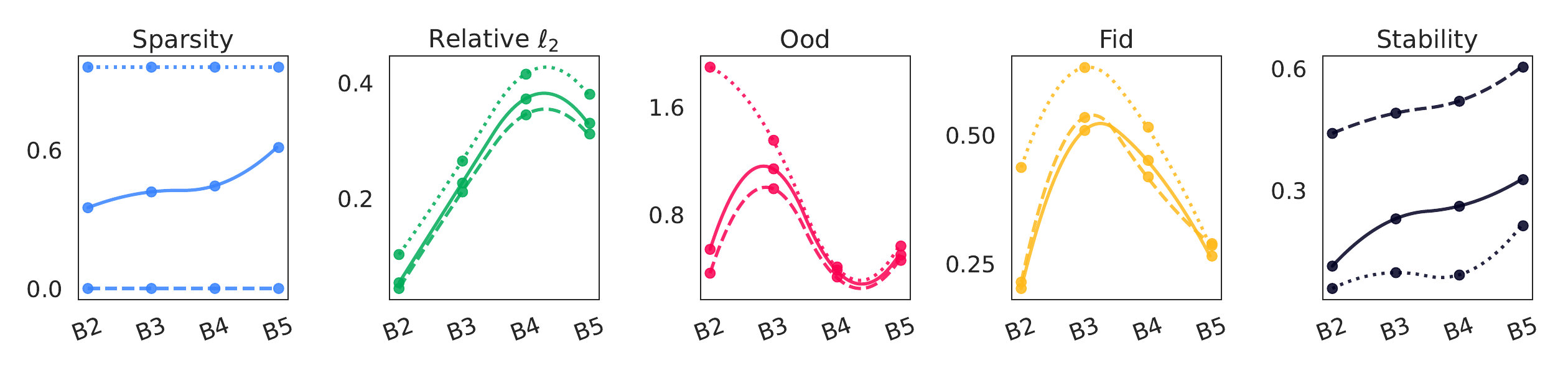}
\end{center}
   \caption{\textbf{Concept extraction metrics across layers.} The concept extraction methods are applied on activations probed on different blocks of a ResNet50 (B2 to B5). Each point is averaged over 10 classes of ImageNet using $16$k images for each class. We evaluate $3$ concept extraction methods: PCA (\dashed), NMF (\full), and KMeans (\dotted).
   }
\label{fig:holistic:metrics_across_layer}

\end{figure}

\paragraph{The Last Layer as a Promising Direction}
The various methods examined, namely \ACE, \ICE, and \CRAFT, generally rely on a deep layer to perform their decomposition without providing quantitative or theoretical justifications for their choice. 
To explore the validity of this choice, we apply the aforementioned metrics to each block's output in a ResNet50 model.
Figure~\ref{fig:holistic:metrics_across_layer} illustrates the metric evolution across different blocks, revealing a trend that favors the last layer for the decomposition. This empirical finding aligns with the practical implementations discussed above.

\subsubsection{Concept importance}\label{sec:importance}

In this section, we leverage our framework to unify concept importance scoring using the existing attribution methods. Furthermore, we demonstrate that specifically in the case of decomposition in the penultimate layer, it exists optimal methods for importance estimation, namely RISE~\cite{petsiuk2018rise}, Integrated Gradients~\cite{sundararajan2017axiomatic}, Gradient-Input~\cite{shrikumar2017learning}, and Occlusion~\cite{zeiler2014visualizing}. We provide theoretical evidence to support the optimality of these methods.

\paragraph{From concept importance to attribution methods}
The dictionary learning formulation allows us to define the concepts $\m{V}$ in such a way that they are optimal to reconstruct the activation, i.e., $\m{A} \approx \m{U} \m{V}^\tr$. Nevertheless, this does not guarantee that those concepts are important for the model's prediction. For example, the ``grass'' concept might be important to characterize the activations of a neural network when presented with a 
St-Bernard image, but it might not be crucial for the network to classify the same image as a St Bernard~\cite{kim2018interpretability, adebayo2018sanity, ghorbani2017interpretation}. The notion of concept importance is precisely introduced to avoid such a confirmation bias and to identify the concepts used to classify among all detected concepts.

We use the notion of Concept ATtribution methods (which we denote as \emph{CAT}s) to assess the concept importance score. The CATs are a generalization of the attribution methods: 
while attribution methods assess the sensitivity of the model output to a change in the pixel space, the concept importance evaluates the sensitivity to a change in the concept space. To compute the CATs methods, it is necessary to link the activation $\v{a} \in \Real^p$ to the concept base $\m{V}$ and the model prediction $\v{y}$. To do so, we feed the second part of the network ($\fb$) with the activation reconstruction ($\v{u} \m{V}^\tr \approx \v{a}$) so that $\v{y} = \fb(\v{u}\m{V}^\tr)$. Intuitively, a CAT method quantifies how a variation of  $\v{u}$ will impact $\v{y}$. 
We denote $\cam_i(\bm{u})$ the $i$-th coordinate of $\cam(\bm{u})$, so that it represents the importance of the $i$-th concept in the representation $\bm{u}$.  Equipped with these notations, we can leverage the sensitivity metrics introduced in standard attribution methods to re-define the current measures of concept importance, as well as introduce the new CATs borrowed from the attribution methods literature:

\scalebox{0.9}{\parbox{\linewidth}{%
\begin{empheq}[left={\cam_{i}(\v{u}) =\empheqlbrace}]{alignat=1}
&\nabla_{\v{u}_i} \fb(\v{u} \m{V}^\tr)
~~
\text{(\small TCAV: \cite{ghorbani2019towards,zhang2021invertible,graziani2021sharpening})}, \nonumber\\
&\displaystyle \frac{ \mathbb{E}_{\mathbf{m}_{\sim i}}( \mathbb{V}_{\mathbf{m}}( \fb( (\v{u} \odot \mathbf{m} ) \m{V}^\tr ) | \mathbf{m}_{\sim i} ) ) }{ \mathbb{V}( \fb( (\v{u} \odot \mathbf{m} ) \m{V}^\tr)) }
\qquad \qquad \qquad \qquad ~~~~~~~~~ \text{\small(Sobol: \CRAFT),}  \nonumber\\
&(\v{u}_i - \v{u}_i')  \times \int_0^1\nabla_{\v{u}_i}\fb((\v{u}' \alpha + (1 - \alpha)(\v{u} - \v{u}'))\m{V}^\tr) d\alpha
\qquad \text{\small(Int.Gradients)}, \nonumber\\
&\displaystyle \underset{\bm{\delta} \sim \mathcal{N}(0, \mathbf{I}\sigma)}{\mathbb{E}}(\nabla_{\v{u}_i} \fb( (\v{u} + \bm{\delta})\m{V}^T) )
\qquad \qquad \qquad \qquad \qquad \qquad ~~ \text{\small(Smoothgrad)}. \nonumber \\
\ldots \nonumber
\end{empheq} 
}}
\vspace{3mm}

The complete derivation of the 7 new CATs is provided in the appendix. 
In the derivations, $\nabla_{\v{u}_i}$ denotes the gradient with respect to the $i$-th coordinate of $\v{u}$, while $\mathbb{E}$ and $\mathbb{V}$ represent the expectation and variance, respectively, $\mathbf{m}$ is a mask of real-valued random variable between $0$ and $1$ (i.e $\mathbf{m}\sim\mathcal{U}([0,1]^p)$). We note that, when we use the gradient (w.r.t to $\v{u}_i$) as an importance score, we end up with the directional derivative used in the TCAV metric~\cite{kim2018interpretability}. In other words, one could say that TCAV is the Saliency of the \textit{Concept Attribution} methods. 

\CRAFT~leverages the Sobol-Hoeffding decomposition (used in sensitivity analysis), to estimate the concept importance. The Sobol indices measure the contribution of a concept as well as its interaction of any order with any other concepts to the output variance. Intuitively, the numerator for the Sobol importance formula is the expected variance that would be left if all variables but $\v{u}_i$ were to be fixed.

\begin{figure}[ht]
\begin{subfigure}[b]{0.49\textwidth}
\includegraphics[width=.99\textwidth]{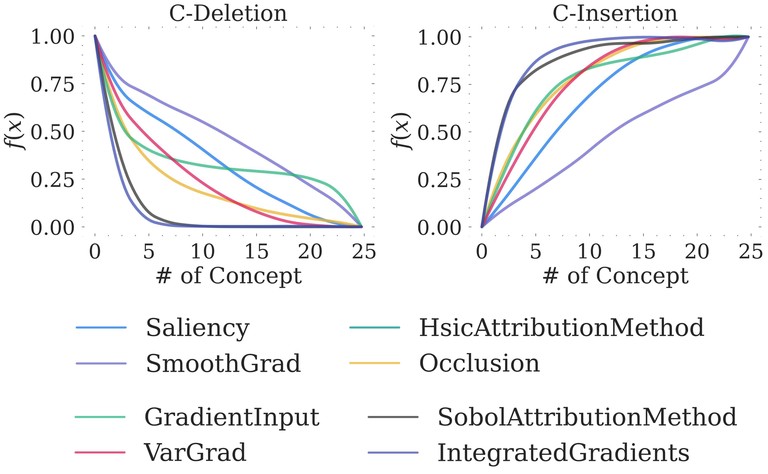}
    \caption{}
\end{subfigure}
\hfil
\begin{subfigure}[b]{0.49\textwidth}
\includegraphics[width=.99\textwidth]{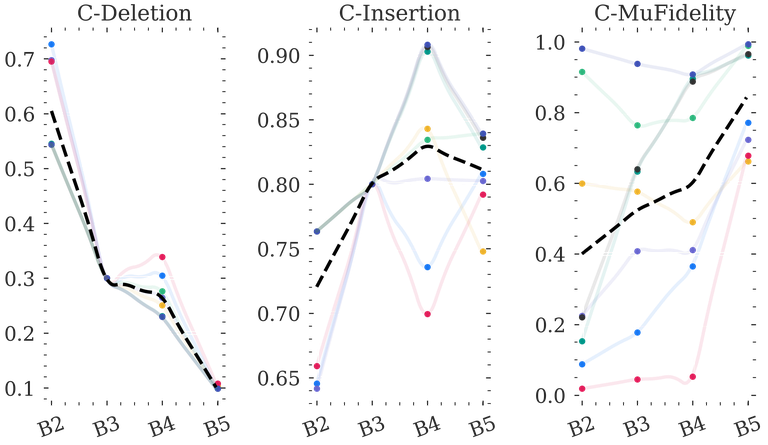}
    \caption{}
\end{subfigure}
\caption{\textbf{(a) C-Deletion, C-Insertion curves.} Fidelity curves for C-Deletion depict the model's score as the most important concepts are removed. The results are averaged across 10 classes of ImageNet using a ResNet50 model.
\textbf{(b) C-Deletion, C-Insertion and C-$\mu$Fidelity across layer.} 
We report the $3$ metrics to evaluate CATs for each block (from B2 to B5) of a ResNet50. 
We evaluate $8$ Concept Attribution methods, all represented with different colors (see legend in  Figure~\ref{fig:holistic:deletion_curves}(a). The average trend of these eight methods is represented by the black dashed line (\dashed). Lower C-Deletion is better, higher C-Insertion and C-$\mu$Fidelity is better. Overall, it appears that the estimation of importance becomes more faithful towards the end of the model.
}
\label{fig:holistic:deletion_curves}
\end{figure}

\paragraph{Evaluation of concept importance methods}

Our generalization of the concept importance score, using the Concept ATtributions (CATs), allows us to observe that current concept-based explainability methods are only leveraging a small subset of concept importance methods. In Appendix~\ref{sup:holistic:all_cams}, we provide the complete derivation of $7$ new CATs based on the following existing attribution methods, notably: Gradient input~\cite{shrikumar2017learning}, Smooth grad~\cite{smilkov2017smoothgrad}, Integrated Gradients~\cite{sundararajan2017axiomatic}, VarGrad~\cite{hooker2018benchmark}, Occlusion~\cite{zeiler2014visualizing}, HSIC~\cite{novello2022making} and RISE~\cite{petsiuk2018rise}.

With the concept importance scoring now formulated as a generalization of attribution methods, we can borrow the metrics from the attribution domain to evaluate the faithfulness~\cite{jacovi2020towards,petsiuk2018rise,aggregating2020} of concept importance methods. In particular, 
we adapt three distinct metrics %
to evaluate the significance of concept importance scores: the C-Deletion~\cite{petsiuk2018rise}, C-Insertion~\cite{petsiuk2018rise}, and C-$\mu$Fidelity~\cite{aggregating2020} metrics.
In C-Deletion, we gradually remove the concepts (as shown in Figure \ref{fig:holistic:deletion_curves}), in decreasing order of importance, and we report the network's output each time a concept is removed. When a concept is removed in C-Deletion, the corresponding coordinate in the representation is set to $\bm{0}$. 
The final C-Deletion metrics are computed as the area under the curve in Figure~\ref{fig:holistic:deletion_curves}. For C-Insertion, this is the opposite: we start from a representation vector filled with zero, and we progressively add more concepts, following an increasing order of importance.

For the C-$\mu$Fidelity, we calculate the correlation between the model's output when concepts are randomly removed and the importance assigned to those specific concepts.
The results across layers for a ResNet50 model are depicted in Figure \ref{fig:holistic:deletion_curves}b. We observe that decomposition towards the end of the model is preferred across all the metrics. As a result, in the next section, we will specifically examine the case of the penultimate layer.

\paragraph{A note on the last layer}
Based on our empirical results, it appears that the last layer is preferable for both improved concept extraction and more accurate estimation of importance. 
Herein, we derive theoretical guarantees about the optimality of concept importance methods in the penultimate layer. %
Without loss of generality, we assume $y \in \Real$ the logits of the class of interest. In the penultimate layer, the score $y$ is a linear combination of activations: $y=\bm{a}\m{W}+\bias$ for weight matrix $\m{W}$ and bias $\bias$. 
In this particular case, all CATs have a closed-form (see appendix~\ref{sup:holistic:closed_form}), that allows us to derive $2$ theorems. The first theorem tackles the CATs optimality for the C-Deletion and C-Insertion methods (demonstration in Appendix~\ref{sup:holistic:matroid}). We observe that the C-Deletion and C-Insertion problems can be represented as weighted matroids. Therefore the greedy algorithms lead to optimal solutions for CATs and a similar theorem could be derived for C-$\mu$Fidelity.
\begin{theorem}[Optimal C-Deletion, C-Insertion in the penultimate layer]
When decomposing in the penultimate layer,~\textbf{Gradient Input}, \textbf{Integrated Gradients}, \textbf{Occlusion}, and \textbf{Rise} yield the optimal solution for the C-Deletion and C-Insertion metrics.
More generally, any method $\cam(\v{u})$ that satisfies the condition 
$\forall (i, j) \in \{1, \ldots, k\}^2, 
(\v{u} \odot \e_i) \m{V}^\tr\m{W} \geq (\v{u} \odot \e_j) \m{V}^\tr \m{W}
\implies 
\cam(\v{u})_i \geq \cam(\v{u})_j 
$ yields the optimal solution.
\end{theorem}
\begin{theorem}[Optimal C-$\mu$Fidelity in the penultimate layer]
When decomposing in the penultimate layer,~\textbf{Gradient Input}, \textbf{Integrated Gradients}, \textbf{Occlusion}, and \textbf{Rise} yield the optimal solution for the C-$\mu$Fidelity metric.
\end{theorem}

\begin{figure}[ht]
    \centering
    \includegraphics[width=0.9\textwidth]{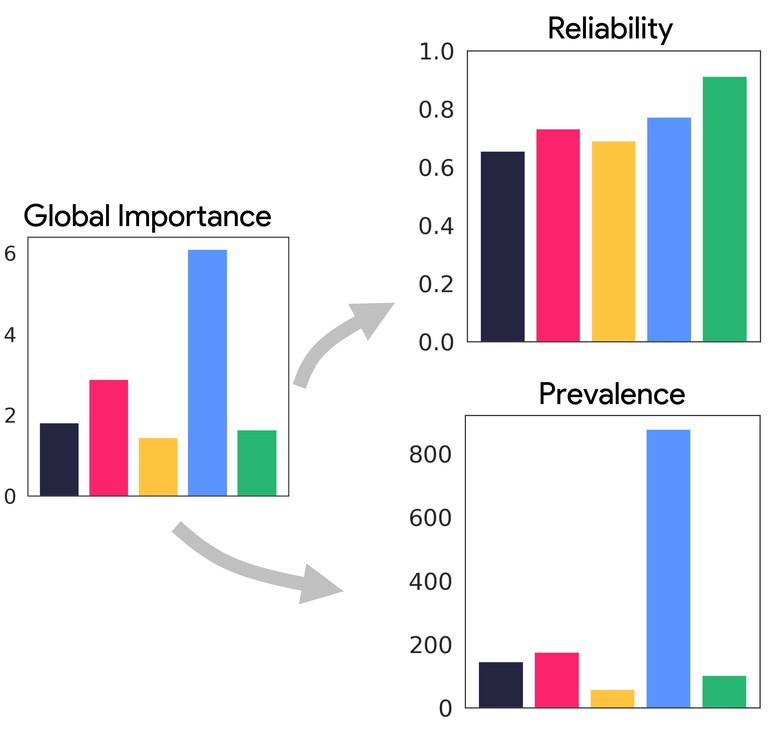}
    \caption{\textbf{From global (class-based) to local (image-based) importance.} Global importance can be decomposed into \textit{reliability} and \textit{prevalence} scores. Prevalence quantifies how frequently a concept is encountered, and reliability indicates how diagnostic a concept is for the class. The bar-charts are computed for the class ``Espresso'' on a ResNet50 (see Figure~\ref{fig:holistic:clustering_graph}, left panel)
    }
    \label{fig:holistic:barchart}
\end{figure}

Therefore, for all $3$ metrics, the concept importance methods based on Gradient Input, Integrated Gradient, Occlusion, and Rise are optimal, when used in the penultimate layer.

In summary, our investigation of concept extraction methods from the perspective of dictionary learning demonstrates that the NMF approach, specifically when extracting concepts from the penultimate layer, presents the most appealing trade-off compared to PCA and K-Means methods. In addition, our formalization of concept importance using attribution methods provided us with a theoretical guarantee for $4$ different CATs. Henceforth, we will then consider the following setup: a NMF on the penultimate layer to extract the concepts, combined with a concept importance method based on Integrated Gradient.

\subsubsection{Unveiling main strategies}

So far, the concept-based explainability methods have mainly focused on evaluating the global importance of concepts, i.e., the importance of concepts for an entire class~\cite{kim2018interpretability,fel2023craft}. This point can be limiting when studying misclassified data points, as we can speculate that the most important concepts for a given class might not hold for an individual sample (local importance). Fortunately, our formulation of concept importance using attribution methods gives us access to importance scores at the level of individual samples (\textit{i.e.,} $\cam(\v{u})$). Here, we show how to use these local importance scores to efficiently cluster data points based on the strategy used for their classification. 

The local (or image-based) importance of concepts can be integrated into global measures of importance for the entire class with the notion of \textit{prevalence} and \textit{reliability} (see Figure~\ref{fig:holistic:barchart}). A concept is said to be prevalent at the class level when it appears very frequently. A \textit{prevalence} score is computed based on the number of times a concept is identified as the most important one, i.e., $\argmax \cam(\v{u})$. At the same time, a concept is said to be reliable if it is very likely to trigger a correct prediction. The \textit{reliability} is quantified using the mean classification accuracy on samples sharing the same most important concept.

\begin{figure}[ht]
\begin{center}
   \includegraphics[width=1\textwidth]{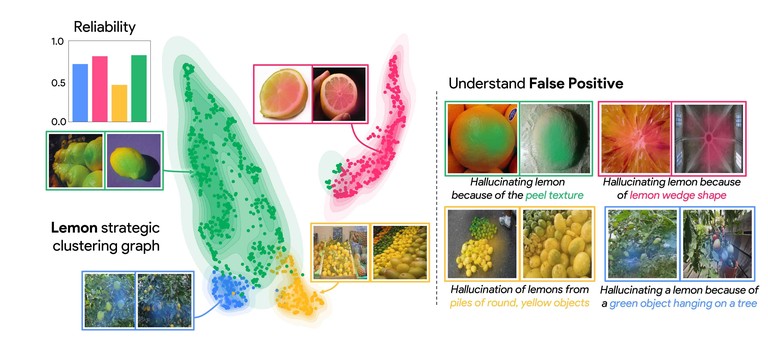}
\end{center}
   \caption{\textbf{Strategic cluster graph for the lemon category.} \textbf{Left}: U-MAP of lemon samples, in the concept space. Each concept is represented with its own color and is exemplified with example belonging to the cluster. The concepts are  \textcolor{red}{$\bullet$} the lemon wedge shape, \textcolor{yellow}{$\bullet$} a pile of round, yellow objects, \textcolor{blue}{$\bullet$} green objects hanging on a tree, and finally \textcolor{green}{$\bullet$} the peel texture, which is the predominant strategy. The reliability of each concept is shown in the top-left bar-chart. \textbf{Right}: 
   Example of images predicted as lemon along with their corresponding explanations. These misclassified images are recognized as lemons through the implementation of strategies that are captured by our proposed strategic cluster graph.
   }
\label{fig:holistic:lemon}
\end{figure}

\paragraph{Strategic cluster graph.} In the strategic cluster graph (Figure~\ref{fig:holistic:clustering_graph} and Figure~\ref{fig:holistic:lemon}), we combine the notions of concept \textit{prevalence} and \textit{reliability} to reveal the main strategies of a model for a given category, more precisely, we reveal their repartition across the different samples of the class.
We use a dimensionality reduction technique (UMAP~\cite{mcinnes2018umap}) to arrange the data points based on the concept importance vector $\cam(\v{u})$ of each sample. Data points are colored according to the associated concept with the highest importance -- $\argmax \cam(\v{u})$. 
Interestingly, one can see in Figure~\ref{fig:holistic:clustering_graph} and Figure~\ref{fig:holistic:lemon} that spatially close points represent samples classified using \textit{similar strategies} -- as they exhibit similar concept importance -- and not necessarily similar embeddings.
For example, for the ``lemon'' object category (Figure \ref{fig:holistic:lemon}), the texture of the lemon peel is the most \textit{prevalent} concept, as it appears to be the dominant concept in $90\%$ of the samples (see the green cluster in Figure~\ref{fig:holistic:lemon}). We also observe that the concept ``pile of round, yellow objects'' is not reliable for the network to properly classify a lemon as it results in a mean classification accuracy of $40\%$ only (see top-left graph in Figure~\ref{fig:holistic:lemon}).

In Figure~\ref{fig:holistic:lemon} (right panel), we have exploited the strategic cluster graph to understand the classification strategies leading to bad classifications. For example, an orange ($1^{st}$ image, $1^{st}$ row) was classified as a lemon because of the peel texture they both share. Similarly, a cathedral roof was classified as a lemon because of the wedge-shaped structure of the structure ($4^{th}$ image, $1^{st}$ row).

\subsection{Discussion}

In this section, we have introduced a theoretical framework that unifies all modern concept-based explainability methods. Breaking down and formalizing the two essential steps in these methods, concept extraction and concept importance scoring, allowed us to better understand the underlying principles driving concept-based explainability. We leveraged this unified framework to propose new evaluation metrics for assessing the quality of extracted concepts. Through experimental and theoretical analyses, we justified the standard use of the last layer of an ANN for concept-based explanation. Finally, we harnessed the parallel between concept importance and attribution methods to gain insights into global concept importance (at the class level) by examining local concept importance (for individual samples). We proposed the strategic cluster graph, which provides insights into the strategy used by an ANN to classify images. We have provided an example use of this approach to better understand the failure cases of a system. Overall, our work demonstrates the potential benefits of the \textbf{dictionary learning} framework for automatic concept extraction and we hope this work will pave the way for further advancements and methodologies in the field au concept-based explainability.

In this research, we deliberately overlooked a particular challenge associated with the automatic concept approach, namely, the comprehensibility of the features extracted by dictionary-based methods. Indeed, relying solely on image segments to elucidate a concept could be restrictive. In the next section, we will examine an alternative approach to visualize concepts with feature visualization.

\clearpage
  
\section{Modern Feature Visualization with MACO}
\label{sec:concepts:maco}

The last section of this chapter will be dedicated to a novel method that will enable us one problem that we identify in \autoref{sec:concepts:craft}: the visualization of concept. Feature visualization -- defined in \autoref{def:intro:feature_viz} -- has gained substantial popularity, particularly after the seminal and influential work of the Clarity team~\cite{olah2017feature}, which established it as a crucial tool for explainability.
However, its widespread adoption has been limited due to a reliance on tricks to generate interpretable images, and corresponding challenges in scaling it to deeper neural networks.
Here, we will introduce \magfv, a simple approach to address these shortcomings.
The main idea is to generate images by optimizing the phase spectrum while keeping the magnitude constant to ensure that generated explanations lie in the space of natural images. Our approach yields significantly better results -- both qualitatively and quantitatively -- and unlocks efficient and interpretable feature visualizations for large state-of-the-art neural networks.
We also show that our approach exhibits an attribution mechanism allowing us to augment feature visualizations with spatial importance.

Overall, our approach unlocks, for the first time, feature visualizations for large, state-of-the-art deep neural networks without resorting to any parametric prior image model.

\begin{figure}[ht]
\begin{center}
   \includegraphics[width=.99\textwidth]{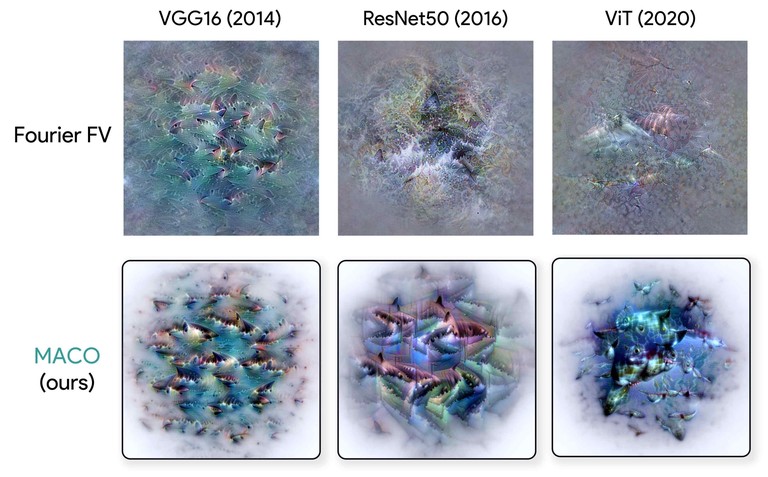}
\end{center}

\caption{\textbf{Comparison between feature visualization methods for ``White Shark'' classification.}
\textbf{(Top)} Standard Fourier preconditioning-based method for feature visualization~\cite{olah2017feature}.
\textbf{(Bottom)} Proposed approach, \magfv, which incorporates a Fourier spectrum magnitude constraint. %
}
\label{fig:maco:logits_fail}

\end{figure}

\subsection{Introduction}

As discussed in \autoref{chap:attributions}, the initial tools in the explainability toolkit were primarily attribution methods~\cite{simonyan2014deep,smilkov2017smoothgrad,selvaraju2017gradcam,fel2021sobol,novello2022making,sundararajan2017axiomatic,zeiler2014visualizing,shrikumar2017learning,Fong_2017,graziani2021sharpening}. We also seen in ~\autoref{sec:attributions:metapred} that those approaches only offer a partial understanding of the learned decision processes as they aim to identify the location of the most discriminative features in an image, the ``\where'', leaving open the ``\what'' question, \textit{i.e.} the semantic meaning of those features.

Feature visualization methods, which aim to bridge this gap, involve formulating and solving an optimization problem to identify an input image that maximizes the activation of a specific target element (be it a neuron, layer, or the entire model)~\cite{zeiler2014visualizing}. Most of the approaches developed in the field fall along a spectrum based on how strongly they regularize the model. At one end of the spectrum, if no regularization is used, the optimization process can search the whole image space, but this tends to produce noisy images and nonsensical high-frequency patterns~\cite{erhan2009visualizing}. To circumvent this issue, researchers have proposed to penalize high-frequency in  the resulting images -- either by reducing the variance between neighboring pixels~\cite{mahendran2015understanding}, by imposing constraints on the image's total variation~\cite{nguyen2016synthesizing,nguyen2017plug,simonyan2014deep}, or by blurring the image at each optimization step~\cite{nguyen2015deep}. However, in addition to rendering images of debatable validity, these approaches also suppress genuine, interesting high-frequency features, including edges. To mitigate this issue, a bilateral filter may be used instead of blurring, as it has been shown to preserve edges and improve the overall result~\cite{tyka2016class}. Other studies have described a similar technique to decrease high frequencies by operating directly on the gradient, with the goal of preventing their accumulation in the resulting visualization~\cite{AudunGoogleNet}. One advantage of reducing high frequencies present in the gradient, as opposed to the visualization itself, is that it resists the amplification of high frequencies while still allowing them to manifest when consistently promoted by the gradient.
This process, known as "preconditioning" in optimization, can greatly simplify the optimization problem. The Fourier transform has been shown to be a successful preconditioner as it forces the optimization to be performed in a decorrelated and whitened image space~\cite{olah2017feature}. 

The emergence of high-frequency patterns in the absence of regularization is associated with a lack of robustness and sensitivity of the neural network to adversarial examples~\cite{szegedy2013intriguing}, and consequently, these patterns are less often observed in adversarially robust models~\cite{engstrom2019adversarial, santurkar2019image, tsipras2018robustness}. An alternative strategy to promote robustness involves enforcing small perturbations, such as jittering, rotating, or scaling, in the visualization process~\cite{mordvintsev2015inceptionism}, which, when combined with a frequency penalty~\cite{olah2017feature}, has been proved to greatly enhance the generated images.

Unfortunately, previous methods in the field of feature visualization have been limited in their ability to generate visualizations for newer architectures beyond VGG, resulting in a lack of interpretable visualizations for larger networks like ResNets~\cite{olah2017feature}. Consequently, researchers have shifted their focus to approaches that leverage statistically learned priors to produce highly realistic visualizations. One such approach involves training a generator, like a GAN~\cite{nguyen2016synthesizing} or an autoencoder~\cite{wang2022traditional, nguyen2017plug}, to map points from a latent space to realistic examples and optimizing within that space. Alternatively, a prior can be learned to provide the gradient (w.r.t the input) of the probability and optimize both the prior and the objective jointly~\cite{nguyen2017plug, tyka2016class}. Another method involves approximating a generative model prior by penalizing the distance between output patches and the nearest patches retrieved from a database of image patches collected from the training data~\cite{wei2015understanding}.
Although it is well-established that learning an image prior produces realistic visualizations, it is difficult to distinguish between the contributions of the generative models and that of the neural network under study. Hence, in this work, we focus on the development of visualization methods that rely on minimal priors to yield the least biased visualizations.

Our proposed approach, called MAgnitude Constrained Optimization (\magfv), builds on the seminal work by Olah et al. We propose a straightforward re-parametrization that essentially relies on exploiting the phase/magnitude decomposition of the Fourier spectrum, to exclusively optimizing the image's phase while keeping its magnitude constant.
Such a constraint is motivated by psychophysics experiments that have shown that humans are more sensitive to differences in phase than in magnitude~\cite{oppenheim1981importance,caelli1982visual,guyader2004image,joubert2009rapid, gladilin2015role}. Our contributions are threefold:

\begin{enumerate}[label=(\textit{\textbf{\roman*}})]

\item{We unlock feature visualizations for large modern CNNs without resorting to any strong parametric image prior (see Figure~\ref{fig:maco:logits_fail}).}

\item{We describe how to leverage the gradients obtained throughout our optimization process to combine feature visualization with attribution methods, thereby explaining both ``\what'' activates a neuron and ``\where'' it is located in an image.}

\item{We introduce new metrics to compare the feature visualizations produced with \magfv~to those generated with other methods.}
\end{enumerate}
As an application of our approach, we propose feature visualizations for FlexViT \cite{beyer2022flexivit} and ViT \cite{Dosovitskiy2021-zy} (logits and intermediate layers;  see Figure~\ref{fig:maco:logits_and_internal}).  We also employ our approach on a feature inversion task to generate images that yield the same activations as target images to better understand what information is getting propagated through the network and which parts of the image are getting discarded by the model (on ViT, see Figure~\ref{fig:maco:inversion}).
Finally, we will make a link with our work introduced in \autoref{sec:concepts:craft} and show how to combine our work with \craft (see Figure~\ref{fig:maco:concepts}). As feature visualization can be used to optimize in directions in the network's representation space, we employ \magfv~to generate concept visualizations, thus allowing us to improve the human interpretability of concepts and reducing the risk of confirmation bias. 

\subsection{Magnitude-Constrained Feature Visualization}

\paragraph{Notations}

Throughout, we consider a general supervised learning setting, with an input space $\sx \subseteq \Real^{h \times w}$, an output space $\sy \subseteq \Real^c$, and a classifier $\f : \sx \to \sy$ that maps inputs $\vx \in \sx$ to a prediction $\v{y} \in \sy$.
Without loss of generality, we assume that $\f$ admits a series of $L$ intermediate spaces $\s{A}_\ell \subseteq \Real^{p_\ell}, 1 < \ell < L$.
In this setup, $\f_\ell : \sx \to \s{A}_\ell$ maps an input to an intermediate activation $\v{v} = (v_1, \ldots, v_{p_\ell})^\intercal \in \s{A}_\ell$ of $\f$.
We respectively denote $\fourier$ and $\fourier^{-1}$ as the 2-D Discrete Fourier Transform (DFT) on $\sx$ and its inverse.

\paragraph{Optimization Criterion.}
The primary goal of a feature visualization method is to produce an image $\vx^\star$ that maximizes a given criterion $\mathcal{L}_{\v{v}}(\vx) \in \Real$; usually some value aggregated over a subset of weights in a neural network $\f$ (neurons, channels, layers, logits).
A concrete example consists in finding a natural "prototypical" image $\vx^\star$ of a class $k \in \llbracket 1, K \rrbracket$ without using a dataset or generative models.
However, optimizing in the pixel space $\Real^{W \times H}$ is known to produce noisy, adversarial-like $\vx^\star$. Therefore, the optimization is constrained using a regularizer $\Omega: \sx \to \Real^+$ to penalize unrealistic images:
\begin{equation}
\vx^\star = \argmax_{\vx \in \sx} \mathcal{L}_{\v{v}}(\vx) - \lambda \Omega(\vx).
\label{eq:maco:general}
\end{equation}
In Eq.~\ref{eq:maco:general}, $\lambda$ is a hyperparameter used to balance the main optimization criterion $\mathcal{L}_{\v{v}}$ and the regularizer $\Omega(\cdot)$. Finding a regularizer that perfectly matches the structure of natural images is hard, so  proxies have to be used instead. Previous studies have explored various forms of regularization spanning from total variation, $\ell_1$, or $\ell_2$ loss~\cite{nguyen2016synthesizing,nguyen2017plug,simonyan2014deep}. More successful attempts rely on the reparametrization of the optimization problem in the Fourier domain rather than on regularization.

\subsubsection{A Fourier perspective}

Mordvintsev et al.~\cite{mordvintsev2018differentiable} noted in their seminal work that one could use differentiable image parametrizations to facilitate the maximization of $\mathcal{L}_{\v{v}}$. Olah et al.~\cite{olah2017feature} proposed to re-parametrize the images using their Fourier spectrum. Such a parametrization allows amplifying the low frequencies using a scalar $\v{w}$. Formally, the prototypal image $\vx^\star$ can be written as $\vx^\star = \fourier^{-1}(\v{z}^\star \odot \v{w})$ with:

$$ \v{z}^\star = \argmax_{\v{z} \in \mathbb{C}^{W \times H}} \mathcal{L}_{\v{v}}(\fourier^{-1}(\v{z} \odot \v{w})).$$

Finding $\vx^\star$ boils down to optimizing a Fourier buffer
$\v{z} = \bm{a} + i \bm{b}$ together with boosting the low-frequency components and then recovering the final image by inverting the optimized Fourier buffer using inverse Fourier transform.

\begin{figure}
\centering
\includegraphics[width=0.9\textwidth]{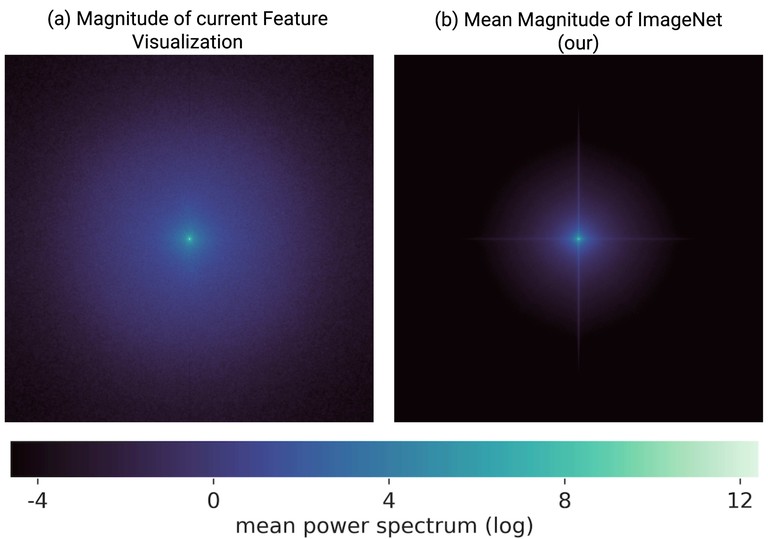};
\caption{\textbf{Comparison between Fourier FV and natural image power spectrum.} In \textbf{(left)}, the power spectrum is averaged over $10$ different logits visualizations for each of the $1000$ classes of ImageNet. The visualizations are obtained using the \textbf{Fourier FV}Fourier FV method to maximize the logits of a ViT network~\citep{olah2017feature}. In \textbf{(right)} the spectrum is averaged over all training images of the ImageNet dataset.}
\label{fig:maco:leakage}
\end{figure}

However, multiple studies have shown that the resulting images are not sufficiently robust, in the sense that a small change in the image can cause the criterion $ \mathcal{L}_{\v{v}}$ to drop. Therefore, it is common to see robustness transformations applied to candidate images throughout the optimization process. In other words, the goal is to ensure that the generated image satisfies the criterion even if it is rotated by a few degrees or jittered by a few pixels. Formally, given a set of possible transformation functions -- sometimes called augmentations -- that we denote $\mathcal{T}$ such that for any transformation $\augmentation \sim \mathcal{T}$, we have $\augmentation(\vx) \in \sx$, the optimization becomes:

$$ 
\v{z}^\star = \argmax_{\v{z} \in \mathbb{C}^{W \times H}}
\mathbb{E}_{\augmentation \sim \mathcal{T}}(\mathcal{L}_{\v{v}}((\augmentation \circ \fourier^{-1})(\v{z} \odot \v{w})).
$$

Empirically, it is common knowledge that the deeper the models are, the more transformations are needed and the greater their magnitudes should be. To make their approach work on models like VGG, Olah et al.~\cite{olah2017feature} used no less than a dozen transformations. However, this method fails for modern architectures, no matter how many transformations are applied. We argue that this may come from the low-frequency scalar (or booster) no longer working with models that are too deep. For such models, high frequencies eventually come through, polluting the resulting images with high-frequency content -- making them impossible to interpret by humans. %
To empirically illustrate this phenomenon, we compute the $k$ logit visualizations obtained by maximizing each of the logits corresponding to the $k$ classes of a ViT using the parameterization used by Olah et al.~ In Figure~\ref{fig:maco:leakage} (left), we show the average of the spectrum of these generated visualizations over all classes: $\frac{1}{k} \sum_{i=1}^k |\fourier(\vx^\star_i)|$. We compare it with the average spectrum of images on the ImageNet dataset (denoted $\mathcal{D}$): $\mathbb{E}_{\vx \sim \mathcal{D}}(|\fourier(\vx)|)$ (Figure~\ref{fig:maco:leakage}, right panel).
We observe that the images obtained through optimization put much more energy into high frequencies compared to natural images. Note that we did not observe this phenomenon in older models such as LeNet or VGG.

In the following section, we introduce our method named~\magfv, which is motivated by this observation. We constrain the magnitude of the visualization to a natural value, enabling natural visualization for any contemporary model, and reducing the number of required transformations to only two.

\subsubsection{\magfv: from Regularization to Constraint}
\begin{figure}[t!]
\center
\includegraphics[width=1\textwidth]{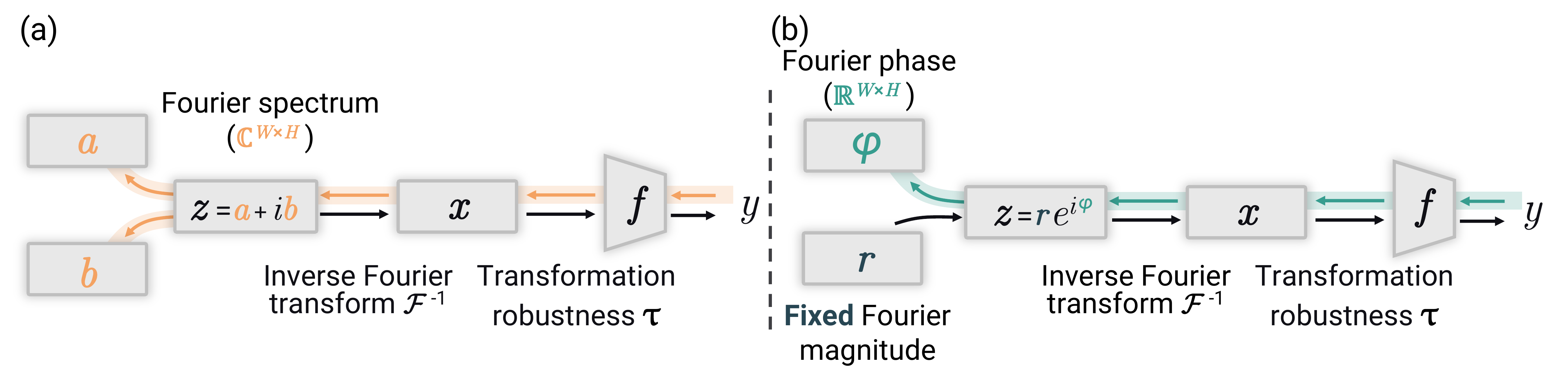}
\caption{\textbf{Overview of the approach:} \textbf{(a)}  Current Fourier parameterization approaches optimize the entire spectrum (yellow arrow). \textbf{(b)}  In contrast,  the optimization flow in our approach (green arrows) goes from the network activation ($\v{y}$) to the phase of the spectrum ($\v{\varphi}$) of the input image ($\vx$).}

\label{fig:maco:method}
\end{figure}

Parameterizing the image in the Fourier space makes it possible to directly manipulate the image in the frequency domain. We propose to take a step further and decompose the Fourier spectrum $\v{z}$ into its polar form $\v{z} = \v{r} e^{i \v{\varphi}}$ instead of its cartesian form $\v{z} = \bm{a} + i \bm{b}$, which allows us to disentangle the magnitude ($\v{r}$) and the phase ($\v{\varphi}$).

It is known that human recognition of objects in images is driven not by magnitude but by phase~\cite{oppenheim1981importance,caelli1982visual,guyader2004image,joubert2009rapid, gladilin2015role}. Motivated by this, we propose to optimize the phase of the Fourier spectrum while fixing its magnitude to a typical value of a natural image (with few high frequencies). In particular, the magnitude is kept constant at the average magnitude computed over a set of natural images (such as ImageNet), so $\v{r} = \mathbb{E}_{\vx \sim \mathcal{D}}(|\fourier(\vx)|)$. Note that this spectrum needs to be calculated only once and can be used at will for other tasks.

\begin{figure}[ht]
\input{assets/maco/algorithm}
\end{figure}

Therefore, our method does not backpropagate through the entire Fourier spectrum but only through the phase (Figure~\ref{fig:maco:method}), thus reducing the number of parameters to optimize by half. Since the magnitude of our spectrum is constrained, we no longer need hyperparameters such as $\lambda$ or scaling factors, and the generated image at each step is naturally plausible in the frequency domain.
We also enhance the quality of our visualizations via two data augmentations: random crop and additive uniform noise.
To the best of our knowledge, our approach is the first to completely alleviate the need for explicit regularization -- using instead a hard constraint on the solution of the optimization problem for feature visualization.
To summarize, we formally introduce our method:

\begin{definition}[\textbf{\magfv}]
The feature visualization results from optimizing the parameter vector $\v{\varphi}$  such that:
$$
\v{\varphi}^\star = \argmax_{\v{\varphi} \in \Real^{W \times H}}
\mathbb{E}_{\augmentation \sim \mathcal{T}}(\mathcal{L}_{\v{v}}((\augmentation \circ \fourier^{-1})(\v{r} e^{i \v{\varphi}})) ~~~\text{where}~~~ \v{r} = \mathbb{E}_{\vx \sim \mathcal{D}}(|\fourier(\vx)|)
$$
The feature visualization is then obtained by applying the inverse Fourier transform to the optimal complex-valued spectrum: $\vx^\star = \fourier^{-1}((\v{r} e^{i \v{\varphi}^\star})$
\end{definition}

\paragraph{Transparency for free:}\label{sec:maco:transparency}
Visualizations often suffer from repeated patterns or unimportant elements in the generated images. This can lead to readability problems or confirmation biases~\cite{borowski2020exemplary}. It is important to ensure that the user is looking at what is truly important in the feature visualization. The concept of transparency, introduced in \cite{mordvintsev2018differentiable}, addresses this issue but induces additional implementation efforts and computational costs.

We propose an effective approach, which leverages attribution methods -- specifically a variant of Smoothgrad seen in \autoref{chap:attributions}) -- that yields a transparency map $\v{\alpha}$ for the associated feature visualization without any additional cost. Our solution takes advantage of the fact that during backpropagation, we can obtain the intermediate gradients on the input $\partial \mathcal{L}_{\v{v}}( \vx) / \partial \vx$ for free as $\frac{\partial \mathcal{L}_{\v{v}}( \vx)}{\partial \v{\varphi}} =  \frac{\partial \mathcal{L}_{\v{v}}( \vx)}{\partial \vx} \frac{\partial \vx}{\partial \v{\varphi}}$. We store these gradients throughout the optimization process and then average them, as done in SmoothGrad, to identify the areas that have been modified/attended to by the model the most during the optimization process. We note that a similar technique has recently been used to explain diffusion models \cite{boutin2023diffusion}. In Algorithm \ref{alg:maco:cap}, we provide pseudo-code for \magfv~and an example of the transparency maps in Figure~\ref{fig:maco:inversion} (third column).

\begin{figure}
    \centering
    \includegraphics[width=0.98\textwidth]{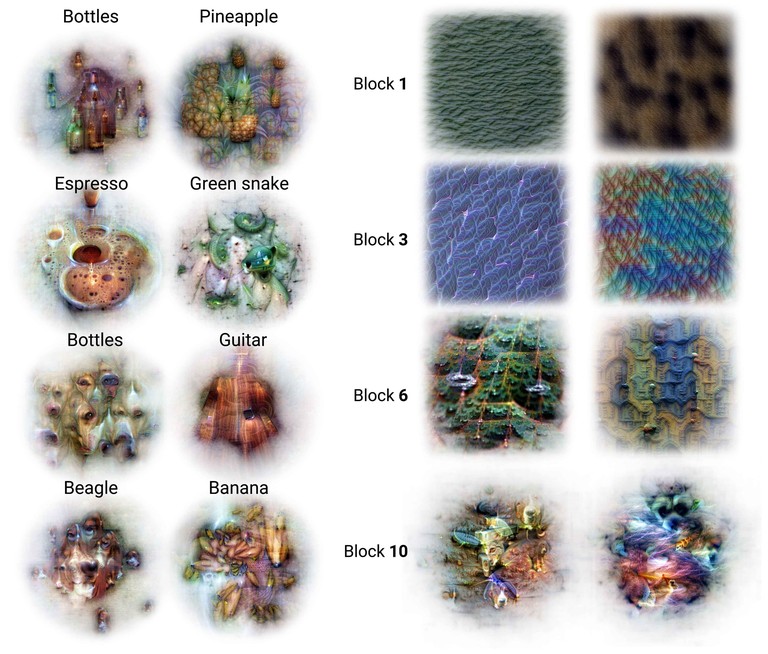}
    \caption{\textbf{(left) Logits and (right) internal representations of FlexiViT.}  \magfv~was used to maximize the activations of \textbf{(left)} logit units and \textbf{(right)} specific channels located in different blocks of the FlexViT (blocks 1, 2, 6 and 10 from left to right).}
    \label{fig:maco:logits_and_internal}
\end{figure}

\subsection{Evaluation}
\label{section:maco:evaluation}
We now describe and compute three different scores to compare the different feature visualization methods: Fourier (Olah et al.), CBR (optimization in the pixel space), and \magfv~(ours). It is important to note that these scores are only applicable to output logit visualizations. We will then demonstrate how we can use our method to perform concept visualization. %
To keep a fair comparison, we restrict the benchmark to methods that do not rely on any learned image priors. Indeed, methods with learned prior will inevitably yield lower FID scores (and lower plausibility score) as the prior forces the generated visualizations to lie on the manifold of natural images.

\paragraph*{Plausibility score.} We consider a feature visualization plausible when it is similar to the distribution of images belonging to the class it represents.
We quantify the plausibility through an OOD metric (Deep-KNN, recently used in~\cite{sun2022out}): it measures how far a feature visualization deviates from the corresponding ImageNet object category images based on their representation in the network's intermediate layers (see Table~\ref{table:maco:ood_fid}).

\paragraph{FID score.} The FID quantifies the similarity between the distribution of the feature visualizations and that of natural images for the same object category. Importantly, the FID measures the distance between two distributions, while the plausibility score quantifies the distance from a sample to a distribution. To compute the FID,  we used images from the ImageNet validation set and used the Inception v3 last layer (see Table~\ref{table:maco:ood_fid}). Additionally, we center-cropped our $512\times 512$ images to $299\times 299$ images to avoid the center-bias problem~\cite{nguyen2016multifaceted}.

\paragraph{Transferability score.} This score measures how consistent the feature visualizations are with other pre-trained classifiers. To compute the transferability score, we feed the obtained feature visualizations into 6 additional pre-trained classifiers (MobileNet~\cite{howard2017mobilenets}, VGG16~\cite{simonyan2014deep}, Xception~\cite{chollet2017xception}, EfficientNet~\cite{tan2019efficientnet}, Tiny ConvNext~\cite{liu2022convnet} and Densenet~\cite{huang2017densely}), and we report their classification accuracy (see Table~\ref{table:maco:transferability}).

All scores are computed using 500 feature visualizations, each of them maximizing the logit of one of the ImageNet classes obtained on the FlexiViT~\cite{beyer2022flexivit}, ViT\cite{kolesnikov2020bit}, and ResNetV2\cite{he2016deep} models. For the feature visualizations derived from Olah et al.~ \cite{olah2017feature}, we used all 10 transformations set from the Lucid library\footnote{\href{https://github.com/tensorflow/lucid}{https://github.com/tensorflow/lucid}}.
CBR denotes an optimization in pixel space and using the same 10 transformations, as described in~\cite{nguyen2015deep}.
For \magfv, $\augmentation$ only consists of two transformations; first we add uniform noise $\bm{\delta} \sim \mathcal{U}([-0.1, 0.1])^{W \times H}$ and crops and resized the image with a crop size drawn from the normal distribution $\mathcal{N}(0.25, 0.1)$, which corresponds on average to 25\% of the image.
We used the NAdam optimizer \cite{dozat2016incorporating} with $lr=1.0$ and $N = 256$ optimization steps. Finally, we used the implementation of \cite{olah2017feature} and CBR which are available in the Xplique library~\cite{fel2022xplique} \footnote{\href{https://github.com/deel-ai/xplique}{https://github.com/deel-ai/xplique}} which is based on Lucid.

\begin{table}[ht]
\centering
        \begin{tabular}{lccc}
            & FlexiViT & ViT & ResNetV2\\
            \hline
            \multicolumn{4}{l}{$\bullet$\;\textbf{Plausibility score} (1-KNN) ($\downarrow$)}\\

            \magfv & {\bf 1473} & {\bf 1097 } & {\bf 1248} \\
            Fourier~\cite{olah2017feature} & 1815 &  1817 & 1837 \\
            CBR~\cite{nguyen2015deep} &  1866 & 1920 & 1933 \\
            \hline
            \multicolumn{4}{l}{$\bullet$\;\textbf{FID Score}  ($\downarrow$)}\\
            \magfv & {\bf 230.68} & {\bf 241.68} & {\bf 312.66} \\
            Fourier~\cite{olah2017feature} &  250.25 & 257.81 & 318.15 \\
            CBR~\cite{nguyen2015deep} &  247.12 & 268.59 & 346.41 \\
            \hline
        \end{tabular}
        
        \caption{Plausibility and FID scores for different feature visualization methods applied on FlexiVIT, ViT and ResNetV2}
    \label{table:maco:ood_fid}
\end{table}

\begin{table}[ht]
\centering
\begin{tabular}{lccc}
    & FlexiViT & ViT & ResNetV2 \\
    \hline
    \multicolumn{4}{l}{$\bullet$\;\textbf{Transferability score($\uparrow$)}: \magfv / Fourier~\cite{olah2017feature}} \\

    MobileNet  & {\bf 68} \slash~38 & {\bf 48}\slash~37  & {\bf 93} \slash~36 \\
    VGG16         & {\bf 64} \slash~30 & {\bf 50} \slash~30 & {\bf 90} \slash~20 \\
    Xception      & {\bf 85} \slash~61 & {\bf 73} \slash~62 & {\bf 97} \slash~64 \\
    Eff. Net  & {\bf 88} \slash~25 & {\bf 63} \slash~25 & {\bf 82} \slash~21 \\
    ConvNext & {\bf 96} \slash~52 & {\bf 84} \slash~55 & {\bf 93} \slash~60\\
    DenseNet      & {\bf 84} \slash~32 & {\bf 66} \slash~31 & {\bf 93} \slash~25 \\
    \hline
    \\
    \end{tabular}
        \caption{Transferability scores for different feature visualization methods applied on FlexiVIT, ViT and ResNetV2.}
        \label{table:maco:transferability}
\end{table}

For all tested metrics, we observe that \magfv~produces better feature visualizations than those generated by Olah et al.~\cite{olah2017feature} and CBR~\cite{nguyen2015deep}. We would like to emphasize that our proposed evaluation scores represent the first attempt to provide a systematic evaluation of feature visualization methods, but we acknowledge that each individual metric on its own is insufficient and cannot provide a comprehensive assessment of a method's performance. However, when taken together, the three proposed scores provide a more complete and accurate evaluation of the feature visualization methods.

\subsubsection{Human psychophysics study}
Ultimately, the goal of any feature visualization method is to demystify the CNN's underlying decision process in the eyes of human users. To evaluate \magfv~'s ability to do this, we closely followed the psychophysical paradigm introduced in~\cite{zimmermann2021well}. In this paradigm, the participants are presented with examples of a model's ``favorite'' inputs (i.e., feature visualization generated for a given unit) in addition to two query inputs. Both queries represent the same natural image, but have a different part of the image hidden from the model by a square occludor. The task for participants is to judge which of the two queries would be ``favored by the model'' (i.e., maximally activate the unit). The rationale here is that a good feature visualization method would enable participants to more accurately predict the model's behavior. Here, we compared four visualization conditions (manipulated between subjects): Olah~\cite{olah2017feature}, \magfv~with the transparency mask (the transparency mask is decribed in \ref{sec:maco:transparency}), \magfv~without the transparency mask, and a control condition in which no visualizations were provided. In addition, the network (VGG16, ResNet50, ViT) was a within-subject variable. The units to be understood were taken from the output layer.

\begin{figure}[ht]
\includegraphics[width=0.9\textwidth]{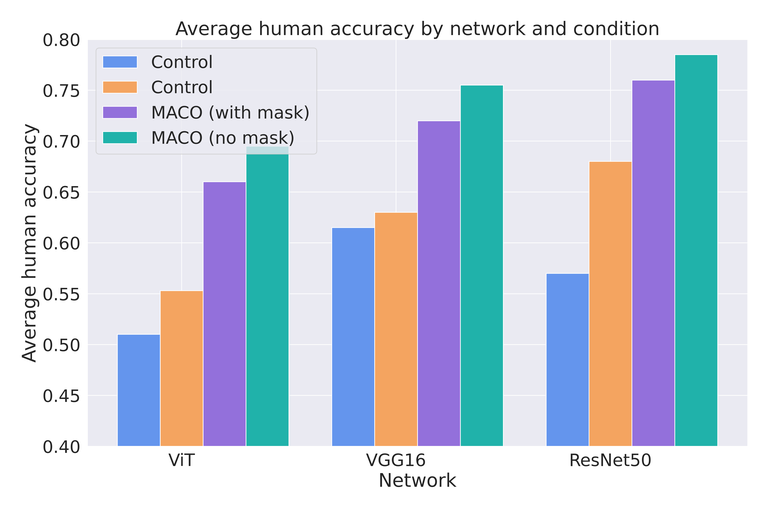}
\caption{\textbf{Human causal understanding of model activations}. We follow the experimental procedure introduced in~\cite{zimmermann2021well} to evaluate Olah and \magfv~visualizations on $3$ different networks. The control condition is when the participant did not see any feature visualization. 
}
\label{fig:maco:human_results}    
\end{figure}

Based on the data of 174 participants on Prolific (\url{www.prolific.com}), we found both visualization and network to significantly predict the logodds of choosing the right query (Fig.~\ref{fig:maco:human_results}). That is, the logodds were significantly higher for participants in both the \magfv~conditions compared to Olah. On the other hand, our tests did not yield a significant difference between Olah and the control condition, or between the two \magfv~conditions. Finally, we found that, overall, ViT was significantly harder to interpret than ResNet50 and VGG16, with no significant difference observed between the latter two networks. Full experiment and analysis details can be found in the supplementary materials, section~\ref{sup:maco:psychophysics}. 

However, it should be noted that investigating the effect on a neuron-by-neuron basis, as in the original setup, may not be advisable for the issues outlined in \autoref{sec:concepts:craft} and referenced in \cite{elhage2022superposition}. Conducting a parallel study that confirms this by utilizing meaningful directions in the latent space -- e.g., with \craft -- instead of individual neurons would be of interest.

\subsubsection{Ablation study}

    \begin{table}%
        \centering
        \begin{tabular}{lccc}
            FlexiViT & Plausibility ($\downarrow$) & FID ($\downarrow$) & logit magnitude ($\uparrow$) \\
            \hline
            \magfv  & 571.68 & 211.0 & 5.12 \\
            - transparency & 617.9 (+46.2) & 208.1 (-2.9) & 5.05 (-0.1)\\
            - crop & 680.1 (+62.2) & 299.2 (-91.1) & 8.18 (+3.1)\\
            - noise & 707.3 (+27.1) & 324.5 (-25.3) & 11.7 (+3.5)\\
            \hline
            Fourier~\cite{olah2017feature} & 673.3 & 259.0 & 3.22\\
            - augmentations & 735.9 (+62.6) &  312.5 (+53.5) & 12.4 (+9.2)\\
        \end{tabular}
        \caption{\textbf{Ablation study on the FlexiViT model:} This reveals that 1. augmentations help to have better FID and Plausibility scores, but lead to lesser salients visualizations (softmax value), 2. Fourier~\cite{olah2017feature} benefits less from augmentations than \magfv.}
        \label{table:maco:ablation}
    \end{table}

    To disentangle the effects of the various components of \magfv, we perform an ablation study on the feature visualization applications. We consider the following components: (1) the use of a magnitude constraint, (2) the use of the random crop, (3) the use of the noise addition, and (4) the use of the transparency mask. We perform the ablation study on the FlexiViT model, and the results are presented in Table~\ref{table:maco:ablation}. We observe an inherent tradeoff between optimization quality (measured by logit magnitude) on one side, and the plausibility (and FID) scores on the other side. This reveals that plausible images which are close to the natural image distribution do not necessarily maximize the logit.
    Finally, we observe that the transparency mask does not significantly affect any of the scores confirming that it is mainly a post-processing step that does not affect the feature visualization itself.

\subsection{Applications}

We demonstrate the versatility of the proposed \magfv~technique by applying it to three different XAI applications:

\paragraph{Logit and internal state visualization.} For logit visualization, the optimization objective is to maximize the activation of a specific unit in the logits vector of a pre-trained neural network (here a FlexiViT\cite{beyer2022flexivit}). The resulting visualizations provide insights into the features that contribute the most to a class prediction (refer to Figure~\ref{fig:maco:logits_and_internal}a). For internal state visualization, the optimization objective is to maximize the activation of specific channels located in various intermediate blocks of the network (refer to Figure~\ref{fig:maco:logits_and_internal}b). This visualization allows us to better understand the kind of features these blocks -- of a FlexiViT\cite{beyer2022flexivit} in the figure -- are sensitive to.

\begin{figure}
    \centering
    \includegraphics[width=1.0\textwidth]{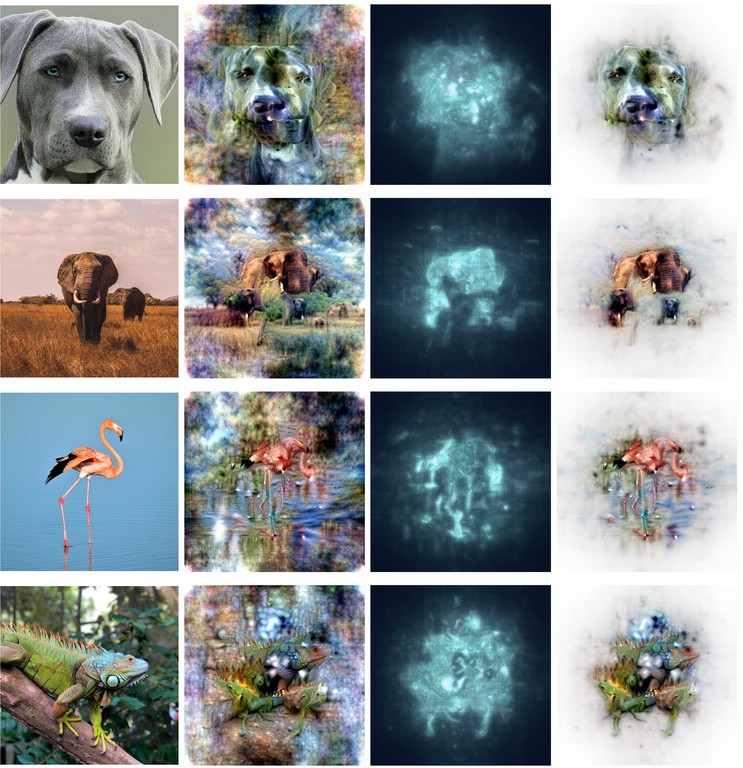}
    \caption{\textbf{Feature inversion.} Images in the second column match the activation pattern of the penultimate layer of a ViT when fed with the images of the first column. In the third column, we show their corresponding attribution-based transparency masks, leading to better feature visualization when applied (fourth column).}
    \label{fig:maco:inversion}
\end{figure}

\paragraph{Feature inversion.} The goal of this application is to find an image that produces an activation pattern similar to that of a reference image. By maximizing the similarity to reference activations, we are able to generate images representing the same semantic information at the target layer but without the parts of the original image that were discarded in the previous stages of the network, which allows us to better understand how the model operates.
Figure~\ref{fig:maco:inversion}a displays the images (second column) that match the activation pattern of the penultimate layer of a VIT when given the images from the first column. We also provide examples of transparency masks based on attribution (third column), which we apply to the feature visualizations to enhance them (fourth column).

\begin{figure}
    \centering
    \includegraphics[width=1.0\textwidth]{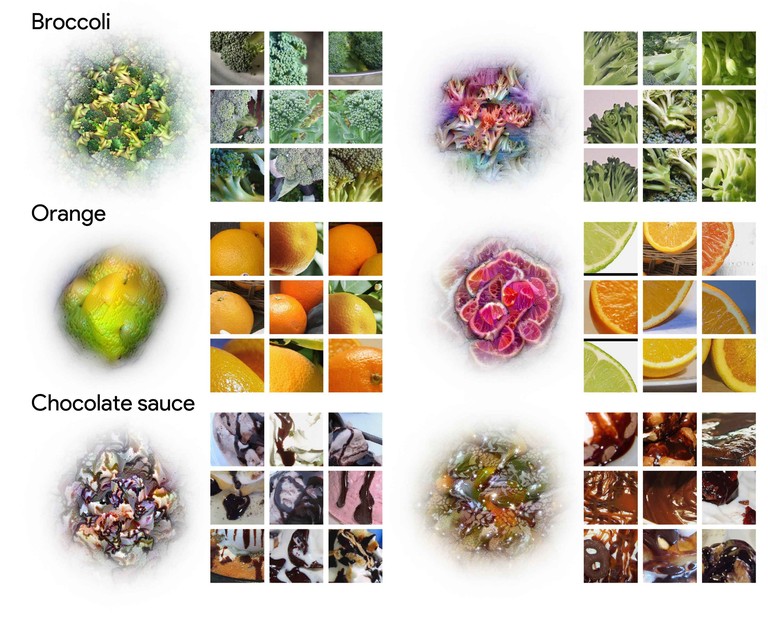}
    \caption{\textbf{Concept visualization.} \magfv~is used to visualize concept vectors extracted with the \craft~ method~\autoref{sec:concepts:craft}. The concepts are extracted from a ResNet50 trained on ImageNet.}
    \label{fig:maco:concepts}
\end{figure}

\paragraph{Concept visualization.} Herein we combine \magfv~with concept-based explainability. Such methods aim to increase the interpretability of activation patterns by decomposing them into a set of concepts~\cite{ghorbani2019towards}. In this work, we leverage our \craft~concept-based explainability method~\cite{fel2023craft}, which uses Non-negative Matrix Factorization to decompose activation patterns into main directions -- that are called concepts --, and then, we apply \magfv~to visualize these concepts in the pixel space. To do so, we optimize the visualization such that it matches the concept activation patterns. In Figure~\ref{fig:maco:concepts}b, we present the top $2$ most important concepts (one concept per column) for five different object categories (one category per row) in a ResNet50 trained on ImageNet. The concepts' visualizations are followed by a mosaic of patches extracted from natural images: the patches that maximally activate the corresponding concept. 

\subsection{Limitations}\label{sec:maco:limitations}
We have demonstrated the generation of realistic explanations for large neural networks by imposing constraints on the magnitude of the spectrum. However, it is important to note that generating realistic images does not necessarily imply effective explanation of the neural networks. The metrics introduced in this section allow us to claim that our generated images are closer to natural images in latent space, that our feature visualizations are more plausible and better reflect the original distribution. However, they do not necessarily indicate that these visualizations helps humans in effectively communicating with the models or conveying information easily to humans.
Furthermore, in order for a feature visualization to provide informative insights about the model, including spurious features, it may need to generate visualizations that deviate from the spectrum of natural images. Consequently, these visualizations might yield lower scores using our proposed metrics.
Simultaneously, several interesting studies have highlighted the weaknesses and limitations of feature visualizations~\cite{borowski2020exemplary,geirhos2023dont,zimmermann2021well}. One prominent criticism is their lack of interpretability for humans, with research demonstrating that dataset examples are more useful than feature visualizations in understanding convolutional neural networks (CNNs)~\cite{borowski2020exemplary}. This can be attributed to the lack of realism in feature visualizations and their isolated use as an explainability technique.
With our approach, \magfv~, we take an initial step towards addressing this limitation by introducing magnitude constraints, which lead to qualitative and quantitative improvements. Additionally, we promote the use of feature visualizations as a supportive and complementary tool alongside other methods such as concept-based explainability, exemplified by \craft. We emphasize the importance of feature visualizations in combating confirmation bias and encourage their integration within a comprehensive explainability framework.

\subsection{Discussion}

In this section, we introduced a novel approach, \magfv, for efficiently generating feature visualizations in modern deep neural networks based on \tbi{i} a hard constraint on the magnitude of the spectrum to ensure that the generated visualizations lie in the space of natural images, and \tbi{ii} a new attribution-based transparency mask to augment these feature visualizations with the notion of spatial importance. This enhancement allowed us to scale up and unlock feature visualizations on large modern CNNs and vision transformers without the need for strong -- and possibly misleading -- parametric priors.
We also complement our method with a set of three metrics to assess the quality of the visualizations. Combining their insights offers a way to compare the techniques developed in this branch of XAI more objectively. We illustrated the scalability of \magfv~ with feature visualizations of large models like ViT, but also feature inversion and, critically, concept visualization.

Indeed, this tool integrates seamlessly with concept extraction methods, enabling the visualization of extracted concepts without resorting to image cropping. This approach offers a clearer, more causal view of the mechanisms that activate a given concept, thereby contributing significantly to our understanding of the internal workings of neural networks.

\clearpage

\section{Conclusion}

The conclusion of this chapter serves as an opportune moment for reflection and synthesis. Our research has led us through an in-depth examination of Hypothesis~\ref{hyp:what}, which posited that existing attribution methods fall short, as they primarily reveal ~\where but overlook the crucial aspect of the \what.

This chapter was dedicated to developing appropriate tools to address this issue. We began by constructing \craft, a method for decomposing the activations of a model into a set of concepts, demonstrating indeed its enhanced utility for human understanding compared to traditional attribution methods.
We decided to go one step further, in \autoref{sec:concepts:holistic}, where we established a theoretical framework that make: \tbi{i} show that concept extraction is \textbf{Dictionary learning}, and \tbi{ii} make a link between attribution methods and concept importance. The formulas used to determine the importance of a pixel, as seen in \autoref{chap:attributions}, are identical to those applied in evaluating the significance of concepts after decomposition. In the final section, we explored concept visualization as a way to visualize concept by introducing \maco. 

To summarize our novel framework, it consists in reinterpreting the intricate latent space of neural networks through a collection of atomic units termed concepts. While these concepts are mathematically abstract, we employed two methods to imbue them with meaning: maximally activating crops and feature visualization techniques. Additionally, it became evident that among these concepts, some offer greater utility than others, with attribution methods precisely identifying the most relevant ones.

\begin{figure}[ht]
    \centering
    \includegraphics[width=0.49\textwidth]{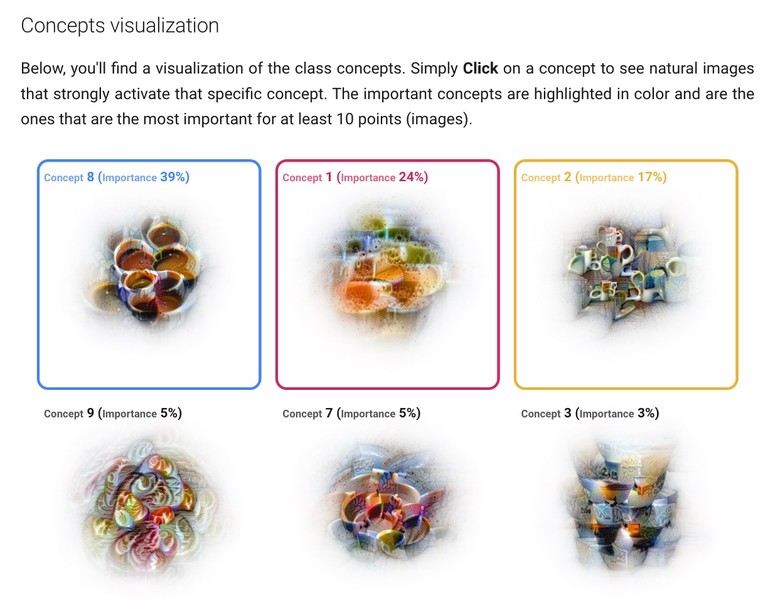}
    \includegraphics[width=0.49\textwidth]{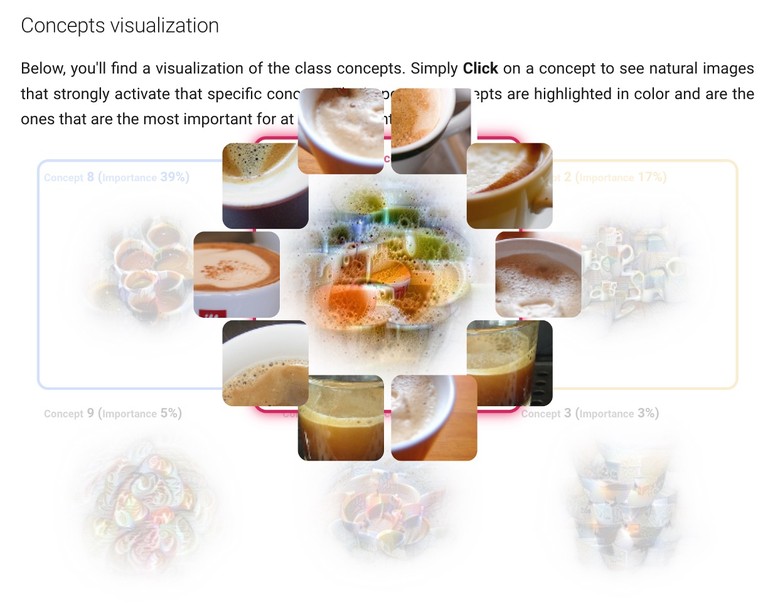}
    \caption{\textbf{LENS Project.} Example of results from the LENS demo for the espresso class. \textbf{(Left)} The page displays the top 10 concepts, ranked from most to least important for the class. These concepts are extracted using \craft~and visualized with \maco; their importance is calculated using the optimal formula found in \autoref{sec:concepts:holistic}. \textbf{(Right)} Clicking on a feature visualization that illustrates a concept reveals the image crops that most strongly activate the concept.}
    \label{fig:concepts:lens}
\end{figure}

\paragraph{A New Synergetic Approach to Explainability.} This new framework is distinct in its ability to \textit{synthesize all existing tools for explainability into a cohesive and synergetic system}. Our goal was to demonstrate the potential of this approach -- and the powerful synergy it creates -- through the visual demonstration offered by \Lens~(illustrated in \autoref{fig:concepts:lens}).

This platform organizes, for each of the 1000 ImageNet classes, the ten most significant concepts, along with their feature visualizations and respective importance. 

\begin{figure}
    \centering
    \includegraphics[width=1.05\textwidth]{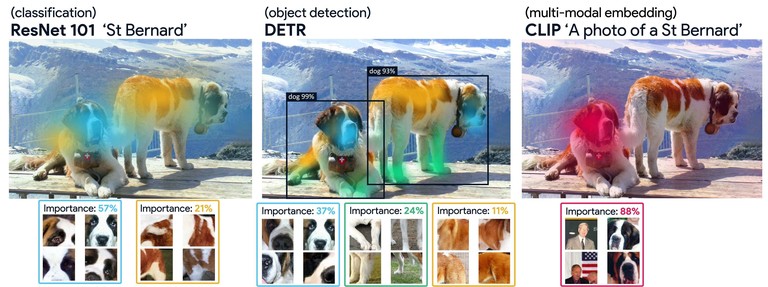}
    \caption{\textbf{Concepts of \craft~on ResNet50 Trained on Different Tasks.} The concepts extracted for classification tasks, specifically for the St. Bernard class, seem to focus on the head and spots of the St. Bernard. In the case of the bounding box (bbox) model, the legs are also deemed important, possibly because they help delineate the edges of the bounding boxes? Interestingly, for the CLIP model, the dog head concept is also activated by human heads, suggesting that despite visual differences (in the pixel space), the concept of 'head' seems present for models trained with language components like CLIP.}
    \label{fig:holistc:conceptbbox}
\end{figure}

\paragraph{Perspective.} While the potential of concept-based methods is clear, it is now critical to establish distinct research directions to fully unlock their potential in the wake of preliminary studies. Four key areas emerge, meriting further exploration:

\begin{itemize}

    \item \textbf{Revisiting Dictionary Learning:} The evident parallels between concept extraction and dictionary learning highlight a pressing need for the XAI community to reassess and tailor dictionary learning methodologies for application in explainability. This adaptation could bridge gaps in our understanding and application of these techniques within XAI.

    \item \textbf{Beyond Classification:} The necessity of extending our investigative scope beyond mere classification tasks is crucial. Diverse models, including bounding box detection, segmentation, generative and Vision-Language models present intricate challenges and vast opportunities for enhancing explainability. Diversifying our focus will enable a general comprehension of AI systems, integrating a broader spectrum of tasks and functionalities, thus deepening the XAI field with richer insights and more adaptable explainability tools. An illustrative example is given in \autoref{fig:holistc:conceptbbox}.
    
    \item \textbf{Exploring Hierarchical Concepts and Compositionality:} Investigating hierarchical concepts and their compositionality also offers a very promising path to deepen our understanding of how neural networks operate. Recent research has highlighted that models can exhibit compositional behaviors~\cite{lepori2024break}. Understanding the ways in which concepts are combined and interact at various abstraction levels could offer a nuanced perspective on decision-making processes within models, paving the way for more refined interpretability strategies.
    
    \item \textbf{Expanding on Synergies:} The demonstrated synergy among different explainability methods within our framework suggests a fertile area for research. A comprehensive examination of how these methods can be cohesively integrated, and the resultant synergistic effects could lead to groundbreaking insights and the development of potent tools for explainability.
    
\end{itemize}

While this framework does not solve all the challenges presented in the \autoref{chap:intro}, and there remains a significant journey toward fully understanding models such as ResNet50 or ViT, it opens a novel avenue. We encourage the academic community to explore the synergies between attribution methods, concepts, and feature visualization for deeper explainability.

\clearpage

\chapter{Conclusion \& Perspective}
\label{chap:conclusion}

\textbf{Much work remains to be done}. Deep Learning is incredibly efficient and the field of Explainable Artificial Intelligence (XAI) is now under intense scrutiny, tasked with the hard task of understanding the inner-working of these models. The fulfillment of this colossal goal remains uncertain, yet our progress is undeniable, gradually providing methods and insights that enhance our understanding of artificial intelligence models.

In that perspective, this thesis aimed to develop a modest suite of explainability methods for computer vision models, making contributions across various chapters.

The \autoref{chap:attributions} dedicated to attribution methods has thoroughly investigated these techniques, introducing a new metric to identify models providing better explanations (\autoref{sec:attributions:mege}), a black-box method based on Sobol indices (\autoref{sec:attributions:sobol}), and an advanced approach utilizing formal methods (\autoref{sec:attributions:eva}). An in-depth human experiment (\autoref{sec:attributions:metapred}) highlighted the usefulness of these methods in simplified scenarios, marking a notable advancement for explainability. However, their efficacy proved limited in more complex contexts, leading us to formulate two hypotheses guiding further research.

The first hypothesis, explored in \autoref{chap:alignment}, focused on aligning vision models to share coherent explanations (\autoref{sec:harmonization}), thereby increasing their accuracy and generalization beyond training explanations. We also sketched an alternative theoretical approach aimed at constraining our neural networks to adhere to a specific functional norm (\autoref{sec:lipschitz}), suggesting a promising alignment with human reasoning. This path would require a more holistic design incorporating improved data, tasks closer to human capabilities, and a more plausible architecture.

The final section of this thesis, \autoref{chap:concepts}, ventured into the emerging domain of concepts-based explainability, proposing an innovative method for their extraction and establishing a theoretical framework that unify concept extraction and dictionary learning as well as concept importance and attribution methods. The chapter culminates with the introduction of a technique for visualizing the extracted concepts, showcased through LENS. This illustration serves as evidence of the synergistic approach achievable by integrating Attribution, Concepts, and Feature Visualization.

\section{Perspective}

It is customary to conclude this manuscript with predictions. However, I must confess a certain reluctance in the art of forecasting, and therefore, I will reformulate this as remarks that have accompanied me throughout this journey. They are fourfold: (i) the paths that seem promising for the future of explainability, (ii) the disease of dimensionality reduction in XAI, (iii) the link between generalization and explainability, and (iv) finally, the human aspect of this research field.

\subsection{Promising Avenues.} 
This work has consistently aimed to build upon established research foundations rather than attempting to invent anew. Given the relative infancy and lack of established benchmarks in our field, it is imperative to construct our understanding based on robust frameworks. An important realization during my investigation into attribution methods was the relevance of Global Sensitivity Analysis (GSA), which is a field that has been tackling analogous challenges for over three decades, offering both tools and insights that are directly applicable. The core challenge addressed by GSA involves identifying and ranking the inputs $\rx_i$ that most significantly influence a model's random output:

$$
\rv{y} = \f(\rvx).
$$

The tools available in GSA range from estimating the impact of each variable in the output to measuring the dependence and interactions between variables themselves. I believe that a cross-disciplinary contributions (XAI and GSA) could enrich our understanding of attribution methods and the broader concept of importance estimation.

To continue on this vein, the exploration in \autoref{chap:concepts}~reveals that current attribution methods often only scratch the surface, suggesting a deeper examination of internal activations is necessary for a thorough understanding. In our case, we have seen that many currently developed methods could be framed as dictionary learning:

$$
\s{R}(\vx) = \min_{\v{u},\m{V}} \norm{ \vx - \v{u}\m{V} }_p + \lambda \Omega(\v{u}).
$$

I believe that this field is incredibly valuable for researchers focused on explainability. Numerous studies have introduced a variety of analysis techniques and extensions, ranging from hierarchical dictionaries to supervised dictionaries, which I find particularly well-suited for explainability.
In summary, we have a solid arsenal at our disposal to tackle the problem of reinterpreting latent space. The insights from this document offer several avenues  --including one discussed below on the disease of dimensionality reduction.

Furthermore, during the last chapter, we advocate for a paradigm shift from \textit{separation} to \textit{integration}, highlighting the importance of interpreting the methods discussed as part of a synergistic framework. Each method -- Attribution, Concept, Feature visualization -- reveal a fragment of the puzzle, suggesting that a comprehensive understanding requires the amalgamation of these complementary insights. The \Lens~demonstration project modestly aspires to guide the field towards this integrated approach, emphasizing the collective interpretation of diverse methodologies to achieve a holistic understanding of the research landscape.

\subsection{Dimensionality Reduction Disease.} In the quest for explainability, there is a legitimate expectation for it to simplify the processes, to reduce the cognitive load of the internal mechanisms governing neural networks. However, it has become apparent that the methods sometimes used to simplify the problem can, paradoxically, obscure more than they elucidate the internal realities or phenomena we aim to explain. Resorting to indiscriminately reducing the dimensions of activations or summarizing a model's decision-making process with a linear model may be useful as a preliminary approach, but I am concerned that these methods might lead to more confusion than clarification. We must now look beyond these initial simplifications and confront the question: \textit{How can we embrace complexity in a non-reductive manner?}

I believe the theory presented in \autoref{sec:concepts:holistic} suggests a path forward: \craft~succeeds not because it reduces dimensions, but \textit{because it expands it}! Indeed, \Lens~encompasses over 10,000 concepts, far exceeding the dimensionality of the model's latent space. If our models can be likened to tangled balls of yarn made up of features, our goal should then be to untangle these to understand the strands and thereby connect the features in a potentially much larger space. In other words, contrary to intuition, I am convinced that explainability must now aim to increase dimensionality (appropriately), rather than reduce it. This expanded approach necessitates developing methodologies that can navigate and articulate the increased complexity, ensuring that the additional dimensions serve to clarify rather than confound our understanding of neural network behaviors. By embracing and effectively managing this complexity, we can move closer to achieving true explainability and reveals the intricate interplay of features and their contributions to model decisions in a comprehensive and nuanced manner.

\subsection{Generalization, Algorithmic Complexity, and Explainability}

\paragraph{Generalization.}
The fate of explainability is intimately linked to the challenge of generalization. The Vapnik-Chervonenkis (VC) Dimension, a fundamental concept in statistical learning, offers a theoretical framework for assessing a model's generalization capability. It quantifies the complexity of a "hypothesis set" or functions a model can learn, suggesting that a high VC-Dimension indicates a model's potential for precise adaptation to training data, potentially leading to overfitting. In theory, a model with a higher VC-Dimension could exhibit reduced generalization capability, becoming less adept at making accurate predictions on unseen data.

However, the empirical performance of deep neural networks, with their often really large number of parameters, challenges this traditional understanding. Despite their complexity, these models have shown remarkable generalization abilities, questioning the adequacy of existing tools and theories, including the VC-Dimension, to fully explain this phenomenon in deep learning.

This realization prompts a deeper exploration into the \textbf{algorithmic dimension of XAI}, emphasizing the importance of acknowledging AI models' computational specificities. The distinction between ideal mathematical operations and their actual computational implementations necessitates a consideration of computational complexity. For instance, the implementation invariance axiom used to build an interpretable surrogate suggests that functions $\f_a$ and $\f_b$ are considered strictly equal if they produce the same output. However, it is possible for two equal functions to have significantly different underlying computational processes (different computational graphs). Early pioneers at the Dartmouth Conference already recognized this specificity and emphasized the need for further research into the algorithmic study of AI.

\paragraph{Algorithmic Complexity.} The field of Algorithmic Information Theory (AIT)~\cite{chaitin1977algorithmic,grunwald2008algorithmic} \footnote{This section was heavily inspired by discussions with Louis Béthune.} appears well-suited for a better understanding of models. AIT formalizes simplicity and complexity from an algorithmic standpoint, positing that an object is simple if it can be concisely described and complex if no such succinct description exists. A central notion in AIT is the \textit{Kolmogorov Complexity}, it is the length of the shortest program that outputs a string $\vx$ when run on a universal Turing machine $U$ and denoted by:

$$K(\vx) = \min_{p} \{ \ell(p) : U(p) = \vx \}.$$

With $\ell(p)$ the length of the program $p$. It can be demonstrated that the definition of $K(\vx)$ is robust with respect to the choice of the universal Turing machine $U$. Specifically, $K(\vx)$ varies by at most an additive constant that is independent of $\vx$ when a different $U$ is selected. This principle, known as the invariance theorem~\cite{solomonoff1964formal,kolmogorov1965three,chaitin1969length}, marks a foundational moment in the development of algorithmic information theory. Despite the non-computability of this measure, which introduces its own set of challenges, it yields significant insights on what the underlying mechanisms are doing, and on the nature of generalization of machine learning models. Other notable concepts include the Algorithmic "Solomonoff" Probability~\cite{solomonoff1997discovery} and the Levin Search~\cite{levin1973universal}, which propose alternative approaches to understanding algorithmic complexity:

\begin{align}
K_{\text{Levin}}(\vx) &\defas \min_{p} \{ \ell(p) + \log(T(p)) : U(p) = \vx \}, ~~
K_{\text{Solomonoff}}(\vx) &\defas \sum_{p : U(p) = \vx} 2^{-\ell(p)}. \nonumber
\end{align}

Where $T(p)$ denote the computational time of the program $p$, that could translate into either the inference time, or the learning time.

\begin{figure}[ht]
    \centering
    \includegraphics[width=0.8\textwidth]{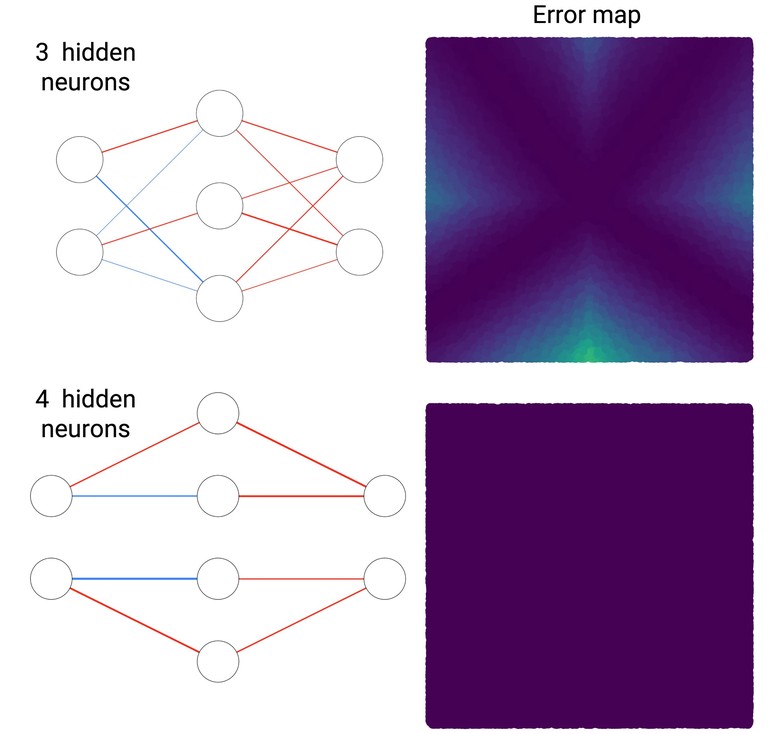}
    \caption{\textbf{Toy Example: Adding Parameters Simplifies the Algorithm.} Contrary to expectations, one might anticipate a more complex algorithm for a model with an increased number of parameters. However, this example demonstrates a DNN trained to predict $\vy = |\vx|$ with $\vx \in \mathbb{R}^2$. \textbf{(Top)} Illustrates a DNN with 3 hidden layers that struggles with generalization, as evidenced by the error map on the right, which depicts $\int_{\mathbb{R}^2}(|\vx| - f(\vx))^2 \mathrm{d} \vx$ and employs a convoluted algorithmic method to approximate $|\vx|$. \textbf{(Bottom)} Conversely, in the lower section, adding a neuron -- thereby increasing the number of parameters --results in not only improved generalization but also a more intuitive algorithm: it effectively divides $\vx$ into two by applying $\text{ReLU}(x_1), \text{ReLU}(-x_1)$ and similarly for $x_2$.}
    \label{fig:conclusion:toy_abs}
    \vspace{-4mm}
\end{figure}

Interestingly, this perspective represents a paradigm shift, focusing not on the \textit{quantity of parameters} but on the \textit{algorithm discovered by the model}. At first glance, some may wonder about the direction of this argument, as it might seem self-evident that an increase in parameters leads to a more complex algorithm. However, I will present an example intended to be instructive, without promising anything beyond offering the intuition that more parameters do not necessarily equate to increased complexity (see \autoref{fig:conclusion:toy_abs})\footnote{Example inspired by the excellent \cite{elhage2022superposition} article.}.

Let's consider the task of learning the function $\vy = |\vx|$ where $(\vx, \vy) \in \mathbb{R}^d$ using a neural network with a single hidden layer, ReLU activations, and $n$ intermediate neurons. Starting with $d=2$ and $n=3$, we have 3 neurons to store intermediate states and then output $(|x_1|, |x_2|)$. The algorithm discovered by the model is complex, and the model fails to generalize well, as evident from the error map in \autoref{fig:conclusion:toy_abs}. However, by adding just one more parameter $n=4$, not only does the model begin to generalize better, but the mechanism it uses also becomes clear and simple: the increase in parameters leads to a reduction in algorithmic complexity. With this example, I aim not to prove a point definitively, but to suggest that the notion of complexity could be much richer than merely the count of parameters\footnote{Code to reproduce is available in \autoref{ap:conclusion:toy_abs}}.

I want to acknowledge that the possible notions of complexity are varied, and other perspectives are possible, ranging from principles of parsimony, symmetry or to the algorithmic complexity I wished to highlight. This diversity in understanding complexity underscores the multifaceted nature of what we deem "complex" and suggests a richer tapestry of factors that can influence the interpretability and functionality of models beyond just their parameter count.

\paragraph{Explainability.} Turning our focus back to explainability, explainability and generalization are poised to evolve in tandem. The development of new explainability tools to study model $\f(\cdot)$, as outlined in this manuscript, marks an initial step. However, we initiated this discussion (\autoref{sec:attributions:mege}) with the study of $\s{A}(.)$, that is, the algorithm generating $\f$, and often, it is these foundational ideas that resurface towards the end, much like a recurring theme. It seems now that the flow should take us into understanding the learning algorithms themselves, rather than merely the models they produce. Do neural networks strive to minimize the complexity of their programs in the vein of Kolmogorov, or perhaps, akin to Levin, seek a balance between program simplicity and execution time? Or is there an element of randomness in program emergence, with certain programs becoming more useful than others in a manner reminiscent of Solomonoff's complexity? What is the relationship between features and programs? What are the inductive biases of our models, and can these be interpreted as routines available in a conditional complexity manner? 

These questions bring the algorithmic aspect of deep learning back into focus, which I believe is a promising direction for continuing this research. Of course, the perspective is certainly not perfectly accurate, yet it has the merit of sparking a new set of questions to which we can now begin to provide some answers, thanks to the tools developed throughout this thesis. 

\vspace{-2mm}
\subsection{The Human Dimension}
\vspace{-2mm}

Explainability also intersects significantly with the human aspect of science. It transcends the mere decoding of models and touches upon our cognitive capacity to comprehend complex systems. Explainability thus challenges our limitations, especially our cognitive boundaries. To put it simply, other forms of intelligence might regard our quest for explainability as unusual, given their potential to directly decipher the weights and biases within AI systems without the intermediary steps humans require.

Echoing Camus, ``To understand the world is to reduce it to the human,'' we might say that to comprehend a neural network means to render it intelligible \textit{to us}, taking into account our cognitive limitations and capabilities. Is it truly possible to distill AI to a level that aligns with our understanding? It seems we are faced with two paths: one where our models harbor an inherent simplicity we have yet to discover, whether it be through symmetry or algorithmic simplicity, and another, more daunting (and exciting) path that recognizes our models as genuinely complex systems. 

The risk, then, lies in not confusing complexity with completeness. Understanding the intricate inner workings of these systems will requires a multidimensional approach, rich in interactions and dependencies. Kolmogorov noted, "The human brain is incapable of creating anything that is truly complex," suggesting that our understanding will be built gradually by discovering, constructing, and assembling simple concepts along the way. Thus, our challenge in comprehending neural networks will require avoiding oversimplification, ensuring we capture both their detailed structures and the fundamental motifs governing their behavior.

\vspace{0mm}
{
\begin{center} 
\Large \adforn{21} 
\end{center}
}
\vspace{0mm}

These insights bridge the human and computational aspects of this challenge. Recognizing the complexity in the face of a \textit{desert} of unknowns, we strive to find equilibrium between our cognitive capabilities and the unique algorithmic features of artificial intelligence. In doing so, we may already be uncovering some \textit{Sparks} of understanding.

\vspace{0mm}

\epigraph{``What makes the desert beautiful'', the little prince said, ``is that it hides a well somewhere...''}{Antoine de Saint-Exupéry}

\clearpage


\clearpage

\appendix

\chapter{Attributions methods}

\section{Algorithmic Stability measure for Explainability}
\begin{algorithm}[h]
\caption{Training procedure to compute $\Setm$ and $\Setp$}
\begin{algorithmic}
\label{alg:mege:procedure}

\REQUIRE $k \in \mathbb{N}_{\geq 2} \ ,\ \s{D} = \{ \s{V}_i \}_{i=1}^k $
\STATE $\Setm \gets{} \{\}$, $\Setp \gets{} \{\}$

\FORALL{$ i \in  \{1,~\ldots{},~k \}  $}
\STATE \textbf{Train} $\f_i$ on $\s{D} \setminus \s{V}_i$
\FORALL{$(\vx, \vy) \in \s{D}$}
\STATE \textit{// generate explanations on all dataset} 
\STATE $ \explanation_{\vx}^{(i)} \gets{} \explainer(\f_i, \vx)$
\ENDFOR
\ENDFOR

\FORALL{$ i \in  \{1,~\ldots{},~k \}$}
\FORALL{$(\vx, \vy) \in \s{V}_i$}
\FORALL{$ j \in  \{1,~\ldots{},~k \mid i \neq j \}$}

\STATE \textit{// $\f_{j}$ was trained on $\vx$, $\f_{i}$ was not} 
\STATE $ \delta_{\vx}^{(i,j)} \gets{} d( \explanation_{\vx}^{(i)}, \explanation_{\vx}^{(j)} )$

\IF{$\f_{i}(\vx) = \vy $ \AND $\f_{j}(\vx) = \vy$}
\STATE \textit{// both model are correct}
\STATE $\Setm \gets{} \Setm \cup \{ \delta_{\vx}^{(i,j)} \}$

\ELSIF{$\f_i(\vx) = \vy $ \OR $\f_j(\vx) = \vy$}
\STATE \textit{// only one model is correct}
\STATE $\Setp \gets{} \Setp \cup \{ \delta_{\vx}^{(i,j)} \}$

\ENDIF
\ENDFOR
\ENDFOR
\ENDFOR

\STATE \textbf{Return} $\Setm, \Setp$

\end{algorithmic}
\end{algorithm}

\subsection{distance over explanations}
\label{ap:mege:distances}

\begin{figure}[h]
  \centering
  \includegraphics[width=0.99\textwidth]{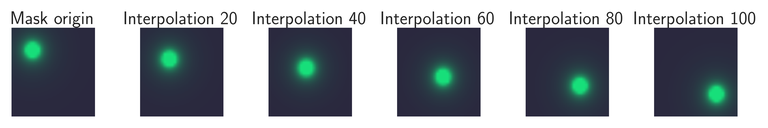}
  \includegraphics[width=0.90\textwidth]{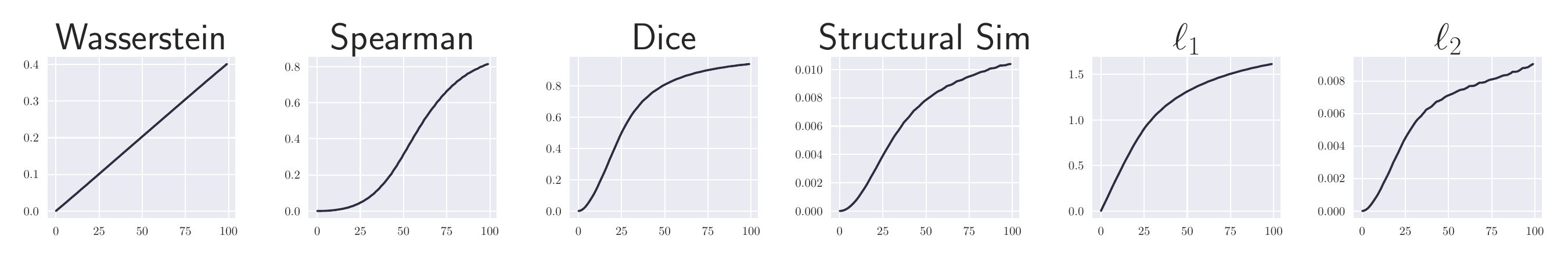}
  \caption{
    Distances with moving interest point. The first line shows the successive interpolations between the baseline image (left), and the target image (right). The second line shows the evolution of the distance between each interpolation and the baseline image.
    }
  \label{dist:move}
\end{figure}

\paragraph{Spatial correlation}
The first test concerns the spatial distance between two areas of interest for an explanation. It is desired that the spatial distance between areas of interest be expressed by the distance used. As a results, two different but spatially close explanations should have a low distance. The test consists in generating several masks representing a point of interest, starting from a left corner of an image of size (32 x 32) and moving towards the right corner by interpolating 100 different masks. The distance between the first image and each interpolation is then measured (see Fig. \ref{dist:move}). 

The different distances evaluated pass this sanity check, i.e. a monotonous growth of the distance, image of the spatial distance of the two points of interest. 

\paragraph{Noise test}

The second test concerns the progressive addition of noise. It is desired that the progressive addition of noise to an original image will affect the distance between the original noise-free image and the noisy image. Formally, with $\vx$ the original image, and $\varepsilon\ \sim\ \mathcal{N}(0, \bm{I}\sigma^2)$ an isotropic Gaussian noise, we wish the distance $d$ to show a monotonic positive correlation $\operatorname{corr}( d(\vx, \vx + \varepsilon), \sigma^2 )$.

In order to validate this prerogative, a Gaussian noise with a progressive intensity $\sigma$ is added to an original image, and the distance between each of the noisy images and the original image is measured. For each value of $\sigma$ the operation is repeated 50 times.

\begin{figure}[h]
  \centering
  \includegraphics[width=0.99\textwidth]{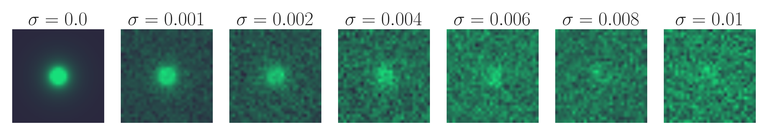}
  \includegraphics[width=0.99\textwidth]{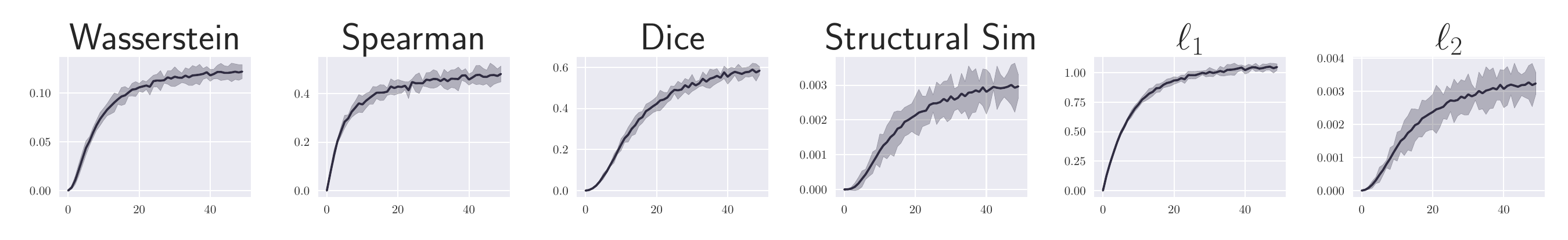}
  \caption{
    Distances with noisy images.
    The first line shows original noise-free image (left) and noisy copies computed by increasing $\sigma$. The second line shows the distances between each noisy image and the baseline image.
    }
  \label{dist:noise}
\end{figure}
Over the different distances tested, they all pass the sanity test : there is a monotonous positive correlation (as seen in Fig. \ref{dist:noise}). Although SSIM and $\ell_2$ have a higher variance.

One will nevertheless note the instability of the Dice score in cases where the areas of interest have a low surface area, as well as a significant computation cost for the Wasserstein distance. For all these reasons, we chose to stay in line with previous work using the absolute value of Spearman rank correlation.

\section{Sobol}
\subsection{Qualitative comparison}

\begin{figure*}[ht]
    \centering
    \includegraphics[width=1.05\linewidth]{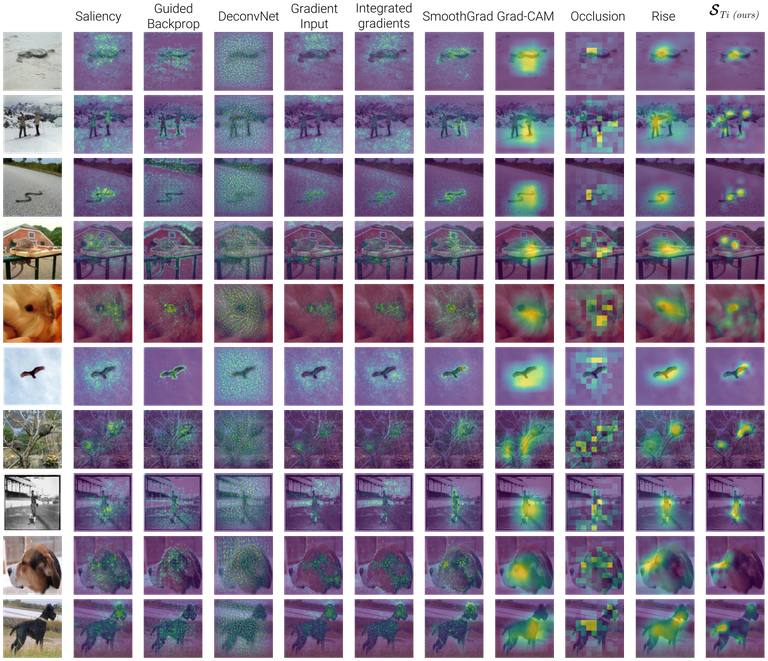}
    \caption{\textbf{Qualitative comparison} with other explainability methods. The heatmaps are normalized and clipped at 2 percentile for Saliency, Guided-Backprop, DeconvNet, Smoothgrad and Integrated-Gradients.
    Explanations are generated from a ResNet50V2.
    }
    \label{app:sobol:qualitative_results}
\end{figure*}

Regarding the visual consistency of our method, Fig.~\ref{app:sobol:qualitative_results} shows a side-by-side comparison between our method and the other methods tested in the Fidelity benchmark. The images are not hand-picked but are the first images from the ImageNet validation set.
To allow better visualization, the gradient-based methods were 2 percentile clipped.
The only black box methods are Occlusion, Rise and $\sob_{T_i}$. We found that $\sob_{T_i}$ consistently provides a sparser map than RISE~\cite{petsiuk2018rise} while being equally consistent.
On the other hand, we found that in general, the gradient-based method provides the sharpest map, but some are prone to failure (fourth row in the Fig.~\ref{app:sobol:qualitative_results}), which is a known problem~\cite{adebayo2018sanity}.

\subsection{Effectiveness of modeling higher-order interactions}

We introduced two approaches, Sobol ($\hat{\sob}_{T_i}$) and Sobol signed ($\hat{\sob}^{\Delta}_{T_i}$), that combine effects of first- and all higher-orders interactions between image regions. For comparison, Occlusion~\cite{zeiler2014visualizing} only accounts for the first order as it removes one region at a time, while RISE~\cite{petsiuk2018rise} accounts for higher-order by removing around 50\% of regions at a time. As seen in Table~\ref{app:sobol:deletion_first_vs_higher}, RISE already surpasses Occlusion on ImageNet in term of Deletion scores, which may indicate that using higher-order information is effective.

To further demonstrates that it is critical to model the higher orders, we evaluate Sobol first-order ($\sob_{i}$) on our Deletion benchmark.
We report that Sobol ($\sob_{T_i}$) reaches lower deletions scores (lower is better) than Sobol first-order ($\sob_{i}$) with 0.121 against 0.170 respectively on ResNet50v2, and similar differences on VGG16, EfficientNet and MobileNetV2.

\begin{table*}[ht]
\centering
\begin{tabular}{lcccc}
\toprule
Method & \textit{ResNet50V2} & \textit{VGG16} & \textit{EfficientNet} & \textit{MobileNetV2} \\
\midrule  
Sobol first-order ($\hat{\sob}_{i}$) & 0.170 & 0.147 & 0.129 & 0.143 \\  
Sobol ($\hat{\sob}_{T_i}$) & \textbf{0.121} & \textbf{0.109} & \textbf{0.104} & \textbf{0.107} \\  
\bottomrule
\end{tabular}
\caption{\textbf{Deletion} scores obtained on 2,000 ImageNet validation set images. Lower is better. 
}\label{app:sobol:deletion_first_vs_higher}
\end{table*}

\subsection{Efficiency of Sobol estimator}\label{app:sobol:efficient}
Regarding the estimation of the Sobol indices, we notice that we can derive a `brute-force' (or often called double-loop method~\cite{sobol2001}) estimator from the definition \ref{def:sobol_indice}:

\begin{equation}
    \label{eq:sobol_double_loop}
    \sob_i = \frac{ \int \big( \int \f(\rvx) \diff \rvx_{\sim i} \big)^2 \diff \rvx_i -  ( \int \f(\rvx) \diff \rvx)^2 }
             {\int \f(\rvx)^2 \diff \rvx - ( \int \f(\rvx) \diff \rvx )^2 }
\end{equation}

However, one the main problems with this estimator is the cost of computation, which can be too heavy, especially with complex models such as large neural networks. This difficulty is particularly true for the calculation of total Sobol indices. 

\begin{figure*}[ht]
    \centering
    \includegraphics[width=0.80\linewidth]{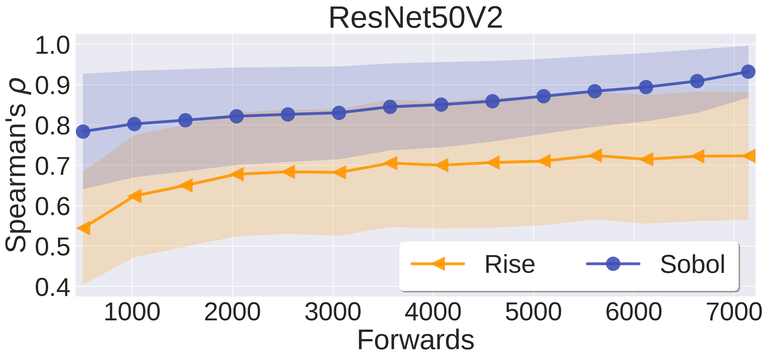}
    \caption{
      Spearman rank correlation of explanations as a function of the number of forwards, compared to an explanation generated with $10,000$ forwards.
      The model used is a ResNet50V2.
    }
    \label{app:sobol:conv:resnet}
\end{figure*}

\begin{figure*}[ht]
    \centering
    \includegraphics[width=0.80\linewidth]{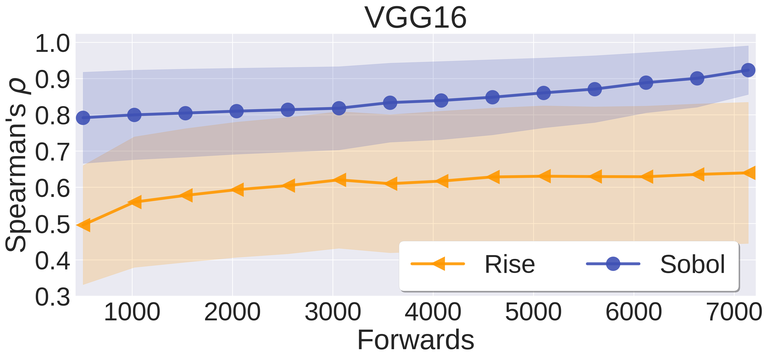}
    \caption{      
    Spearman rank correlation of explanations as a function of the number of forwards, compared to an explanation generated with $1,0000$ forwards.
    The model used is a VGG16.
      }
      \label{app:sobol:conv:vgg}
\end{figure*}

\begin{figure*}[ht]
    \centering
    \includegraphics[width=0.80\linewidth]{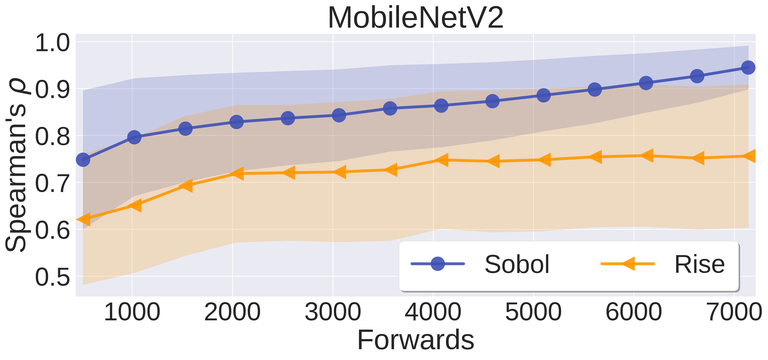}
    \caption{
      Spearman rank correlation of explanations as a function of the number of forwards, compared to an explanation generated with $1,0000$ forwards.
      The model used is a MobileNetV2.
    }
    \label{app:sobol:conv:mobilenet}
\end{figure*}

Since the perturbation masks are used to approximate these integrals, an efficient way to proceed is to generate those masks from a low discrepancy sequences, also called Quasi-random sequences. These sequences allow to efficiently integrate functions on the hypercube $[0, 1]^d$. In fact, they have a faster convergence rate compared to ordinary Monte Carlo methods~\cite{gerber2015} (with $\pred$ sufficiently regular). This difference being due to the use of a deterministic sequence that covers $[0, 1]^d$ more uniformly.
In our experiments we used Sobol sequences~\cite{sobol1967sequence}, we refer the readers to~\cite{leobacher2014introduction} for more informations.
The efficiency of the estimator and the sampling is shown on Figures \ref{app:sobol:conv:resnet}, \ref{app:sobol:conv:vgg} and \ref{app:sobol:conv:mobilenet} where our estimator consistently converges faster than RISE~\cite{petsiuk2018rise}.


We also perform an ablation study of the number of forwards on the Deletion benchmark. In Table~\ref{app:sobol:deletion_first_vs_higher}, we show that competitive scores can be obtained with lower number of forwards such as 0.151 in Deletion score with 492 forwards instead of 0.121 with 3936 forwards which is our default number of forwards.

\begin{table*}[ht]
\centering
\begin{tabular}{cc}
\toprule
Number of samples & Deletion scores \\
\midrule  
492 & 0.151 \\
984 & 0.140 \\
1476 & 0.132 \\
1968 & 0.123 \\
2460 & 0.121 \\
2952 & 0.120 \\
3444 & 0.120 \\
3936 & 0.121 \\
\bottomrule
\end{tabular}
\caption{\textbf{Deletion} scores averaged over 2,000 images of ImageNet validation set using ResNet50V2 and Sobol ($\hat{\sob}_{T_i}$). Lower is better. 
}\label{app:sobol:deletion_ablation}
\end{table*}

\subsection{Sanity check}

\begin{figure*}[ht]
    \centering
    \includegraphics[width=0.95\linewidth]{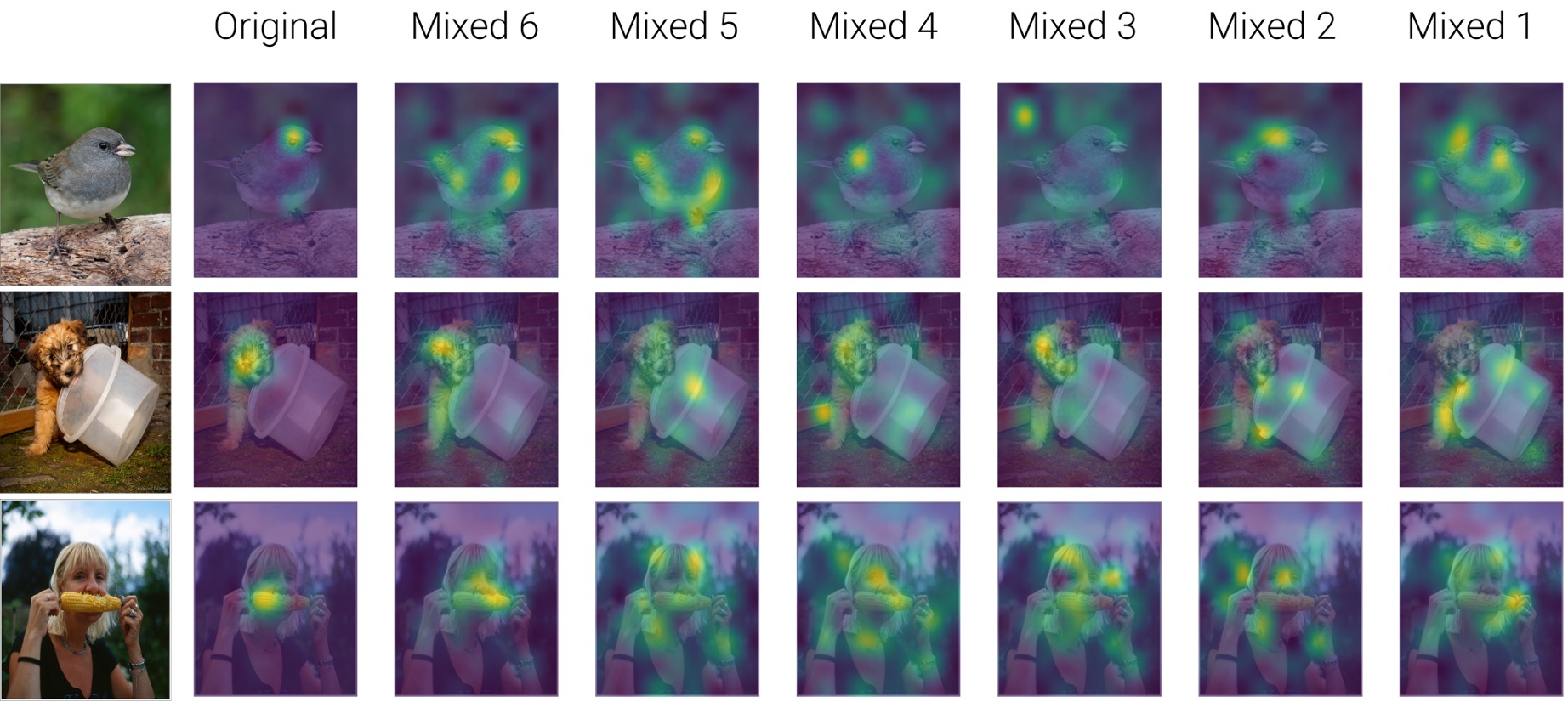}
    \caption{\textbf{Sanity Check} model weights are progressively reinitialized from Mixed 6 to Mixed 1 in InceptionV3~\cite{szegedy2015going}, demonstrating our method’s sensitivity to model weights.}
    \label{sobol:app:sanitycheck}
\end{figure*}

We followed the procedure used by~\cite{adebayo2018sanity}, namely the progressive reset of the network weights. We used an Inception V3~\cite{szegedy2015going} model, each images shows the $\sob_{T_i}$ explanation for the network in which the upper layers (from logits) were reset. 
Fig.~\ref{sobol:app:sanitycheck} shows that our method passes the sanity check: it turns out to be sensitive to the modification of the model weights.

\subsection{Word Deletion}
\label{sobol:app:word_deletion}

For the bidirectional LSTM, the word embedding is in $\mathbb{R}^{300}$ and is initialized with the pre-trained GloVe embedding. The layer has a hidden size of $64$ (bidirectional architectures: $32$ dimensions per direction). The resulting document representation is projected to $64$ dimensions then $2$ dimensions using fully connected layers, followed by a softmax and reached an accuracy of $89\%$ on the test dataset.

For the BERT-based models, we use the Transformers library from HuggingFace~\cite{wolf2020transformers}
and more specifically the bert-base-uncased model.
The final layer is tuned to minimize cross-entropy,
 with Adam optimizer~\cite{kingma2014adam}
and initial learning rate of $1e^{-3}$
to reach an accuracy of $92$\% on the test dataset.

The observation that local perturbation: with the majority of words present, gets a better score is verified by playing on the threshold of the perturbation function. By decreasing the percentage of words removed on average we observe that a better deletion score is obtained.

\begin{table}[ht]
\centering
\begin{tabular}{l cccc}
\toprule
 & $\hat{\sob}_{T_i}\Delta$ $50$\%  & $\hat{\sob}_{T_i}^\Delta$ $90$\% & $\hat{\sob}_{T_i}\Delta$ $95$\% & Occlusion \\
\midrule
Deletion & 0.598 & 0.553 & \textbf{0.527} &  \underline{0.531} \\
\bottomrule
\end{tabular}

\caption{\textbf{Word deletion scores} on the Bert based model when the perturbation threshold is modified to control the average presence of words in each generated perturbated input. Lower is better. 
}\label{app:sobol:word_deletion_bis}

\end{table}

\section{EVA}
\subsection{Qualitative comparison}

\begin{figure*}[!ht]
  \centering
  \includegraphics[width=0.99\textwidth]{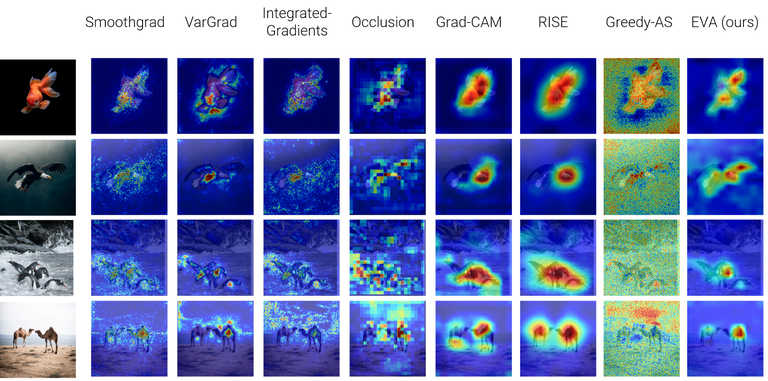}
  \caption{\textbf{Qualitative comparison} with other attribution methods. To allow for better visualization, the gradient-based methods (Saliency, Gradient-Input, SmoothGrad, Integrated-Gradient, VarGrad) are clipped at the 2nd percentile. For more results and details on each method and choice of hyperparameters, see Appendix.
  }
  \vspace{-2mm}
  \label{fig:eva:imagenet_explanations}
\end{figure*}

Regarding the visual consistency of our method, Figure~\ref{fig:eva:imagenet_explanations} shows a side-by-side comparison between our method and the attribution methods  tested in our benchmark. 
To allow better visualization, the gradient-based
methods were 2 percentile clipped.

\subsection{Ablation studies}
\label{ap:eva:benchmarks}

\begin{table*}[t]
    \centering
    \scalebox{0.75}{
        \begin{tabular}{l C{0mm} P{0mm} P{5mm}P{5mm}P{5mm}P{5mm}P{5mm} P{1mm} P{5mm}P{5mm}P{9mm}P{5mm}P{5mm} P{1mm} P{5mm}P{5mm}P{5mm}P{5mm}P{5mm}}
        \toprule
        &&& \multicolumn{5}{c}{MNIST} &&  \multicolumn{5}{c}{Cifar-10} &&  \multicolumn{5}{c}{ImageNet}  \\
        \cmidrule(lr){4-8} \cmidrule(lr){10-14} \cmidrule(lr){16-20}
        
        &&& Del.$\downarrow$ & Ins.$\uparrow$ & Fid.$\uparrow$ & Rob.$\downarrow$ & Time 
        && Del.$\downarrow$ & Ins.$\uparrow$ & Fid.$\uparrow$ & Rob.$\downarrow$ & Time 
        && Del.$\downarrow$ & Ins.$\uparrow$ & Fid.$\uparrow$ & Rob.$\downarrow$ & Time
        \\
        
        \midrule
        Greedy-AS &   && .260  & .497  & .110  & \textbf{.061}  & 335
                 && .205 & .264 & -.003 & \textbf{.013} & 4618
                 && .088 & .047 & .023 & \textbf{.612} & 180056  \\
        \midrule 
        Greedy-AO &&& .237 & .572 & .244 & \underline{.063} & 290
                 && \textbf{.162} & .283 & .041 & .024 & 2874
                 && .086 & .050 & .023 & \underline{.752} & 26762 \\
        \evaEmp  &&& \underline{.101} & .621 & \underline{.378} & .067 & 14.4 
                 && .184 & .270 & \textbf{.397} & \underline{.022} & 186.6
                 && \multirow{ 2}{*}{.070} & \multirow{ 2}{*}{\textbf{.289}} & \multirow{ 2}{*}{.048} & \multirow{ 2}{*}{.758} & \multirow{ 2}{*}{6454} \\ 
        \textbf{\eva} ~(ours)     & && \textbf{.089} & \textbf{.736} & \textbf{.428} & \underline{.069} & 1.29 
                 && \textbf{.164} & \underline{.290} & \textbf{.352} & \underline{.025} & 12.7 
                 \\
        
        \bottomrule \\
        \end{tabular}
    }
    \caption{
Results on Deletion (Del.), Insertion (Ins.), $\mu$Fidelity (Fid.) and \rsr~ (Rob.) metrics. 
Time in seconds corresponds to the generation of 100 explanations on an Nvidia P100.
Note that \eva~is the only method with guarantees that the entire set of possible perturbations has been exhaustively searched.
Verified perturbation analysis with IBP + Forward + Backward is used for MNIST, with Forward only for  CIFAR-10 and with our hybrid strategy described in Section.\ref{sec:eva:scaling} for ImageNet. 
Grad-CAM and Grad-CAM++ are not calculated on the MNIST dataset since the network  only has dense layers. 
Greedy-AO is the equivalent of Greedy-AS but with the \AO estimator. 
The first and second best results are  in \textbf{bold} and \underline{underlined}, respectively. 
}
    \label{tab:eva:ablation_ao}
    \vspace{-3mm}
\end{table*}

For a more thorough understanding of the impact of the different components that made EVA - the adversarial overlap and the use of verification tools- we proposed different ablation versions of EVA which are the following:
(\textbf{\textit{i}}) Empirical EVA, (\textbf{\textit{ii}}) GreedyAO which is the equivalent of Greedy-AS but with the $\AO$ estimator. This allow us to perform ablation on the proposed $\AO$~estimator. Results can be found in Table~\ref{tab:eva:ablation_ao}.
\subsection{Empirical EVA.}
    
In this section, we describe the ablation consisting in estimating \eva~ without any use of verified perturbation analysis -- thus without any guarantees.

A first intuitive approach would be to replace verification perturbation analysis with adversarial attacks (as used in \textit{Greedy-AS}~\cite{hsieh2020evaluations}); we denote this approach as \textit{Greedy-AO}. 
In addition, we go further with a purely statistical approach based on a uniform sampling of the domain; we denote this approach \evaEmp. 
    
This estimator proves to be a very good alternative in terms of computation time but also with respect to the considered metrics as shown in Section ~\ref{sec:eva:experiments}. Unfortunately the lack of guarantee makes it not as relevant as \eva.
Formally, it consists in directly estimating empirically \AO  using $N$ randomly sampled perturbations.
    
\begin{equation}\label{eq:eva:adv_empirique}
    \AOemp(\vx, \ball) = 
    \max_{\substack{\v{\delta}_1,\cdots \v{\delta}_i,\cdots \v{\delta}_N  \overset{\mathrm{iid}}{\sim} U(\ball)\\c'\neq{}c}} \pred_{c'}(\vx + \v{\delta}_i) - \pred_c(\vx + \v{\delta}_i).
\end{equation}
    
We then denote accordingly \evaEmp which uses $\AOemp$:

\begin{equation}
    \label{eq:eva:tod_estimator_emp}
    \evaEmp(\vx, \v{u}, \ball) = \AOemp(\vx, \ball) - \AOemp(\vx, \ballu)
\end{equation}

\subsection{\eva~and Robustness-Sr}

We show here that the explanations generated by \eva~  provide an optimal solution from a certain stage to the $\rsr$ metric proposed by~\cite{hsieh2020evaluations}. We admit a unique closest adversarial perturbation $\v{\delta}^* = \min ||\v{\delta}||_p : \f(\vx + \v{\delta}) \neq \f(\vx)$, and we define $\varepsilon$, the radius of $\ball$ as $\varepsilon = ||\v{\delta}||_p$. 
Note that $||\v{\delta}||_p$ can be obtained by binary search using the verified perturbation analysis method.

We briefly recall the $\rsr$ metric. With $\vx = (x_1, ..., x_d)$, the set $\mathcal{U} = \{1, ..., d\}$, $\v{u}$ a subset of $\mathcal{U}$ : $\v{u} \subseteq \mathcal{U}$ and $\overline{\v{u}}$ its complementary. Moreover, we denote the minimum distance to an adversarial example $\varepsilon^*_{\v{u}}$: 
$$ \varepsilon^*_{\v{u}} = \big\{ \min || \v{\delta} ||_p ~:~ \f(\vx + \v{\delta}) \neq \f(\vx), \v{\delta}_{\overline{\v{u}}} = 0  \big\} $$ 

The $\rsr$ score is the AUC of the curve formed by the points $\{ (1,  \varepsilon^{*}_{(1)}), ..., (d,  \varepsilon^{*}_{(d)})  \}$ where $\varepsilon^{*}_{(k)}$ is the minimum distance to an adversarial example for the $k$ most important variables.
From this, we can deduce that $||\v{\delta}^*|| \leq \varepsilon^*_{\v{u}}$, $\forall \v{u} \subseteq \{1, ..., d\}$.

The goal here is to minimize this score, which means for a number of variables $|\v{u}| = k$, finding the set of variables $\v{u}^*$ such that $\varepsilon^*_{\v{u}}$ is minimal. We call this set the \textit{optimal set at $k$}. 

\begin{definition}
The \textit{optimal set at $k$} is the set of variables $\v{u}^{*}_k $ such that 
$$ \v{u}^{*}_k = \underset{ \v{u} \subseteq \mathcal{U},~ |\v{u}| = k}{\argmin ~~ \varepsilon^*_{\v{u}} }. $$
\end{definition}

We note that finding the minimum cardinal of a variable to guarantee a decision is also a standard research problem  ~\cite{ignatiev2019abduction, ignatiev2019relating} and is called subset-minimal explanations. 

Intuitively, the optimal set is the combination of variables that allows finding the closest adversarial example.
Thus, minimizing $\rsr$ means finding the optimal set $\v{u}^*$ for each $k$. 
Note that this set can vary drastically from one step to another, it is therefore potentially impossible for attribution to satisfy this optimality criterion at each step.
Nevertheless, an optimal set that is always reached at some step is the one allowing to build $\v{\delta}^*$.
We start by defining the notion of an essential variable before showing the optimality of $\v{\delta}^*$.

\begin{definition}
Given an adversarial perturbation $\v{\delta}$, we call \textit{essentials variables} $\v{u}$ all variables such that $|\v{\delta}_{i}| > 0, i \in \v{u}$. Conversely, we call \textit{inessentials variables} variables that are not essential.
\end{definition}

For example, if $\v{\delta}^*$ has $k$ \textit{essential variables}, it is reachable by modifying only $k$ variables. 
This allows us to characterize the optimal set at step $k$.

\begin{proposition} 
\label{prop:eva:uoptimal}
Let $\v{u}$ be the set of essential variables of $\v{\delta}^*$, then $\v{u}$ is an optimal set for $k$, with $k \in [\![|\v{u}|,d]\!] $.
\end{proposition}

\begin{proof}
Let $\v{v}$ be a set such that $ \varepsilon^*_{\v{v}} < \varepsilon^*_{\v{u}} $, then $ \varepsilon^*_{\v{v}} < || \v{\delta}^* || $ which is a contradiction.
\end{proof}

Specifically, as soon as we have the variables allowing us to build $\v{\delta}^*$, then we reach the minimum possible for $\rsr$.
We will now show that \eva~allows us to reach this in $|\v{u}|$ steps, with $|\v{u}| \leq d$ by showing (1) that $\v{\delta}^*$ \textit{essential variables} obtain a positive attribution and (2) that $\v{\delta}^*$ \textit{inessential variables} obtain a zero attribution.

\begin{proposition}
\label{prop:eva:ess}
All essential variables \(\v{u}\) w.r.t \(\v{\delta}^*\) have a strictly positive importance score \(\eva(\v{u}) > 0\). 
\end{proposition}

\begin{proof}
Let us assume that $i$ is \textit{essential} and $\eva(i) = 0$, then $\bm{F}(\ball) = \bm{F}(\ball_i)$ which implies
$$
\max_{\substack{\v{\delta} \in \ball\\c'\neq{}c}} \f_{c'}(\vx + \v{\delta}) - \f_c(\vx + \v{\delta}) = 
\max_{\substack{\v{\delta}' \in \ball_i\\c'\neq{}c}} \f_{c'}(\vx + \v{\delta}') - \f_c(\vx + \v{\delta}')
$$
by uniqueness of the adversarial perturbation, $ \v{\delta} = \v{\delta}' $ which is a contradiction as $\v{\delta}' \notin \ball_i$ since $\v{\delta}'_i \neq 0$ by definition of an \textit{essential variable}. Thus $x_i$ cannot be \textit{essential}, which is a contradiction.
\end{proof}

Essentially, if the variable $i$ is necessary to reach $\v{\delta}^*$, then removing it prevents the adversarial example from being reached and lowers the \adv, giving a strictly positive attribution.

\begin{proposition}
\label{prop:eva:iness}
All inessential variables \(\v{v}\) w.r.t. \(\v{\delta}^*\) have a zero importance score \(\eva(\v{v}) = 0\). 
\end{proposition}

\begin{proof}
With $i$ being an \textit{inessential} variable, then $\v{\delta}^*_i = 0$. It follow that $\v{\delta}^* \in \ball_i \subseteq \ball$. Thus
\begin{align*} 
\bm{F}(\ball) &= \max_{\substack{\v{\delta} \in \ball\\c'\neq{}c}} \f_{c'}(\vx + \v{\delta}) - \f_c(\vx + \v{\delta}) \\
              &= \f_{c'}(\vx + \v{\delta}^*) - \f_c(\vx + \v{\delta}^*)
\end{align*}
as $\v{\delta}^*$ is the unique adversarial perturbation in $\ball$, similarly 
\begin{align*} 
\bm{F}(\ball_i) &= \max_{\substack{\v{\delta}' \in \ball\\c'\neq{}c}} \f_{c'}(\vx + \v{\delta}') - \f_c(\vx + \v{\delta}') \\
              &= \f_{c'}(\vx + \v{\delta}^*) - \f_c(\vx + \v{\delta}^*)
\end{align*}
thus $\bm{F}(\ball) = \bm{F}(\ball_i)$ and $\eva(i) = 0$.
\end{proof}

Finally, since \eva~ranks the \textit{essential variables} of $\v{\delta}^*$ before the \textit{inessential variables}, and since $\v{\delta}^*$ is the \textit{optimal set} from the step $|\v{u}|$ to the last one $d$, then \eva~provide the \textit{optimal set}, at least from the step $|\v{u}|$.

\begin{theorem}\textbf{\eva~provide the optimal set from step $|\v{u}|$ to the last step.}
\label{app:eva:rsr}
With $\v{u}$ the essential variables of $\v{\delta}^*$, \eva~will rank the $\v{u}$ variables first and provide the optimal set from the step $|\v{u}|$ to the last step. 
\end{theorem}

\begin{proof}
Let $\v{u}$ denote the \textit{essential variables} of $\v{\delta}^*$ and $\v{v}$ the \textit{inessential variables}. Then according to Proposition~\ref{prop:eva:ess} and Proposition~\ref{prop:eva:iness}, $\forall i \in \v{u}, \forall j \in \v{v}: \eva(i) > \eva(j)$. It follow that $\v{u}$ are the most important variables at step $|\v{u}|$. Finally, according to Proposition~\ref{prop:eva:uoptimal}, $\v{u}$ is the optimal set for $k$, with $k \in [\![|\v{u}|,d]\!]$.
\end{proof}

\begin{figure}[ht]
  \centering
  \includegraphics[width=0.8\textwidth]{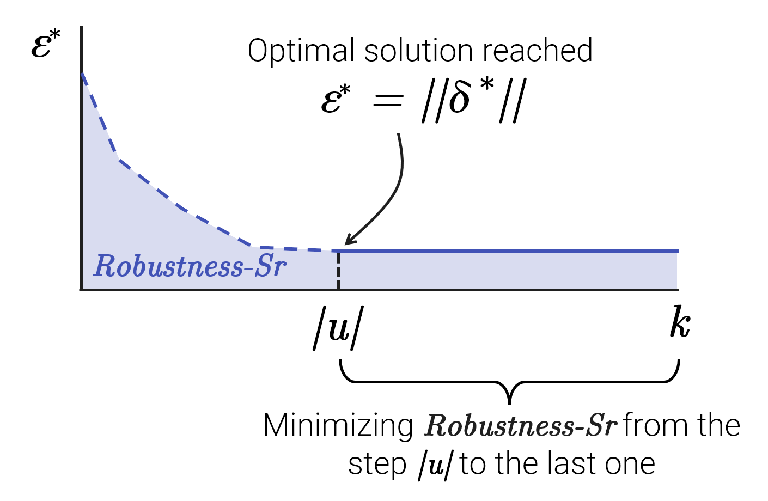}
  \caption{\textbf{\eva~yield optimal subset of variable from step $|\v{u}|$.} $\rsr$ measures the AUC of the distances to the nearest adversary for the $k$ most important variables. With $\v{\delta}^*$ the nearest reachable adversarial perturbation around $\vx$, then \eva~yield the optimal set -- the variables allowing to reach the nearest adversarial example for a given cardinality -- at least from $||\v{u}|| \leq d$ step to the last one, $\v{u}$ being the so-called essential variables.
  }
  \label{fig:eva:eva_optimal}
\end{figure}

\subsection{\eva~and \textit{Stability}}

Stability is one of the most crucial properties of an explanation. Several metrics have been proposed~\cite{aggregating2020,yeh2019infidelity} and the most common one consists in finding around a point $\vx$, another point $\v{z}$ (in a radius $\radiusbis$) such that the explanation changes the most according to a given distance between explanation $d$ and a distance over the inputs $\rho$:

$$
\textit{Stability}(\vx, \explainer) = \max_{\v{z} : \rho(\v{z}, \vx) \leq \radius} d( \explainer(\vx), \explainer(\v{z}) ) 
$$

and $\explainer$ an explanation functional.
It can be shown that the proposed ~\eva~ estimator is bounded by the stability of the model as well as by the radii $\radius$ and $\radiusbis$, $\radius$ being the radius of $\ball$ and $\radiusbis$ the radius of stability.
From here, we assume $d$ and $\rho$ are the $\ell_2$ distance.

Let assume that $\f$ is $L$-lipschitz. We recall that a function $\f$ is said $L$-lipschitz over $\mathcal{X}$ if and only if $\forall (\vx, \v{z}) \in \mathcal{X}^2, || \f(\vx) - \f(\v{z}) || \leq L || \vx - \v{z} ||$.

\begin{theorem}\textbf{\eva~ has bounded Stability}
\label{app:eva:stab}
Given a $L$-lipschitz predictor $\f$, $\radius$ the radius of $\ball$ and $\radiusbis$ the Stability radius, then
$$
\textit{Stability}(\vx, \eva) \leq 4L(\radius + \radiusbis)
$$
\end{theorem}

\begin{proof}
With $c' \neq c$ we denote the so-called \textit{margin} $\vm(\vx) = \f_{c'}(\vx) - \f_{c}(\vx)$. We note that by additivity of the Lipschitz constant $\vm$ is 2$L$-Lipschitz.
\begin{align*} 
&\textit{Stability}(\vx, \eva) = \max_{\v{z} : \rho(\v{z}, \vx) \leq \radiusbis} || \eva(\vx), \eva(\v{z}) || \\
  &= \max_{\v{z} : \rho(\v{z}, \vx) \leq \radiusbis} 
  ||\max_{\v{\delta}} \vm(\vx + \v{\delta}) 
  - \max_{\v{\delta}_{\v{u}}} \vm(\vx + \v{\delta}_{\v{u}}) \\
  &~~~~- \max_{\v{\delta}} \vm(\v{z} + \v{\delta})
  + \max_{\v{\delta}_{\v{u}}} \vm(\v{z} + \v{\delta}_{\v{u}}) || \\
  &\leq \max_{\v{z} : \rho(\v{z}, \vx) \leq \radiusbis} 
  ||\max_{\v{\delta}} \vm(\vx + \v{\delta}) 
  - \max_{\v{\delta}} \vm(\v{z} + \v{\delta}) || \\
  &~~~~+ || \max_{\v{\delta}_{\v{u}}} \vm(\v{z} + \v{\delta}_{\v{u}}) 
  - \max_{\v{\delta}_{\v{u}}} \vm(\vx + \v{\delta}_{\v{u}}) || \\
  &= \max_{\v{\xi} : ||\v{\xi}|| \leq \radiusbis} 
  ||\max_{\v{\delta}} \vm(\vx + \v{\delta}) 
  - \max_{\v{\delta}} \vm(\vx + \v{\delta} + \v{\xi}) || \\
  &~~~~+ || \max_{\v{\delta}_{\v{u}}} \vm(\vx + \v{\delta}_{\v{u}} + \v{\xi}) 
  - \max_{\v{\delta}_{\v{u}}} \vm(\vx + \v{\delta}_{\v{u}}) || \\
  &\leq 2L (||\v{\delta}|| + ||\v{\xi}||) + 2L (||\v{\delta}|| + ||\v{\xi}||)\\
  &= 4L (\radius + \radiusbis)
\end{align*}

\end{proof}

\subsection{Targeted explanations}
\label{ap:eva:targeted}

\begin{figure}[t!]
  \centering
  \includegraphics[width=0.9\textwidth]{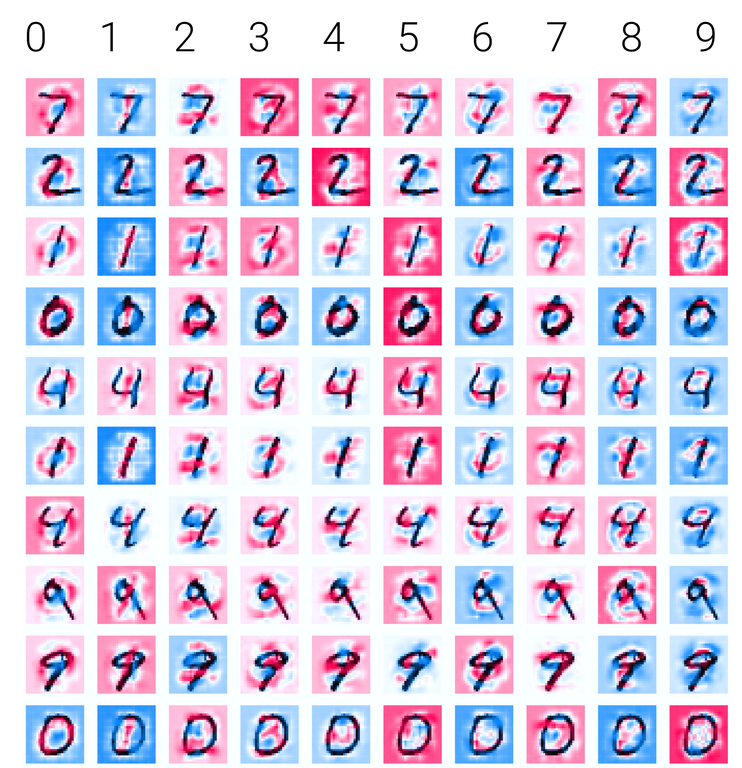}
  \caption{\textbf{Targeted Explanations} Attribution-generated explanations for a decision other than the one predicted. Each column represents the class explained, e.g., the first column looks for an explanation for the class `0' for each of the samples. As indicated in section~\ref{sec:eva:targeted_explanations}, the red areas indicate that a black line should be added and the blue areas that it should be removed. More examples are available in the Appendix.
  }
  \label{fig:eva:ap_targeted}
\end{figure}

In order to generate targeted explanations, we split the calls to $\eva(\cdot, \cdot)$ in two: the first one with `positive' perturbations from $\ball^{(+)}$ (only positive noise), a call with `negative' perturbations from $\ball^{(-)}$ (only negative-valued noise) as defined in Section~\ref{sec:eva:targeted_explanations}. 

We then get two explanations, one for positive noise
$\explanation^{(+)}_{\v{u}} = \bm{F}_c(\ball^{(+)}(\vx)) - \bm{F}_c(\ball^{(+)}_{\v{u}}(\vx))$, the other for negative noise $\explanation^{(-)}_{\v{u}} = \bm{F}_c(\ball^{(-)}(\vx)) - \bm{F}_c(\ball^{(-)}_{\v{u}}(\vx))$. Intuitively, high importance for $\explanation^{(+)}_{\v{u}}$  means that the model is sensitive to the addition of a white line. Conversely, high importance for $\explanation^{(-)}_{\v{u}}$  means that removing it changes the decision model. These two explanations being opposed, we construct the final explanation as $\explanation_{\v{u}} = \explanation^{(+)}_{\v{u}} - \explanation^{(-)}_{\v{u}}$. More examples of results are given in Fig.~\ref{fig:eva:ap_targeted}.

\section{Metapred}

\section{Human experiments}
\label{ap:protocole}
\subsection{Experimental design}
\begin{figure*}[ht]
    \centering
    \includegraphics[width=0.85\textwidth]{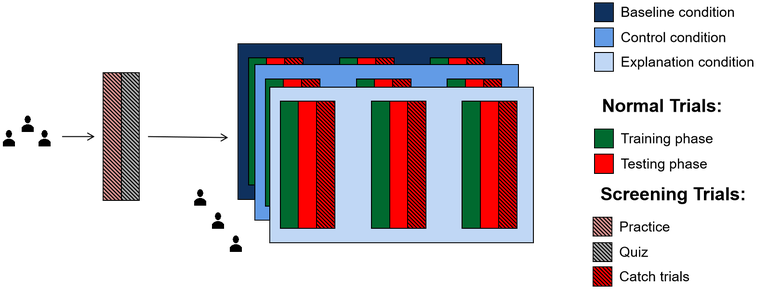}
    \caption{\textbf{Experimental design.} 
    First, every participant goes through a practice session (fig \ref{fig:practice}) to make sure they understand how to use attribution methods to infer the rules used by a model, and a quiz (fig \ref{fig:quiz}) to make sure they actually read and understand the instructions. Then, participants are split into the different conditions -- every participant will only go through one condition. The 3 possible conditions are: an Explanation condition where an explanation is provided to human participants during their training phase, a Baseline condition where no explanation was provided to the human participants, and a Control condition where a non-informative explanation was provided.
    The main experiment was divided into 3 training sessions each followed by a brief test. In each individual training trial, an image was presented with the associated prediction of the model, either alone for the baseline condition or together with an explanation for the experimental and control condition. After a brief training phase (5 samples), participants' ability to predict the classifier's output was evaluated on 7 new samples (only the image, no explanation) during a test phase. To filter out uncooperative participants we also add a catch trial (fig \ref{fig:catch}) in each test session.}
    \label{fig:design}
\end{figure*}

Figure \ref{fig:design} summarizes the experimental design used for our experiments. The participants that went through our experiments are users from the online platform Amazon Mechanical Turk (AMT). Through this platform, users stay anonymous, hence, we do not collect any sensitive personal information about them. We prioritized users with a Master qualification (which is a qualification attributed by AMT to users who have proven to be of excellent quality) or normal users with high qualifications (number of HIT completed $=10 000$ and HIT accepted $> 98 \%$). 

Before going through the experiment, participants are asked to read and agree to a consent form, which specifies: the objective and procedure of the experiment, as well as the time expected to completion ($\sim 5$ - $8$ min) with the reward associated ($\$1.4$), and finally, the risk, benefits, and confidentiality of taking part in this study. 
There are no anticipated risks and no direct benefits for the participants taking part in this study.

\paragraph{Controlling for prior class knowledge}

To control for users' own semantic knowledge, we balanced the samples shown to participants so that the classifiers were correct/incorrect 50\% of the time. This way, the baseline (participants who try to simply predict the true class label of an image as opposed to learning to predict the model's outputs) is at 50\%. Any higher score reflects a certain understanding of the rules used by the model.

\subsection{Pruning out uncooperative participants}

\paragraph{3-stage screening proccess.}

\begin{figure*}
    \centering
    \includegraphics[width=0.65\textwidth]{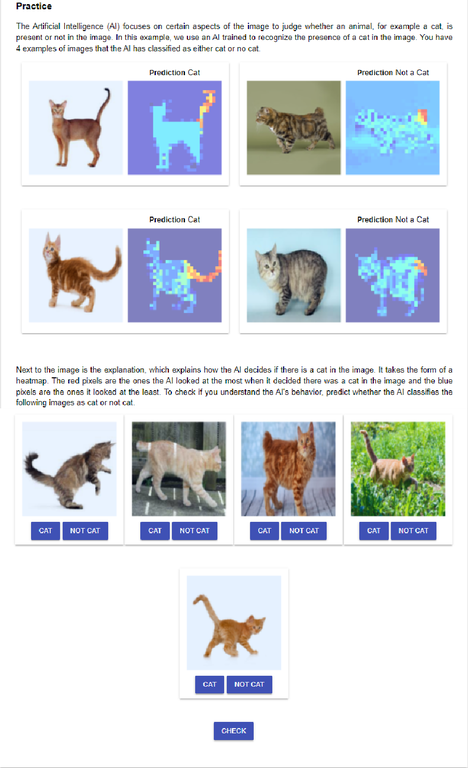}
    \caption{\textbf{Practice session.} Through a practice session, which is a simplified version of the main experiment, we evaluate if users understand how to read and use explanations. Participants that failed to predict correctly any of the 5 cat test images on the first try were excluded from further analysis.}
    \label{fig:practice}
\end{figure*}

\begin{figure*}
    \centering
    \includegraphics[width=0.7\textwidth]{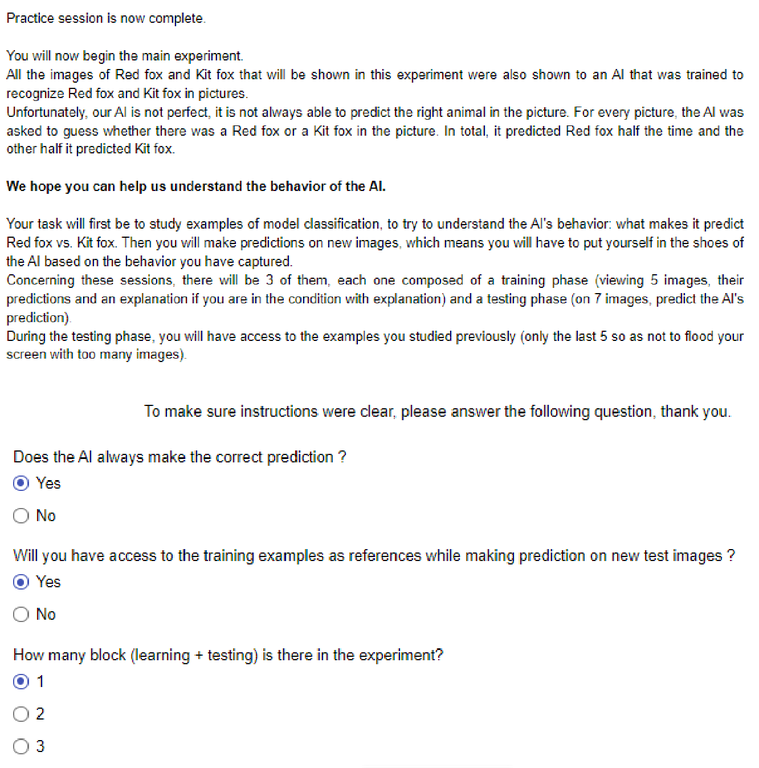}
    \caption{\textbf{Quiz.} Through a quiz, we make sure that users read and understood the instructions. Participants that did not answer correctly every question on the first try were excluded from further analysis.}
    \label{fig:quiz}
\end{figure*}

\begin{figure*}
    \centering
    \includegraphics[width=0.7\textwidth]{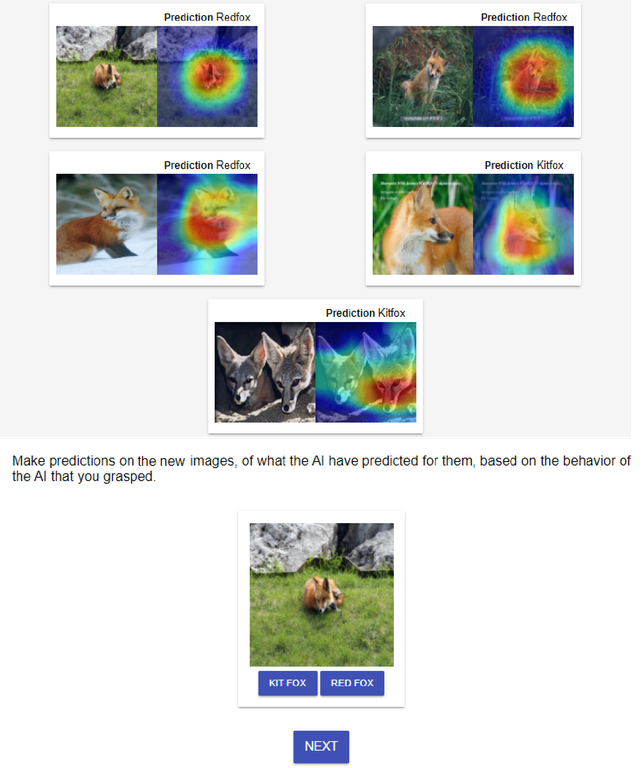}
    \caption{\textbf{Catch trial.} We use a reservoir (to store all the examples of the current training session) that participants can refer to during the testing phase to minimize memory load. At the top of the screen is the reservoir, at the bottom of the screen is a trial from the testing phase. We take advantage of the reservoir to introduce a catch trial. We added a trial in the testing phase of each session where the input image corresponded to one of the training samples used in the current session: since the answer is still on the screen (or a scroll away) we expect participants to be correct on these catch trials. Participants that failed any of the 3 catch trials (one per session) were excluded from further analysis.}
    \label{fig:catch}
\end{figure*}

To prune out uncooperative participants, we subjected them to a 3-stage screening process. First, participants completed a short practice session to make sure they understood the task and how to use the attribution methods to infer the rules used by the model (fig \ref{fig:practice}). 
Second, we asked participants to answer a few questions regarding the instructions provided to make sure they actually read and understood them (fig \ref{fig:quiz}). 
Third, during the main experiment, we took advantage of the reservoir to introduce a catch trial (fig \ref{fig:catch}). The reservoir is the place where we store the training example of the current session, which can be accessed during the testing phase. We added a trial in the testing phase of each session where the input image corresponded to one of the training samples used in the current session: since the answer is still on the screen (or a scroll away) we expect participants to be correct on these catch trials. Participants that failed any of the 3 screening processes were excluded from further analysis.

\clearpage

\subsection{More results}
\paragraph{Reaction time.}
We explored whether the usefulness of a method is reflected in the reaction time of participants -i.e., the more useful the explanation the faster the participants are able to grasp the strategy of the model-. Table \ref{tab:time} shows the reaction time of participants across methods, across datasets. We do not find any trend linking reaction time with usefulness.
\begin{table}[h]
\vspace{2mm}
\centering
\begin{tabular}{lccc}
\toprule
 Method & \textit{Husky vs. Wolf} & \textit{Leaves} & \textit{ImageNet} \\
\midrule
Saliency~\cite{simonyan2014deep}                  & \underline{207.7} & \textbf{212.9} & 202.3 \\
Integ.-Grad.~\cite{sundararajan2017axiomatic}     & 213.1 & 216.5 & 218.5 \\
SmoothGrad~\cite{smilkov2017smoothgrad}           & 215.8 & \textbf{268.8} & 243.9 \\
GradCAM~\cite{selvaraju2017gradcam}               & \textbf{168.9} & 154.6 & 268.9\\
Occlusion~\cite{zeiler2014visualizing}            & 221.2 & 229.2 & 274.4 \\
Grad.-Input~\cite{shrikumar2017learning}               & \underline{210.4} & \underline{238.1} & 208.0 \\
\bottomrule
\end{tabular}
\vspace{2mm}
\caption{\textbf{ Average total \textit{time} per method per dataset (in second).} For each dataset, we \textbf{bold} the most useful method, and we \underline{underline} the least useful method.}
\label{tab:time}
\vspace{-2mm}
\end{table}
\clearpage

\clearpage

\section{Why do the best methods for the use cases Bias detection and Identifying an expert strategy (leaves) differ?}
\label{ap:differ}

The most interesting case is Saliency, which is the worst method on the bias dataset but the best on the “leaves” dataset. On the bias dataset, the model seems to focus on the background (i.e., a coarse feature), and on the “leaves'' dataset the model seems to focus either on the margin or on the vein of the leaf (i.e., very fine features). We hypothesize that different methods suit different granularity of features (coarse vs fine). \cite{smilkov2017smoothgrad}~make the hypothesis that “the saliency maps are faithful descriptions of what the network is doing” but because “the derivative of the score function with respect to the input [is] not [...] continuously differentiable”, the saliency map can appear noisy. Because of this local discontinuity of the gradient, a large patch of important pixels is often portrayed in the saliency map as a collection of smaller patches of important pixels (i.e., a coarse feature vs multiple individual fine features) which can make it hard to identify if the strategy is the coarse feature or a more complex interaction of the smaller features. In the bias dataset, because the model relies on the background, the Saliency maps appear very noisy and the explanation ends-up not being useful. We note that SmoothGrad, which proposes to fix that discontinuity, is useful. On the other hand, on the leaves dataset, the model uses very fine features, therefore the Saliency maps suffer less from the discontinuity, it does not appear noisy, Saliency is useful. We also note that in this case, SmoothGrad is not better than Saliency, which can arguably be attributed to the fact that we do not need to fix the discontinuity of the gradient. Conversely, because the granularity of both Grad-CAM (the feature map is much smaller than image size) and Occlusion (the patch size is much bigger than a pixel) is too high, the heatmaps they offer on the “leaves” dataset are too coarse to specifically highlight the fine features and it seems to take more time for the subjects to pick-up on them. But on the biased dataset, Grad-CAM and Occlusion are the best performing methods.

\section{Why do attribution methods fail?}

\subsection{Faithfulness}

\begin{figure*}[h]
    \includegraphics[width=0.48\textwidth]{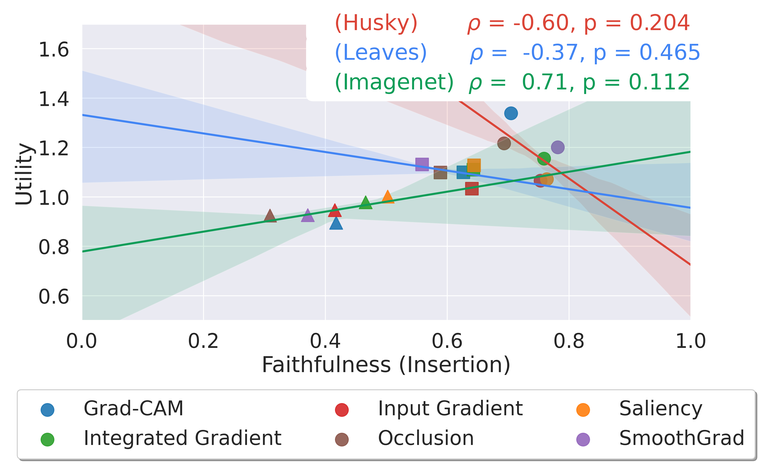}
    \includegraphics[width=0.48\textwidth]{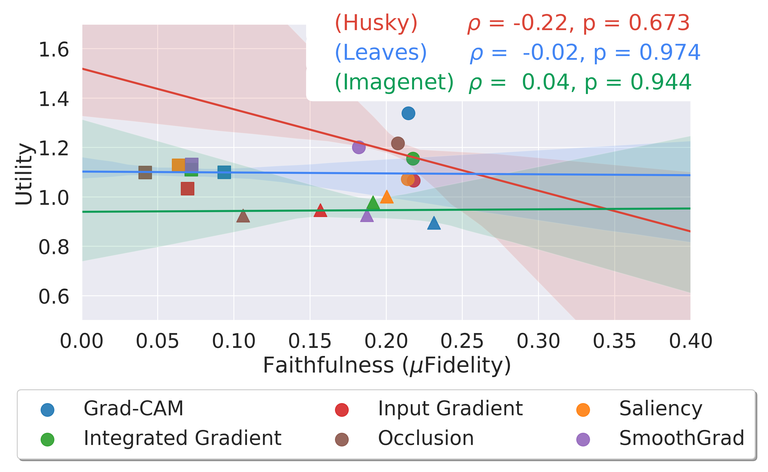}
    \caption{\textbf{\metric~vs Insertion correlation \& \metric~vs $\mu$Fidelity correlation}
        The results suggest that every faithfulness metrics tested are poor predictors of the practical usefulness of attribution methods. 
        Concerning the ImageNet dataset (triangle marker), the \metric~scores are insignificant since none of the methods improves the baseline.
        }
    \label{fig:insertion_mufidelity}
    \vspace{-4mm}
\end{figure*}

While the Deletion\cite{petsiuk2018rise} measure is the most commonly used faithfulness metric, for completeness we also consider 2 others faithfulness metric available in the Xplique library\cite{fel2022xplique}: Insertion\cite{petsiuk2018rise} and $\mu$Fidelity\cite{aggregating2020}. 
Fig \ref{fig:insertion_mufidelity} shows the correlation between either measure and our \metric. We find them to be no better predictor of the practical usefulness of attribution methods than the Deletion measure. \\

\subsection{Perceptual Similarity}

\begin{figure*}[ht]
    \includegraphics[width=0.95\textwidth]{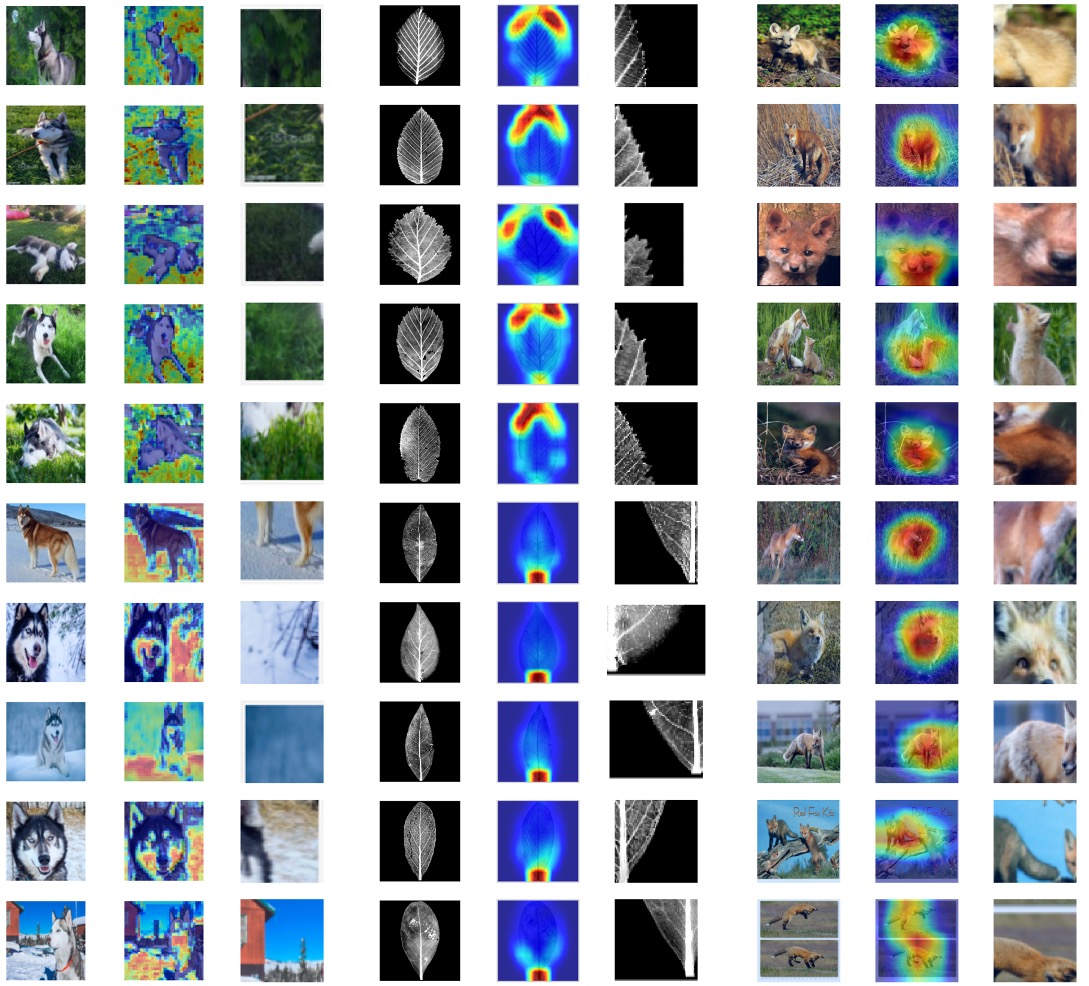}
    \caption{\textbf{Examples of extracted patches.} The perceptual similarity score is performed on the locations considered most important by the attribution methods. Examples of patches extracted for the three datasets with the \expgc~ method.}
    \label{fig:patchs_examples}
    \vspace{-4mm}
\end{figure*}

\begin{table}[h]
\vspace{2mm}
\centering
\begin{tabular}{lccc}
\toprule
 Method & \textit{Husky vs. Wolf} & \textit{Leaves} & \textit{ImageNet} \\
\midrule
Saliency~\cite{simonyan2014deep}                  & 0.304 & 0.334 & \textbf{0.378} \\ 
Integ.-Grad.~\cite{sundararajan2017axiomatic}     & 0.292 & \textbf{0.411} & \textbf{0.388} \\
SmoothGrad~\cite{smilkov2017smoothgrad}           & 0.285 & 0.286 & \textbf{0.384} \\
GradCAM~\cite{selvaraju2017gradcam}               & 0.241 & 0.312 & \textbf{0.38} \\
Occlusion~\cite{zeiler2014visualizing}            & 0.282 & 0.277 & \textbf{0.41} \\
Grad.-Input~\cite{shrikumar2017learning}               & 0.309 & \textbf{0.44} & \textbf{0.378} \\
\bottomrule
\end{tabular}
\vspace{2mm}
\caption{\textbf{\textit{Perceptual Similarity} scores.} The perceptual similarity of highlighted regions by a given attribution method for both classes is measured, for each method, for each dataset. The perceptual similarity scores that are higher than $0.378$ (the minimum score on ImageNet) are \textbf{bolded}. Higher is more similar.}
\label{tab:similarity}
\vspace{-2mm}
\end{table}
Tab \ref{tab:similarity} shows the Perceptual Similarity scores obtained for each method, on every dataset. We observe that on ImageNet, where attribution methods do not help, the perceptual similarity scores are clearly higher than on the two other datasets, where attribution methods help. \\
Fig \ref{fig:patchs_examples} shows examples of patches for each dataset using \expgc.

\chapter{Alignment}
\section{Psychophyics}\label{si_sec:psychophysics}

The psychophysics experiments of \textsection{4.2} were implemented with the psiTurk framework \cite{Gureckis2016-if} and custom javascript functions. Each trial sequence was converted to a HTML5-compatible video for the fastest reliable presentation time possible in a web browser. Videos were cached before each trial to optimize reliability of experiment timing within the web browser. A photo-diode verified the reliability of stimulus timing in our experiment was consistently accurate within $\sim10\mathrm{ms}$ across different operating system, web browser, and display type configurations.

\paragraph{Participants:} We recruited 199 participants from Amazon Mechanical Turk (\url{mturk.com}) for the experiments. Participants were based in the United States, used either the Firefox or Chrome browser on a non-mobile device, and had a minimal average approval rating of 95\% on past Mechanical Turk tasks. 

\paragraph{Stimuli:} Experiment images were taken from the \textit{Clicktionary} dataset~\cite{Linsley2017-qe}. Images were sampled from 5 target and 5 distractor categories: border collie, sorrel (horse), great white shark, bald eagle, and panther; trailer truck, sports car, speedboat, airliner, and school bus. Images were presented to human participants (and DNNs) either intact or with a perceptual phase scrambled mask that exposed a proportion of their most important visual features, as described in the main text. Images were cast to greyscale to control for trivial color-based cues for classification and blend the scrambled mask background into the foreground. Responses to intact images were used to normalize the performance of each observer on masked images relative to their maximum performance on these images.

\begin{figure}[h!]
\begin{center}
   \includegraphics[width=1\linewidth]{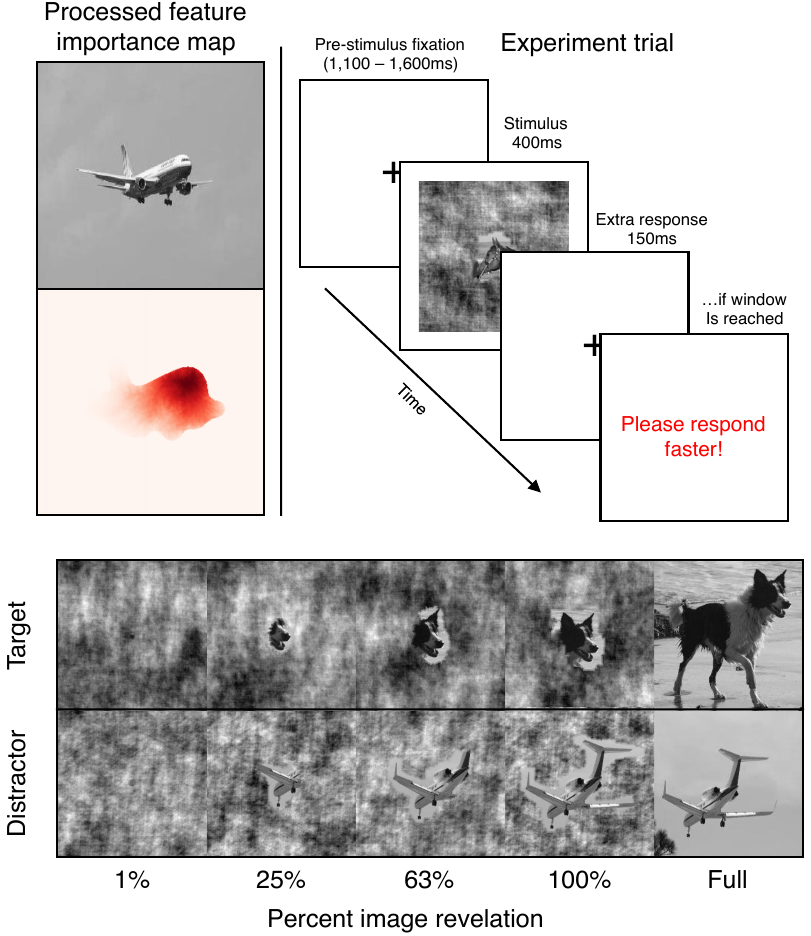}
\end{center}
   \caption{\textbf{Overview of the psychophysics paradigm.} Participants performed a rapid animals vs. vehicles categorization paradigm (top). Stimuli were created using feature importance maps derived from humans or DNNs via a ``stochastic flood-fill'' algorithm that revealed image regions of different sizes centered on important features. Sample stimuli are shown (bottom) for different percentages of image revelation. Note that 100\% revelation corresponds to all non-zero pixels in a feature importance map.}
\label{si_fig:psychophysics}
\end{figure}

\begin{figure}[h!]
\begin{center}
   \includegraphics[width=1\linewidth]{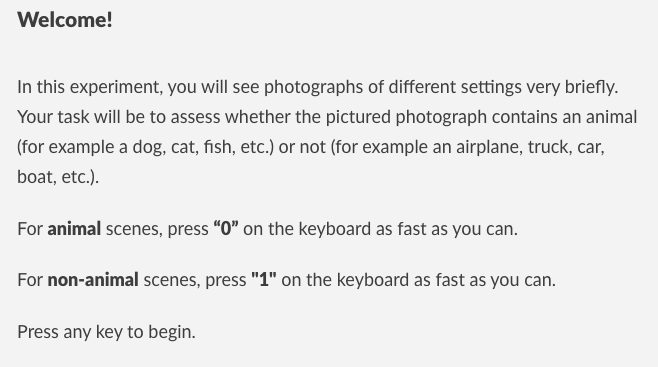}
\end{center}
   \caption{\textbf{Psychophysics experiment instructions.}}
\label{si_fig:instructions}
\end{figure}

Image masks were created for each image to reveal only a proportion of the most important visual features. For each image, we created masks that revealed between 1\% and 100\% (at log-scale spaced intervals) of the object pixels in the corresponding image's \textit{Clicktionary} feature importance map. We generated these masks in two steps. First, we computed a phase-scrambled version of the image~\cite{Oppenheim1981-sl, Thomson1999-rp}. Next, we used a novel ``stochastic flood-fill'' algorithm to reveal a contiguous region of the most important visual features in the image according to humans. Our flood-fill algorithm was seeded on the pixel deemed most important by humans in the image, then grew outwards anisotropically and biased towards pixels with higher feature importance scores (Figure~\ref{si_fig:psychophysics}). The revealed region was always centered on the image. Each participant saw every category exemplar only once, with its amount of image revelation randomly selected from all possible configurations. 

After providing online consent, participants were instructed to complete a rapid visual categorization task in which they had to classify stimuli revealing a portion of the most diagnostic object features (Fig.~\ref{si_fig:instructions}). Each experimental trial began with a cross for participants to fixate for a variable time (1,100–1,600ms), then a stimulus for 400ms, then another cross and additional time for participants to render a decision. Participants were instructed to provide a decision after the first fixation cross, but that they only had 650ms to answer. If they were too slow to respond they were told to respond faster and the trial was discarded.

\begin{figure}[h!]
\begin{center}
   \includegraphics[width=0.5\linewidth]{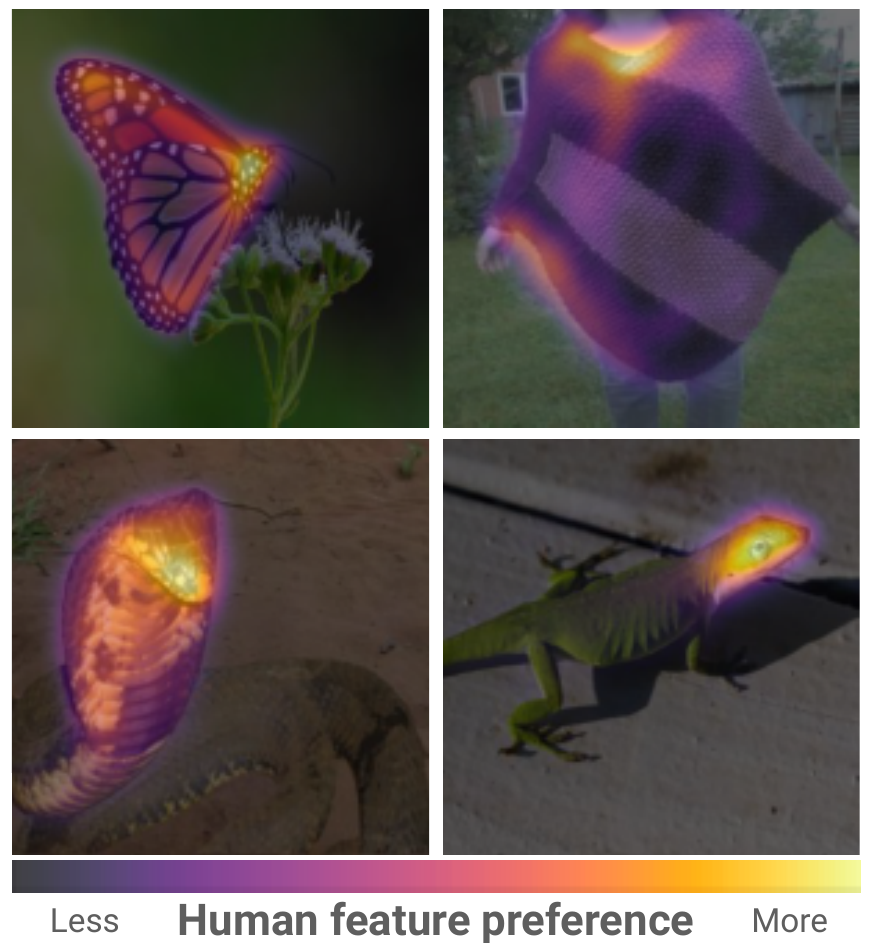}
\end{center}
   \caption{\textbf{Example \textit{ClickMe} feature importance maps on ImageNet images.}}
\label{si_fig:clickme_examples}
\end{figure}

\section{Additional Results}
\subsection{\textit{ClickMe}}
The \textit{ClickMe} game by \cite{Linsley2019-ew} was used to identify category diagnostic features in ImageNet images. These feature importance maps largely focus on object regions rather than context, and in contrast to segmentation maps select features on the ``front'' or ``face'' of objects (Fig.~\ref{si_fig:clickme_examples}).

\begin{figure}[h!]
\begin{center}
   \includegraphics[width=1\linewidth]{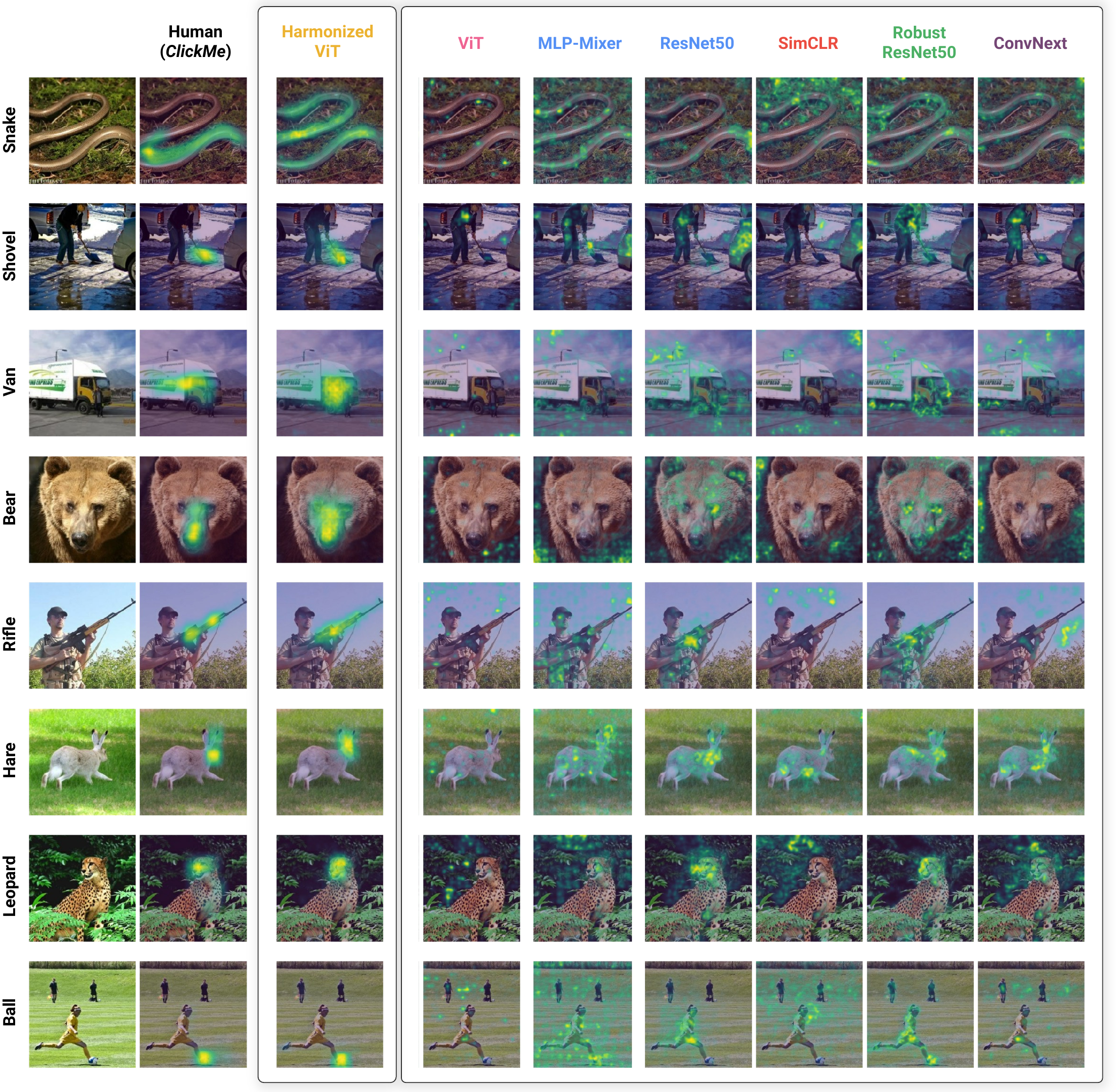}
\end{center}
   \caption{\textbf{Feature importance maps of humans, harmonized, and unharmonized models on ImageNet}.}
 \label{fig:qualitative_figure_big}
\end{figure}

\begin{figure}[h!]
\begin{center}
   \includegraphics[width=1\linewidth]{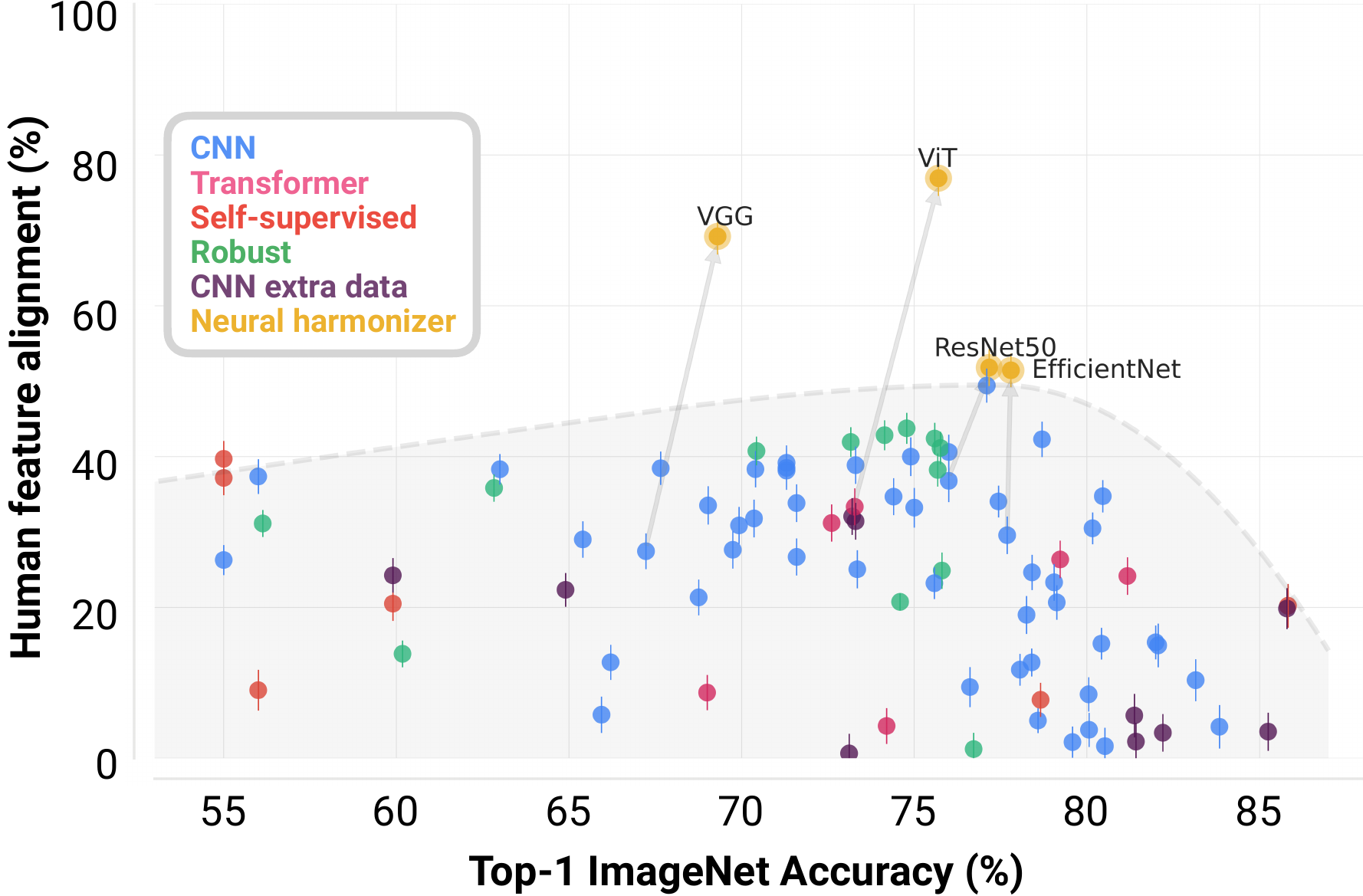}
\end{center}
   \caption{\textbf{The neural harmonizer's effect is robust across image scales.} Here, we show that the trade-off between ImageNet accuracy and alignment with humans holds across downsizing by a factor of \textit{16}. The Neural harmonizer once again yields the model with the best alignment with humans. Grey-shaded area captures the trade-off between accuracy and alignment in standard DNNs. Error bars are bootstrapped standard deviations over feature alignment.}
\label{si_fig:clickme_results_16}
\end{figure}

As discussed in the main text, we found a trade-off between DNN top-1 ImageNet accuracy and the alignment of their feature importance maps with humans importance maps from \textit{ClickMe}. This trade-off persists across multiple scales of feature importance maps, including 16$\times$ (Fig.~\ref{si_fig:clickme_results_16}) sub-sampled maps, meaning that simple smoothing is not sufficient to fix the trade-off.

\subsection{\textit{ViT attention}} While in the main text we investigate alignment between humans and models using gradient feature importance visualizations, the attention maps in transformer  models like the ViT provide another avenue for investigation. To understand whether or not attention maps from ViT are more aligned with humans than their gradient-based decision explanation maps, we computed attention rollouts for harmonized and unharmonized ViTs~\cite{Abnar2020-sb}. We found that both versions of the ViT had similar correlations between their attention rollouts and human \textit{ClickMe} maps: 0.38 for the harmonized ViT and 0.393 for the unharmonized model. This surprising result suggests that the harmonizer affects the process by which ViTs integrate visual information into their decisions rather than how they allocate attention. Through manipulating ViT decision making processes, the harmonizer can induce the large changes in gradient-based visualizations and psychophysics that we describe in the main text.

\subsection{\textit{Correlations between measurements of human visual strategies}}
Our results rely on three independent datasets measuring different features of human visual strategies: \textit{ClickMe}, \textit{Clicktionary}, and the psychophysics experiments we introduce in this manuscript. The fact that all three evoke similar trade-offs between top-1 accuracy and human alignment is a surprising result that deserves further attention. We investigated these trade-offs by measuring the correlation between human alignment on each dataset, with and without models trained with the neural harmonizer. We found that correlations between datasets were lower across the board when neural harmonizer models were not included. Each correlation improved when the neural harmonizer models were included in the calculation. This finding indicates that the neural harmonizer successfully aligned visual strategies between humans and DNNs, and was not merely benefiting from either \textit{where} humans versus DNNs considered important visual features to be or \textit{how} humans versus DNNs incorporated those features into their decisions.

\begin{figure}[h!]
\begin{center}
   \includegraphics[width=1\linewidth]{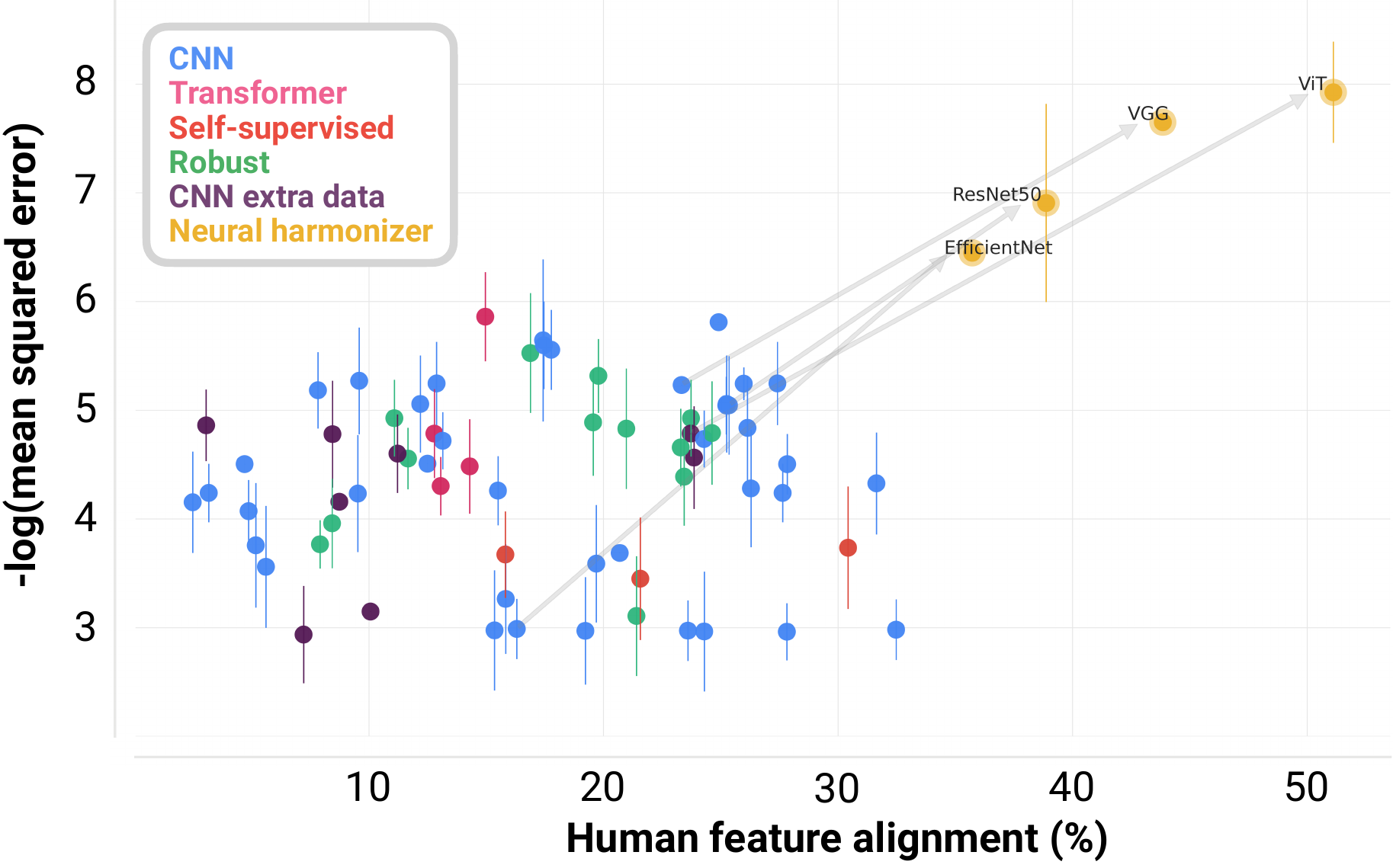}
\end{center}
   \caption{\textbf{The association between \textit{Clicktionary} alignment versus psychophysics alignment.} These scores are significantly correlated, $\rho=0.53, p<0.001$.}
\label{si_fig:clicktionary_vs_psych}
\end{figure}

\begin{figure}[h!]
\begin{center}
   \includegraphics[width=1\linewidth]{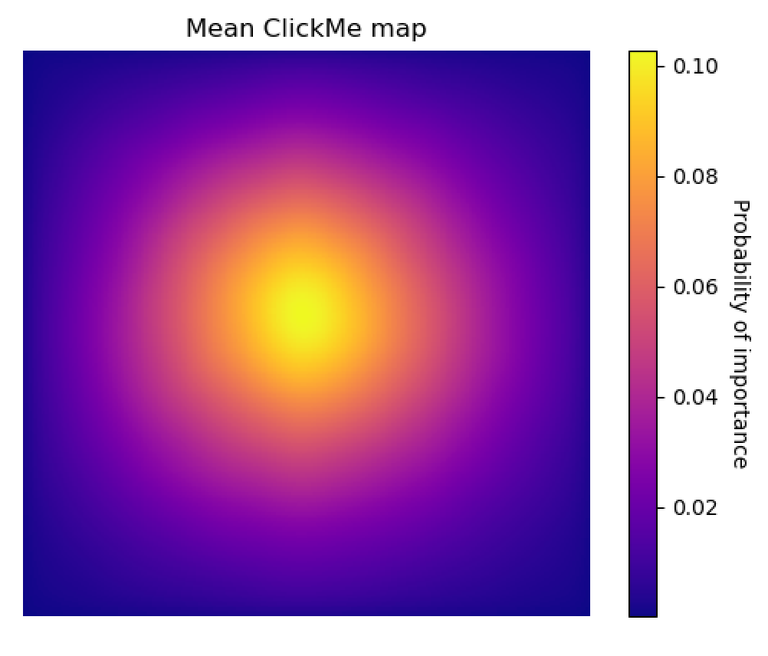}
\end{center}
   \caption{\textbf{The mean of \textit{ClickMe} feature importance maps exhibits a center bias, likely due to the positioning of objects in ImageNet images rather than a purely spatial bias of human participants (compare to individual maps shown in \ref{si_fig:clickme_examples})}.}
\label{si_fig:center_bias}
\end{figure}

\clearpage

\section{Theoretical consideration.}

We recall the stated theorem concerning the alignment of predictions given aligned explanations. The theorem asserts the consistency of predictions up to a constant when the explanations generated by any pair of aligned predictors are equivalent.

\begin{theorem}[$\explainer$-Aligned Imply Aligned Predictions]
\label{app:harmonization:thm}
Given a function space where $\fspace : \sx \to \sy$, with $\sx = (0, 1]^d$ denoting the input space and $\sy \subseteq \Real$ the output space, an explanation functional $\explainer : \fspace \times \sx \to \sx$. Assume $(\f, \fbis)$ are two predictors that are $\explainer$-Aligned within this space. For any explanation functional $\explainer$ from the set $\{ \explainer_{\text{Sa}}, \explainer_{\text{GI}}, \explainer_{\text{IG}}, \explainer_{\text{OC}} \}$, the alignment of explanations imply the alignment of predictions up to a constant difference:
\[
\forall \vx \in \sx, ~~ \f(\vx) = \fbis(\vx) + \kappa
\]
where $\kappa$ is a constant independent of $\vx$.
\end{theorem}

\begin{proof}

\textbf{Saliency:} Let $\vx_0 \in \sx$ serve as a reference point in the input space, and define the constant $\kappa = \f(\vx_0) - \fbis(\vx_0)$. Given the alignment of explanations by $\explainer_{\text{Sa}}$, we have:

\begin{align}
\explainer_{\text{Sa}}(\f, \vx) &= \explainer_{\text{Sa}}(\fbis, \vx), \\
 \grad \f(\vx) &= \grad \fbis(\vx).
\end{align}

From the equality of gradients, the fundamental theorem of calculus permits us to deduce:

\begin{align}
\int_{\vx_0}^{\vx} \grad \f(\vx') \mathrm{d} \vx' &= \int_{\vx_0}^{\vx} \grad \fbis(\vx') \mathrm{d} \vx', \\
 \f(\vx) - \f(\vx_0) &= \fbis(\vx) - \fbis(\vx_0), \\
 \f(\vx) &= \fbis(\vx) + \kappa.
\end{align}

Similar reasoning could be applied to \textbf{Gradient-Input}.

\textbf{Integrated-Gradients:} we recall that $\sx = (0, 1]^d$, $\sy \subseteq \Real$, for $\explainer_{\text{IG}}$, we have:

\begin{align}
\explainer_{\text{IG}}(\f, \vx) &= \explainer_{\text{IG}}(\fbis, \vx) \\
(\vx - \vx_0) \int_{0}^{1} \grad \f((1 - \alpha) \vx_0 + &\alpha(\vx - \vx_0)) \diff \alpha = \\
(\vx - \vx_0) \int_{0}^{1} \grad & \fbis((1 - \alpha) \vx_0 + \alpha(\vx - \vx_0)) \diff \alpha.
\end{align}

With $\tilde{\vx} = (1 - \alpha) \vx_0 + \alpha(\vx - \vx_0)$, 

\begin{align}
(\vx - \vx_0) \int_{0}^{1} \grad \f(\tilde{\vx}) \diff \alpha &= 
(\vx - \vx_0) \int_{0}^{1} \grad  \fbis(\tilde{\vx}) \diff \alpha \\
\f(\vx) - \f(\vx_0) &= \fbis(\vx) - \fbis(\vx_0) \\
\f(\vx) &= \fbis(\vx) + \kappa.
\end{align}

\textbf{Occlusion:} by definition,

\begin{align}
\explainer_{\text{OC}}(\f, \vx) &= \explainer_{\text{OC}}(\fbis, \vx) \\
\f(\vx) - \f(\vx_{[x_i = x_0]}) &= \fbis(\vx) - \fbis(\vx_{[x_i = x_0]}).
\end{align}

To say it simply, the difference when one feature is set to a baseline state is the same between the two predictor. However, we observe that if we removed another pixel, the equality still hold, (one could pose $\vx' = \vx_{[x_i = x_0]}$. Intuitively, we are building a discrete path from any image $\vx$ to the image full of the baseline vector $\vx_0$ by flipping each element $x_i$ of $\vx$ one by one.

\begin{align}
\f(\vx') - \f(\vx_{[x_j = x_0]}) &= \fbis(\vx') - \fbis(\vx_{[x_j = x_0]}).
\end{align}

Thus, by recursion:

\begin{align}
\f(\vx) - \f(\vx_{[x_i = x_0]}) + \f(\vx_{[x_i = x_0]}) - \f(\vx_{[x_i = x_0, x_j = x_0]}) + \ldots - \f(\vx_0)  = \\ \fbis(\vx) - \fbis(\vx_{[x_i = x_0]}) + \fbis(\vx_{[x_i = x_0]}) - \fbis(\vx_{[x_i = x_0, x_j = x_0]}) + \ldots - \fbis(\vx_0).
\end{align}

We reduce the following telescopic sum:

\begin{align}
\f(\vx) - \f(\vx_0)  &= \fbis(\vx) - \fbis(\vx_0) \\
\f(\vx) &= \fbis(\vx) + \kappa.
\end{align}

As whatever the $\vx$ choosen, $\vx_0$, the last element of the telescopic sum is always the same is constant, the vector in $\Real^d$ full of the baseline state $x_0$. 

This comprehensive approach across different explanation functionals substantiates the theorem, confirming that explanation alignment under any of these functionals imply prediction alignment up to a constant.

\end{proof}

\chapter{Concepts}

\section{CRAFT}
\subsection{Limitations}
\label{apx:craft:limitations}

Although we believe concept-based XAI to be a promising research direction, it isn't without pitfalls. It is capable of producing explanations that are ideally easy to understand by humans, but to what extent is a question that remains unanswered. The fact that there is no way to mathematically measure this prevents researchers from easily comparing the different techniques in the literature other than through time consuming and expensive experiments with human subjects. We think that developing a metric should be one of the field's priorities.%

With \craft, we address the question of \what~by showing a cluster of the images that better represent each concept. However, we recognize that it's not perfect: in some cases, concepts are difficult to clearly define -- put a label on what it represents --, and might induce some confirmation and selection bias. Feature visualization~\cite{olah2017feature} might help in better illustrating the specific concept (as done in appendix \ref{app:craft:feature-viz-val}), but we believe there's still space for improvement. For instance, an interesting idea could be to leverage image captioning methods to describe the clusters of image crops, as textual information could help humans in better understanding clusters.

Although we believe \craft~to be a considerable step in the good direction for the field of concept-based XAI, it also have some pitfalls. Namely, we chose the NMF as the activation factorization, which, while drastically improving the quality of extracted concepts, also comes with it's own caveats. For instance, it is known to be NP-hard to compute exactly, and in order to make it scalable, we had to use a tractable approximation by alternating the optimization of $\m{U}$ and $\m{W}$ through ADMM~\cite{boyd2011distributed}. This approach might indeed yield non-unique solutions. Our experiments (section \ref{sec:craft:expSobol}), have shown a low variance on between the runs, which comforts us about the stability of our results.%
However the absence of formal guarantee for uniqueness must be kept in mind: this subject is still an active topic of research and improvement could be expected in the near future. Namely, sparsity constraints and regularization seem to be promising paths.
Naturally, we also need enough samples of the class under study to be available for the factorization to construct a relevant concept bank, which might affect the quality of the explanations on frugal applications where data is very scarce. %

\subsection{Additional results}
\label{apx:craft:more-craft}

\subsubsection{Qualitative comparison with ACE}\label{apx:qualitative}

\begin{figure*}[ht]
  \includegraphics[width=0.99\textwidth]{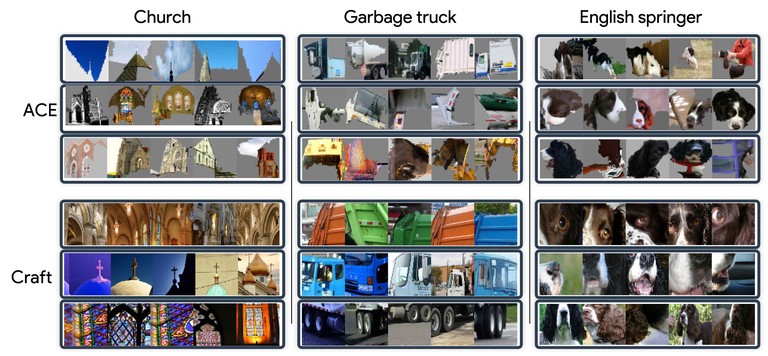}
  \caption{ \textbf{Qualitative comparison.} We compare concepts found by our method (top) to those extracted with ACE~\cite{ghorbani2019towards} (bottom) for the classes \textit{Church}, \textit{Garbage truck} and \textit{English springer} from ILSVRC2012~\cite{imagenet_cvpr09}. %
  }
  \label{fig:app:craft:qualitative}
\end{figure*}

Figure~\ref{fig:app:craft:qualitative} compares the examples of concepts found by CRAFT against those found by ACE~\cite{ghorbani2019towards} for 3 classes of Imagenette. 
For each class the concepts are ordered by importance (the highest being the most important). 
ACE uses a clustering technique and TCAV to estimate importance, while CRAFT uses the method introduced in \ref{sec:craft:method} and Sobol to estimate importance. These examples illustrate one of the weaknesses of ACE: the segmentation used can introduce biases through the baseline value used~\cite{sturmfels2020visualizing,fong2017meaningful}. The concepts found by CRAFT seem distinct: (vault, cross, stained glass) for the Church class, (dumpster, truck door, two-wheeler) for the garbage truck, and (eyes, nose, fluffy ears) for the English Springer.

\subsubsection{Most important concepts.} 
\begin{figure*}[ht]
    \centering
      \includegraphics[width=\textwidth]{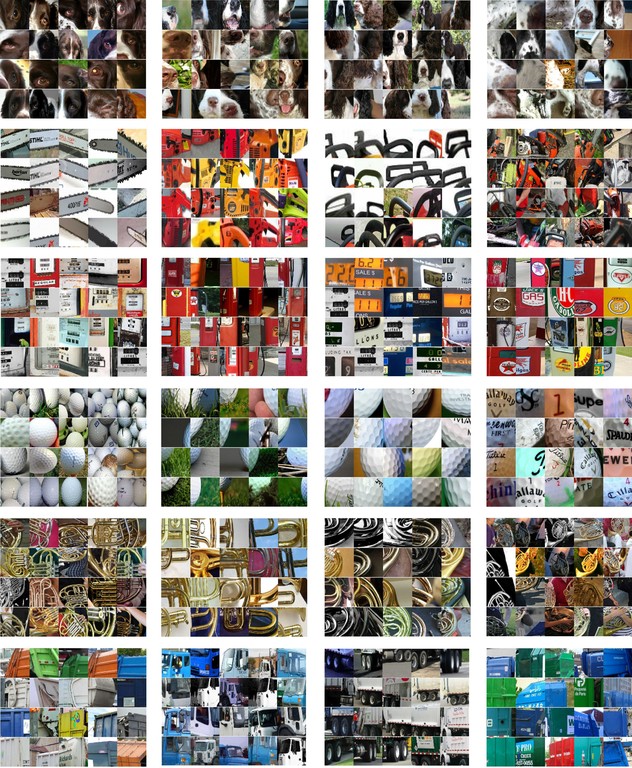}
      \caption{ \textbf{CRAFT most important concepts}. The 4 most important concepts ranked by importance (left to right) for the following classes: `English springer', `Chain saw',  `Gas pump', `Golf ball', `French horn',  and `Garbage truck'.
      }
      \label{fig:craft:segments_all}
\end{figure*}

We show more example of the 4 most importants concepts for 6 classes: `Chain saw', `English springer', `Gas pump', `Golf ball', `French horn' and `Garbage Truck' (Figure~\ref{fig:craft:segments_all}).%

\clearpage

\subsubsection{Feature Visualization validation} \label{app:craft:feature-viz-val}

\begin{figure}[ht]
\centering
  \includegraphics[width=0.45\textwidth]{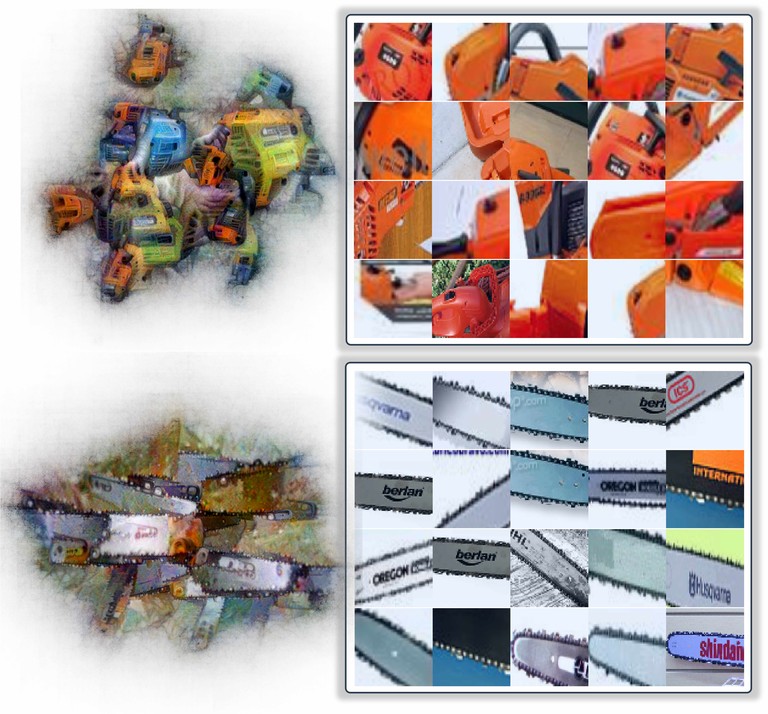}
  \caption{ \textbf{Feature visualization for chainsaw CRAFT concepts.}
  }
  \label{fig:craft:feature_viz_chainsaw}
\end{figure}

\begin{figure}[ht]
\centering
  \includegraphics[width=0.45\textwidth]{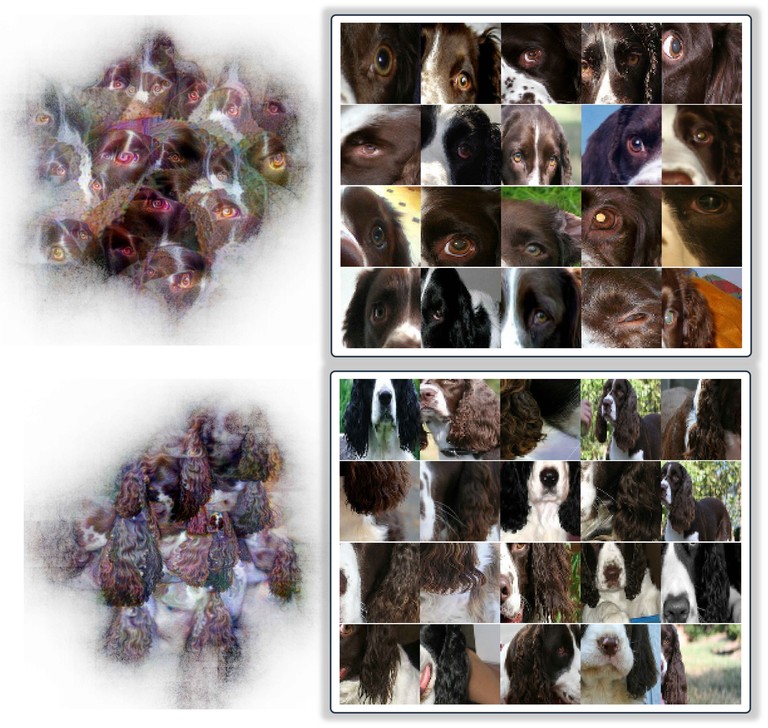}
  \caption{ \textbf{Feature visualization for english springer CRAFT concepts.}
  }
  \label{fig:craft:feature_viz_englishspringer}
\end{figure}

\begin{figure}[ht]
\centering
  \includegraphics[width=0.45\textwidth]{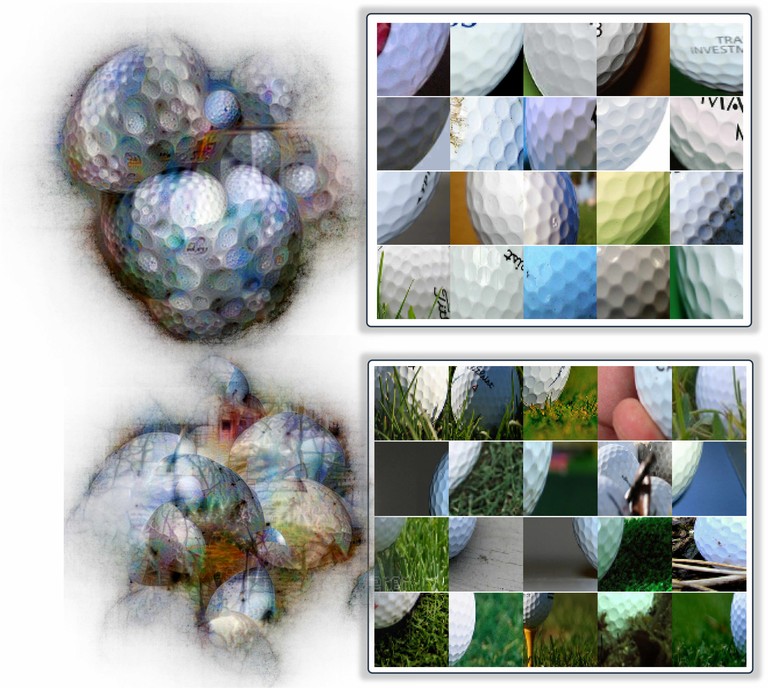}
  \caption{ \textbf{Feature visualization for golf CRAFT concepts.} 
  }
  \label{fig:craft:feature_viz_golf}
\end{figure}

Another way of interpreting concepts -- as per~\cite{kim2018interpretability} -- is to employ feature visualization methods: through optimization, find an image that maximizes an activation pattern.
In our case, we used the set of regularization and constraints proposed by \cite{olah2017feature}, which allow us to successfully obtain realistic images. In Figures~[\ref{fig:craft:feature_viz_chainsaw}-\ref{fig:craft:feature_viz_golf}], we showcase these synthetic images obtained through feature visualization, along with the segments that maximize the target concept. We observe that they do reflect the underlying concepts of interest.

Concretely, to produce those feature visualization, we are looking for an image $\vx^*$ that is optimized to correspond to a concept from the concept bank $\m{W}_i$. We use the so called `dot-cossim' loss proposed by ~\cite{olah2017feature}, which give the following objective:

\begin{equation*}
    \vx^* = \argmax_{\vx \in \mathcal{X}} ~ \langle \bm{g}(\vx), \m{W}_i \rangle 
\frac{\langle \bm{g}(\vx), \m{W}_i \rangle^2}{||\bm{g}(\vx)|| ~ ||\m{W}_i||  } - \mathcal{R}(\vx)    
\end{equation*}

With $\mathcal{R}(\cdot)$, the regularizations applied to $\vx$  -- the default regularizations in the \textbf{Xplique} library~\cite{xplique}. As for the specific parameters, we used Fourier preconditioning on the image with a decay rate of $0.8$ and an Adam optimizer ($lr = 1e-1$). 

\clearpage
\newpage

\subsection{Backpropagating through the NMF block}

\subsubsection{Alternating Direction Method of Multipliers (ADMM) for NMF}

We recall that NMF decomposes the positive features vector $\m{A} \in \mathbb{R}^{n \times p}$ of $n$ examples lying in dimension $p$, into a product of positive low rank matrices $\m{U}(\m{A})\in\mathbb{R}^{n\times r}$ and $\m{W}(\m{A})\in\mathbb{R}^{p\times r}$ (with $r<<\min(n,p)$), i.e the solution to the problem:
\begin{align}\label{apeq:craft:nmf}
\min_{\m{U}\geq 0,\m{W}\geq 0} & \frac{1}{2}\|\m{A}-\m{U}\m{W}^T\|^2_F. %
\end{align}

For simplicity we used a non-regularized version of the NMF objective, following Algorithms 1 and 3 in paper~\cite{huang2016flexible}, based on ADMM~\cite{boyd2011distributed}. This algorithm transforms the non-linear equality constraints into indicator functions $\bm{\delta}$. Auxiliary variables $\tilde{\m{U}},\tilde{\m{W}}$ are also introduced to separate the optimization of the objective on the one side, and the satisfaction of the constraint on $\m{U}, \m{W}$ on the other side. The equality constraints $\tilde{\m{U}}=\m{U},\tilde{\m{W}}=\m{W}$ are linear and easily handled by the ADMM framework through the associated dual variables $\bar{\m{U}},\bar{\m{W}}$. In our case, the problem in Equation~\ref{apeq:craft:nmf} is transformed into:
  
\begin{equation}
\begin{aligned}\label{apeq:craft:admm}
\min_{\m{U},\tilde{\m{U}}, \m{W},\tilde{\m{W}}} & \frac{1}{2}\|\m{A}-\tilde{\m{U}} \tilde{\m{W}}^T\|^2_F+\bm{\delta}(\m{U})+\bm{\delta}(\m{W}), 
\\
~ ~ s.t. ~ ~ &\tilde{\m{U}}=\m{U}, \tilde{\m{W}}=\m{W} \\
     \text{with} ~ & \bm{\delta}(\bm{H})=\begin{cases}
                            0 \text{ if } \bm{H} \geq 0,\\
                            +\infty \text{ otherwise.}
                            \end{cases}
\end{aligned}
\end{equation}

Note that $\tilde{\m{U}}$ and $\m{U}$ (resp. $\tilde{\m{W}}$ and $\m{W}$) seem redundant: they are meant to be equal thanks to constraints $\tilde{\m{U}}=\m{U}, \tilde{\m{W}}=\m{W}$. This is standard practice within ADMM framework: introducing redundancies allows to disentangle the (unconstrained) optimization of the objective on one side (with $\tilde{\m{U}}$ and $\tilde{\m{W}}$) and constraint satisfaction on the other side with $\m{U}$ and $\m{W}$. During the optimization process the variables $\tilde{\m{U}},\m{U}$ (resp. $\tilde{\m{W}},\m{W}$) are different, and only become equal in the limit at convergence. The dual variables $\bar{\m{U}},\bar{\m{W}}$ control the balance between optimization of the objective $\frac{1}{2}\|\m{A}-\tilde{\m{U}} \tilde{\m{W}}^T\|^2_F$ and constraint satisfaction $\tilde{\m{U}}=\m{U}, \tilde{\m{W}}=\m{W}$. The constraints are simplified at the cost of a non-smooth (and even a non-finite) objective function $\frac{1}{2}\|\m{A} -\bar{\m{U}} \bar{\m{W}}^T\|^2_F+\bm{\delta}(\m{U})+\bm{\delta}(\m{W})$ due to the term $\bm{\delta}(\m{U})+\bm{\delta}(\m{W})$. ADMM proceeds to create a so-called \textit{augmented Lagrangian} with $l_2$ regularization $\rho>0$:
\begin{equation}
    \begin{aligned}
    \Lagrangian&(\m{A},\m{U},\m{W},\tilde{\m{U}},\tilde{\m{W}},\bar{\m{U}},\bar{\m{W}})=\\
    &\frac{1}{2}\|\m{A}-\tilde{\m{U}}\tilde{\m{W}}^T\|^2_F+\bm{\delta}(\m{U})+\bm{\delta}(\m{W})\\
    &+\bar{\m{U}}^T(\tilde{\m{U}}-\m{U})+\bar{\m{W}}^T(\tilde{\m{W}}-\m{W})\\
    &+\frac{\rho}{2}\left(\|\tilde{\m{U}}-\m{U}\|_2^2+\|\tilde{\m{W}}-\m{W}\|_2^2\right).
    \end{aligned}
\end{equation}

This regularization ensures that the dual problem is well posed and that it remain convex, even with the non smooth and infinite terms $\bm{\delta}(\m{U})+\bm{\delta}(\m{W})$. Once again, this is standard practice within ADMM framework. The (regularized) problem associated to this Lagrangian is decomposed into a sequence of convex problems that alternate minimization over the $\m{U},\tilde{\m{U}},\bar{\m{U}}$ and the $\m{W},\tilde{\m{W}},\bar{\m{W}}$ triplets.
  
\begin{align}\label{apeq:craft:pairnnls}
\m{U}_{t+1}&=\argmin_{\m{U}=\tilde{\m{U}}} \frac{1}{2}\|\m{A}-\tilde{\m{U}}\m{W}_t^T\|^2_F+\bm{\delta}(\m{U})+\frac{\rho}{2}\|\tilde{\m{U}}-\m{U}\|_2^2. %
\\
\m{W}_{t+1}&=\argmin_{\m{W}=\tilde{\m{W}}} \frac{1}{2}\|\m{A}-\m{U}_t\tilde{\m{W}}^T\|^2_F+\bm{\delta}(\m{W})+\frac{\rho}{2}\|\tilde{\m{W}}-\m{W}\|_2^2.%
\end{align}

This guarantees a monotonic decrease of the objective function $\|\m{A}-\tilde{\m{U}}_t\tilde{\m{W}}_t^T\|_F^2$. Each of these sub-problems is thus solved with ADMM separately, by alternating minimization steps of $\frac{1}{2}\|\m{A}-\tilde{\m{U}}\m{W}_t^T\|^2_F+\bar{\m{U}}^T(\tilde{\m{U}}-\m{U})+\frac{\rho}{2}\|\m{U}-\tilde{\m{U}}\|_2^2$ over $\tilde{\m{U}}$ (\textbf{\textit{i}}), with minimization steps of $\bm{\delta}(\m{U})+\frac{\rho}{2}\|\m{U}-\tilde{\m{U}}\|_2^2$ over $\m{U}$ (\textbf{\textit{ii}}), and gradient ascent steps (\textbf{\textit{iii}}) on the dual variable $\bar{\m{U}}\gets \bar{\m{U}}+(\tilde{\m{U}}-\m{U})$. A similar scheme is used for $\m{W}$ updates. Step (\textbf{\textit{i}}) is a simple convex quadratic program with equality constraints, whose KKT~\cite{karush1939minima,kuhn1951nonlinear} conditions yield a linear system with a Positive Semi-Definite (PSD) matrix. Step (\textbf{\textit{ii}}) is a simple projection of $\tilde{\m{U}}$ onto the convex set $\bm{\delta}^{-1}(\bm{0})$. Finally, step (\textbf{\textit{iii}}) is inexpensive.

Concretely, we solved the quadratic program using Conjugate Gradient, from \textit{jax.scipy.sparse.linalg.cg}. This indirect method only involves \textit{matrix-vector} products and can be more GPU-efficient than methods that are based on matrix factorization (such as Cholesky decomposition). Also, we re-implemented the pseudo code of~\cite{huang2016flexible} in \textit{Jax} for a fully GPU-compatible program. We used the primal variables $\m{U}_0,\m{W}_0$ returned by \textit{sklearn.decompose.nmf} as a \textit{warm start} for ADMM and observe that the high quality initialization of these primal variables considerably speeds up the convergence of the dual variables.

\subsubsection{Implicit differentiation}\label{app:craft:implicit}

The Lagrangian of the NMF problem reads $\mathcal{L}(\m{U},\m{W},\bar{\m{U}},\bar{\m{W}})=\frac{1}{2}\|\m{A}-\m{U}\m{W}^T\|_F^2-\bar{\m{U}}^T\m{U}-\bar{\m{W}}^T\m{W}$, with dual variables $\bar{\m{U}}$ and $\bar{\m{W}}$ associated to the constraints $\m{U}\geq 0, \m{W} \geq 0$. It yields a function $\bm{F}$ based on the KKT conditions~\cite{karush1939minima,kuhn1951nonlinear} whose optimal tuple $\m{U},\m{W},\bar{\m{U}},\bar{\m{W}}$ is a root.  
  
For single NNLS problem (for example, with optimization over $\m{U}$) the KKT conditions are:

\begin{equation} %
    \begin{cases}
    \nabla_{\m{U}}\left(\frac{1}{2}\|\m{A}-\tilde{\m{U}} \tilde{\m{W}}^T\|^2_F+\bar{\m{U}}^T(-\m{U})\right)    =0, \text{ stationarity,}\\
    -\m{U}\leq 0, \text{ primal feasability,}\\
    \bar{\m{U}} \odot \m{U}   =0, \text{ complementary slackness,}\\
    \bar{\m{U}}   \geq 0, \text{ dual feasability.}\\
\end{cases}
\label{apeq:craft:optimality_fun}
\end{equation}

By stacking the KKT conditions of the NNLS problems the we obtain the so-called \textit{optimality function} $\bm{F}$:

\begin{equation} %
    \bm{F}((\m{U},\m{W},\bar{\m{U}},\bar{\m{W}}),\m{A})=\begin{cases}
    (\m{U}\m{W}^T-\m{A})\m{W}-\bar{\m{U}}    ,& \\ %
    (\m{W}\m{U}^T-\m{A}^T)\m{U}-\bar{\m{W}}  ,& \\ %
    \bar{\m{U}} \odot \m{U}   ,& \\ %
    \bar{\m{W}} \odot \m{W}   .& \\ %
\end{cases}
\label{eq:craft:optimality_fun_2}
\end{equation}

The implicit function theorem~\cite{griewank2008evaluating} allows us to use implicit differentiation~\cite{krantz2002implicit,griewank2008evaluating,bell2008algorithmic} to efficiently compute the Jacobians $\frac{\partial \m{U}}{\partial \m{A}}$ and $\frac{\partial \m{W}}{\partial \m{A}}$ without requiring to back-propagate through each of the iterations of the NMF solver:
\begin{equation}
    \frac{\partial (\m{U},\m{W},\bar{\m{U}},\bar{\m{W}})}{\partial \m{A}}=-(\partial_1 \bm{F})^{-1}\partial_2 \bm{F}.
\end{equation}

Implicit differentiation requires access to the dual variables of the optimization problem in equation~\ref{eq:craft:nmf}, which are not computed by Scikit-learn's popular implementation. Scikit-learn uses Block coordinate descent algorithm~\cite{cichocki2009fast,fevotte2011algorithms}, with a randomized SVD initialization. Consequently, we leverage our implementation in Jax based on ADMM~\cite{boyd2011distributed}.

Concretely, we perform a two-stage backpropagation \textit{Jax (2)}$\to$\textit{Tensorflow (1)} to leverage the advantage of each framework. The lower stage (1) corresponds to feature extraction $\m{A}=\v{h}_l(\vx)$ from crops of images $\vx$, and upper stage (2) computes NMF $\m{A} \approx \m{U}\m{W}^T$.  
  
We use the \textit{Jaxopt}~\cite{blondel2021implicitdiff} library that allows efficient computation of $\frac{\partial (\m{U},\m{W},\bar{\m{U}},\bar{\m{W}})}{\partial \m{A}}=-(\partial_1 \bm{F})^{-1}\partial_2 \bm{F}$. The matrix $(\partial_1 \bm{F})^{-1}$ is never explicitly computed -- that would be too costly. Instead, the system $\partial_1 \bm{F}\frac{\partial (\m{U},\m{W},\bar{\m{U}},\bar{\m{W}})}{\partial \m{A}}=-\partial_2 \bm{F}$ is solved with Conjugate Gradient through the use of Jacobian Vector Products (JVP) $\bm{v}\mapsto (\partial_1 \bm{F})\bm{v}$.  
  
The chain rule yields:
$$\frac{\partial \m{U}}{\partial \vx}=\frac{\partial \m{A}}{\partial \vx}\frac{\partial \m{U}}{\partial \m{A}}.$$

Usually, most Autodiff frameworks (e.g Tensorflow, Pytorch, Jax) handle it automatically. Unfortunately, combining two of those framework raises a new difficulty since they are not compatible. Hence, we re-implement manually the two stages auto-differentiation.  
  
Since $r$ is far smaller ($r=25$ in all our experiments) than input dimension $\vx$ (typically $224\times 244$ for ImageNet images), back-propagation is the preferred algorithm in this setting over forward-propagation. We start by computing sequentially the gradients $\nabla_{\vx} \m{U}_i$ for all concepts $1\leq i\leq r$. This amounts to compute $\bm{v}=\nabla_{\m{A}} \m{U}_i$ with Implicit Differentiation in Jax, convert the Jax array $\bm{v}$ into Tensorflow tensor, and then to compute $\nabla_{\vx} \m{U}_i=\frac{\partial \m{A}}{\partial \vx}\nabla_{\m{A}} \m{U}_i=\nabla_{\vx} (\v{h}_l(\vx) \cdot \bm{v})$. The latter is easily done in Tensorflow. Finally we stack the gradients $\nabla_{\vx} \m{U}_i$ to obtain the Jacobian $\frac{\partial \m{U}}{\partial \vx}$.

\subsection{Sobol indices for concepts} \label{apdx:sobol}

We propose to formally derive the Sobol indices for the estimation of the importance of concepts.
Let us define a probability space  $(\Omega, \mathcal{F}, \mathbb{P})$ of possible concept perturbations. In order to build these concept perturbations, we start from an original vector of concepts coefficient\footnote{We choose to name $\vx$ the concept coefficient vector here, instead to avoid any confusion with $\v{u}$ that will be the set of indices.} $\v{x} \in \mathbb{R}^r$ and use i.i.d. stochastic masks $\rv{m} = (\r{m}_1, ..., \r{m}_r) \sim \mathcal{U}([0, 1]^r)$, as well as a perturbation operator $\bm{\tau}(\cdot)$ to create stochastic perturbation of $\vx$ that we call concept perturbation $\rvx = \bm{\tau}(\vx, \rv{m})$.

Concretely, to create our concept perturbation we consider the inpainting function as our perturbation operator (as in \cite{ribeiro2016lime, petsiuk2018rise, fel2021sobol}) : $\bm{\tau}(\vx, \rv{m}) = \vx \odot \rv{m} + (\bm{1} - \rv{m}) \mu$ with $\odot$ the Hadamard product and $\mu \in \mathbb{R}$ a baseline value, here zero.
For the sake of notation, we will note $\pred$ the function mapping a random concept perturbation $\rvx$ from an intermediat layer to the output $\pred(\rvx)$ (e.g., the final layer if we do the concept extraction on the penultimate layer).
We denote the set $\mathcal{U} = \{1, ..., r\}$, $\bm{u}$ a subset of $\mathcal{U}$, its complementary $\sim \bm{u}$ and $\mathbb{E}(\cdot)$ the expectation over the perturbation space.
Finally, we assume that $\pred \in \mathbb{L}^2(\mathcal{F}, \mathbb{P})$ i.e. $|\mathbb{E}(\pred(\rvx))| < + \infty$.

The Hoeffding decomposition allows us to express the function $\pred$ into summands of increasing dimension, denoting $\pred_{\bm{u}}$ the partial contribution of the concepts $\rvx_{\bm{u}} = (\r{x}_i)_{i\in \bm{u}}$ to the score $\pred(\rvx)$:
\begin{equation}
    \label{eq:craft:anova}
    \begin{aligned}
    \bm{f}(\rvx) &= \bm{f}_{\emptyset}
     + \sum_i^r \bm{f}_i(\r{x}_i)
     + \sum_{1 \leqslant i < j \leqslant r} \bm{f}_{i,j}(\r{x}_i, \r{x}_j)
     + \cdots 
     + \bm{f}_{1,...,r}(\rvx) \\
    &= \sum_{\substack{\bm{u} \subseteq \mathcal{U}}} \bm{f}_{\bm{u}}(\rvx_{\bm{u}}).
    \end{aligned}
\end{equation}

Eq.~\ref{eq:craft:anova} consists of $2^r$ terms and is unique under the following orthogonality constraint:
\begin{equation}
    \label{eq:craft:anova_ortho}
    \begin{aligned}
    \forall (\bm{u},\bm{v}) \subseteq \mathcal{U}^2 \; s.t. \;  \bm{u} \neq \bm{v}, \;\; \mathbb{E}\big(\bm{f}_{\bm{u}}(\rvx_{\bm{u}}) \bm{f}_{\bm{v}}(\rvx_{\bm{v}})\big) = 0.
    \end{aligned}
\end{equation}

Furthermore, orthogonality yields the characterization $\bm{f}_{\bm{u}}(\rvx_{\bm{u}}) = \mathbb{E}(\bm{f}(\rvx)|\rvx_{\bm{u}}) - \sum_{\bm{v}\subset \bm{u}}\bm{f}_{\bm{v}}(\rvx_{\bm{v}})$ and allows us to decompose the model variance as:
\begin{equation}
    \label{eq:craft:var_decomposition}
    \begin{aligned}
        \V(\pred(\rvx)) &= \sum_i^r \V(\bm{f}_i(\r{x}_i)) 
        +\sum_{1 \leqslant i < j \leqslant r} \V(\bm{f}_{i,j}(\r{x}_i, \r{x}_j))
        + ... + \V(\bm{f}_{1,...,r}(\rvx)) \\
        &=\sum_{\substack{\bm{u} \subseteq \mathcal{U}}} \V(\bm{f}_{\bm{u}}(\rvx_{\bm{u}})).
        \end{aligned}
\end{equation}

Building from Eq.~\ref{eq:craft:var_decomposition}, it is natural to characterize the influence of any subset of concepts $\bm{u}$ as its own variance w.r.t. the total variance. This yields, after normalization by $\V(\bm{f}(\rvx))$, the general definition of Sobol' indices.
\begin{definition}[Sobol indices~\cite{sobol1993sensitivity}]
The sensitivity index $\mathcal{S}_{\bm{u}}$ which measures the contribution of the concept set $\rvx_{\bm{u}}$ to the model response $\bm{f}(\rvx)$ in terms of fluctuation is given by:
\begin{equation}\label{eq:craft:sobol_indice}
\begin{aligned}
    \mathcal{S}_{\bm{u}}  &= \frac{ \V(\bm{f}_{\bm{u}}(\rvx_{\bm{u}})) }{ \V(\pred(\rvx)) }\\
    &= \frac{ \V(\mathbb{E}(\bm{f}(\rvx) | \rvx_{\bm{u}})) - \sum_{\bm{v}\subset \bm{u}}\V(\mathbb{E}(\bm{f}(\rvx) | \rvx_{\bm{v}} ))}{ \V(\bm{f}(\rvx)) }.
\end{aligned}
\end{equation}
\end{definition}

Sobol indices give a quantification of the importance of any subset of concepts with respect to the model decision, in the form of a normalized measure of the model output deviation from $\bm{f}(\rvx)$. Thus, Sobol indices sum to one : $\sum_{\bm{u} \subseteq \mathcal{U}} \mathcal{S}_{\bm{u}} = 1$. 

\vspace{2mm}
Furthermore, the framework of Sobol' indices enables us to easily capture higher-order interactions between features. Thus, we can view the Total Sobol indices defined in \ref{eq:craft:total_sobol} as the sum of of all the Sobol indices containing the concept $i$ : $\mathcal{S}^{T}_i = \sum_{\bm{u} \subseteq \mathcal{U}, i \in \bm{u}} \mathcal{S}_{\bm{u}}$. Concretely, we estimate the total Sobol indices using the Jansen estimator~\cite{janon2014asymptotic} and Quasi-Monte carlo Sequence (Sobol $LP_{\tau}$ sequence).

\clearpage

\subsection{Human experiments}\label{app:craft:human-exp}

We first describe how participants were enrolled in our studies, then the general experimental design they went through.

\subsubsection{Utility evaluation}
\label{app:craft:utility}

\paragraph{Participants}
The participants that went through our experiments are users from the online platform Amazon Mechanical Turk (AMT), specifically, we recruit users with high qualifications (number of HIT completed $=5 000$ and HIT accepted $> 98 \%$). All participants provided informed consent electronically in order to perform the experiment ($\sim 5 - 8$ min), for which they received 1.4\$.\\

For the \textit{Husky vs. Wolf} scenario, $n=84$ participants passed all our screening and filtering process, respectively $n=32$ for CRAFT, $n=22$ for ACE and $n=22$ for CRAFTCO.

For the \textit{Leaves} scenario, after filtering, we analyzed data from $n=87$ participants, respectively $n=32$ for CRAFT, $n=24$ for ACE and $n=31$ for CRAFTCO.

For the \textit{"Kit Fox" vs. "Red Fox"} scenario, the results come from $n=79$ participants who passed all our screening processes, respectively $n=22$ for CRAFT, $n=31$ for ACE and $n=26$ for CRAFTCO.

\paragraph{General study design}
We followed the experimental design described in \autoref{sec:meta_pred}, in which explanations are evaluated according to their ability to help training participants at getting better at predicting their models' decisions on unseen images.

Each of those participants are only tested on a single condition to avoid possible experimental confounds. 

The main experiment is divided into 3 training sessions (with 5 training samples in each) each followed by a brief test. In each individual training trial, an image was presented with the associated prediction of the model, together with an explanation. After a brief training phase (5 samples), participants' ability to predict the classifier's output was evaluated on 7 new samples during a test phase. During the test phase, no explanation was provided.
We also use the reservoir that subjects can refer to during the testing phase to minimize memory load as a confounding factor.

We implement the same 3-stage screening process: First we filter participants not successful at the practice session done prior to the main experiment used to teach them the task, then we have them go through a quiz to make sure they understood the instructions. Finally, we add a catch trial in each testing phase --that users paying attention are expected to be correct on-- allowing us to catch uncooperative participants.

\begin{figure*}[hb]
    \centering
    \begin{subfigure}{0.9\textwidth}
        \centering
        \includegraphics[width=0.45\textwidth]{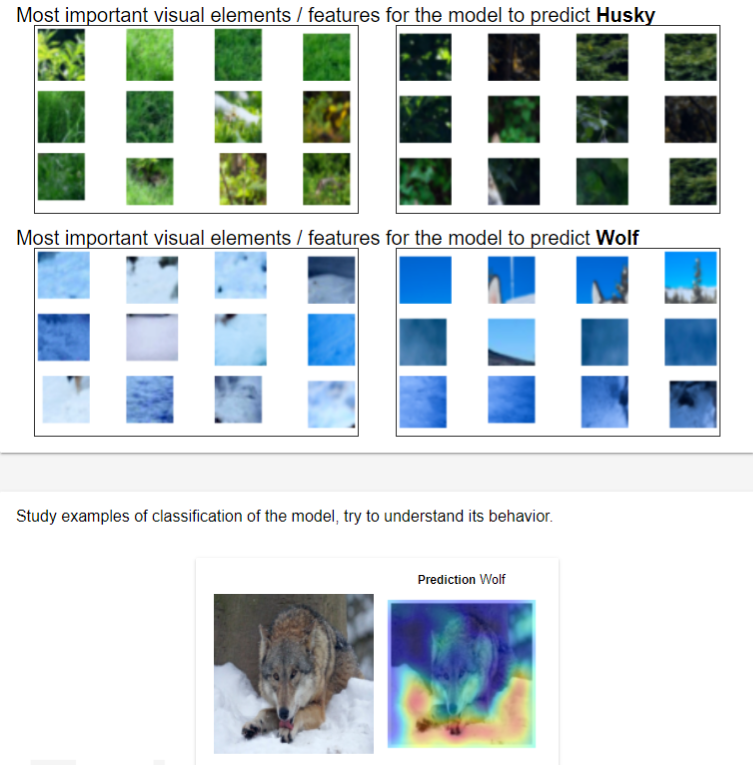}
        \includegraphics[width=0.45\textwidth]{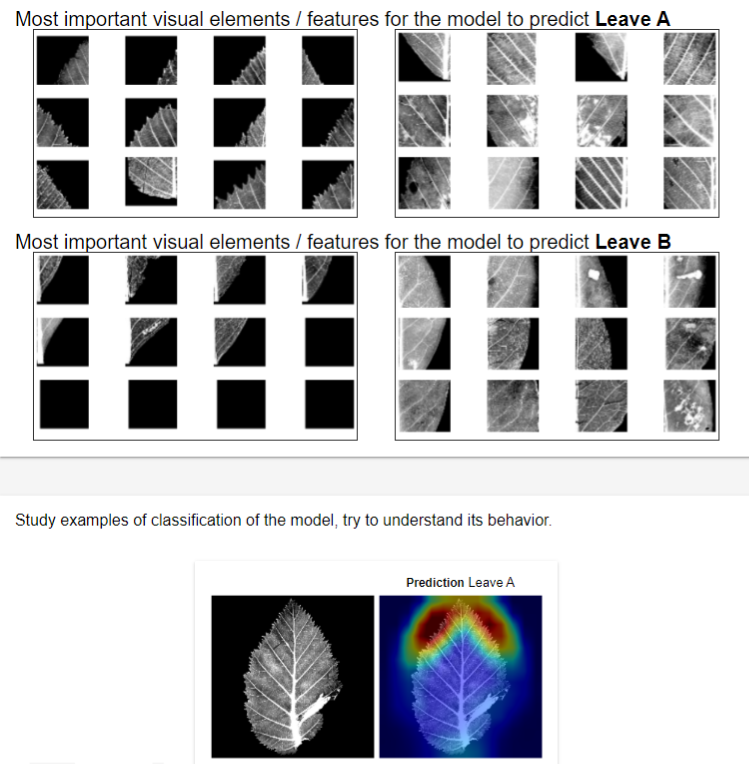}
        \caption{\textbf{Utility experiment.} Training trials taken from the \textit{Husky vs. Wolf} scenario (left) and the \textit{Leaves} scenario (right).}
        \label{fig:craft:website_utility}
    \end{subfigure}
\end{figure*}

\subsubsection{Validation of Recursivity}

\paragraph{Participants} Behavioral accuracy data were gathered from $n=73$ participants. All participants provided informed consent electronically in order to perform the experiment ($\sim 4 - 6$ min). The protocol was approved by the University IRB and was carried out in accordance with the provisions of the World Medical Association Declaration of Helsinki. 
For each of the 2 experiment tested, we had prepared filtering criteria for uncooperative people (namely based on time), but all participants passed these filters.

\paragraph{General study design}

\begin{figure*}[ht]
    \centering
    \begin{subfigure}{0.95\textwidth}
        \includegraphics[width=\textwidth]{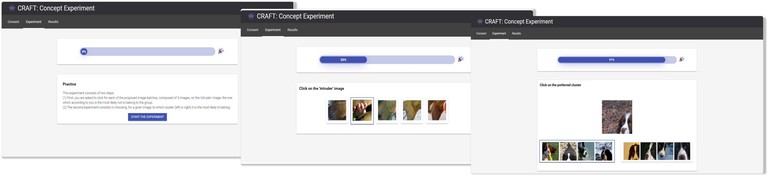}
        \caption{\textbf{Recursivity Experiment Website.}}
        \label{fig:craft:website}
      
    \end{subfigure}
    \begin{subfigure}{0.95\textwidth}
        \centering
        \includegraphics[width=0.32\textwidth]{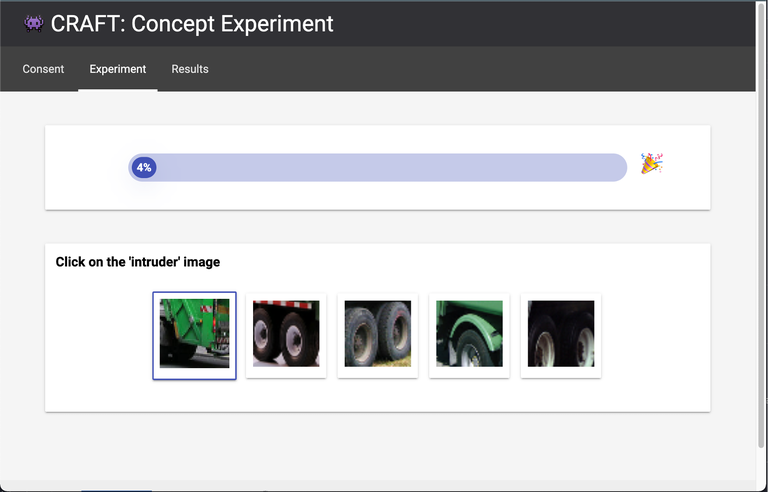}
        \includegraphics[width=0.32\textwidth]{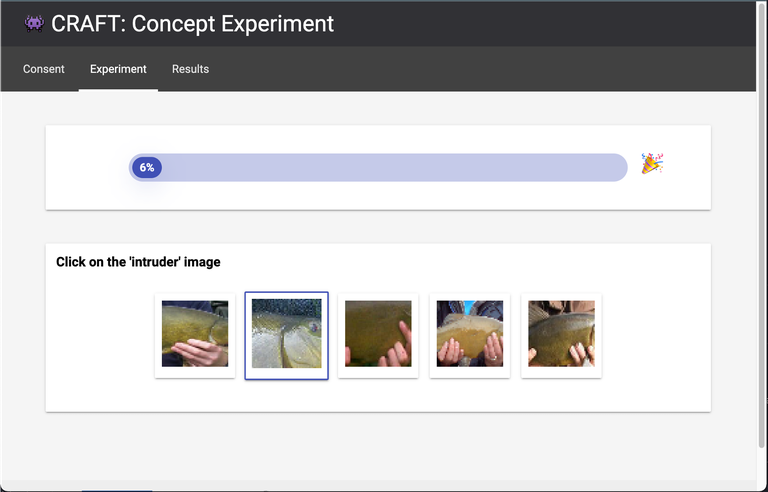}
        \includegraphics[width=0.32\textwidth]{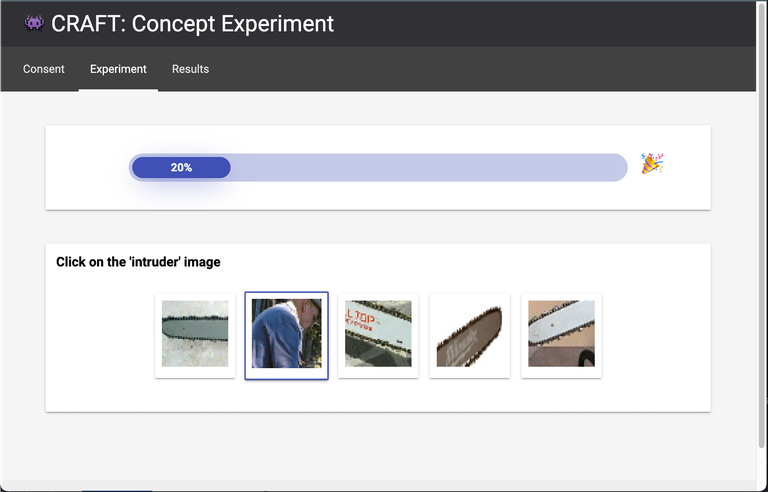}
        \caption{\textbf{Binary choice experiment.}}
        \label{fig:craft:website_intruder}
    \end{subfigure}
    
    \begin{subfigure}{0.95\textwidth}
        \centering
        \includegraphics[width=0.32\textwidth]{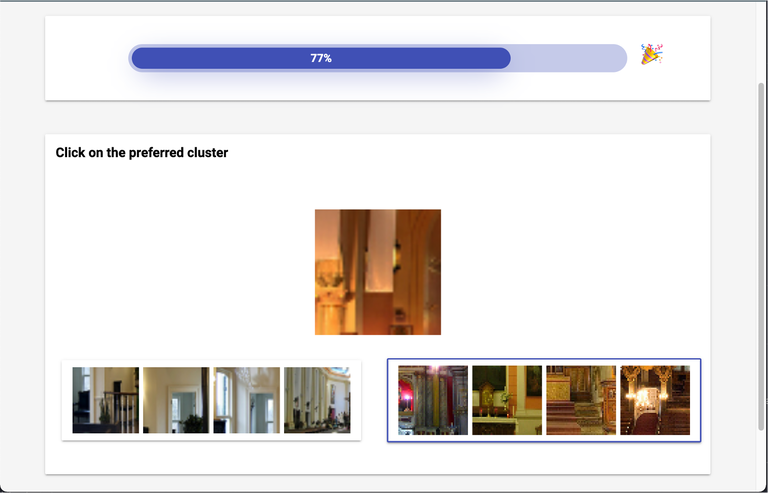}
        \includegraphics[width=0.32\textwidth]{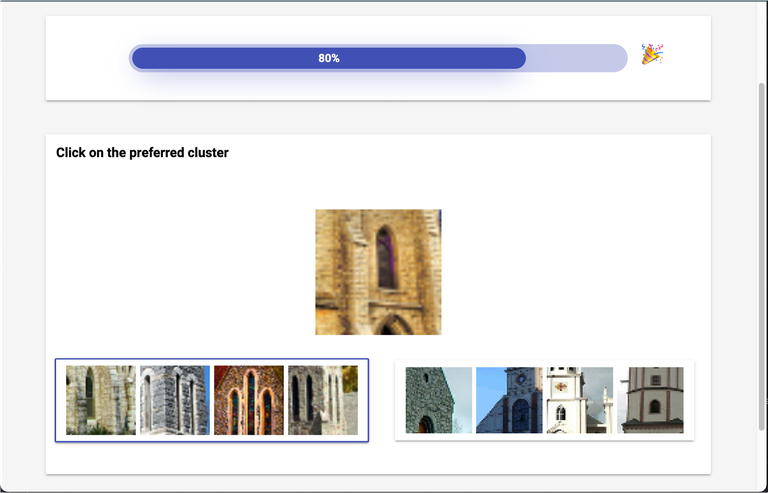}
        \includegraphics[width=0.32\textwidth]{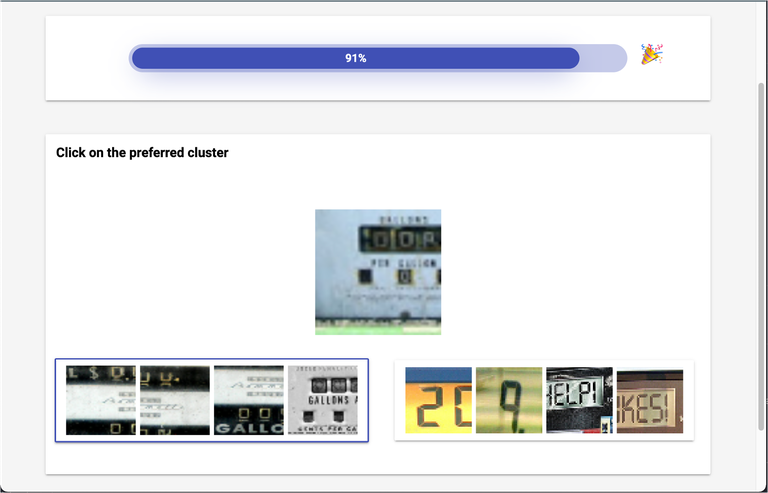}
        \caption{\textbf{Intruder experiment.}}
        \label{fig:craft:website_choice}
    \end{subfigure}
\end{figure*}

For the first experiment -- consisting in finding the intruder among elements of the same concept and an element from a different concept (but of the same class, see Figure~\ref{fig:craft:website_intruder}) -- the order of presentation is randomized across participants so that it does not bias the results.
Moreover, in order to avoid any bias coming from the participants themselves (one group being more successful than the other) all participants went through both conditions of finding intruders in batches of images coming from either concepts or sub-concepts.
Concerning experiment 2, the order was also randomized (see Figure~\ref{fig:craft:website_choice}).

The participants had to successively find 30 intruders (15 block concepts and 15 block sub-concepts) for experiment 1 and then make 15 choices (sub-concept vs concept) for experiment 2, see Figure~\ref{fig:craft:website}.

The expert participants are people working in machine learning (researchers, software developers, engineers) and have participated in the study following an announcement in the authors' laboratory/company. The other participants (Laymen) have no expertise in machine learning.

\clearpage

\subsection{Fidelity experiments}\label{app:craft:fidelity}

\begin{figure}[ht]
  \includegraphics[width=.99\linewidth]{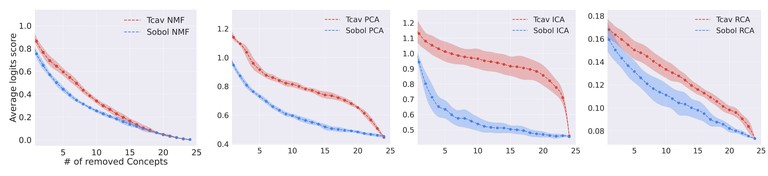}
  \includegraphics[width=.99\linewidth]{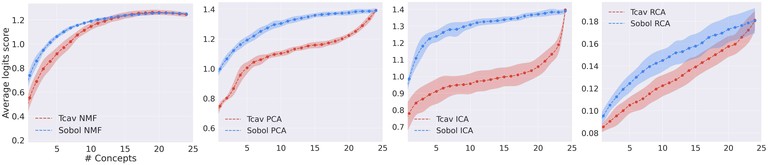}
  \caption{ \textbf{(1) Deletion curves} for different concept extraction methods, Sobol outperforms TCAV not only for NMF to correctly estimate concept importance (lower is better). \textbf{(2) Insertion curves} for different concept extraction methods, Sobol outperforms TCAV to correctly estimate concept importance (higher is better).}
  \label{fig:craft:deletion_full}
\end{figure}

For our experiments on the concept importance measure, we focused on certain classes of ILSRVC2012~\cite{imagenet_cvpr09} and used a ResNet50V2~\cite{he2016deep} that had already been trained on this dataset. Just like in~\cite{ghorbani2017interpretation, zhang2021invertible}, we measure the insertion and deletion metrics for our concept extraction technique -- as well as concepts vectors extracted using PCA, ICA and RCA as dimensionality reduction algorithms, see Figure~\ref{fig:craft:deletion_full} -- and we compare them when we add/remove the concepts as ranked by the TCAV score~\cite{kim2018interpretability} and by the Sobol importance score. As originally explained in~\cite{petsiuk2018rise}, the objective of these metrics is to add/remove parts of the input according to how much an explainability method considers that it is influential and looking at the speed at which the logit for the predicted class increases/decreases.

In particular, for our experimental evaluations, we have randomly chosen 100000 images from ILSVRC2012~\cite{imagenet_cvpr09} and computed the deletion and insertion metrics for 5 different seeds -- for a total of half a million images. In Figure~\ref{fig:craft:deletion_full}, the shade around the curves represent the standard deviation over these 5 experiments.

\clearpage

\subsection{Sanity Check}
\label{app:craft:sanity-checks}

\begin{figure}[h]
    \centering
    \includegraphics[width=0.95\linewidth]{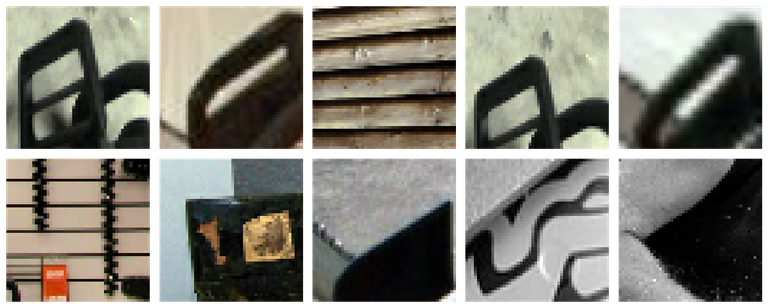}
    \includegraphics[width=0.95\linewidth]{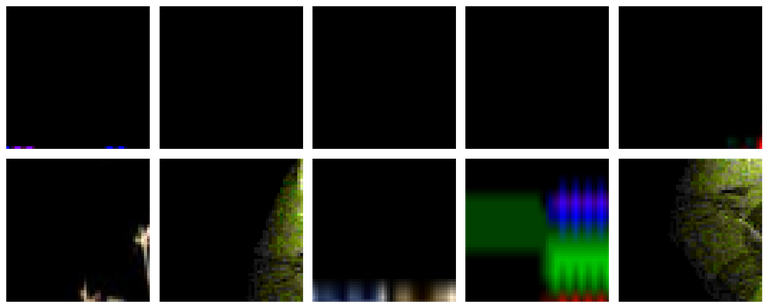}
    \includegraphics[width=0.95\linewidth]{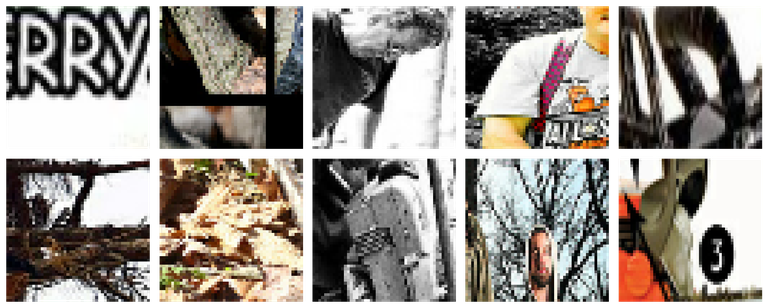}
    \caption{\textbf{Sanity check of the method:} we ran the method on a Resnet50 with randomized weights, and extracted the 3 most relevant concepts for the class `Chain saw'. When weights are randomized, concepts are mainly based on color histograms.}
    \label{fig:craft:sanity_check}
\end{figure}

Following the work from~\cite{adebayo2018sanity}, we performed a sanity check on our method, by running the concept extraction pipeline on a randomized model. This procedure was performed on a ResNet-50v2 model with randomized weights. As showcased in Figure~\ref{fig:craft:sanity_check}, the concepts drastically differ from trained models, thus proving that CRAFT passes the sanity check.

\section{Holistic}
\subsection{Attribution methods for Concepts}\label{sup:holistic:all_cams}

In the following section, we will re-derive the different attribution methods in the literature. We use the Xplique library and adapted each methods~\cite{fel2022xplique}.
We quickly recall that we seek to estimate the importance of each concept for a set of concept coefficients $\v{u} = (\v{u}_1, \ldots, \v{u}_k) \in \mathbb{R}^k$ in the concept basis $\m{V} \in \mathbb{R}^{p \times k}$. This concept basis is a re-interpretation of a latent space (in $\mathbb{R}^{p}$) and the function $\fb: \mathbb{R}^{p} \to \mathbb{R}$ is a signal used to compute importance from (e.g., logits value, cosine similarity with a sentence...). Each Attributions method will map a set of concept values to an importance score $\cam: \mathbb{R}^k \to \mathbb{R}^k$, a greater score $\cam(\v{u})_i$ indicates that a concept $\v{u}_i$ is more important. 

\textbf{Saliency (SA)}~\cite{simonyan2013deep} was originally a visualization technique based on the gradient of a class score relative to the input, indicating in an infinitesimal neighborhood, which pixels must be modified to most affect the score of the class of interest. In our case, it indicates which concept in an infinitesimal neighborhood has the most influence on the output:

$$ \cam^{(SA)}(\v{u}) = \nabla_{\v{u}} \fb(\v{u} V^\tr) .$$

\textbf{Gradient $\odot$ Input (GI)}~\cite{shrikumar2017learning} is based on the gradient of a class score relative to the input, element-wise with the input, it was introduced to improve the sharpness of the attribution maps. A theoretical analysis conducted by~\cite{ancona2017better} showed that Gradient $\odot$ Input is equivalent to $\epsilon$-LRP and DeepLIFT~\cite{shrikumar2017learning} methods under certain conditions -- using a baseline of zero, and with all biases to zero. In our case, it boils down to:

$$ \cam^{(GI)}(\v{u}) = \v{u} \odot \nabla_{\v{u}} \fb(\v{u} \m{V}^\tr) .$$

\textbf{Integrated Gradients (IG)}~\cite{sundararajan2017axiomatic} consists of summing the gradient values along the path from a baseline state to the current value. The baseline $\v{u}_0$ used is zero. This integral can be approximated with a set of $m$ points at regular intervals between the baseline and the point of interest. In order to approximate from a finite number of steps, we use a trapezoidal rule and not a left-Riemann summation, which allows for more accurate results and improved performance (see~\cite{sotoudeh2019computing} for a comparison). For all the experiments $m = 30$.

$$ \cam^{(IG)}(\v{u}) = (\v{u} - \v{u}_0) \int_0^1 \nabla_{\v{u}} \fb((\v{u}_0 + \alpha(\v{u} - \v{u}_0))\m{V}^\tr) \dif\alpha. $$

\textbf{SmoothGrad (SG)}~\cite{smilkov2017smoothgrad} is also a gradient-based explanation method, which, as the name suggests, averages the gradient at several points corresponding to small perturbations (drawn i.i.d from an isotropic normal distribution of standard deviation $\sigma$) around the point of interest. The smoothing effect induced by the average helps to reduce the visual noise, and hence improves the explanations. In our case, the attribution is obtained after averaging $m$ points with noise added to the concept coefficients. For all the experiments, we took $m = 30$ and $\sigma = 0.1$.

$$ \cam^{(SG)}(\v{u}) = \underset{\bm{\delta} \sim \mathcal{N}(0, \mathbf{I}\sigma)}{\mathbb{E}}(\nabla_{\v{u}} \fb( \v{u} + \bm{\delta}) ).
$$

\textbf{VarGrad (VG)}~\cite{hooker2018benchmark} was proposed as an alternative to SmoothGrad as it employs the same methodology to construct the attribution maps: using a set of $m$ noisy inputs, it aggregates the gradients using the variance rather than the mean. For the experiment, $m$ and $\sigma$ are the same as SmoothGrad. Formally:

$$ \cam^{(VG)}(\v{u}) = \underset{\bm{\delta} \sim \mathcal{N}(0, \mathbf{I}\sigma)}{\mathbb{V}}(\nabla_{\v{u}} \fb( \v{u} + \bm{\delta}) ).
$$

\textbf{Occlusion (OC)}~\cite{zeiler2013visualizing} is a simple -- yet effective -- sensitivity method that sweeps a patch that occludes pixels over the images using a baseline state and use the variations of the model prediction to deduce critical areas. In our case, we simply omit each concept one-at-a-time to deduce the concept's importance. For all the experiments, the baseline state $\v{u}_0$ was zero.

$$ \cam^{(OC)}(\v{u})_i = \fb(\v{u} \m{V}^\tr) - \fb(\v{u}_{[i = \v{u}_0]} \m{V}^\tr)  $$

\textbf{Sobol Attribution Method (SM)}~\cite{fel2021sobol} then used for estimating concept importance in \cite{fel2023craft} is a black-box attribution method grounded in Sensitivity Analysis. Beyond modeling the individual contributions of image regions, Sobol indices provide an efficient way to capture higher-order interactions between image regions and their contributions to a neural network’s prediction through the lens of variance. In our case, the score for a concept $\v{u}_i$ is the expected variance that would be left if all variables but $i$ were to be fixed : 

$$ \cam^{(SM)}(\v{u})_i = \frac{ \mathbb{E}( \mathbb{V}( \fb( (\v{u} \odot \mathbf{M} ) \m{V}^\tr ) | \mathbf{M}_{\sim i} ) ) }{ \mathbb{V}( \fb( (\v{u} \odot \mathbf{M} ) \m{V}^\tr)) } . $$

With $\mathbf{M} \sim \mathcal{U}([0, 1])^k$. For all the experiments, the number of designs was $32$ and we use the Jansen estimator of the Xplique library.

\textbf{HSIC Attribution Method (HS)}~\cite{novello2022making} seeks to explain a neural network's prediction for a given input image by assessing the dependence between the output and patches of the input. In our case, we randomly mask/remove concepts and measure the dependence between the output and the presence of each concept through $N$ binary masks. Formally:

$$ \cam^{(HS)}(\v{u}) = \frac{1}{(N-1)^2} \mathrm{Tr}(KHLH). $$

With $H, L, K \in \mathbb{R}^{N \times N}$ and $K_{ij} = k(\mathbf{M}_i, \mathbf{M}_j)$, $L_{ij} = l(\vy_i, \vy_j)$ and $H_{ij} = \delta(i=j)-N^{-1}$. Here, $k(\cdot, \cdot)$ and $l(\cdot, \cdot)$ denote the chosen kernels and $\mathbf{M} \sim \{0, 1\}^p$ the binary mask applied to the input $\v{u}$.

\textbf{RISE (RI)}~\cite{petsiuk2018rise} is also a black-box attribution method that probes the model with multiple version of a masked input to model the most important features. Formally, with $\bm{m} \sim \mathcal{U}([0, 1])^k$. : 

$$ \cam^{(RI)}_i(\v{u}) =  
\mathbb{E}(\fb( \v{u} \odot \bm{m} ) | \bm{m}_i = 1).
$$

\subsection{Closed-form of Attributions for the last layer}\label{sup:holistic:closed_form}

Without loss of generality, we focus on the decomposition in the last layer, that is $\v{a} = \v{u}\m{V}^\tr$ with parameters $(\m{W}, \bias)$ for the weight and the bias respectively, hence we obtain $\vy = (\v{u} \m{V}^\tr)\m{W} + \bias$ with $\m{W} \in \mathbb{R}^{p}$ and $\bias \in \mathbb{R}$.

We start by deriving the closed form of Saliency (SA) and naturally Gradient-Input (GI):

\begin{flalign*}
\cam^{(SA)}(\v{u}) 
&= \nabla_{\v{u}} \fb(\v{u} \m{V}^\tr)
= \nabla_{\v{u}} (\v{u} \m{V}^\tr \m{W} + \bias) &\\
&= \m{W}^\tr \m{V}&.
\end{flalign*}
\begin{flalign*}
\cam^{(GI)}(\v{u}) 
&= \nabla_{\v{u}} \fb(\v{u} \m{V}^\tr) \odot \v{u} 
= \nabla_{\v{u}} (\v{u} \m{V}^\tr \m{W} + \bias) \odot \v{u} &\\
&= \m{W}^\tr \m{V} \odot \v{u} &.
\end{flalign*}

We observe two different forms that will in fact be repeated for the other methods, for example with Integrated-Gradient (IG) which will take the form of Gradient-Input, while SmoothGrad (SG) will take the form of Saliency.

\begin{flalign*}
\cam^{(IG)}(\v{u})
 &= (\v{u} - \v{u}_0) \odot \int_0^1 \nabla_{\v{u}} \fb((\v{u}_0 + \alpha (\v{u} - \v{u}_0)) \m{V}^\tr) \dif \alpha &\\
 &= \v{u} \odot \int_0^1 \nabla_{\v{u}}((\alpha \v{u})) \m{V}^\tr\m{W} + \bias + (\alpha-1)\v{u}_0\m{V}^\tr\m{W}) \dif \alpha &\\
 &= \v{u} \odot \int_0^1 \alpha\m{W}^\tr \dif \alpha = \v{u} \odot \m{W}^\tr \m{V} \left[\frac{1}{2}\alpha^2\right]_0^1\\
 &= \frac{1}{2}\v{u} \odot \m{W}^\tr \m{V}.
\end{flalign*}

\begin{flalign*}
\cam^{(SG)}(\v{u})
&= \underset{\bm{\delta} \sim \mathcal{N}(0, \mathbf{I}\sigma)}{\mathbb{E}}(\nabla_{\v{u}} \fb( \v{u} + \bm{\delta}) ) 
= \underset{\bm{\delta} \sim \mathcal{N}(0, \mathbf{I}\sigma)}{\mathbb{E}}(\nabla_{\v{u}}( (\v{u} + \bm{\delta}) \m{V}^\tr\m{W} + \bias) ) & \\
& = \underset{\bm{\delta} \sim \mathcal{N}(0, \mathbf{I}\sigma)}{\mathbb{E}}(\nabla_{\v{u}}(\v{u} \m{V}^\tr\m{W})) & \\
& = \m{W}^\tr \m{V} &.
\end{flalign*}

The case of VarGrad is specific, as the gradient of a linear system being constant, its variance is null.

\begin{flalign*}        
\cam^{(VG)}(\v{u})
&= \underset{\bm{\delta} \sim \mathcal{N}(0, \mathbf{I}\sigma)}{\mathbb{V}}(\nabla_{\v{u}} \fb( \v{u} + \bm{\delta}) )
= \underset{\bm{\delta} \sim \mathcal{N}(0, \mathbf{I}\sigma)}{\mathbb{V}}(\nabla_{\v{u}} ( (\v{u} + \bm{\delta}) \m{V}^\tr \m{W} + \bias) ) & \\
&= \underset{\bm{\delta} \sim \mathcal{N}(0, \mathbf{I}\sigma)}{\mathbb{V}}(\m{W}^\tr \m{V}) &\\
&= 0&.
\end{flalign*}

Finally, for Occlusion (OC) and RISE (RI), we fall back on the Gradient Input form (with multiplicative and additive constant for RISE).

\begin{flalign*}
\cam^{(OC)}_i(\v{u})
&= \fb(\v{u} \m{V}^\tr) - \fb(\v{u}_{[i = \v{u}_0]} \m{V}^\tr)
= \v{u} \m{V}^\tr\m{W} + \bias - (\v{u}_{[i = \v{u}_0]} \m{V}^\tr\m{W} + \bias) & \\
& = (\sum_{j}^{r} \v{u}_j \m{V}_j^\tr)\m{W} - (\sum_{j \neq i}^{r} \v{u}_j \m{V}_j^\tr)\m{W} &\\
& = \v{u}_i \m{V}_i^\tr \m{W} &
\end{flalign*}
thus $\cam^{(OC)}(\v{u}) = \v{u} \odot \m{W}^\tr \m{V}$

\begin{flalign*}
\cam^{(RI)}_i(\v{u})
&= \mathbb{E}(\fb( \v{u} \odot \bm{m} ) | \bm{m}_i = 1)
= \mathbb{E}( (\v{u}\odot\bm{m}) \m{V}^\tr\m{W} + \bias | \bm{m}_i = 1) & \\
& = \bias + \sum_{j \neq i}^r \v{u}_j \mathbb{E}(\bm{m}_j) \m{V}_j^\tr\m{W} + \v{u}_i \m{V}_i^\tr\m{W} &\\
& = \bias + \frac{1}{2} (\v{u}\m{V}^\tr\m{W} + \v{u}_i \m{V}_i^\tr\m{W})&
\end{flalign*}

\subsection{Fidelity optimality}\label{sup:holistic:fidelity_theorem}

Before showing that some methods are optimal with regard to C-Deletion and C-Insertion, we start with a first metric that studies the fidelity of the importance of concepts: $\mu$Fidelity, whose definition we recall

$$
\mu F = \underset{\substack{S \subseteq \{1, \ldots, k\} \\ |S| = m} }{\rho}(
\sum_{i \in S} \cam(\v{u})_i,
\fb(\v{u}) - \fb(\v{u}_{[\v{u}_i = \v{u}_0, i \in S]})
)
$$

With $\rho$ the Pearson correlation and $\v{u}_{[\v{u}_i = \v{u}_0, i \in S]}$ means that all $i$ components of $\v{u}$ are set to zero.

\begin{theorem}[Optimal $\mu$Fidelity in the last layer]
When decomposing in the last layer,~\textbf{Gradient Input}, \textbf{Integrated Gradients}, \textbf{Occlusion}, and \textbf{Rise} yield the optimal solution for the $\mu$Fidelity metric.
In a more general sense, any method $\cam(\v{u})$ that is of the form
$\cam_{i}(\v{u}) = a (\v{u}_i\m{V}_i^\tr \m{W}) + b $ with $a \in \mathbb{R}^+, b \in \mathbb{R}$ yield the optimal solution, thus having a correlation of 1.
\end{theorem}
\begin{proof}
In the last layer case, $\mu$Fidelity boils down to:

\begin{flalign*}
\mu F &= \underset{\substack{S \subseteq \{1, \ldots, k\} \\ |S| = m} }{\rho}\big(\sum_{i \in S} \cam(\v{u})_i,
\v{u} \m{V}^\tr \m{W} + \bias - ( \sum_{i \notin S} \v{u}_i \m{V}_i^\tr \m{W}) - \bias
\big) & \\
&= \underset{\substack{S \subseteq \{1, \ldots, k\} \\ |S| = m} }{\rho}\big(\sum_{i \in S} \cam(\v{u})_i,
\sum_{i \in S} \v{u}_i \m{V}_i^\tr \m{W}
\big) &
\end{flalign*}

We recall that for \textbf{Gradient Input}, \textbf{Integrated Gradients}, \textbf{Occlusion}, $\cam_i(\v{u}) \propto \v{u}_i \m{V}_i^\tr \m{W}$, thus 
\begin{flalign*}
\mu F &= \underset{\substack{S \subseteq \{1, \ldots, k\} \\ |S| = m} }{\rho}\big(
\sum_{i \in S} \v{u}_i \m{V}_i^\tr \m{W},
\sum_{i \in S} \v{u}_i \m{V}_i^\tr \m{W}
\big) = 1 &
\end{flalign*}
For \textbf{RISE}, we get the following characterization:
\begin{flalign*}
\mu F &= \underset{\substack{S \subseteq \{1, \ldots, k\} \\ |S| = m} }{\rho}\big(
\sum_{i \in S} \bias + \frac{1}{2} (\v{u}\m{V}^\tr\m{W} + \v{u}_i \m{V}_i^\tr\m{W})
,
\sum_{i \in S} \v{u}_i \m{V}_i^\tr \m{W}
\big) & \\
&= \underset{\substack{S \subseteq \{1, \ldots, k\} \\ |S| = m} }{\rho}\big(
|S|(\bias + \frac{1}{2} (\v{u}\m{V}^\tr\m{W})) + 
\sum_{i \in S} \frac{1}{2} \v{u}_i \m{V}_i^\tr\m{W}
,
\sum_{i \in S} \v{u}_i \m{V}_i^\tr \m{W}
\big) & \\
&= \underset{\substack{S \subseteq \{1, \ldots, k\} \\ |S| = m} }{\rho}\big(
a(  
\sum_{i \in S} \v{u}_i \m{V}_i^\tr\m{W}) + b
,
\sum_{i \in S} \v{u}_i \m{V}_i^\tr \m{W}
\big)  = 1 & \\
\end{flalign*}

with $a = \frac{1}{2}, b = m(\bias + \frac{1}{2} (\v{u}\m{V}^\tr\m{W}))$. 

\end{proof}

\subsection{Optimality for C-Insertion and C-Deletion}\label{sup:holistic:matroid}

In order to prove the optimality of some attribution methods on the C-Insertion and C-Deletion metrics, we will use the Matroid theory of which we recall some fundamentals.

Matroids were introduced by Whitney in 1935~\cite{whitney1992abstract}. 
It was quickly realized that they unified properties of various domains such as graph theory, linear algebra or geometry. 
Later, in the '60s, a connection was made with combinatorial optimization, nothing that they also played a central role in combinatorial optimization. 

The power of this tool is that it allows us to show easily that greedy algorithms are optimal with respect to some criterion on a broad range of problems. Here, we show that insertion is a greedy algorithm (since the concepts inserted are chosen sequentially based on the model score).

For the rest of this section, we assume $E = \{ e_1, \ldots, e_k \}$ the set of the canonical vectors in $\mathbb{R}^k$, with $e_i$ being the element associated with the $i^{th}$ concept.

\begin{definition}[Matroid] A matroid $M$ is a tuple $(E, \mathcal{J})$, where E is a finite ground set and $\mathcal{J} \subseteq 2^E$ is the power set of $E$, a collection of independent sets, such that:

\begin{enumerate}
  \item $\mathcal{J}$ is nonempty, $\emptyset \in \mathcal{J}$.
  \item $\mathcal{J}$ is downward closed; i.e., if $S \in \mathcal{J}$ and $S' \subseteq S$, then $S' \in \mathcal{J}$ 
  \item If $S, S' \in \mathcal{J}^2$ and $|S| < |S'|$, then $\exists s \in S' \setminus S$ such that $S \cup \{s\} \in \mathcal{J}$
\end{enumerate}

\end{definition}

In particular, we will need uniform matroids: 

\begin{definition}[Uniform Matroid] 
\label{def:matroid}
Let $E$ be a set of size $k$ and let $n \in \{1, \ldots, k \}$. If $\mathcal{J}$ is the collection of all subsets of $E$ of size at most $n$, then $(E, \mathcal{J})$ is a matroid, called a uniform matroid and denoted $M^{(n)}$.
\end{definition}

Finally, we need to characterize the concept set chosen at each step.

\begin{definition}[Base of Matroid] 
Let $M = (E, \mathcal{J})$ be a matroid. A subset $B$ of $E$ is called a basis of $M$ if and only if:
\begin{enumerate}
  \item $B \in \mathcal{J}$
  \item $\forall e \in E \setminus B, ~ B \cup \{e\} \notin \mathcal{J}$
\end{enumerate}
Moreover, we denote $\mathcal{B}(M)$ the set of all the basis of $M$.
\end{definition}

At each step, the insertion metric selects the concepts of maximum score given a cardinality constraint. At each new step, the concepts from the previous step are selected and it add a new concept from the whole available set, the one not selected so far with the highest score.  
This criterion requires an additional ingredient: the \emph{weight} associated to each element of the matroid - here an element of the matroid is a concept.

\paragraph{Ponderated Matroid}

Let $M^{(n)} = (E, \mathcal{J})$ be a uniform matroid and $w : E \to \mathbb{R}$ a weighting function associated to an element of $E$ (a concept).
The goal of C-Insertion at step $n$ is to find a basis (a set of concepts) $B^\star$ subject to $|B| = n$, that maximizes the weighting function : 

$$
\forall B \in \mathcal{J}, ~~ \sum_{e \in B^\star} w(e) \geq \sum_{e \in B} w(e).
$$

Such a basis is called the basis of maximum weights (MW) of the weighted matroid $M^{(n)}$. We will see that the greedy algorithm associated with this weighting function gives the optimal solution to the MW problem on C-Insertion. First, let's define the \emph{Greedy algorithm}.

\begin{algorithm}[ht]
\caption{Greedy algorithm}\label{alg:greedy_matroide}
\begin{algorithmic}
  \REQUIRE A $n$-uniform weighted matroid $M^{(n)} = (E, \mathcal{J}, w)$
  \STATE Sort the concepts by their weight $w(e_i)$ in non-increasing order, and store them in a list $\bar{e}$ such that~${\forall (i, j) \subseteq \{1, \ldots, k\}^2, w(\bar{e}_i) \geq w(\bar{e}_j) ~ \text{if} ~ i < j}$.
  \STATE $B^{\star} = \{\}$
  \FOR{$k = 1$ to $n$}
    \STATE $B^{\star} = B^{\star} \cup \bar{e}_k$ %
  \ENDFOR
  \STATE \textbf{Return} $B^{\star}$
\end{algorithmic}
\end{algorithm}

\begin{theorem}[Greedy Algorithm is an optimal solution to MW.] Let $M = (E, \mathcal{J}, w)$ a weighted matroid. The greedy Algorithm~\ref{alg:greedy_matroide} returns a maximum basis of $M$.
\end{theorem}

\begin{proof}
First, by definition, $B^\star$ is a basis and thus an independent set, i.e., $B^\star \in \mathcal{B}(M)$ (as $\forall (e,e') \in E^2, ~ \langle e,e' \rangle = 0$).
Now, suppose by contradiction that there exists a base $B'$ with a weight strictly greater than $B^\star$. We will obtain a contradiction with respect to the augmentation axiom of the matroid definition.
Let $e_1, \ldots, e_k$ be the elements of $M$ sorted such that $w(e_i) > w(e_j)$ whenever $i < j$. 
Let $n$ be the rank of our weighted uniform matroid $M^{(n)}$. 
Then we can write $B^\star = (e_{i_1}, \ldots, e_{i_n})$ and $B' = (e_{j_1}, \ldots, e_{j_n})$ with $j_k < j_l$ and $i_k < i_l$ for any $k < l$.

Let $\ell$ be the smallest positive integer such that $i_\ell$ > $j_\ell$. In particular, $\ell$ exists and is at most $n$ by assumption. Consider the independent set $S_{\ell-1} = \{e_{i_1}, \ldots e_{\ell-1}\}$ (in particular, $S_{\ell-1} = \emptyset$ if $\ell =1$). According to the augmentation axiom (Definition \ref{def:matroid}, I3), there exist $k \in \{1, \ldots, \ell \}$ such that $S_{\ell-1} + e_{j_k} \in \mathcal{J}$ and $e_{j_k} \notin S_{\ell-1}$. However, $j_k \leq j_\ell < i_\ell$, thus $w(e_{j_k}) \leq w(e_{j_\ell}) <w(e_{i_\ell})$. This contradicts the definition of the greedy algorithm.
\end{proof}

Now, we notice that for the last layer, Insertion is a weighted matroid. We insist that this result is \emph{only true for the concepts in the penultimate layer}, as our demonstrations rely on the linearity of the decomposition. Here, the weight is given by the score of the model, which is a linear combination of concepts.

\begin{theorem}[Optimal Insertion in the last layer]
When decomposing in the last layer,~\textbf{Gradient Input}, \textbf{Integrated Gradients}, \textbf{Occlusion}, and \textbf{Rise} yield the optimal solution for the C-Insertion metric.
In a more general sense, any method $\cam(\v{u})$ that  satisfies the condition 
$\forall (i, j) \in \{1, \ldots, k\}^2, 
(\v{u} \odot \e_i) \m{V}^\tr\m{W} \geq (\v{u} \odot \e_j) \m{V}^\tr \m{W}
\implies 
\cam(\v{u})_i \geq \cam(\v{u})_j 
$ yield the optimal solution.
\end{theorem}

\begin{proof}
Each $n$ step of the C-Insertion algorithm corresponds to the $n$-uniform weighted matroid with weighting function $w(e_i) = (\v{u} \odot e_i) \m{V}^\tr\m{W} + b = \v{u}_i \m{V}^\tr\m{W} + b$. Therefore, any $\cam(\cdot)$ method that produces the same ordering as $w(\cdot)$ will yield the optimal solution. 
It easily follows that \textbf{Gradient Input}, \textbf{Integrated Gradients}, \textbf{Occlusion} are optimal as they all boil down to $\cam_i(\v{u}) = \v{u}_i \m{V}^\tr\m{W}+b$.
Concerning RISE, suppose that $w(e_i) \geq w(e_j)$, then $\v{u}_i \m{V}_i^\tr\m{W} + b \geq \v{u}_j \m{V}_j^\tr\m{W} + b$, and  
$\cam_i^{(RI)}(\v{u}) - \cam_j^{(RI)}(\v{u})
= \bias + \frac{1}{2} (\v{u}\m{V}^\tr\m{W} + \v{u}_i \m{V}_i^\tr\m{W}) - \bias + \frac{1}{2} (\v{u}\m{V}^\tr\m{W} + \v{u}_j \m{V}_j^\tr\m{W})
= \v{u}_i \m{V}_i^\tr\m{W} - \v{u}_j \m{V}_j^\tr\m{W}
\geq 0.
$ Thus, RISE importance will order in the same manner and is also optimal.
\end{proof}

\begin{corollary}[Optimal Deletion in the last layer]
When decomposing in the last layer,~\textbf{Gradient Input}, \textbf{Integrated Gradients}, \textbf{Occlusion}, and \textbf{Rise} yield the optimal solution for the C-Deletion metric.
\end{corollary}
\begin{proof}
It is simply observed that the C-Deletion problem seeks a minimum weight basis and corresponds to the same weighted matroid with weighting function $w'(\cdot) = -w(\cdot)$.
\end{proof}

\subsection{Sparse Autoencoder}

As a remainder, a general method (as it encompasses both PCA and K-means) to obtain the loading-dictionary pair and achieve a matrix reconstruction $\mathbf{A} = \mathbf{U} \mathbf{V}^\tr$ is to train a neural network to obtain $\mathbf{U}$ from $\mathbf{A}$ such that the reconstruction of $\mathbf{A}$ is linear in $\mathbf{U}$. This can be formally represented as:

$$
(\bm{\psi}^\star, \mathbf{V}^\star) = \arg\min_{\bm{\psi},\mathbf{V}} \| \mathbf{A} - \bm{\psi}(\mathbf{A}) \mathbf{V}^\top \|_F^2
$$

Here, $\mathbf{U}^\star = \bm{\psi}^\star(\mathbf{A}).$ An interesting characteristic of NMF and K-means is the non-linear relationship between $\mathbf{A}$ and $\mathbf{U}$. Specifically, the transformation from $\mathbf{A}$ to $\mathbf{U}$ is non-linear, while the transformation from $\mathbf{U}$ to $\mathbf{A}$ is linear, as explained in \cite{fel2022xplique}, which need to introduce a method based on implicit differentiation to obtain the gradient of $\mathbf{U}$ with respect to $\mathbf{A}$. Indeed, the sequence of operations to optimize $\mathbf{U}$ causes us to lose information about which elements of $\mathbf{A}$ contributed to obtaining $\mathbf{U}$. We believe that this non-linear relationship (absent in PCA) may be an essential ingredient for effective concept extraction.

Finally, as described in this article, other characteristics that appear to make it interpretable include its compositionality (due to non-extreme sparsity), good reconstruction, and positivity, which aids in interpretation. Thus, the architecture of $\bm{\psi}$ used for Figure~\ref{fig:holistic:qualitative_comparison} consists of a sequence of dense layers and batch normalization with ReLU activation to obtain positive scores and sparsity similar to NMF, without imposing constraints on $\mathbf{V}$. More formally, $\bm{\psi}$ is a sequence of layers as follows:

$$
\textsc{Dense(128) - BatchNormalization - ReLU}
$$
$$
\textsc{Dense(64) - BatchNormalization - ReLU}
$$
$$
\textsc{Dense(10) - BatchNormalization - ReLU}
$$

While the vector $\m{V}$ is initialized using a truncated SVD~\cite{fathi2023initialization}. We used Adam optimizer\cite{kingma2014adam} with a learning rate of $1e^{-3}$. However, it's worth noting that there is a wealth of literature on dictionary learning that remains to be explored for the task of concept extraction~\cite{dumitrescu2018dictionary}.

\section{MACO}

In this section, we provide additional results for logit and internal feature visualizations, and feature inversion. 

For all of the following visualizations, we used the same parameters as in the main paper.
For the feature visualizations derived from~\cite{olah2017feature}, we used all 10 transformations set from the Lucid library\footnote{\href{https://github.com/tensorflow/lucid}{https://github.com/tensorflow/lucid}}.
For \magfv, $\augmentation$ only consists of two transformations; first we add uniform noise $\bm{\delta} \sim \mathcal{U}([-0.1, 0.1])^{W \times H}$ and crops and resized the image with a crop size drawn from the normal distribution $\mathcal{N}(0.25, 0.1)$, which corresponds on average to 25\% of the image.
We used the NAdam optimizer \cite{dozat2016incorporating} with a $lr=1.0$ and $N = 256$ optimization steps. Finally, we used the implementation of \cite{olah2017feature} and CBR which are available in the Xplique library~\cite{fel2022xplique} \footnote{\href{https://github.com/deel-ai/xplique}{https://github.com/deel-ai/xplique}} which is based on Lucid.

\subsubsection{Logit and Internal State Visualization}\label{app:maco:fviz}

\begin{figure}[ht]
    \centering
    \includegraphics[width=0.99\textwidth]{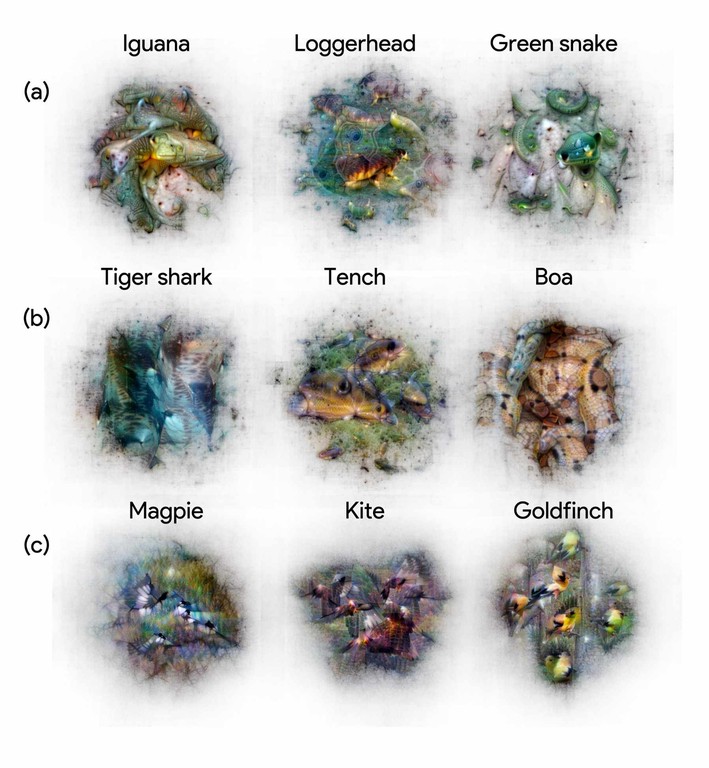}
    \caption{\textbf{Feature visualizations on FlexiViT, ViT and ResNet50.} We compare the feature visualizations from \magfv~generated for \textbf{(a)} FlexiViT, \textbf{(b)} ViT and \textbf{(c)} ResNet50 on a set of different classes from ImageNet. We observe that the visualizations get more abstract as the complexity of the model increases.}
    \label{fig:supp-qualitative}
\end{figure}

\begin{figure}[ht]
    \centering
    \includegraphics[width=0.99\textwidth]{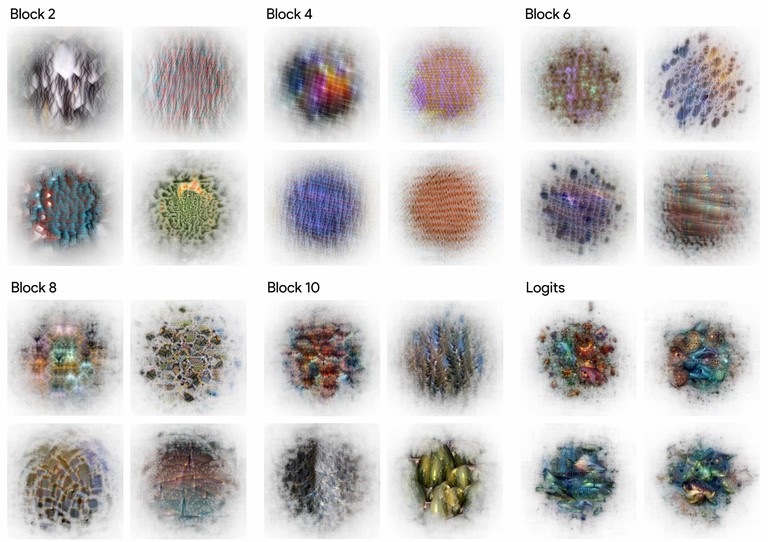}
    \caption{\textbf{Logits and internal representation of a ViT.} Using \magfv, we maximize the activations of specific channels in different blocks of a ViT, as well as the logits for 4 different classes.}
    \label{fig:supp-internal}
\end{figure}

\begin{figure}[ht]
    \centering
    \includegraphics[width=0.99\textwidth]{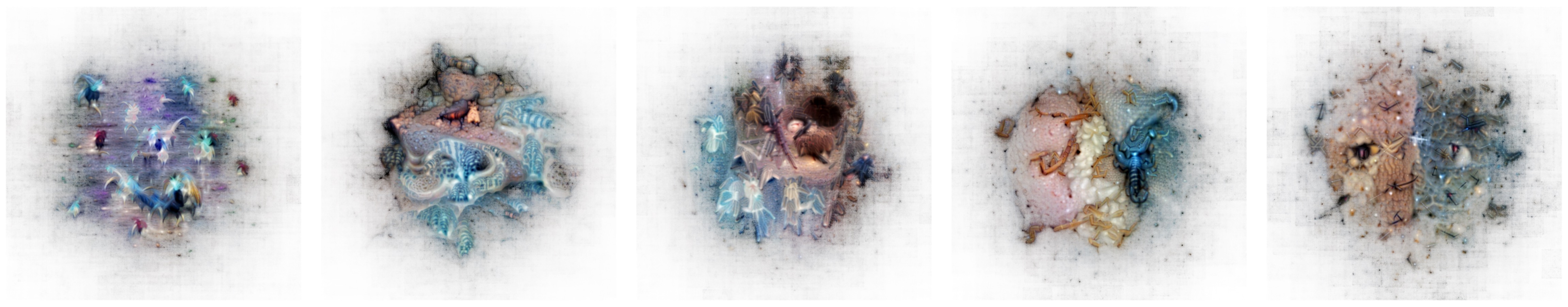}
    \caption{\textbf{Hue invariance.} Through feature visualization, we are able to determine the presence of hue invariance on our pre-trained ViT model manisfesting itself through phantom objects in them. This can be explained the data-augmentation that is typically employed for training these models.}
    \label{fig:hue-inv}
\end{figure}

\subsubsection{Feature Inversion}\label{app:maco:inversion}

\begin{figure}[ht]
    \centering
    \includegraphics[width=0.99\textwidth]{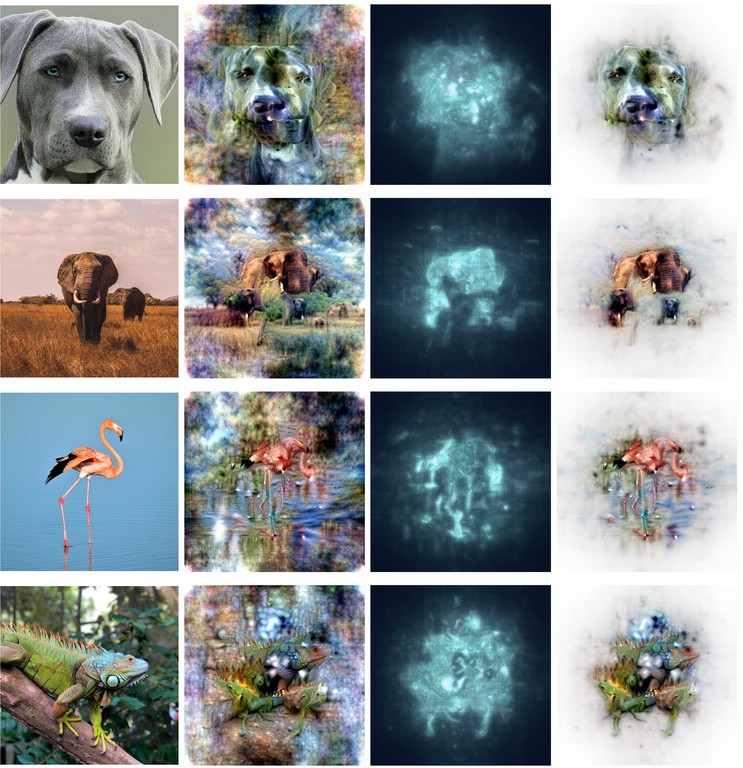}
    \caption{\textbf{Feature inversion and Attribution-based transparency.} We performed feature inversion on the images on the first column to obtain the visualizations (without transparency) on the second column. During the optimization procedure, we saved the intensity of the changes to the image in pixel space, which we showcase on the third column, we used this information to assign a transparency value, as exhibited in the final column.}
    \label{fig:supp-inv}
\end{figure}

\subsection{Human psychophysical study}\label{sup:maco:psychophysics}

To evaluate \magfv~'s ability to improve humans' causal understanding of a CNN's activations, we conducted a psychophysical study closely following the paradigm introduced in \cite{zimmermann2021well}. In this paradigm, participants are asked to predict which of two query inputs would be favored by the model (i.e., maximally activate a given unit), based on example "favorite" inputs serving as a reference (i.e., feature visualizations for that unit). The two queries are based on the same natural image, but differ in the location of an occludor which hides part of the image from the model.

\paragraph{Participants.} We recruited a total of 191 participants for our online psychophysics study using Prolific (www.prolific.com) [September 2023]. As compensation for their time (roughly 7 minutes), participants were paid 1.4\$. Of those who chose to disclose their age, the average age was 39 years old ($SD = 13$). Ninety participants were men, 86 women, 8 non-binary and 7 chose not to disclose their gender. The data of 17 participants was excluded from further analyses because they performed significantly below chance ($p < .05$, one-tailed).

\paragraph{Design.} Participants were randomly assigned to one of four Visualization conditions: Olah~\cite{olah2017feature}, \magfv~with mask, \magfv~without mask, or a control condition in which no visualizations were provided. Furthermore, we varied Network (VGG16, ResNet50, ViT) as a within-subjects variable. The specific units whose features to visualize were taken from the output layer, meaning they represented concrete classes. The classes were: Nile crocodile, peacock, Kerry Blue Terrier, Giant Schnauzer, Bernese Mountain Dog, ground beetle, ringlet, llama, apiary, cowboy boot, slip-on shoe, mask, computer mouse, muzzle, obelisk, ruler, hot dog, broccoli, and mushroom. For every class, we included three natural images to serve as the source image for the query pairs. This way, a single participant would see all 19 classes crossed with all 3 networks, without seeing the same natural image more than once (which image was presented for which network was randomized across participants). The main experiment thus consisted of 57 trials, with a fully randomized trial order.

\paragraph{Stimuli.} The stimuli for this study included 171 ((4-1)x3x19) reference stimuli, each displaying a 2x2 grid of feature visualizations, generated using the respective visualization method. The query pairs were created from each of the 57 (19x3) source images by placing a square occludor on them. In one member of the pair, the occludor was placed such that it minimized the activation of the unit. In the other member of the pair, the occludor was placed on an object of a different class in the same image or a different part of the same object. Here, we deviated somewhat from the query geneation in \cite{zimmermann2021well}, where the latter occludor was placed where it maximized the activation of the unit. However, we observed that this often resulted in the occludor being on the background, making the task trivial. Indeed, a pilot study ($N=42$) we ran with such occludor placement showed that even the participants in the control condition were on average correct in $83\%$ of the trials. 

\begin{figure}[ht]
    \centering
    \includegraphics[width=0.99\textwidth]{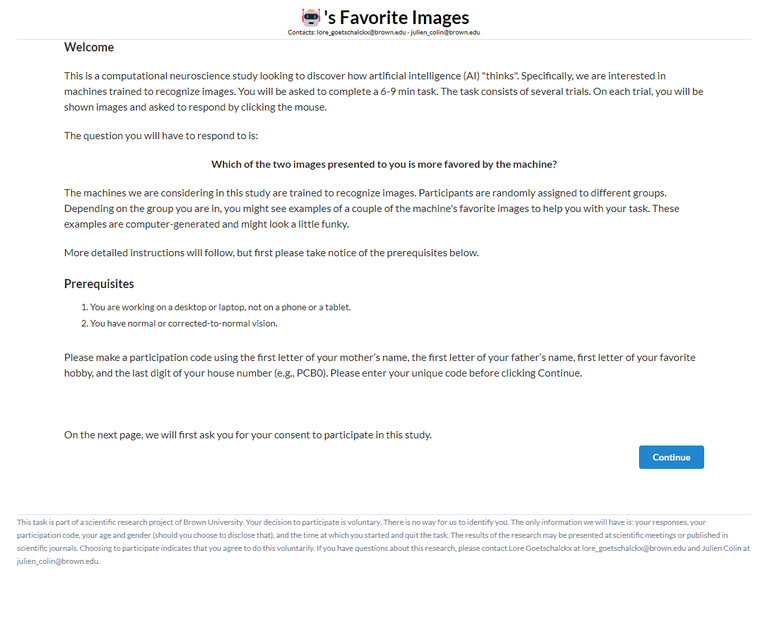}
    \caption{\textbf{Welcome page.} This is a screenshot of the first page participants saw when entering our online psychophysics study.}
    \label{fig:psychophysics-welcome}
\end{figure}

\begin{figure}[ht]
    \centering
    \includegraphics[width=0.99\textwidth]{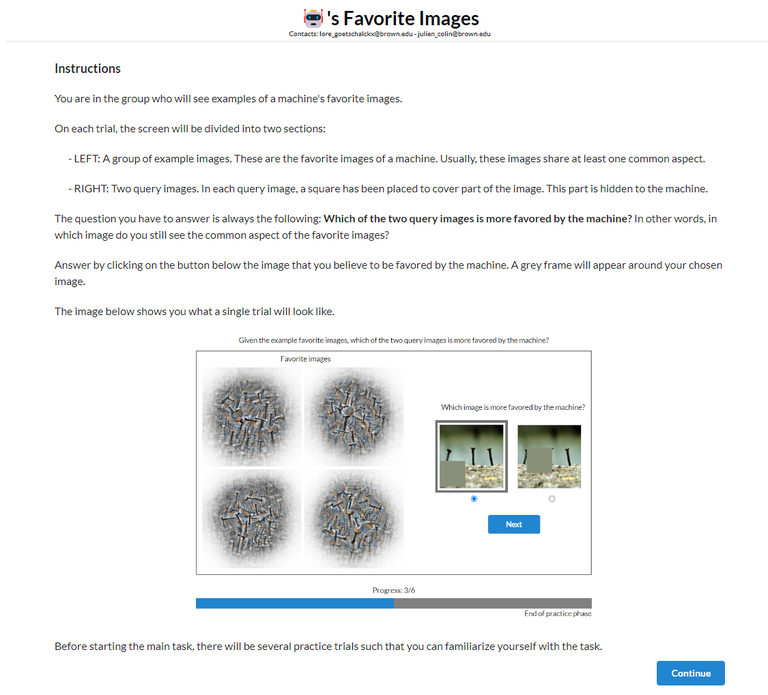}
    \caption{\textbf{Instructions page.} After providing informed consent, participants in our online psychophysics task received more detailed instructions, as shown here.}
    \label{fig:psychophysics-instructions}
\end{figure}

\begin{figure}[ht]
    \centering
    \includegraphics[width=0.99\textwidth]{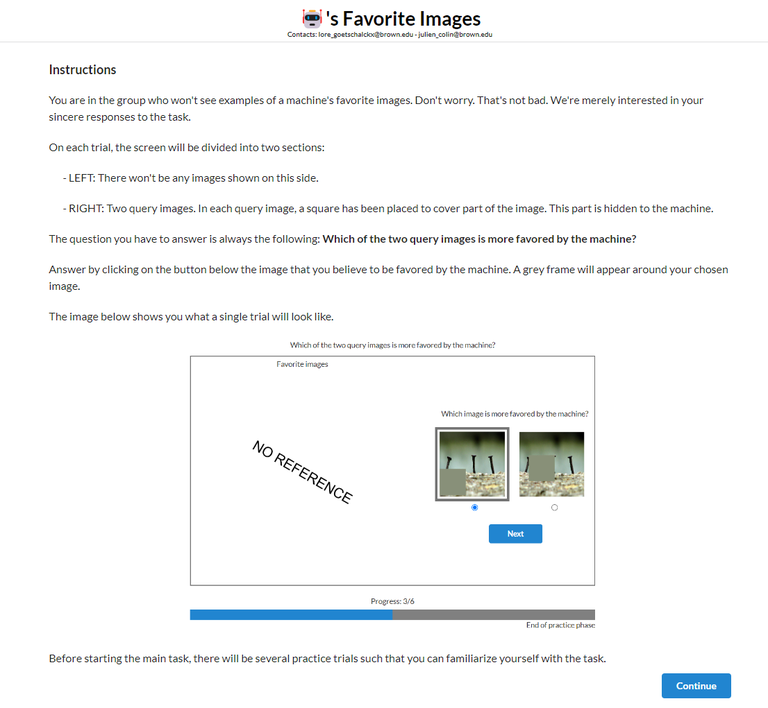}
    \caption{\textbf{Instructions page for control condition.} After providing informed consent, participants in our online psychophysics task received more detailed instructions, as shown here. If they were randomly assigned to the control condition, they were informed that they would not see examples of the machine's favorite images. }
    \label{fig:psychophysics-instructions-control}
\end{figure}

\begin{figure}[ht]
    \centering
    \includegraphics[width=0.99\textwidth]{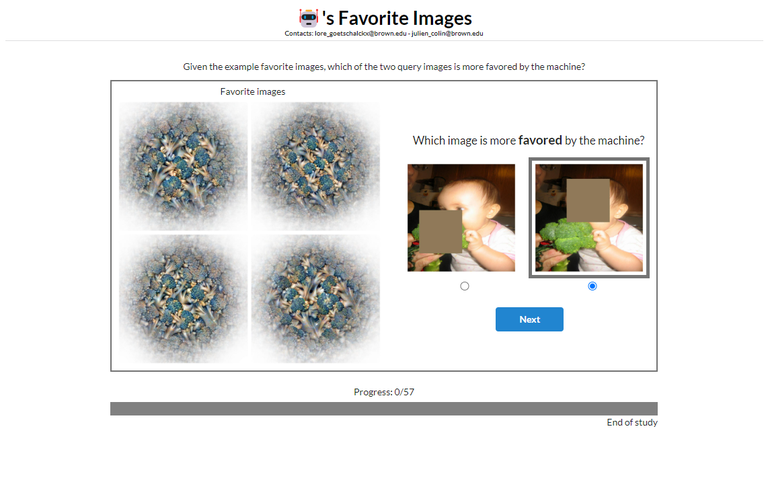}
    \caption{\textbf{Example trial.} On every trial of our psychophysics study, participants were asked to select which of two query images would be favored by the machine. They were shown examples of the machine's favorite inputs (i.e., feature visualizations) on the left side of the screen.}
    \label{fig:psychophysics-trial}
\end{figure}

\begin{figure}[ht]
    \centering
    \includegraphics[width=0.99\textwidth]{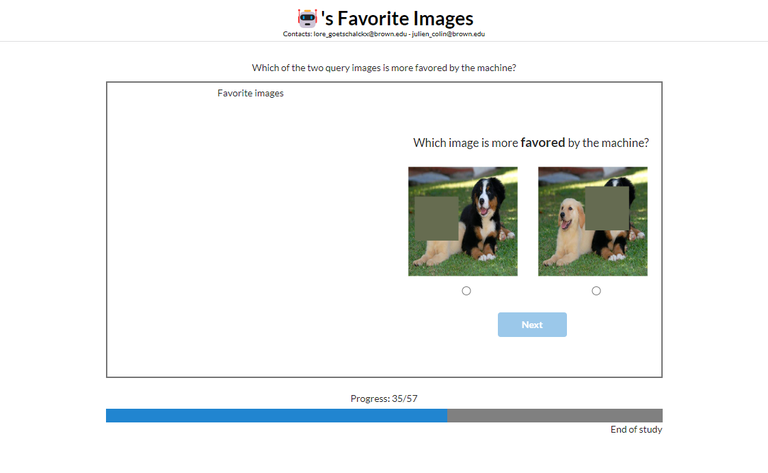}
    \caption{\textbf{Example trial in the control condition.} On every trial of our psychophysics study, participants were asked to select which of two query images would be favored by the machine. In the control condition, they were not shown examples of the machine's favorite inputs and the left side of the screen remained empty.}
    \label{fig:psychophysics-trial-control}
\end{figure}

\paragraph{Task and procedure.} The protocol was approved by the University IRB and was carried out in accordance with the provisions of the World Medical Association Declaration of Helsinki. Participants were redirected to our online study through Prolific and first saw a page explaining the general purpose and procedure of the study (Fig.~\ref{fig:psychophysics-welcome}). Next, they were presented with a form outlining their rights as a participant and actively had to click ``I agree'' in order to give their consent. More detailed instructions were given on the next page (Fig.~\ref{fig:psychophysics-instructions}, Fig.~\ref{fig:psychophysics-instructions-control}). Participants were instructed to answer the following question on every trial: ``Which of the two query images is more favored by the machine?''. The two query images were presented on the right-hand side of the screen. The feature visualizations were displayed on the left-hand side of the screen (Fig.~\ref{fig:psychophysics-trial}). In the control condition, the left-hand side remained blank (Fig.~\ref{fig:psychophysics-trial-control}). Participants could make their response by clicking on the radio button below the respective query image. They first completed a practice phase, consisting of six trials covering two additional classes, before moving on to the main experiment. For the practice trials, they received feedback in the form of a green (red) frame appearing around their selected query image if they were correct (incorrect). No such feedback was given during the main experiment.

\paragraph{Analyses and results.} We analyzed the data through a logistic mixed-effects regression analysis, with trial accuracy (1 vs. 0) as the dependent variable. The random-effects structure included a by-participant random intercept and by-class random intercept. We compared two regression models, both of which had Visualization and Network as a fixed effect, but only one also fitted an interaction term between the two. Based on the Akaike Information Criterion (AIC), the former, less complex model was selected ($AIC=11481 vs. 11482$). Using this model, we then analyzed all pairwise contrasts between the levels of the Visualization variable. We found that the logodds of choosing the correct query were overall significantly higher in both \magfv~conditions compared to the control condition: $\beta_{\magfv~Mask}-\beta_{Control} = 0.69, SE=0.13, z=5.38, p<.0001;\beta_{\magfv~NoMask}-\beta_{Control} = 0.92, SE=0.13, z=7.07, p<.0001.$ Moreover, \magfv~visualizations helped more than Olah visualizations: $\beta_{\magfv~Mask}-\beta_{Olah} = 0.43, SE=0.13, z=3.31, p=.005;\beta_{\magfv~NoMask}-\beta_{Olah} = 0.66, SE=0.13, z=4.99, p<.0001.$ No other contrasts were statistically significant (at a level of $p < .05$). $P$-values were adjusted for multiple comparisons with the Tukey method. Finally, we also examined the pairwise contrasts for the Network variable. We found that ViT was the hardest model to interpret overall: $\beta_{ResNet50}-\beta_{ViT} = 0.49, SE=0.06, z=8.65, p<.0001;\beta_{VGG16}-\beta_{ViT} = 0.35, SE=0.06, z=6.38, p<.0001.$ There was only marginally significant evidence that participants could better predict ResNet50's behavior in this task than VGG16: $\beta_{ResNet50}-\beta_{VGG16} = 0.13, SE=0.06, z=2.30, p=0.056.$

Taken together, these results suggest that \magfv~indeed helps humans causally understand a CNN's activations and that it outperforms Olah's method \cite{olah2017feature} on this criterion.

\clearpage

\chapter{Conclusion}

\section{Toy example}
\label{ap:conclusion:toy_abs}

\begin{figure}
\begin{lstlisting}
import tensorflow as tf
import numpy as np

N = 100_000  # nb of points
d = 2        # dimension of input
n = 4        # internal dimension

x = (np.random.rand(N, d) * 2.0) - 1.0  # sample on [-1, 1]
y = np.abs(x)

nn = tf.keras.Sequential([
    tf.keras.layers.Dense(n, use_bias=False),
    tf.keras.layers.Activation('relu'),
    tf.keras.layers.Dense(d, use_bias=False)
])

nn.compile(loss='mse', optimizer=tf.keras.optimizers.Adam(1e-3))
nn.fit(x, y, epochs=10)
\end{lstlisting}
\caption{\textbf{Pythonic implementation of the toy example.}}
\end{figure}

\clearpage

\end{document}